\documentclass{article}
\usepackage[a4paper,margin=1in]{geometry}
\usepackage{times}
\usepackage{natbib}
\usepackage{microtype}
\usepackage{float}
\usepackage{caption}
\usepackage{fancyhdr}
\usepackage[utf8]{inputenc}
\usepackage[T1]{fontenc}
\usepackage{amsmath,amssymb}
\usepackage{amsthm}
\theoremstyle{definition}
\newtheorem{definition}{Definition}

\usepackage{graphicx,booktabs,enumitem,xspace,multirow,longtable,listings,array,xcolor,wrapfig}
\newcolumntype{P}[1]{>{\raggedright\arraybackslash}p{#1}}   
\usepackage{hyperref}
\hypersetup{colorlinks=true,linkcolor=black,citecolor=black,urlcolor=blue}   
\usepackage{url}

\newcommand{\ActionBudget}{48\xspace}  
\newcommand{\AdmissionAnineteenStatus}{pass\xspace}  
\newcommand{\Alpha}{0.05\xspace}  
\newcommand{\AtwoChangedDismissedLog}{8\xspace}  
\newcommand{\AtwoChangedN}{8\xspace}  
\newcommand{\AtwoCleanN}{2\xspace}  
\newcommand{\AtwoCleanPass}{2\xspace}  
\newcommand{\AtwoEffect}{0.125\xspace}  
\newcommand{\AtwoInstructionDistinct}{7\xspace}  
\newcommand{\AtwoInstructionIdenticalInstances}{8\xspace}  
\newcommand{\AtwoInstructionInstances}{8\xspace}  
\newcommand{\AtwoManifestSha}{\texttt{b66756ff}\xspace}  
\newcommand{\AtwoModuleSha}{\code{17f6c7cd72f2}\xspace}  
\newcommand{\AtwoModuleWhere}{Block~A2's task module (\AtwoModuleSha), absent from the working tree, was recovered from a backup and verified by its digest, and ships in the supplement\xspace}  
\newcommand{\AtwoRederivedSignatures}{5\xspace}  
\newcommand{\AtwoReplaced}{1\xspace}  
\newcommand{\AtwoSeedList}{8--15\xspace}  
\newcommand{\AtwoSuccessN}{7\xspace}  
\newcommand{\AtwoSuccessReread}{7\xspace}  
\newcommand{\AtwoVOneExposed}{8\xspace}  
\newcommand{\BCRCReachText}{An instance favours at most one adjacent contrast (CUE-A must pass to favour CUE--OFF and fail to favour LIVE--CUE), so the reading that both reject under the block-local Holm analysis (R-C) needs at least 7 favourable discordances on one and 6 on the other, 13 in all, with none adverse: with 16 triples it is reachable only if at least 13 of them are favourable.\xspace}  
\newcommand{\BCRuleSourceSha}{\texttt{493abaaeeb4b}\xspace}  
\newcommand{\BCSAttempts}{24\xspace}  
\newcommand{\BCSCellsCounted}{21\xspace}  
\newcommand{\BCSCellsRegistered}{51\xspace}  
\newcommand{\BCSExecSentence}{ \emph{Execution record (no interpretation).} Block~BCS counted \BCSCellsCounted{} of its \BCSCellsRegistered{} registered cells in \BCSAttempts{} attempts (\BCSNotCounted{} not counted) before its registered launch cutoff, 12:00~UTC on 25~September. No attempt was logged between 19:30~UTC on 24~September and 04:34~UTC on 25~September: an edit to the paper generator had changed the source digest of the sealed reading rule, so the block's launch gate refused to resume it. The rule was restored byte-identically and the block resumed from its frozen search log, re-running no counted row (\code{REPAIR\_\allowbreak{}r33/\allowbreak{}BCS\_\allowbreak{}HALT\_\allowbreak{}DEVIATION.md}). The registered terminal rule had made the runner's failure terminal, so the resumption is a disclosed deviation (Appendix~\ref{app:disclosures}).\xspace}  
\newcommand{\BCSHaltPointer}{\paragraph{Block BCS: halt and resumption (a disclosed deviation).} An operator error in a paper file, not a rollout outcome, halted Block~BCS for about nine hours: an edit changed the source digest of the sealed reading rule, and the block's launch gate refused to resume it. Under the registered terminal rule the runner's failure was terminal; instead the rule was restored byte-identically and the block resumed from its frozen search log, re-running no counted row, with its cells, order and cutoff unchanged. The operator had seen the verdicts of its first four attempts; the resumption depended on none of them. The block ended incomplete at its registered experiment cutoff, so both of its paper-wide slots enter as $p=1$ placeholders (Appendix~\ref{app:bc}).\xspace}  
\newcommand{\BCSNotCounted}{3\xspace}  
\newcommand{\BGCDispositionDetail}{The terminal disposition recorded before the experiment cutoff lists the unmet launch preconditions: its release manifest, its admission link and its inline-delivery dry run.\xspace}  
\newcommand{\BPairs}{10\xspace}  
\newcommand{\BPairsWords}{ten\xspace}  
\newcommand{\BSAmendAddenda}{11 and 12\xspace}  
\newcommand{\BSInvalidUnderRule}{8\xspace}  
\newcommand{\BSManifestNow}{\code{ead3fa67655a}\xspace}  
\newcommand{\BSPreAmendCounted}{1\xspace}  
\newcommand{\BSPreAmendManifest}{\code{9aa433099a46}\xspace}  
\newcommand{\BSPreAmendRows}{4\xspace}  
\newcommand{\BSReserveLeft}{3\xspace}  
\newcommand{\BSReserveTotal}{6\xspace}  
\newcommand{\BSReserveUsed}{3\xspace}  
\newcommand{\BSeedList}{20--29\xspace}  
\newcommand{\BlindAdmClaims}{27\xspace}  
\newcommand{\BlindAdmClinic}{9\xspace}  
\newcommand{\BlindAdmitted}{36\xspace}  
\newcommand{\BlindAmbiguous}{1\xspace}  
\newcommand{\BlindAppendixTextR}{\emph{This is neither the sealed challenge (Table~\ref{tab:challenge}) nor the post-hoc author-built checks of Appendix~\ref{app:posthoc}, and it changes neither result.} \emph{Design.} GPT-6 (ChatGPT, a new temporary chat with memory off) received a brief of verbatim excerpts of the registered rules (the claims pre-registration and the clinic addenda that govern Block~BC) and an archive schema (file list, what each log records, one type skeleton per file); the brief held no code, no paper text, no earlier case, class or finding, and no model outcome, and three redactions removed a historical defect label, a defect narrative and changelog identifiers. It returned \BlindSpecs{} case specifications (\BlindSpecsPerWorkflow{} per workflow: \BlindSpecClaimsCorrupt{} corruptions and \BlindSpecClaimsControls{} controls each), each with its expected verdict, its rule basis and any accepted alternative; they were parsed without edits, and GPT-6 marked \BlindAmbiguous{} as ambiguous. \emph{Implementation and seal.} The base-selection, exclusion and scoring rules were committed by digest at \BlindRulesUTC, before any answer text had been retrieved (the retrieval log records none until \BlindFirstTextUTC). The authors then implemented each specification by literal application of its named edits (a specification that could not be applied as written was to be excluded, not reinterpreted) on a copy of a counted prospective archive (claims: Blocks CLM-B and CLM-BS; clinic: Block~BC), editing every copy of a fact and recomputing recorded digests. The answer, specifications, cases, expectations and all implementation, run and comparison code were sealed at \BlindSealUTC; no tool under test had run on any case before then, so no case could be altered to make a tool fail. A second GPT-6 temporary chat---a model, not a human adjudicator---compared every implementation with its specification and judged \BlindAdmitted{} of the \BlindImplemented{} implemented ones faithful; cases it did not confirm were excluded under the frozen admission rule, and the fidelity claim remains the authors'; its decision was recorded at \BlindFidelityUTC. The authors checked every admitted expectation against the rule text and disputed none; expectations are copied verbatim and none was edited, and under the frozen rules a dispute would have been reported beside the case without changing any score. The run began at \BlindRunUTC, used no model and no world, and the tool and case files did not change during it. \emph{Exclusions (pre-declared rules).} Of the specifications, \BlindNoBase{} had no base archive meeting their stated requirement: \BlindNoBaseClinic{} clinic ones (\BlindNoBaseOffSucc{} need a successful OFF-A episode, and Block~BC has none) and \BlindNoBaseClaims{} claims one; the brief, being blind, did not say which outcomes each level contains, and the fidelity reviewer judged \BlindUnfaithful{} (\BlindUnfaithfulIDs) unfaithful; none of these ran. Admitted: \BlindAdmClaims{} claims cases (\BlindCorruptClaims{} corruptions, \BlindUncountedClaims{} of them expected uncounted, and \BlindControlsClaims{} controls) and \BlindAdmClinic{} clinic cases (\BlindCorruptClinic{} corruptions, all expected uncounted, and \BlindControlsClinic{} controls). \emph{Tools.} \emph{claims}: combined pipeline (classifier, void gate), evaluator with classifier, claims validity re-derivation (C8), independent scorer, simple raw-log baseline (r28), repaired raw-log checker (r29); \emph{clinic}: combined pipeline (driver row, void gate), evaluator with classifier (refresh), BC validity re-derivation, independent scorer, simple raw-log clinic baseline (r29). The clinic evaluator path re-applies what the harness computes from archived evidence, with the world readouts as archived. Two disclosed deviations: the runner crashed in its own bookkeeping on the case with a deleted trajectory before any claims tool had produced output; the one-line fix changed only that bookkeeping (both runner digests recorded; the tool files are identical across the two invocations), the claims part then ran once, and the clinic part, which had already run once, was not rerun; and the brief described the agent-attributed access log as the same list as the trajectory's, so three cases were implemented on the trajectory's copy with the slice re-derived, which the fidelity reviewer was told and confirmed.\par\paragraph{Results and every miss.} Table~\ref{tab:blind-tools} gives each tool; Table~\ref{tab:blind-cases} every case. The combined pipeline accepted all \BlindControls{} controls (\BlindControlsClaims{} claims, \BlindControlsClinic{} clinic) and, of the \BlindCorrupt{} corruptions, rejected or flagged \BlindPipeCaught{} (including the wrong-class case below), gave no verdict on \BlindPipeErr{} and counted none as clean: no false acceptance and \BlindPipeErr{} abstentions, which is not \BlindCorrupt/\BlindCorrupt{} detection. It missed three: on KC01, KC08 (missing trajectory, missing required artifact) the classifier alone returned the expected \code{harness\_error}, but the evaluator raised a file-not-found error, so the pipeline reached no verdict; on KC13 (missing required ON review panel) it rejected the case as \code{invalid\_intervention} rather than \code{observation\_not\_delivered}, because the evaluator's panel-geometry validity term precedes the delivery row. The claims validity re-derivation counted \BlindClaimsCEightFa{} corruptions as clean (KC06, KC07, KC09, KC11, KC12, KC13, KC14, KC15, KC20); each falls under a countedness row it does not re-derive by design, since it re-derives intervention validity only. The independent scorer counted \BlindClaimsScorerFa{} claims corruptions as clean (KC02, KC03, KC04, KC05, KC06, KC11, KC14, KC15, KC20): it does not read the stop reason, recorded infrastructure errors, read-target violations, per-call model identity, harness text across steps, access-log episode stamps or the arm payload. In the clinic, Block~BC's validity re-derivation and the scorer both counted LC07 (supplied-frame failure on silent arm) as clean: the validity script re-derives frame supply from the frame file, which the case left in place, and the scorer's input check gates only rows the driver counted; the operational driver voided the case as expected, and in a block that disagreement triggers the void gate. The simple raw-log baselines counted \BlindClaimsBaselineFa{} of the \BlindUncountedClaims{} claims corruptions expected uncounted as clean (the repaired checker the same cases) and \BlindClinicBaselineFa{} of the \BlindUncountedClinic{} clinic ones; none reads stop reasons, infrastructure errors, read targets, prompts, episode stamps or arm payloads. The clinic baseline accepted every control but never reports the booked-slot term I1 as passing, so it meets no exact expectation.\par\paragraph{Limits.} Coverage is uneven: the clinic side has \BlindCorruptClinic{} corruptions, all expected uncounted, and no outcome-invariant corruption, because every such specification needed a base outcome Block~BC does not contain; \BlindClaimsLZero{} of the \BlindAdmClaims{} claims cases use the two L0 archives, and every clinic base is an Opus Block~BC archive. The results are per case, not rates. The authors chose the brief's excerpts, wrote the implementation and chose the tools, having read the earlier challenges; the implementation was sealed before any tool ran, and a model, not an independent team, reviewed its fidelity. What was blind is exactly the case design: specifications and expected verdicts were written without the implementation, the paper, earlier cases or model outcomes. The controls are a constructed case set, not a false-positive-rate estimate, and claims and clinic counts are never pooled into a rate. GPT-6's expectations are its readings of the registered rules.\xspace}  
\newcommand{\BlindBaseFA}{19\xspace}  
\newcommand{\BlindClaimsBaselineFa}{14\xspace}  
\newcommand{\BlindClaimsCEightFa}{9\xspace}  
\newcommand{\BlindClaimsLZero}{19\xspace}  
\newcommand{\BlindClaimsScorerFa}{9\xspace}  
\newcommand{\BlindClinicBaselineFa}{5\xspace}  
\newcommand{\BlindControls}{10\xspace}  
\newcommand{\BlindControlsClaims}{7\xspace}  
\newcommand{\BlindControlsClinic}{3\xspace}  
\newcommand{\BlindCorrupt}{26\xspace}  
\newcommand{\BlindCorruptClaims}{20\xspace}  
\newcommand{\BlindCorruptClinic}{6\xspace}  
\newcommand{\BlindFidelityUTC}{2026-09-24 19:08Z\xspace}  
\newcommand{\BlindFirstTextUTC}{2026-09-24 18:51Z\xspace}  
\newcommand{\BlindImplemented}{38\xspace}  
\newcommand{\BlindMainThree}{\paragraph{Rule-only case-design challenge.} A fresh GPT-6 chat given only rule excerpts and the archive schema (no code, paper, earlier cases or outcomes) specified \BlindSpecs{} cases with expected verdicts. We implemented them on archive copies under frozen rules; a second chat reviewed fidelity, and \BlindAdmitted{} admitted cases ran once. The pipeline accepted all \BlindControls{} controls and, of \BlindCorrupt{} corruptions, caught \BlindPipeCaught{} (one under another class), gave no verdict on \BlindPipeErr{} and counted none as clean (outcome scorer alone: \BlindScorerFA; raw-log baselines: \BlindBaseFA). Only the case design was blind, and clinic coverage is thin (Appendix~\ref{app:blind}).\xspace}  
\newcommand{\BlindNoBase}{18\xspace}  
\newcommand{\BlindNoBaseClaims}{1\xspace}  
\newcommand{\BlindNoBaseClinic}{17\xspace}  
\newcommand{\BlindNoBaseOffSucc}{13\xspace}  
\newcommand{\BlindPipeCaught}{24\xspace}  
\newcommand{\BlindPipeErr}{2\xspace}  
\newcommand{\BlindRulesUTC}{2026-09-24 18:44Z\xspace}  
\newcommand{\BlindRunUTC}{2026-09-24 19:08Z\xspace}  
\newcommand{\BlindScorerFA}{10\xspace}  
\newcommand{\BlindSealUTC}{2026-09-24 19:00Z\xspace}  
\newcommand{\BlindSpecClaimsControls}{8\xspace}  
\newcommand{\BlindSpecClaimsCorrupt}{20\xspace}  
\newcommand{\BlindSpecs}{56\xspace}  
\newcommand{\BlindSpecsPerWorkflow}{28\xspace}  
\newcommand{\BlindUncountedClaims}{16\xspace}  
\newcommand{\BlindUncountedClinic}{6\xspace}  
\newcommand{\BlindUnfaithful}{2\xspace}  
\newcommand{\BlindUnfaithfulIDs}{LC14, LC15\xspace}  
\newcommand{\BlockBCConfig}{Claude CLI, frames by file read\xspace}  
\newcommand{\BlockBCContradictions}{0\xspace}  
\newcommand{\BlockBCControls}{3\xspace}  
\newcommand{\BlockBCCounted}{51\xspace}  
\newcommand{\BlockBCCountedFail}{31\xspace}  
\newcommand{\BlockBCCountedSucc}{20\xspace}  
\newcommand{\BlockBCCueOffHarm}{0\xspace}  
\newcommand{\BlockBCCueOffHolmLocal}{0.5\xspace}  
\newcommand{\BlockBCCueOffNetCI}{$+0.13$ [$-0.09$, $0.36$]\xspace}  
\newcommand{\BlockBCCueOffP}{0.5\xspace}  
\newcommand{\BlockBCCueOffRescue}{2\xspace}  
\newcommand{\BlockBCCueOffTie}{14\xspace}  
\newcommand{\BlockBCDisposition}{completed\xspace}  
\newcommand{\BlockBCEndpointFull}{51\xspace}  
\newcommand{\BlockBCFRXCueALinkFollowed}{1\xspace}  
\newcommand{\BlockBCFRXCueAN}{16\xspace}  
\newcommand{\BlockBCFRXCueAPassFrac}{2/16\xspace}  
\newcommand{\BlockBCFRXCueARecovered}{1\xspace}  
\newcommand{\BlockBCFRXLiveAN}{16\xspace}  
\newcommand{\BlockBCFRXLiveAPassFrac}{15/16\xspace}  
\newcommand{\BlockBCFRXLiveARecovered}{15\xspace}  
\newcommand{\BlockBCFRXLiveIPassFrac}{3/3\xspace}  
\newcommand{\BlockBCFRXOffAPassFrac}{0/16\xspace}  
\newcommand{\BlockBCInstances}{16\xspace}  
\newcommand{\BlockBCLiveCueHarm}{0\xspace}  
\newcommand{\BlockBCLiveCueHolmLocal}{0.000488\xspace}  
\newcommand{\BlockBCLiveCueHolmPaper}{0.00293\xspace}  
\newcommand{\BlockBCLiveCueNetCI}{$+0.81$ [$0.49$, $0.92$]\xspace}  
\newcommand{\BlockBCLiveCueP}{0.000244\xspace}  
\newcommand{\BlockBCLiveCuePairs}{16\xspace}  
\newcommand{\BlockBCLiveCueRescue}{13\xspace}  
\newcommand{\BlockBCLiveCueTie}{3\xspace}  
\newcommand{\BlockBCLiveOffHarm}{0\xspace}  
\newcommand{\BlockBCLiveOffNetCI}{$+0.94$ [$0.64$, $0.99$]\xspace}  
\newcommand{\BlockBCLiveOffRescue}{15\xspace}  
\newcommand{\BlockBCLiveOffTie}{1\xspace}  
\newcommand{\BlockBCMinPairs}{13\xspace}  
\newcommand{\BlockBCModel}{Claude Opus~5\xspace}  
\newcommand{\BlockBCOffGate}{8\xspace}  
\newcommand{\BlockBCOrderWord}{frozen\xspace}  
\newcommand{\BlockBCReadingBody}{Claude Opus~5 passed OFF-A \BlockBCFRXOffAPassFrac, CUE-A \BlockBCFRXCueAPassFrac and LIVE-A \BlockBCFRXLiveAPassFrac on \BlockBCInstances instances (Table~\ref{tab:tests} gives every contrast). At the registered block-local level, displaying the current location increased success relative to this cue-and-link panel. The CUE--OFF contrast did not reject; this does not establish absence of a cue effect. Contradiction, salience, and reduced verification effort remain possible explanations. The LIVE--CUE rejection survives the paper-wide Holm analysis over \HolmPaperM claim-bearing tests (adjusted $p=\BlockBCLiveCueHolmPaper$). The LIVE-I controls passed \BlockBCFRXLiveIPassFrac; nothing is attributed to the panel from unmatched controls. The independent scorer reconstructs appointment availability from archived rows; calendar compatibility is not re-derived but rests on stated design conditions (the generator and state preservation), whose checks are reported separately. Under that condition, every outcome term is resolved and agreeing with the evaluator on \BlockBCEndpointFull of \BlockBCCounted counted rollouts (Table~\ref{tab:coverage}).\xspace}  
\newcommand{\BlockBCReadingFull}{Over the \BlockBCInstances instances (seeds \BlockBCSeedList), Claude Opus~5 passed OFF-A \BlockBCFRXOffAPassFrac, CUE-A \BlockBCFRXCueAPassFrac and LIVE-A \BlockBCFRXLiveAPassFrac. LIVE versus CUE (primary): rescue \BlockBCLiveCueRescue, harm \BlockBCLiveCueHarm, tie \BlockBCLiveCueTie of \BlockBCLiveCuePairs pairs, net \BlockBCLiveCueNetCI (marginal 95\,\% interval), $p=\BlockBCLiveCueP$ (block-local Holm $p=\BlockBCLiveCueHolmLocal$). CUE versus OFF (key secondary): \BlockBCCueOffRescue/\BlockBCCueOffHarm/\BlockBCCueOffTie, net \BlockBCCueOffNetCI, $p=\BlockBCCueOffP$ (block-local Holm $p=\BlockBCCueOffHolmLocal$). LIVE versus OFF (supporting, descriptive, outside the paper-wide family): \BlockBCLiveOffRescue/\BlockBCLiveOffHarm/\BlockBCLiveOffTie, net \BlockBCLiveOffNetCI. At the registered block-local level, displaying the current location increased success relative to this cue-and-link panel. The CUE--OFF contrast did not reject; this does not establish absence of a cue effect. Contradiction, salience, and reduced verification effort remain possible explanations. The LIVE--CUE rejection survives the paper-wide Holm analysis over \HolmPaperM claim-bearing tests (adjusted $p=\BlockBCLiveCueHolmPaper$). The LIVE--OFF comparison is a supporting descriptive analysis; it supports neither a confirmatory replication claim nor a claim that the effect differs from Block~B. The LIVE-I controls passed \BlockBCFRXLiveIPassFrac; nothing is attributed to the panel from unmatched controls. The independent scorer reconstructs appointment availability from archived rows; calendar compatibility is not re-derived but rests on stated design conditions (the generator and state preservation), whose checks are reported separately. Under that condition, every outcome term is resolved and agreeing with the evaluator on \BlockBCEndpointFull of \BlockBCCounted counted rollouts (Table~\ref{tab:coverage}). Independent code re-derived this block's validity verdicts (\BlockBCVindPath), agreeing with the archive on \BlockBCVindAgree attempts. Strict recovery occurred in \BlockBCFRXCueARecovered of \BlockBCFRXCueAN CUE-A and \BlockBCFRXLiveARecovered of \BlockBCFRXLiveAN LIVE-A runs, and \BlockBCFRXCueALinkFollowed of \BlockBCFRXCueAN CUE-A runs followed the panel's link (descriptive; never a reading condition).\xspace}  
\newcommand{\BlockBCReadingIDs}{R-A, A19\xspace}  
\newcommand{\BlockBCRegistered}{ADDENDUM~18\xspace}  
\newcommand{\BlockBCSConfig}{Claude CLI, frames by file read\xspace}  
\newcommand{\BlockBCSDisposition}{incomplete at the registered experiment cutoff\xspace}  
\newcommand{\BlockBCSModel}{Claude Sonnet~5\xspace}  
\newcommand{\BlockBCSReadingFull}{Block~BCS started but did not complete (\BlockBCSDisposition). It contributes no inferential result; its rows are listed in Appendix~\ref{app:bc} as a record only.\xspace}  
\newcommand{\BlockBCSReadingIDs}{INCOMPLETE\xspace}  
\newcommand{\BlockBCSRegistered}{ADDENDUM~19\xspace}  
\newcommand{\BlockBCSStateCell}{incomplete at the registered cutoff: \BCSCellsCounted{} of \BCSCellsRegistered{} cells counted; no inference\xspace}  
\newcommand{\BlockBCSVindStatus}{re-derived\xspace}  
\newcommand{\BlockBCSeedList}{42--49, 51--55, 57--59\xspace}  
\newcommand{\BlockBCVerdictFail}{31\xspace}  
\newcommand{\BlockBCVerdictSucc}{20\xspace}  
\newcommand{\BlockBCVindAgree}{53 of 53\xspace}  
\newcommand{\BlockBCVindAgreeSlash}{53/53\xspace}  
\newcommand{\BlockBCVindPath}{\code{audit/\allowbreak{}failure\_\allowbreak{}mining/\allowbreak{}P3v2/\allowbreak{}VALIDITY\_\allowbreak{}INDEPENDENT\_\allowbreak{}blockBC.json}\xspace}  
\newcommand{\BlockBCVindStatus}{re-derived\xspace}  
\newcommand{\BlockBFFRXOffAN}{4\xspace}  
\newcommand{\BlockBFRXOffAPassFrac}{0/10\xspace}  
\newcommand{\BlockBFRXOnAPass}{10\xspace}  
\newcommand{\BlockBFRXOnAPassFrac}{10/10\xspace}  
\newcommand{\BlockBGCConfig}{Codex CLI, frames inline\xspace}  
\newcommand{\BlockBGCCounted}{\textit{not run}\xspace}  
\newcommand{\BlockBGCDisposition}{launch preconditions unmet by the experiment cutoff\xspace}  
\newcommand{\BlockBGCModel}{GPT-6\xspace}  
\newcommand{\BlockBGCReadingFull}{Block~BGC was not run (\BlockBGCDisposition). It contributes no inferential result; its paper-wide slots are non-rejection placeholders.\xspace}  
\newcommand{\BlockBGCReadingIDs}{NOTRUN\xspace}  
\newcommand{\BlockBGCRegistered}{ADDENDUM~19\xspace}  
\newcommand{\BlockBGCVindStatus}{\textit{not re-derived}\xspace}  
\newcommand{\BlockBGtwoConfig}{Codex CLI, frames inline\xspace}  
\newcommand{\BlockBGtwoCounted}{\textit{not run}\xspace}  
\newcommand{\BlockBGtwoDisposition}{fallback not triggered\xspace}  
\newcommand{\BlockBGtwoModel}{GPT-6\xspace}  
\newcommand{\BlockBGtwoReadingFull}{Block~BG2 was not run (\BlockBGtwoDisposition).\xspace}  
\newcommand{\BlockBGtwoReadingIDs}{NOTRUN\xspace}  
\newcommand{\BlockBGtwoRegistered}{ADDENDUM~19\xspace}  
\newcommand{\BlockBP}{0.002\xspace}  
\newcommand{\BlockBPairs}{10\xspace}  
\newcommand{\BlockBSFRXOffAPassFrac}{0/10\xspace}  
\newcommand{\BlockBSFRXOnAPassFrac}{6/10\xspace}  
\newcommand{\BlockBSP}{0.031\xspace}  
\newcommand{\BlockBSVoided}{3\xspace}  
\newcommand{\BlockClmBAttempts}{24\xspace}  
\newcommand{\BlockClmBAuditGate}{passed\xspace}  
\newcommand{\BlockClmBAuditTerms}{4\xspace}  
\newcommand{\BlockClmBClmLzeroPassFrac}{1/1\xspace}  
\newcommand{\BlockClmBClmOffAN}{10\xspace}  
\newcommand{\BlockClmBClmOffAPassFrac}{10/10\xspace}  
\newcommand{\BlockClmBClmOnAPassFrac}{10/10\xspace}  
\newcommand{\BlockClmBClmOnIPassFrac}{3/3\xspace}  
\newcommand{\BlockClmBConfig}{Claude CLI, frames by file read\xspace}  
\newcommand{\BlockClmBContradictions}{0\xspace}  
\newcommand{\BlockClmBControlsCredible}{credible\xspace}  
\newcommand{\BlockClmBCopyLimitControls}{not credible\xspace}  
\newcommand{\BlockClmBCopyLimitHarm}{0\xspace}  
\newcommand{\BlockClmBCopyLimitOffFail}{10\xspace}  
\newcommand{\BlockClmBCopyLimitP}{0.0156\xspace}  
\newcommand{\BlockClmBCopyLimitReading}{R1\xspace}  
\newcommand{\BlockClmBCopyLimitRescue}{7\xspace}  
\newcommand{\BlockClmBCopyLimitTie}{3\xspace}  
\newcommand{\BlockClmBCopyLimitXupN}{3\xspace}  
\newcommand{\BlockClmBCopyLimitXupOn}{0\xspace}  
\newcommand{\BlockClmBCounted}{24\xspace}  
\newcommand{\BlockClmBCountedFail}{0\xspace}  
\newcommand{\BlockClmBCountedSucc}{24\xspace}  
\newcommand{\BlockClmBDecideCheck}{registered decide() = mirror = stored reading\xspace}  
\newcommand{\BlockClmBDecisionTerm}{24\xspace}  
\newcommand{\BlockClmBDisposition}{completed\xspace}  
\newcommand{\BlockClmBEndpointFull}{24\xspace}  
\newcommand{\BlockClmBEndpointUnresolved}{0\xspace}  
\newcommand{\BlockClmBHolmSlot}{6\xspace}  
\newcommand{\BlockClmBInstancePhrase}{\BlockClmBInstances frozen generated instances (seeds \BlockClmBSeedList)\xspace}  
\newcommand{\BlockClmBInstances}{10\xspace}  
\newcommand{\BlockClmBInstancesWords}{ten\xspace}  
\newcommand{\BlockClmBInterrupted}{0\xspace}  
\newcommand{\BlockClmBModel}{Claude Opus~5\xspace}  
\newcommand{\BlockClmBOffBack}{8\xspace}  
\newcommand{\BlockClmBOffPass}{10\xspace}  
\newcommand{\BlockClmBOffReread}{10\xspace}  
\newcommand{\BlockClmBOffStale}{0\xspace}  
\newcommand{\BlockClmBOnBack}{8\xspace}  
\newcommand{\BlockClmBOnOffHarm}{0\xspace}  
\newcommand{\BlockClmBOnOffHolmLocal}{---\xspace}  
\newcommand{\BlockClmBOnOffNetCI}{$0.00$ [$-0.28$, $0.28$]\xspace}  
\newcommand{\BlockClmBOnOffP}{1.00\xspace}  
\newcommand{\BlockClmBOnOffPairs}{10\xspace}  
\newcommand{\BlockClmBOnOffRescue}{0\xspace}  
\newcommand{\BlockClmBOnOffTie}{10\xspace}  
\newcommand{\BlockClmBOnReread}{10\xspace}  
\newcommand{\BlockClmBOrderWord}{frozen\xspace}  
\newcommand{\BlockClmBPaperFamilyP}{1.00\xspace}  
\newcommand{\BlockClmBReadingFull}{Over \BlockClmBOnOffPairs sealed instance pairs (seeds \BlockClmBSeedList), Claude Opus~5 met the full approval endpoint on \BlockClmBClmOffAPassFrac without the panel and \BlockClmBClmOnAPassFrac with it; of the pairs, \BlockClmBOnOffRescue favoured the panel, \BlockClmBOnOffHarm the arm without it, and \BlockClmBOnOffTie neither; net \BlockClmBOnOffNetCI (marginal 95\,\% interval), $p=\BlockClmBOnOffP$. Controls: ON-I \BlockClmBClmOnIPassFrac, L0 \BlockClmBClmLzeroPassFrac. Without the panel, the agent succeeded on \BlockClmBOffPass/\BlockClmBClmOffAN instances under silent supersession; the registered rescue-opportunity threshold was not met. The scripted copy-limit shortcut, run on the same cells before any model, is assigned \BlockClmBCopyLimitReading under the same rules. Every outcome term the independent scorer re-derives is resolved and agreeing with the evaluator on \BlockClmBEndpointFull of \BlockClmBCounted counted rollouts. The standard-library validity re-derivation agrees with the archive on \BlockClmBVindAgree attempts.\xspace}  
\newcommand{\BlockClmBReadingIDs}{R2\xspace}  
\newcommand{\BlockClmBRegistered}{PRECOMMIT\_\allowbreak{}CLAIMS\xspace}  
\newcommand{\BlockClmBReplacementsSpent}{0\xspace}  
\newcommand{\BlockClmBSAttempts}{25\xspace}  
\newcommand{\BlockClmBSAuditGate}{passed\xspace}  
\newcommand{\BlockClmBSAuditTerms}{4\xspace}  
\newcommand{\BlockClmBSClmLzeroPassFrac}{1/1\xspace}  
\newcommand{\BlockClmBSClmOffAN}{10\xspace}  
\newcommand{\BlockClmBSClmOffAPassFrac}{1/10\xspace}  
\newcommand{\BlockClmBSClmOnAN}{10\xspace}  
\newcommand{\BlockClmBSClmOnAPassFrac}{9/10\xspace}  
\newcommand{\BlockClmBSClmOnIPassFrac}{3/3\xspace}  
\newcommand{\BlockClmBSConfig}{Claude CLI, frames by file read\xspace}  
\newcommand{\BlockClmBSContradictions}{0\xspace}  
\newcommand{\BlockClmBSControlsCredible}{credible\xspace}  
\newcommand{\BlockClmBSCopyLimitReading}{R1\xspace}  
\newcommand{\BlockClmBSCounted}{24\xspace}  
\newcommand{\BlockClmBSCountedFail}{10\xspace}  
\newcommand{\BlockClmBSCountedSucc}{14\xspace}  
\newcommand{\BlockClmBSDecideCheck}{registered decide() = mirror = stored reading\xspace}  
\newcommand{\BlockClmBSDecisionTerm}{24\xspace}  
\newcommand{\BlockClmBSDisposition}{completed\xspace}  
\newcommand{\BlockClmBSEndpointFull}{24\xspace}  
\newcommand{\BlockClmBSEndpointUnresolved}{0\xspace}  
\newcommand{\BlockClmBSHolmSlot}{7\xspace}  
\newcommand{\BlockClmBSInterrupted}{1\xspace}  
\newcommand{\BlockClmBSInterruptedWhere}{pair9 CLM-ON-A, attempt 0\xspace}  
\newcommand{\BlockClmBSInterruptedWhy}{operationally interrupted before the archive was written\xspace}  
\newcommand{\BlockClmBSModel}{Claude Sonnet~5\xspace}  
\newcommand{\BlockClmBSOffBack}{0\xspace}  
\newcommand{\BlockClmBSOffReread}{1\xspace}  
\newcommand{\BlockClmBSOffStale}{9\xspace}  
\newcommand{\BlockClmBSOnBack}{10\xspace}  
\newcommand{\BlockClmBSOnFailCorrect}{1\xspace}  
\newcommand{\BlockClmBSOnFailTerms}{final-report\xspace}  
\newcommand{\BlockClmBSOnOffHarm}{0\xspace}  
\newcommand{\BlockClmBSOnOffHolmLocal}{---\xspace}  
\newcommand{\BlockClmBSOnOffHolmPaper}{0.0781\xspace}  
\newcommand{\BlockClmBSOnOffNetCI}{$+0.80$ [$0.37$, $0.92$]\xspace}  
\newcommand{\BlockClmBSOnOffP}{0.00781\xspace}  
\newcommand{\BlockClmBSOnOffPairs}{10\xspace}  
\newcommand{\BlockClmBSOnOffRescue}{8\xspace}  
\newcommand{\BlockClmBSOnOffTie}{2\xspace}  
\newcommand{\BlockClmBSOnReread}{8\xspace}  
\newcommand{\BlockClmBSPaperFamilyP}{0.00781\xspace}  
\newcommand{\BlockClmBSReadingFull}{Over \BlockClmBSOnOffPairs sealed instance pairs (seeds \BlockClmBSSeedList), Claude Sonnet~5 met the full approval endpoint on \BlockClmBSClmOffAPassFrac without the panel and \BlockClmBSClmOnAPassFrac with it; of the pairs, \BlockClmBSOnOffRescue favoured the panel, \BlockClmBSOnOffHarm the arm without it, and \BlockClmBSOnOffTie neither; net \BlockClmBSOnOffNetCI (marginal 95\,\% interval), $p=\BlockClmBSOnOffP$. Controls: ON-I \BlockClmBSClmOnIPassFrac, L0 \BlockClmBSClmLzeroPassFrac. The review-step panel increased full-task success at the registered local level. The OFF-A failure categories and observed pre-commit revisions are reported separately (Appendix~\ref{app:claims}). The OFF-A--ON-A contrast rejects under its registered block-local analysis but not under the paper-wide Holm analysis (adjusted $p=\BlockClmBSOnOffHolmPaper$ over \HolmPaperM tests). The scripted copy-limit shortcut, run on the same cells before any model, is assigned \BlockClmBSCopyLimitReading under the same rules. Every outcome term the independent scorer re-derives is resolved and agreeing with the evaluator on \BlockClmBSEndpointFull of \BlockClmBSCounted counted rollouts. The standard-library validity re-derivation agrees with the archive on \BlockClmBSVindAgree attempts.\xspace}  
\newcommand{\BlockClmBSReadingIDs}{R1\xspace}  
\newcommand{\BlockClmBSReadingMainS}{Over \BlockClmBSOnOffPairs sealed instance pairs (seeds \BlockClmBSSeedList), Claude Sonnet~5 met the full approval endpoint on \BlockClmBSClmOffAPassFrac without the panel and \BlockClmBSClmOnAPassFrac with it; of the pairs, \BlockClmBSOnOffRescue favoured the panel, \BlockClmBSOnOffHarm the arm without it, and \BlockClmBSOnOffTie neither; net \BlockClmBSOnOffNetCI (marginal 95\,\% interval), $p=\BlockClmBSOnOffP$. Controls: ON-I \BlockClmBSClmOnIPassFrac, L0 \BlockClmBSClmLzeroPassFrac. The review-step panel increased full-task success at the registered local level. The OFF-A failure categories and observed pre-commit revisions are reported separately (Appendix~\ref{app:claims}). The OFF-A--ON-A contrast rejects under its registered block-local analysis but not under the paper-wide Holm analysis (adjusted $p=\BlockClmBSOnOffHolmPaper$ over \HolmPaperM tests). That family was finalized after the claims outcomes (Appendix~\ref{app:multiplicity}). The pre-registered derived-decision gate passed on all three threshold-crossing cases. Every outcome term the independent scorer re-derives is resolved and agreeing with the evaluator on \BlockClmBSEndpointFull of \BlockClmBSCounted counted rollouts. The standard-library validity re-derivation agrees with the archive on \BlockClmBSVindAgree attempts.\xspace}  
\newcommand{\BlockClmBSRegistered}{PRECOMMIT\_\allowbreak{}CLAIMS ADDENDUM~1\xspace}  
\newcommand{\BlockClmBSReplacementsSpent}{0\xspace}  
\newcommand{\BlockClmBSSeedList}{0--9\xspace}  
\newcommand{\BlockClmBSStaleRescues}{8\xspace}  
\newcommand{\BlockClmBSStepCap}{0\xspace}  
\newcommand{\BlockClmBSStoredNetCI}{$+0.80$ [$0.39$, $0.91$]\xspace}  
\newcommand{\BlockClmBSUncounted}{1\xspace}  
\newcommand{\BlockClmBSVerdictFail}{10\xspace}  
\newcommand{\BlockClmBSVerdictSucc}{14\xspace}  
\newcommand{\BlockClmBSVindAgree}{25 of 25\xspace}  
\newcommand{\BlockClmBSVindAgreeSlash}{25/25\xspace}  
\newcommand{\BlockClmBSVindCountedValid}{24 of 24\xspace}  
\newcommand{\BlockClmBSVindStatus}{re-derived\xspace}  
\newcommand{\BlockClmBSXupN}{3\xspace}  
\newcommand{\BlockClmBSXupOn}{3\xspace}  
\newcommand{\BlockClmBSeedList}{0--9\xspace}  
\newcommand{\BlockClmBStaleRescues}{0\xspace}  
\newcommand{\BlockClmBStepCap}{0\xspace}  
\newcommand{\BlockClmBStoredNetCI}{$0.00$ [$-0.28$, $0.28$]\xspace}  
\newcommand{\BlockClmBUncounted}{0\xspace}  
\newcommand{\BlockClmBVerdictFail}{0\xspace}  
\newcommand{\BlockClmBVerdictSucc}{24\xspace}  
\newcommand{\BlockClmBVindAgree}{24 of 24\xspace}  
\newcommand{\BlockClmBVindAgreeSlash}{24/24\xspace}  
\newcommand{\BlockClmBVindCountedValid}{24 of 24\xspace}  
\newcommand{\BlockClmBVindStatus}{re-derived\xspace}  
\newcommand{\BlockClmBXupN}{3\xspace}  
\newcommand{\BlockClmBXupOn}{3\xspace}  
\newcommand{\BonfPaperAtwo}{1.00\xspace}  
\newcommand{\BonfPaperB}{0.0234\xspace}  
\newcommand{\BonfPaperBCCueOff}{1.00\xspace}  
\newcommand{\BonfPaperBCLiveCue}{0.00293\xspace}  
\newcommand{\BonfPaperBCSCueOff}{\textit{not admissible}\xspace}  
\newcommand{\BonfPaperBCSLiveCue}{\textit{not admissible}\xspace}  
\newcommand{\BonfPaperBGCCueOff}{\textit{not admissible}\xspace}  
\newcommand{\BonfPaperBGCOffiOffa}{\textit{not admissible}\xspace}  
\newcommand{\BonfPaperBS}{0.375\xspace}  
\newcommand{\BonfPaperClmB}{1.00\xspace}  
\newcommand{\BonfPaperClmBS}{0.0938\xspace}  
\newcommand{\BonfPaperSlotTen}{\textit{not admissible}\xspace}  
\newcommand{\CensusFrameUnread}{80\xspace}  
\newcommand{\CensusReadChars}{300\xspace}  
\newcommand{\CensusSteps}{654\xspace}  
\newcommand{\CertDistinctChecks}{65\xspace}  
\newcommand{\CertFRXOffAChecks}{40\xspace}  
\newcommand{\CertGovAtwoMax}{48\xspace}  
\newcommand{\CertGovAtwoMin}{30\xspace}  
\newcommand{\CertGovChecksMax}{50\xspace}  
\newcommand{\CertGovChecksMin}{30\xspace}  
\newcommand{\CertGovN}{9\xspace}  
\newcommand{\CertGovUnrecorded}{3\xspace}  
\newcommand{\CertN}{7\xspace}  
\newcommand{\ChallengeAppendixText}{Built by a different model (GPT-6, through the Codex CLI) from real Block~A2, B and BS episodes, with its expectations written per tool before its harness: \ChallengeN{} cases, \ChallengeValid{} valid controls and \ChallengeCorrupt{} corruptions, one per defect class. It is unseen only by the Claude agents that wrote and ran the pipeline, not by its builder: our brief named the defect classes, and the builder could read the whole repository (the code under test, the protocol and the paper), so its cases are white-box to it. Its seal record (\ChallengeSealSha) and the \ChallengeSealFiles{} file digests it holds are those ADDENDUM~19 \S7 froze before any Claude agent read the challenge (\S7 discloses that a Claude agent's repository-wide search printed lines of its comparison and expectation-writing scripts, never its expectations); the build re-verifies them, and every one of the \ChallengeFirstRunFiles{} outputs against the hash list written at the run before anyone read them (\ChallengeFirstRunSha). This is its first and only run, and its first output is the result: no case-specific patch was made (the \ChallengeToolSources{} tool source files its run manifest pins are byte-identical now, so no patch followed), no expectation was edited, and the misses are reported, not repaired. Three tools ran on every case, and no invocation ended in a tool error or was skipped. The sealed sentence at the head of this paragraph gives the registered pipeline's numbers, the outcome scorer's and the validity re-derivation's jointly; Section~\ref{sec:caught} reports each layer, and Table~\ref{tab:challenge} gives each tool per case class. The \ChallengeMissed{} misses, as its comparison output words them: (1)~\texttt{C01\_\allowbreak{}success\_\allowbreak{}without\_\allowbreak{}delivery} (success-before-validity: notice), the validity re-derivation: ``validity: valid (expected observation\_\allowbreak{}not\_\allowbreak{}delivered)''; (2)~\texttt{C02\_\allowbreak{}panel\_\allowbreak{}without\_\allowbreak{}delivery} (missing intervention panel delivery), the validity re-derivation: ``validity: valid (expected observation\_\allowbreak{}not\_\allowbreak{}delivered)''; (3)~\texttt{C08\_\allowbreak{}commit\_\allowbreak{}before\_\allowbreak{}change} (commit before change), the validity re-derivation: ``validity: valid (expected invalid\_\allowbreak{}intervention); registered\_\allowbreak{}validity: valid (expected invalid\_\allowbreak{}intervention)''; (4)~\texttt{C15\_\allowbreak{}truncated\_\allowbreak{}trajectory} (truncated trajectory), the validity re-derivation: ``registered\_\allowbreak{}validity: valid (expected delivery\_\allowbreak{}ambiguous/\allowbreak{}verifier\_\allowbreak{}failure/\allowbreak{}infra\_\allowbreak{}failure)''. Under the rules as implemented the validity re-derivation missed three (C01, C02, C08), under the rules as registered two (C08, C15); one (C08) under both. The evaluator's archived verdict path met \ChallengeEvalMet{} of its frozen expectations and missed \ChallengeEvalMissed{} (it met \ChallengeEvalValidMet{} of the \ChallengeValid{} valid controls and \ChallengeEvalCorruptMet{} of the \ChallengeCorrupt{} corruptions); these misses stand as failed expectations. Architecturally, its archive-replay entry point re-derives the delivery checkpoints from the raw logs and re-runs the classifier, but keeps the archived run-validity and invariant claims, which the corruptions leave in place; its verdict equalled the archived one on every case; this is why the outcome scorer and the validity re-derivation are separate layers. The challenge bounds defect detection on one family; it does not establish universal correctness, and it selects no cell and gates no block.\xspace}  
\newcommand{\ChallengeClause}{A second, sealed archive challenge built by a different model met \ChallengeMet of \ChallengeN frozen expectations on its first and only run (\ChallengeCorruptRejected of \ChallengeCorruptInstantiated instantiated corruptions rejected; \ChallengeValidAccepted of \ChallengeValidInstantiated valid cases accepted); its \ChallengeMissed misses are listed in Appendix~\ref{app:certificates} as its comparison output words them.\xspace}  
\newcommand{\ChallengeCorrupt}{18\xspace}  
\newcommand{\ChallengeCorruptInstantiated}{18\xspace}  
\newcommand{\ChallengeCorruptRejected}{14\xspace}  
\newcommand{\ChallengeEvalCorruptMet}{0\xspace}  
\newcommand{\ChallengeEvalMet}{10\xspace}  
\newcommand{\ChallengeEvalMissed}{18\xspace}  
\newcommand{\ChallengeEvalValidMet}{10\xspace}  
\newcommand{\ChallengeFirstRunFiles}{337\xspace}  
\newcommand{\ChallengeFirstRunSha}{\texttt{9d0c898bc8bb}\xspace}  
\newcommand{\ChallengeLedgerClause}{These are the registered pipeline's numbers, the outcome scorer's and the validity re-derivation's jointly (a case is met only if both meet it); by tool, the outcome scorer met \ChallengeScorerMet, the validity re-derivation \ChallengeVindMet{} and the evaluator's archive replay \ChallengeEvalMet{} of the \ChallengeN{} (Table~\ref{tab:challenge}), and the misses are reported, not repaired.\xspace}  
\newcommand{\ChallengeMainThree}{\paragraph{Sealed challenge.} A separate model built a run-once \ChallengeN-case challenge, white-box to its builder but unseen by the Claude agents that built the pipeline; the outcome scorer met \ChallengeScorerMet/\ChallengeN{} sealed expectations, validity re-derivation \ChallengeVindMet/\ChallengeN{} and archived evaluator replay \ChallengeEvalMet/\ChallengeN, unrepaired (Table~\ref{tab:layers}, Appendix~\ref{app:certificates}); not blind validation.\xspace}  
\newcommand{\ChallengeMet}{24\xspace}  
\newcommand{\ChallengeMissed}{4\xspace}  
\newcommand{\ChallengeN}{28\xspace}  
\newcommand{\ChallengeScorerMet}{28\xspace}  
\newcommand{\ChallengeSealFiles}{7\xspace}  
\newcommand{\ChallengeSealSha}{\texttt{b9f9000f0393}\xspace}  
\newcommand{\ChallengeToolSources}{7\xspace}  
\newcommand{\ChallengeValid}{10\xspace}  
\newcommand{\ChallengeValidAccepted}{10\xspace}  
\newcommand{\ChallengeValidInstantiated}{10\xspace}  
\newcommand{\ChallengeVindCorruptMet}{14\xspace}  
\newcommand{\ChallengeVindGaps}{missing observation delivery (2); commit before change (1); truncated trajectory (1)\xspace}  
\newcommand{\ChallengeVindMet}{24\xspace}  
\newcommand{\ChallengeVindValidMet}{10\xspace}  
\newcommand{\ClmActionBudget}{40\xspace}  
\newcommand{\ClmBSPaperHolmR}{\xspace}  
\newcommand{\ClmBundleFiles}{29\xspace}  
\newcommand{\ClmBundleSha}{\texttt{063b7ae44e11}\xspace}  
\newcommand{\ClmCopyLimitTally}{10 OFF-A failures, 7 rescues, 0 harms, 3 ties\xspace}  
\newcommand{\ClmGAddendumSha}{\texttt{bf3999dbd602e432}\xspace}  
\newcommand{\ClmGAttempts}{25\xspace}  
\newcommand{\ClmGBlockedSteps}{0\xspace}  
\newcommand{\ClmGBuildCutoff}{23:00Z\xspace}  
\newcommand{\ClmGCaption}{Gemini 3.1 Pro Preview received each frame inline in a stateless REST request with no tools; the Claude models were instructed to read each frame file with their file-read tool. Provider, model, invoker and image-delivery channel differ together; rows describe model–invoker configurations, not vendor effects.\xspace}  
\newcommand{\ClmGChecks}{42\xspace}  
\newcommand{\ClmGControls}{credible\xspace}  
\newcommand{\ClmGCopyLimit}{R1 (7/0/3, $p=0.0156$; controls not credible)\xspace}  
\newcommand{\ClmGCounted}{24\xspace}  
\newcommand{\ClmGEndpointFull}{24\xspace}  
\newcommand{\ClmGFirstReqUTC}{21:47Z\xspace}  
\newcommand{\ClmGFrameReadB}{357/357\xspace}  
\newcommand{\ClmGFrameReadBS}{217/297\xspace}  
\newcommand{\ClmGFrozenUTC}{2026-09-24 21:36Z\xspace}  
\newcommand{\ClmGGenConfig}{temperature 1.0, top-$p$ 0.95, top-$k$ 64, thinking level ``high'', high media resolution\xspace}  
\newcommand{\ClmGHarm}{0\xspace}  
\newcommand{\ClmGIncidentEpisode}{41 of 48\xspace}  
\newcommand{\ClmGIncidentFrames}{3 of 915\xspace}  
\newcommand{\ClmGInfraStep}{22\xspace}  
\newcommand{\ClmGJournalRows}{626\xspace}  
\newcommand{\ClmGLastReqUTC}{23:34Z\xspace}  
\newcommand{\ClmGLimitR}{All prospective inferential claims of the paper rest on Claude-model blocks; the only non-Anthropic block (CLM-BG) is a separately registered configuration follow-up on reused instances.\xspace}  
\newcommand{\ClmGLzeroBack}{0\xspace}  
\newcommand{\ClmGLzeroLabels}{correct 1\xspace}  
\newcommand{\ClmGLzeroPassFrac}{1/1\xspace}  
\newcommand{\ClmGLzeroReread}{0\xspace}  
\newcommand{\ClmGLzeroStepCap}{0\xspace}  
\newcommand{\ClmGLzeroWilson}{[0.21, 1.00]\xspace}  
\newcommand{\ClmGMaxSteps}{40\xspace}  
\newcommand{\ClmGModel}{Gemini 3.1 Pro Preview (\texttt{gemini-3.1-pro-preview})\xspace}  
\newcommand{\ClmGModelVersion}{\texttt{3.1-pro-preview-01-2026}\xspace}  
\newcommand{\ClmGNetCI}{$+0.50$ [$0.12$, $0.76$]\xspace}  
\newcommand{\ClmGOffABack}{0\xspace}  
\newcommand{\ClmGOffALabels}{no commit 1, stale approval 9\xspace}  
\newcommand{\ClmGOffAPassFrac}{0/10\xspace}  
\newcommand{\ClmGOffAReread}{0\xspace}  
\newcommand{\ClmGOffAStepCap}{1\xspace}  
\newcommand{\ClmGOffAWilson}{[0.00, 0.28]\xspace}  
\newcommand{\ClmGOnABack}{9\xspace}  
\newcommand{\ClmGOnALabels}{correct 5, no commit 5\xspace}  
\newcommand{\ClmGOnAPassFrac}{5/10\xspace}  
\newcommand{\ClmGOnAReread}{6\xspace}  
\newcommand{\ClmGOnAStepCap}{5\xspace}  
\newcommand{\ClmGOnAWilson}{[0.24, 0.76]\xspace}  
\newcommand{\ClmGOnIBack}{0\xspace}  
\newcommand{\ClmGOnILabels}{correct 3\xspace}  
\newcommand{\ClmGOnIPassFrac}{3/3\xspace}  
\newcommand{\ClmGOnIReread}{0\xspace}  
\newcommand{\ClmGOnIStepCap}{0\xspace}  
\newcommand{\ClmGOnIWilson}{[0.44, 1.00]\xspace}  
\newcommand{\ClmGP}{0.0625\xspace}  
\newcommand{\ClmGPairs}{10\xspace}  
\newcommand{\ClmGPolicySteps}{603\xspace}  
\newcommand{\ClmGPriorResultUTC}{2026-09-24 04:06Z\xspace}  
\newcommand{\ClmGReading}{R3\xspace}  
\newcommand{\ClmGReplacements}{1\xspace}  
\newcommand{\ClmGReplacementsMax}{6\xspace}  
\newcommand{\ClmGRescue}{5\xspace}  
\newcommand{\ClmGRescuedOff}{stale approval 5\xspace}  
\newcommand{\ClmGResultsR}{On the ten registered reused claims instances, under the frozen Gemini 3.1 Pro Preview configuration, OFF-A failed on 10/10 instances, but review-step re-exposure did not meet the registered effect criterion (5 rescues, 0 harms, 5 ties; exact paired p = 0.0625); the effect is not shown for these instances under this configuration at this budget, and this is not evidence of no effect.\xspace}  
\newcommand{\ClmGTie}{5\xspace}  
\newcommand{\ClmGTrigFired}{OFF-A 10/10, ON-A 10/10, ON-I 3/3\xspace}  
\newcommand{\ClmGTrigOnly}{10 pairs, 5 rescues, 0 harms\xspace}  
\newcommand{\ClmGTsaCutoff}{23:30Z\xspace}  
\newcommand{\ClmGTsaUTC}{21:36:27Z\xspace}  
\newcommand{\ClmGVindAgree}{25\xspace}  
\newcommand{\ClmGVindValid}{24\xspace}  
\newcommand{\ClmGXupOn}{1/3\xspace}  
\newcommand{\ClmInstanceControls}{ON-I at seeds 0 (1400.00, copy-limit pass), 1 (1550.00, copy-limit pass) and 4 (3856.18, copy-limit fail); L0 at seed 5 (2240.59, copy-limit fail)\xspace}  
\newcommand{\ClmInstrChecked}{48\xspace}  
\newcommand{\ClmInstrExample}{CLM-BS, OFF-A, seed~0\xspace}  
\newcommand{\ClmInstrRequestSha}{18cd71847bed\xspace}  
\newcommand{\ClmInstructionPhrase}{as it stands when you approve\xspace}  
\newcommand{\ClmLooseParserText}{\emph{Report-format sensitivity.} One CLM-BS ON-A episode (seed~9) approved the correct amount and failed only the report term, because its single reply line starts with prose before \code{RESULT:}, which the frozen parser requires at the start of a line. Under a looser reading that accepts \code{RESULT:} anywhere in the line, the pair would be a further rescue (9 discordant pairs, exact $p=0.00391$); the registered, stricter reading is the one reported.\xspace}  
\newcommand{\ClmOpusMain}{Opus succeeded \BlockClmBClmOffAPassFrac{} in both OFF-A and ON-A; because every OFF-A instance already succeeded, the registered rescue-opportunity gate was not met and no panel-effect claim is made for Opus. Opus requested the policy record after the change in all \BlockClmBClmOffAN{} OFF-A runs, versus \BlockClmBSOffReread/\BlockClmBSClmOffAN{} for Sonnet; these process counts are descriptive, and the instruction itself may encourage checking.\xspace}  
\newcommand{\ClmOrderClause}{In CLM-BS the arm run first and the arm run second each succeeded on 5 of 10 pairs (by order: Appendix~\ref{app:claimsspec}); five pairs per order cannot exclude a period effect.\xspace}  
\newcommand{\ClmPageTextCap}{3000\xspace}  
\newcommand{\ClmRegisteredROne}{The certified evaluation protocol transfers to a second domain with a derived decision, and the review-step intervention increases full-task success\xspace}  
\newcommand{\ClmRuleText}{a block with more than 6 replacements or failed readiness gives reading R4 and no result; the rescue question is open only if OFF-A fails on at least 5 pairs (else R2); R1 needs $p<0.05$, more rescues than harms and a minimum number of rescues, which together require at least 6 (else R3); controls are credible only if every L0 control passes and at least 2 ON-I controls pass; the derived-decision gate needs ON-A to pass on at least 2 threshold-crossing-increase pairs\xspace}  
\newcommand{\ClmScorerScope}{The independent scorer's scope covers all \BlockClmBSAuditTerms{} registered outcome terms.\xspace}  
\newcommand{\ConclNullClause}{These contrasts estimate within-configuration effects; unrejected comparisons do not establish absence or equivalence.\xspace}  
\newcommand{\ConclThree}{The prospective claims and clinic experiments, not the historical predecessor blocks, carry the model-effect conclusions (Section~\ref{sec:experiments}). In clinic booking, at the block-local level the value display increased success relative to the cue-and-link panel (LIVE--CUE block-local Holm $p=\BlockBCLiveCueHolmLocal$, paper-wide $p=\BlockBCLiveCueHolmPaper$; CUE--OFF block-local $p=\BlockBCCueOffHolmLocal$). In insurance-claim approval Opus succeeded \BlockClmBClmOffAPassFrac{} without review-step re-exposure, and for Sonnet the panel increased full-task success at the registered local level (block-local $p=\BlockClmBSOnOffP$; retrospective paper-wide Holm $p=\HolmPaperClmBS$). \ConclNullClause{} The auditing challenges and the post-hoc, author-built raw-log comparisons show concretely that intervention, delivery and scoring checks can disagree on the same corrupted archive, so each layer must be tested rather than assumed correct.\xspace}  
\newcommand{\CorrElevenRows}{5\xspace}  
\newcommand{\CorrElevenSig}{4\xspace}  
\newcommand{\CorrSixteenFlips}{7\xspace}  
\newcommand{\CorrSixteenSig}{4\xspace}  
\newcommand{\CorrTenCounterfactual}{25\xspace}  
\newcommand{\CountedSuccessN}{46\xspace}  
\newcommand{\CovAllResolved}{32\xspace}  
\newcommand{\CovAttempts}{82\xspace}  
\newcommand{\CovAttributionGap}{18\xspace}  
\newcommand{\CovCertIOneUnresolved}{41\xspace}  
\newcommand{\CovCertIThreeUnresolved}{15\xspace}  
\newcommand{\CovCertPaperIOneUnresolved}{0\xspace}  
\newcommand{\CovCertRuns}{76\xspace}  
\newcommand{\CovClinicFalse}{27\xspace}  
\newcommand{\CovClinicNoRow}{5\xspace}  
\newcommand{\CovClinicResolved}{73\xspace}  
\newcommand{\CovClinicTrue}{46\xspace}  
\newcommand{\CovEpisodes}{78\xspace}  
\newcommand{\CovIOneUnresolved}{46\xspace}  
\newcommand{\CovInvContradictions}{0\xspace}  
\newcommand{\CovSlotEvidence}{0\xspace}  
\newcommand{\DefectIRows}{3\xspace}  
\newcommand{\DefectKRows}{3\xspace}  
\newcommand{\DiscBCNew}{16\xspace}  
\newcommand{\DiscBCOld}{10\xspace}  
\newcommand{\DiscDeadlineNew}{09:00\xspace}  
\newcommand{\DiscDeadlineOld}{06:00\xspace}  
\newcommand{\DiscInterruptedWord}{One\xspace}  
\newcommand{\DiscTemplatesWord}{Two\xspace}  
\newcommand{\DiscTimeoutNew}{120\xspace}  
\newcommand{\DiscTimeoutOld}{30\xspace}  
\newcommand{\DistractorIsNewClinic}{6\xspace}  
\newcommand{\DistractorNewRows}{23\xspace}  
\newcommand{\ExtAppendixText}{\emph{This challenge differs from the rule-only one (Appendix~\ref{app:blind}) in one respect that matters: the authors neither implemented nor reviewed the cases.} \emph{Constructor and sandbox.} One headless Gemini CLI call (\code{gemini-3.1-pro-preview}, default approval mode; it could neither run a shell nor write files) worked in a sandbox holding only: the four rule and schema files of the rule-only challenge's brief (byte copies, same redactions), \ExtBasesOffered{} byte-verified copies of counted base archives (\ExtClaimsBasesOffered{} claims, three each from CLM-B, CLM-BS and CLM-BG, including failed episodes; \ExtClinicBasesOffered{} Block~BC clinic ones), and an instruction file asking for at least 20 corruptions and 8 unusual-but-valid controls per workflow, a stdlib-only generator that edits every copy of a changed fact and recomputes recorded digests, and an expectation file. The sandbox held no code of ours, no earlier case, class or finding, no paper and no outcome table; a leak check of the instructions, prompt and every base file found no case identifier, tool name or earlier-round label outside verbatim rule text. \emph{Instantiation and seal.} The agent returned the generator and \ExtDesigned{} expectations between markers; both were saved byte for byte and the generator ran once, unedited. It built \ExtRun{} cases; \ExtExcludedIDs{} is excluded because its own generation raised, and no repair round was used. The cases, expectations, transcript, runner and scorer were sealed at \ExtSealUTC, before any tool under test touched any case (the only earlier tool run was a loader smoke test on unmodified base copies). The run began at \ExtRunUTC, used no model and no world; tool and case files did not change during it. The runner is the rule-only challenge's with only its loader adapted, and the scoring functions are that challenge's, copied verbatim. \par\paragraph{Results.} Table~\ref{tab:external-tools} gives each tool. Against the simple raw-log baselines (claims: the r28 baseline; clinic: the r29 baseline) the pipeline caught \ExtClaimsPipeOnly{} claims corruptions the baseline missed and the baseline none the pipeline missed; in the clinic \ExtClinicPipeOnly{} against \ExtClinicBaseOnly{} (EB12, whose edit added a key the schema lacks; the baseline scores it only because it never reports the booked-slot term I1 as passing). Every miss: the pipeline counted EA08, EA09, EA19, EA20 (claims) and EB01, EB03, EB10, EB12, EB14, EB16, EB17, EB18 (clinic) as clean. EB14 and EB18 are undisputed gaps: EB14 changed a collateral appointment's status only in the final state rows, disagreeing with the archived I3 pass, and EB18 set the trajectory's event-log target version equal to its source version while the event log and step copies were unchanged; the pipeline does not cross-check these copies. Three claims misses (EA08, EA19, EA20) edit only an archived derived verdict, which the pipeline silently re-derives instead of flagging the disagreement. It gave no verdict on EA01 and EA18 (a deleted trajectory or claims file: the evaluator raised), and rejected EA11, EB05, EB08, EB09, EB19, EB20 under a class other than the registered one (clinic: the void gate's \code{G0\_void} where the rules name the infrastructure or harness verdict). It did not count EA26, EA27, EB27, controls that change one copy of a fact: the scorer reported the disagreement and the void gate voided the case, whereas the agent expected them counted. \emph{Disputes (scores unchanged).} Written after scoring from a mechanical diff of each case against its base: six edit only a derived field that the refresh re-derives or the rules say is never read (EA08, EA19, EA20, EB03, EB10, EB17); three equal their base (EA09, EA28, EB28); three expect an outcome the unchanged evidence contradicts (EA24, EB24, EB25); two add a key the schema lacks (EB12, EB26); two set a stop reason the rules do not register (EB01, EB16); two are controls that change one copy of a fact, against the instructions (EA26, EA27); one is a control the rules leave open (EB27). On the undisputed cases alone the pipeline caught \ExtClaimsUndPipeCaught/\ExtClaimsUndCor{} claims corruptions with \ExtClaimsUndPipeFA{} counted as clean and \ExtClaimsUndPipeErr{} without verdict (baseline \ExtClaimsUndBaseCaught{} and \ExtClaimsUndBaseFA), and \ExtClinicUndPipeCaught/\ExtClinicUndCor{} clinic ones with \ExtClinicUndPipeFA{} counted as clean (baseline \ExtClinicUndBaseCaught{} and \ExtClinicUndBaseFA), accepting every undisputed control.\par\paragraph{Limits.} The design is shallow: one call, one base per workflow (a CLM-B Opus success and a Block~BC success) although several were requested, one-to-three-word case descriptions, and no case marked ambiguous; \ExtDisputes{} of \ExtDesigned{} expectations are disputable. Most corruptions flip a single field the harness records (stop reason, infrastructure errors, read-target violations, episode marker, a policy-input fault or digest) that the operational validity table reads by design; the raw-log baselines were built to audit intervention validity and the I1/I2 terms and do not read those fields, so part of the gap reflects their narrower scope, and a baseline that reads the validity-table fields was not tested. The constructor is an LLM agent, not an independent research team, and Gemini is also a policy under test elsewhere in this paper (CLM-BG, the same model); the authors chose the rule excerpts, the base archives, the instructions and the tools. The results are per case on this suite, not detection rates, and claims and clinic counts are not pooled.\xspace}  
\newcommand{\ExtBasesOffered}{15\xspace}  
\newcommand{\ExtClaimsBaseCaught}{1\xspace}  
\newcommand{\ExtClaimsBaseFA}{15\xspace}  
\newcommand{\ExtClaimsBasesOffered}{9\xspace}  
\newcommand{\ExtClaimsCor}{20\xspace}  
\newcommand{\ExtClaimsPipeCaught}{14\xspace}  
\newcommand{\ExtClaimsPipeFA}{4\xspace}  
\newcommand{\ExtClaimsPipeOnly}{13\xspace}  
\newcommand{\ExtClaimsUndBaseCaught}{1\xspace}  
\newcommand{\ExtClaimsUndBaseFA}{11\xspace}  
\newcommand{\ExtClaimsUndCor}{16\xspace}  
\newcommand{\ExtClaimsUndPipeCaught}{14\xspace}  
\newcommand{\ExtClaimsUndPipeErr}{2\xspace}  
\newcommand{\ExtClaimsUndPipeFA}{0\xspace}  
\newcommand{\ExtClinicBaseCaught}{3\xspace}  
\newcommand{\ExtClinicBaseFA}{15\xspace}  
\newcommand{\ExtClinicBaseOnly}{1\xspace}  
\newcommand{\ExtClinicBasesOffered}{6\xspace}  
\newcommand{\ExtClinicCor}{19\xspace}  
\newcommand{\ExtClinicPipeCaught}{11\xspace}  
\newcommand{\ExtClinicPipeFA}{8\xspace}  
\newcommand{\ExtClinicPipeOnly}{9\xspace}  
\newcommand{\ExtClinicUndBaseCaught}{2\xspace}  
\newcommand{\ExtClinicUndBaseFA}{10\xspace}  
\newcommand{\ExtClinicUndCor}{13\xspace}  
\newcommand{\ExtClinicUndPipeCaught}{11\xspace}  
\newcommand{\ExtClinicUndPipeFA}{2\xspace}  
\newcommand{\ExtConstructorModel}{gemini-3.1-pro-preview\xspace}  
\newcommand{\ExtControls}{16\xspace}  
\newcommand{\ExtCorTotal}{39\xspace}  
\newcommand{\ExtDesigned}{56\xspace}  
\newcommand{\ExtDisputes}{19\xspace}  
\newcommand{\ExtExcludedIDs}{EB11\xspace}  
\newcommand{\ExtMainR}{\paragraph{External challenge.} A Gemini CLI agent given only the rules, schema and \ExtBasesOffered{} archive copies designed \ExtDesigned{} cases \emph{and} wrote their generator, which we ran unread, sealed and scored once per tool. The pipeline caught \ExtClaimsPipeCaught/\ExtClaimsCor{} claims and \ExtClinicPipeCaught/\ExtClinicCor{} clinic corruptions (raw-log baselines \ExtClaimsBaseCaught, \ExtClinicBaseCaught), counted \ExtClaimsPipeFA{} and \ExtClinicPipeFA{} as clean (baselines \ExtClaimsBaseFA, \ExtClinicBaseFA) and rejected \ExtPipeFR{} of \ExtControls{} controls (baselines $\le$1). We dispute \ExtDisputes{} expectations; it used one base per workflow, and an LLM agent is no independent team (Appendix~\ref{app:external}).\xspace}  
\newcommand{\ExtPipeFATotal}{12\xspace}  
\newcommand{\ExtPipeFR}{3\xspace}  
\newcommand{\ExtRun}{55\xspace}  
\newcommand{\ExtRunUTC}{2026-09-25 09:59:09Z\xspace}  
\newcommand{\ExtSealUTC}{2026-09-25 09:59:06Z\xspace}  
\newcommand{\FeasAtwoN}{34\xspace}  
\newcommand{\FeasAtwoOk}{34\xspace}  
\newcommand{\FeasBN}{44\xspace}  
\newcommand{\FeasBOffAReferenceN}{10\xspace}  
\newcommand{\FeasBOffAReferenceSolved}{10\xspace}  
\newcommand{\FeasBOk}{44\xspace}  
\newcommand{\FreezeStatusText}{ADDENDA~18 and~19 and the claims pre-registration are frozen (each file verifies against its sha256 record) and the interpretation templates are sealed\xspace}  
\newcommand{\FreshBases}{24\xspace}  
\newcommand{\FreshCorr}{15\xspace}  
\newcommand{\FreshCtl}{9\xspace}  
\newcommand{\FreshN}{24\xspace}  
\newcommand{\FreshPipeMiss}{7\xspace}  
\newcommand{\FreshReconFR}{0\xspace}  
\newcommand{\FreshReconMiss}{0\xspace}  
\newcommand{\FreshRepMiss}{6\xspace}  
\newcommand{\FreshRunUTC}{10:03:34Z\xspace}  
\newcommand{\FreshSealUTC}{10:03:30Z\xspace}  
\newcommand{\FreshSummarySentence}{In summary (moved from the main text), on these \FreshN{} post-hoc, author-built fresh cases with expectations sealed before their single run, the reconstruction missed none of \FreshCorr{} corruptions, the frozen pipeline \FreshPipeMiss{} (truncation, stale pages, access-log duplicate commits) and the repaired raw-log checker \FreshRepMiss{} (panel delivery, commit ordering). ``Fresh'' means unused archive bases, not independent or blind evidence: the cases were built by the authors after the fault classes were known.\xspace}  
\newcommand{\HeldoutCorrupt}{5\xspace}  
\newcommand{\HeldoutCorruptInstantiated}{5\xspace}  
\newcommand{\HeldoutCorruptRejected}{5\xspace}  
\newcommand{\HeldoutMet}{8\xspace}  
\newcommand{\HeldoutMissed}{2\xspace}  
\newcommand{\HeldoutN}{10\xspace}  
\newcommand{\HeldoutRunDate}{2026-09-23\xspace}  
\newcommand{\HeldoutValid}{5\xspace}  
\newcommand{\HeldoutValidAccepted}{4\xspace}  
\newcommand{\HeldoutValidInstantiated}{4\xspace}  
\newcommand{\HeldoutValidNotInstantiated}{1\xspace}  
\newcommand{\HistReplacedN}{four\xspace}  
\newcommand{\HistoryWindow}{48\xspace}  
\newcommand{\HolmBCProspectiveSentence}{Block~BC's family was frozen before any BC outcome, so its registered block-local Holm adjustment is retained as prospective multiplicity control.\xspace}  
\newcommand{\HolmClaimsRetroSentence}{The claims contrasts were individually pre-registered before rollout; because the paper-wide family was finalized only after those outcomes existed, its adjustment of the claims tests is retrospective and is reported as a sensitivity analysis, not prospective multiplicity control.\xspace}  
\newcommand{\HolmFamily}{3\xspace}  
\newcommand{\HolmFixedClause}{fixed in the frozen ADDENDUM~19\xspace}  
\newcommand{\HolmMainSentence}{Each prospective block uses its pre-registered local test and multiplicity rule. Appendix~\ref{app:multiplicity} additionally reports a fixed \HolmPaperMWords-slot paper-wide Holm sensitivity family. Because that family was finalized after the claims outcomes existed, its adjustment of the claims and historical tests is retrospective; BC's block-local Holm family was frozen before any BC outcome and is its prospective multiplicity control.\xspace}  
\newcommand{\HolmPaperAtwo}{1.00\xspace}  
\newcommand{\HolmPaperB}{0.0215\xspace}  
\newcommand{\HolmPaperBBound}{0.0234\xspace}  
\newcommand{\HolmPaperBCCueOff}{1.00\xspace}  
\newcommand{\HolmPaperBCLiveCue}{0.00293\xspace}  
\newcommand{\HolmPaperBCPairs}{16\xspace}  
\newcommand{\HolmPaperBCSCueOff}{\textit{not admissible}\xspace}  
\newcommand{\HolmPaperBCSLiveCue}{\textit{not admissible}\xspace}  
\newcommand{\HolmPaperBGCCueOff}{\textit{not admissible}\xspace}  
\newcommand{\HolmPaperBGCOffiOffa}{\textit{not admissible}\xspace}  
\newcommand{\HolmPaperBS}{0.281\xspace}  
\newcommand{\HolmPaperBSMin}{0.0625\xspace}  
\newcommand{\HolmPaperClmB}{1.00\xspace}  
\newcommand{\HolmPaperClmBS}{0.0781\xspace}  
\newcommand{\HolmPaperFirst}{0.00417\xspace}  
\newcommand{\HolmPaperFirstSplits}{$k{:}0$ for $k=9,\ldots,16$ and $k{:}1$ for $k=12,\ldots,15$\xspace}  
\newcommand{\HolmPaperM}{12\xspace}  
\newcommand{\HolmPaperMWords}{twelve\xspace}  
\newcommand{\HolmPaperRCText}{Under ordered outcomes, the adjacent gains divide at most 16 favorable discordances. Neither 8--0 contrast passes the first Holm threshold, but both can survive later steps if sufficiently many other family tests reject: with six smaller family $p$-values rejecting first, two 8--0 contrasts ($p=0.00781$ each) have Holm-adjusted $p=0.0469$. Paper-wide support therefore depends on the complete family.\xspace}  
\newcommand{\HolmPaperSlotTen}{\textit{not admissible}\xspace}  
\newcommand{\HolmPaperStatus}{12 of 12 resolved\xspace}  
\newcommand{\HolmPaperSurvivalFacts}{(7) Every slot of the family is resolved; the adjusted values are in Table~\ref{tab:tests}.\xspace}  
\newcommand{\IndepFullBFRXOnA}{0\xspace}  
\newcommand{\InjAckStep}{16\xspace}  
\newcommand{\InjCommitStep}{41\xspace}  
\newcommand{\InjRepeatStep}{20\xspace}  
\newcommand{\InjSeed}{10\xspace}  
\newcommand{\InstructionSha}{\texttt{66260588}\xspace}  
\newcommand{\LambdaShape}{0.5\xspace}  
\newcommand{\LeakAuditWhere}{it is in the supplement\xspace}  
\newcommand{\LeakBrowserAttempts}{8\xspace}  
\newcommand{\LeakCanaries}{5\xspace}  
\newcommand{\LeakDate}{2026-09-21\xspace}  
\newcommand{\LeakLeakedFrom}{0\xspace}  
\newcommand{\LeakModelReads}{5\xspace}  
\newcommand{\LeakOsAttempts}{15\xspace}  
\newcommand{\LeakPolicyUser}{\texttt{envos-policy}\xspace}  
\newcommand{\LeakVerdict}{\textsc{pass}\xspace}  
\newcommand{\LiveUsedText}{no macro the manuscript uses is read from a live entry\xspace}  
\newcommand{\MBClaimsScopeBase}{6\xspace}  
\newcommand{\MBClaimsScopeN}{12\xspace}  
\newcommand{\MBClaimsScopeProt}{10\xspace}  
\newcommand{\MBClinicBaseLines}{143\xspace}  
\newcommand{\MBClinicCorr}{18\xspace}  
\newcommand{\MBClinicCtl}{10\xspace}  
\newcommand{\MBClinicJoint}{14\xspace}  
\newcommand{\MBClinicScopeN}{12\xspace}  
\newcommand{\MBClinicScopeProt}{11\xspace}  
\newcommand{\MBClinicTopSourceN}{14\xspace}  
\newcommand{\MBDevBC}{16\xspace}  
\newcommand{\MBDevN}{92\xspace}  
\newcommand{\MBRunUTC}{10:01Z\xspace}  
\newcommand{\MBSealUTC}{2026-09-24 10:00Z\xspace}  
\newcommand{\ManifestBPaths}{29\xspace}  
\newcommand{\ManifestImages}{2\xspace}  
\newcommand{\ManifestPaths}{29\xspace}  
\newcommand{\MatchedAppendixText}{\paragraph{The same comparison on both workflows.} After the sealed challenge and the check above, we asked what the protocol adds over a fair simple checker on \emph{identical} evidence and scored properties. For clinic we wrote \code{clinic\_baseline.py} (\MBClinicBaseLines{} lines, standard library), the counterpart of the claims baseline: it asks whether the session row changed exactly once, after its old location was served in the same login session and inside the agent's own recorded request, and whether any served page contradicts the change (on L0, whether nothing changed); whether exactly one commit and one patient row followed the change; whether the booked clinic equals the session's location at the commit (the nearest-slot term is not archived, so it reports fail or unresolved, never pass); and whether the report matches the committed record. It does not check delivery, the arm served, collateral rows, the confirm state or archive completeness. It was developed only on \MBDevN{} clinic-family run archives that are not sources of any challenge case, sealed (with the table's decision definitions) at \MBSealUTC, and run once on the challenge's cases at \MBRunUTC; nothing was tuned afterwards. Before writing it, its author had read the challenge's expected verdicts, the sealed tools' recorded outputs and the unmodified control archives, but not the mutation code. \emph{Disclosure:} \MBDevBC{} of the \MBDevN{} development archives were Block~BC archives written while that block was running; the development check compared the baseline only with their archived evaluator scores, no number from them appears in this paper, and BC's reading is selected mechanically by its frozen rule. The sealed challenge's joint figure, \MBClinicJoint{} of \MBClinicCorr{} corruptions, is the set both validity readings reject; the protocol rows of Table~\ref{tab:matched} take one reading at a time. The claims baseline's in-scope misses reflect implementation choices a careful checker need not make (nearest-request attribution, timestamp ordering, no event log, counting history rows rather than change requests); the clinic baseline, which reads more raw sources, matched the protocol within its scope. The added value this shows is breadth of checked properties and, on claims, a more robust causal chain than one simple checker; it does not show that a simple checker cannot match the causal chain.\xspace}  
\newcommand{\MatchedCaption}{\textbf{The protocol against a simple raw-log baseline} (post hoc, author-built, not blind; separate from Table~\ref{tab:layers}). Same raw evidence and scored properties. Controls accepted; corruptions rejected (not counted, or counted with the corrupted invariant flagged), in brackets those in the baseline's declared scope. Clinic protocol rows are derived from the sealed tools' recorded outputs; the claims pipeline voids \code{C09b} as invalid rather than failing its single-commit term. Every case: Appendix~\ref{app:posthoc}.\xspace}  
\newcommand{\MatchedMainFour}{\paragraph{Against a simple raw-log baseline.} This comparison is post hoc, author-built and not blind: after the sealed result, which stands unchanged, we wrote a simple checker rebuilding each verdict from the same raw state, access log and served pages, with the pipeline in view, and sealed each before its one run (Table~\ref{tab:matched}). No protocol path rejected a control; the clinic baseline rejected one. Within the baseline's declared scope both rejected \MBClinicScopeProt{} of \MBClinicScopeN{} clinic corruptions; on claims the pipeline rejected \MBClaimsScopeProt{} and the baseline \MBClaimsScopeBase{} of \MBClaimsScopeN. A careful simple checker thus reproduced most narrow clinic checks; \envos adds breadth (delivery, served arm, collateral state) that such a checker could be extended to cover, and the baseline alone caught one case (Appendix~\ref{app:posthoc}). As we built both cases and checkers, only the external challenge above tests independent construction, and there the pipeline counted \ExtClaimsPipeFA{} of \ExtClaimsCor{} claims and \ExtClinicPipeFA{} of \ExtClinicCor{} clinic corruptions as clean.\xspace}  
\newcommand{\MemoMaxChars}{300\xspace}  
\newcommand{\ModelAttempts}{82\xspace}  
\newcommand{\ModelCounted}{78\xspace}  
\newcommand{\ModelRunDirs}{82\xspace}  
\newcommand{\MutationDevChecks}{11\xspace}  
\newcommand{\MutationsDev}{11\xspace}  
\newcommand{\MutationsDevControls}{2\xspace}  
\newcommand{\MutationsDevCorruptions}{9\xspace}  
\newcommand{\MutationsHeldout}{10\xspace}  
\newcommand{\MutationsHeldoutSha}{\texttt{d1ac0f83}\xspace}  
\newcommand{\NDefectsBlockB}{4\xspace}  
\newcommand{\NDefectsModelAttributable}{7\xspace}  
\newcommand{\NDefectsTotal}{11\xspace}  
\newcommand{\NeutralCertFiles}{10\xspace}  
\newcommand{\NeutralInstances}{18\xspace}  
\newcommand{\NeutralMailCalendarHits}{0\xspace}  
\newcommand{\NewcombeCorrectedSha}{\texttt{54f6d791e7c4}\xspace}  
\newcommand{\NewcombeErratumApplies}{0\xspace}  
\newcommand{\NewcombeErratumChanged}{0\xspace}  
\newcommand{\NewcombeErratumMaxDiff}{below $10^{-9}$\xspace}  
\newcommand{\NewcombeErratumNWords}{five\xspace}  
\newcommand{\NonReadCallsSonnet}{1591\xspace}  
\newcommand{\OrigBSFRXOnAFrac}{6/7\xspace}  
\newcommand{\OrigBSP}{0.0312\xspace}  
\newcommand{\OswClmBLzeroFailed}{0\xspace}  
\newcommand{\OswClmBLzeroHandled}{0\xspace}  
\newcommand{\OswClmBLzeroN}{1\xspace}  
\newcommand{\OswClmBLzeroPartial}{4/4 (1.00)\xspace}  
\newcommand{\OswClmBLzeroSucc}{1\xspace}  
\newcommand{\OswClmBLzeroUntested}{1\xspace}  
\newcommand{\OswClmBOffAFailed}{0\xspace}  
\newcommand{\OswClmBOffAHandled}{10\xspace}  
\newcommand{\OswClmBOffAN}{10\xspace}  
\newcommand{\OswClmBOffAPartial}{40/40 (1.00)\xspace}  
\newcommand{\OswClmBOffASucc}{10\xspace}  
\newcommand{\OswClmBOffAUntested}{0\xspace}  
\newcommand{\OswClmBOnAFailed}{0\xspace}  
\newcommand{\OswClmBOnAHandled}{10\xspace}  
\newcommand{\OswClmBOnAN}{10\xspace}  
\newcommand{\OswClmBOnAPartial}{40/40 (1.00)\xspace}  
\newcommand{\OswClmBOnASucc}{10\xspace}  
\newcommand{\OswClmBOnAUntested}{0\xspace}  
\newcommand{\OswClmBOnIFailed}{0\xspace}  
\newcommand{\OswClmBOnIHandled}{0\xspace}  
\newcommand{\OswClmBOnIN}{3\xspace}  
\newcommand{\OswClmBOnIPartial}{12/12 (1.00)\xspace}  
\newcommand{\OswClmBOnISucc}{3\xspace}  
\newcommand{\OswClmBOnIUntested}{3\xspace}  
\newcommand{\OswClmBSLzeroFailed}{0\xspace}  
\newcommand{\OswClmBSLzeroHandled}{0\xspace}  
\newcommand{\OswClmBSLzeroN}{1\xspace}  
\newcommand{\OswClmBSLzeroPartial}{4/4 (1.00)\xspace}  
\newcommand{\OswClmBSLzeroSucc}{1\xspace}  
\newcommand{\OswClmBSLzeroUntested}{1\xspace}  
\newcommand{\OswClmBSOffAFailed}{9\xspace}  
\newcommand{\OswClmBSOffAHandled}{1\xspace}  
\newcommand{\OswClmBSOffAN}{10\xspace}  
\newcommand{\OswClmBSOffAPartial}{30/40 (0.75)\xspace}  
\newcommand{\OswClmBSOffASucc}{1\xspace}  
\newcommand{\OswClmBSOffAUntested}{0\xspace}  
\newcommand{\OswClmBSOnAFailed}{0\xspace}  
\newcommand{\OswClmBSOnAHandled}{10\xspace}  
\newcommand{\OswClmBSOnAN}{10\xspace}  
\newcommand{\OswClmBSOnAPartial}{39/40 (0.98)\xspace}  
\newcommand{\OswClmBSOnASucc}{9\xspace}  
\newcommand{\OswClmBSOnAUntested}{0\xspace}  
\newcommand{\OswClmBSOnIFailed}{0\xspace}  
\newcommand{\OswClmBSOnIHandled}{0\xspace}  
\newcommand{\OswClmBSOnIN}{3\xspace}  
\newcommand{\OswClmBSOnIPartial}{12/12 (1.00)\xspace}  
\newcommand{\OswClmBSOnISucc}{3\xspace}  
\newcommand{\OswClmBSOnIUntested}{3\xspace}  
\newcommand{\OswNoGoR}{A pre-registered screenshot-only coordinate-action crossover of the claims workflow stopped at its engineering admission gate before any model rollout: viewport fit failed (18 of 144 page checks did not fit) and legibility failed (4 of 404 values unrecovered at $1536\times864$). It therefore contributes no behavioral evidence, and interface robustness remains untested.\xspace}  
\newcommand{\PSignBOneAgainst}{0.021\xspace}  
\newcommand{\PSignSixZero}{0.0312\xspace}  
\newcommand{\PairedAtwoNet}{$-0.13$\xspace}  
\newcommand{\PairedAtwoNetCI}{$-0.13$ [$-0.47$, $0.22$]\xspace}  
\newcommand{\PairedAtwoP}{1\xspace}  
\newcommand{\PairedBNetCI}{$+1.00$ [$0.61$, $1.00$]\xspace}  
\newcommand{\PairedBP}{0.00195\xspace}  
\newcommand{\PairedBSNetCI}{$+0.60$ [$0.20$, $0.83$]\xspace}  
\newcommand{\PairedBSP}{0.0312\xspace}  
\newcommand{\PairedBTreatOnly}{10\xspace}  
\newcommand{\PreAddenda}{19\xspace}  
\newcommand{\PreControlFloor}{6/8\xspace}  
\newcommand{\PrecommitSha}{\texttt{c280ce46}\xspace}  
\newcommand{\PrecommitTaskTextSha}{\texttt{cdce6a1d}\xspace}  
\newcommand{\ProtocolVersion}{2\xspace}  
\newcommand{\PtwoRecertFailed}{P2-C 8; P2-L0 9; P2-L1 18\xspace}  
\newcommand{\QconfigNote}{\xspace}  
\newcommand{\ReadingRuleSha}{\texttt{b23e52dbffcf}\xspace}  
\newcommand{\ReadingSealStatus}{sealed\xspace}  
\newcommand{\ReadingTemplatesSha}{\texttt{af62f98dba4e}\xspace}  
\newcommand{\RelatedClaimsAnchor}{The prospective claims blocks do not show this for Opus~5, which succeeded on \BlockClmBClmOffAPassFrac{} silent-change instances without the panel; there the panel contrast is Sonnet~5's (\BlockClmBSClmOffAPassFrac{} without, \BlockClmBSClmOnAPassFrac{} with), for this task and interface.\xspace}  
\newcommand{\ReproBCSentence}{ The anonymous supplement includes the raw text archives, search logs, results, independent scorer outputs and validity re-derivations for every prospective block reported in the main text, including BC; a no-model reproduction target recomputes the cells of Table~\ref{tab:results}, the per-instance grids of Figure~\ref{fig:results}, all registered paired tests and the associated audit counts from those packaged artifacts.\xspace}  
\newcommand{\RequeryOffAFrac}{0/24\xspace}  
\newcommand{\ReserveBlock}{6\xspace}  
\newcommand{\ReserveCell}{2\xspace}  
\newcommand{\RewardVersion}{2\xspace}  
\newcommand{\ScorerAuditDate}{2026-09-21\xspace}  
\newcommand{\ScorerBehaviour}{2\xspace}  
\newcommand{\ScorerContradictions}{0\xspace}  
\newcommand{\ScorerEvaluatorImports}{0\xspace}  
\newcommand{\ScorerRunsAudited}{76\xspace}  
\newcommand{\ScorerRunsWithUnresolved}{15\xspace}  
\newcommand{\ScorerSourceSha}{\texttt{a113a2a8}\xspace}  
\newcommand{\ScorerUnresolved}{46\xspace}  
\newcommand{\SensAtwoMatchesResultFile}{yes\xspace}  
\newcommand{\SensBMatchesResultFile}{yes\xspace}  
\newcommand{\SensBSMatchesResultFile}{yes\xspace}  
\newcommand{\SlotAllRecords}{1\xspace}  
\newcommand{\SlotBRescues}{10\xspace}  
\newcommand{\SlotCommitted}{73\xspace}  
\newcommand{\SlotComplete}{0\xspace}  
\newcommand{\SlotFPCommitStep}{36\xspace}  
\newcommand{\SlotFPLastRead}{26\xspace}  
\newcommand{\SlotFPRunLabel}{A2, S-L2, seed 15, attempt 1\xspace}  
\newcommand{\SlotFPSupported}{1\xspace}  
\newcommand{\SlotFPSupportedWords}{one\xspace}  
\newcommand{\SlotFPWrites}{2\xspace}  
\newcommand{\SlotOneRecord}{78\xspace}  
\newcommand{\SlotSeedRecords}{4\xspace}  
\newcommand{\SlotSuccessComplete}{0\xspace}  
\newcommand{\SlotSuccessCompleteFP}{1\xspace}  
\newcommand{\SonnetPanelShotSuccess}{6\xspace}  
\newcommand{\SonnetPanelTextFail}{3\xspace}  
\newcommand{\StackBArms}{32\xspace}  
\newcommand{\StackBControlArms}{5\xspace}  
\newcommand{\StackBEnvsWords}{three\xspace}  
\newcommand{\StackBKMax}{7\xspace}  
\newcommand{\StackBKMin}{1\xspace}  
\newcommand{\StackBKThreeArms}{24\xspace}  
\newcommand{\StackBOpusRollouts}{93\xspace}  
\newcommand{\StackBPhenArms}{27\xspace}  
\newcommand{\StackBRescoreField}{93\xspace}  
\newcommand{\StackBRescoreFlips}{2\xspace}  
\newcommand{\StackBSharedArms}{22\xspace}  
\newcommand{\StackBSonnetArms}{23\xspace}  
\newcommand{\StackBUnionArms}{33\xspace}  
\newcommand{\SupersededBlockEpisodes}{12\xspace}  
\newcommand{\SupersededBlockNotTriggered}{3\xspace}  
\newcommand{\SuppBCRuns}{53\xspace}  
\newcommand{\SuppBCSRuns}{24\xspace}  
\newcommand{\SuppClaimsAfter}{\xspace}  
\newcommand{\SuppClaimsAttemptDirs}{49\xspace}  
\newcommand{\SuppCodeObjects}{123\xspace}  
\newcommand{\SuppFiles}{24399\xspace}  
\newcommand{\SuppIrrecoverableText}{two code versions pinned by the archived pre-image Block~A manifest (the harness \texttt{e4e215d1} and the task module \texttt{1eeaab16})\xspace}  
\newcommand{\SuppRecordsShip}{the frozen pre-registrations, the release manifests and the ledger (\code{claims.json}) of the paper build the package was made from ship as files in the supplement\xspace}  
\newcommand{\SuppReproChalChecks}{31\xspace}  
\newcommand{\SuppReproClaimsChecks}{85\xspace}  
\newcommand{\SuppReproLedgerN}{2319\xspace}  
\newcommand{\SuppReproProsChecks}{174\xspace}  
\newcommand{\SuppReproResult}{In one offline run of this package (no network, world or model), the claims target passed all \SuppReproClaimsChecks{} of its checks, every recomputed count, test, interval, reading and audit count equal to this ledger's entry, and the challenge target all \SuppReproChalChecks{}, the re-run comparison's output identical to the shipped result; the packaged ledger's verification re-derived all \SuppReproLedgerN{} of its entries. Its prospective target passed all \SuppReproProsChecks{} of its checks: Table~\ref{tab:results}'s cells and Figure~\ref{fig:results}'s per-instance grids for the claims blocks and Block~BC, Block~BC's paired tests, Holm adjustments, Newcombe intervals and audit counts, recomputed from the raw run directories, equal this ledger's entries; for Block~BCS, which stopped unfinished at its registered cutoff and reports no test, its independent scorer and validity re-derivation were re-run and equal their records, and its attempts, counted cells and per-instance table were recomputed. All \SuppReproTablesOk{} historical archive re-analysis re-derivations reproduced; every generated table (61 files) re-renders byte for byte from the shipped records.\xspace}  
\newcommand{\SuppReproTablesOk}{9\xspace}  
\newcommand{\SuppRuns}{82\xspace}  
\newcommand{\SuppStackBFiles}{149\xspace}  
\newcommand{\SuppSummary}{the environment, task-module, certifier, scorer, harness, invoker and test sources; the pre-registrations, certificates, block specifications, release manifests and search logs; the raw text archives of the historical, claims (CLM-BG included) and Blocks~BC and~BCS model episodes (no screenshot frame ships, for size); the held-out battery and the sealed archive challenge after its one run; and a reproduction script that needs no network, world or model\xspace}  
\newcommand{\SuppZipMB}{94.6\xspace}  
\newcommand{\SweepArmed}{46\xspace}  
\newcommand{\SweepCseventeenChecked}{31\xspace}  
\newcommand{\SweepCseventeenVacuous}{15\xspace}  
\newcommand{\SweepGapCommit}{22.4 to 199.7 s (median 75.6)\xspace}  
\newcommand{\SweepGapFire}{11.6 to 54.5 s (median 19.9)\xspace}  
\newcommand{\SweepN}{48\xspace}  
\newcommand{\SweepOldFrameUnread}{1\xspace}  
\newcommand{\SweepVacuousBreakdown}{9 OFF-A, 4 ON-I and 2 ON-A\xspace}  
\newcommand{\TestsHistClinic}{31\xspace}  
\newcommand{\TestsHistFull}{0\xspace}  
\newcommand{\TestsHistSucc}{31\xspace}  
\newcommand{\VFirstCurrent}{46\xspace}  
\newcommand{\VFirstSuperseded}{7\xspace}  
\newcommand{\VFirstVoidCurrent}{8\xspace}  
\newcommand{\VFirstVoidList}{A2 S-L2 s9; A2 S-L2 s11; A2 S-C s12; A2 S-C s13; A2 S-C s14; A2 S-C s15; BS FRX-ON-I s21; BS FRX-ON-I s22\xspace}  
\newcommand{\VFirstVoidSuperseded}{1\xspace}  
\newcommand{\ValBaseCtl}{10\xspace}  
\newcommand{\ValBaseLines}{108\xspace}  
\newcommand{\ValBaseMissed}{11\xspace}  
\newcommand{\ValBaseOnly}{1\xspace}  
\newcommand{\ValBaseOnlyIds}{\code{C09a}\xspace}  
\newcommand{\ValBaseScope}{2\xspace}  
\newcommand{\ValBaseStop}{6\xspace}  
\newcommand{\ValBoth}{5\xspace}  
\newcommand{\ValBothIds}{\code{C03}, \code{C09b}, \code{C12}, \code{C17}, \code{C18}\xspace}  
\newcommand{\ValCEightCtl}{10\xspace}  
\newcommand{\ValCachedCounted}{16\xspace}  
\newcommand{\ValCases}{29\xspace}  
\newcommand{\ValControls}{10\xspace}  
\newcommand{\ValCorrupt}{19\xspace}  
\newcommand{\ValEvalCtl}{10\xspace}  
\newcommand{\ValExpectedSha}{1450457f8225\xspace}  
\newcommand{\ValMissesText}{\emph{Every miss, with its mechanism.} \code{C15} (truncated trajectory): all five counted it as a correct success. The step list was cut after the change and before any approval step, leaving the report, the logs and the harness's state archive intact; every completeness check (the manifest's artifact check, the scorer's evidence check, C8's loader) looks only at the steps that are listed, and none checks that every archived state snapshot and the final step appear in the list; in the sealed challenge the historical validity re-derivation also missed its truncated-trajectory case. \code{C17p} (stale pages): all five counted it. Every policy-record page served after the change still showed the starting limit although the database row changed; the validity term that checks served pages looks only at steps up to the change, and the certifier predicate that would catch it runs only during certification, not in the classifier (the clinic blocks' re-derivation has the equivalent check). \code{C09a} (two committing requests against one approval event): the scorer, the evaluator/classifier and the pipeline scored I2 as passed, because the claims I2 term requires at least one committing request, not exactly one; the baseline, which requires exactly one, caught it. That I2 must fail here is our normative reading, since the pre-registration phrases I2 in terms of approval events. \code{C09b} (two approvals in the world): C8, the scorer, the evaluator/classifier and the pipeline rejected it as an invalid intervention, which would replace the episode instead of counting it as an I2 failure: the before-commit term binds a single approval and is undefined for two in all three implementations, so they agree and the block's void gate would not flag it; only the baseline gave the sealed class. Approval is terminal in the task, so a real double approval would need a portal fault. C8 also counted the three panel and delivery corruptions (\code{C02}, \code{C02b}, \code{C07}), which its declared scope, intervention validity only, does not cover; the classifier and the scorer reject all three.\xspace}  
\newcommand{\ValModelsWords}{two\xspace}  
\newcommand{\ValNeither}{2\xspace}  
\newcommand{\ValNeitherIds}{\code{C15}, \code{C17p}\xspace}  
\newcommand{\ValNormativeList}{\code{V09}, \code{C09a}, \code{C17p}\xspace}  
\newcommand{\ValNormativeWords}{three\xspace}  
\newcommand{\ValPipeCtl}{10\xspace}  
\newcommand{\ValPipeMissed}{3\xspace}  
\newcommand{\ValPipeOnly}{11\xspace}  
\newcommand{\ValPipeOnlyIds}{\code{C02}, \code{C02b}, \code{C04}, \code{C05}, \code{C06}, \code{C07}, \code{C08a}, \code{C08b}, \code{C10}, \code{C11}, \code{C16}\xspace}  
\newcommand{\ValPipeStop}{16\xspace}  
\newcommand{\ValPipeWrong}{1\xspace}  
\newcommand{\ValResultsText}{No tool rejected a control (C8, scorer, evaluator/classifier, baseline, pipeline: \ValCEightCtl, \ValScorerCtl, \ValEvalCtl, \ValBaseCtl, \ValPipeCtl{} of \ValControls). Of the \ValCorrupt{} corruptions, the pipeline kept \ValPipeStop{} from a clean score (\ValPipeWrong{} with the wrong class) and missed \ValPipeMissed; the baseline kept \ValBaseStop, missed \ValBaseMissed{} and left \ValBaseScope{} outside its scope. \emph{Incremental coverage.} Stopped by the pipeline only (\ValPipeOnly): \ValPipeOnlyIds; by the baseline only (\ValBaseOnly): \ValBaseOnlyIds; by both (\ValBoth): \ValBothIds; by neither (\ValNeither): \ValNeitherIds. The baseline therefore does not catch everything the pipeline catches: it never reads prompts or review panels (the delivery cases), orders events by timestamp rather than by the world's own history, attributes the change to the nearest agent request, does not read the event log, and does not assess I3 or I4. The one case only it caught is a duplicate commit that the claims I2 term, which requires at least one committing request, accepts.\xspace}  
\newcommand{\ValRunUTC}{2026-09-24 09:14Z\xspace}  
\newcommand{\ValScorerCtl}{10\xspace}  
\newcommand{\ValSealUTC}{2026-09-24 09:13Z\xspace}  
\newcommand{\ValSourcesWords}{six\xspace}  
\newcommand{\ValTopSourceN}{16\xspace}  
\newcommand{\ValTrustText}{\emph{Trust boundary.} On the archive as stored, with the archived evaluator verdict not recomputed, the classifier alone counts \ValCachedCounted{} of the \ValCorrupt{} corruptions: it rejects only the three its own archive checks see (\code{C02}, \code{C02b}, \code{C07}), because its validity row reads the archived verdict. For such an archive the backstop is the scorer's and C8's recomputation through the block's void gate, which voids the whole block rather than replacing the attempt; this is inferred from their recomputed verdicts in this run, not run separately on the stored copies.\xspace}  
\newcommand{\ValXColsWords}{five\xspace}  
\newcommand{\ValXRows}{26\xspace}  
\newcommand{\ValidityCounted}{4\xspace}  
\newcommand{\ValidityFirstOf}{53\xspace}  
\newcommand{\ValidityFirstVoid}{9\xspace}  
\newcommand{\ValidityNotCounted}{7\xspace}  
\newcommand{\VindAgree}{82\xspace}  
\newcommand{\VindAttempts}{82\xspace}  
\newcommand{\VindCountedImpl}{78\xspace}  
\newcommand{\VindCountedReg}{71\xspace}  
\newcommand{\VindDisagree}{0\xspace}  
\newcommand{\VindNoticeFirstList}{S-C s12 and S-C s15\xspace}  
\newcommand{\VindPostHoc}{2\xspace}  
\newcommand{\VindPriorFrames}{14\xspace}  
\newcommand{\VindPriorFramesAtwoS}{3\xspace}  
\newcommand{\VindPriorFramesBS}{11\xspace}  
\newcommand{\VindProspPathClaims}{\code{audit/\allowbreak{}failure\_\allowbreak{}mining/\allowbreak{}CLAIMS/\allowbreak{}VALIDITY\_\allowbreak{}INDEPENDENT\_\allowbreak{}block\textit{ID}.json}\xspace}  
\newcommand{\VindProspPathClinic}{\code{audit/\allowbreak{}failure\_\allowbreak{}mining/\allowbreak{}P3v2/\allowbreak{}VALIDITY\_\allowbreak{}INDEPENDENT\_\allowbreak{}block\textit{ID}.json}\xspace}  
\newcommand{\VindProspStatusR}{complete for BC, CLM-B and CLM-BS\xspace}  
\newcommand{\VindProxyAgree}{78\xspace}  
\newcommand{\VindReadings}{14\xspace}  
\newcommand{\VindRegDiffers}{7\xspace}  
\newcommand{\VindRegList}{FRX-ON-A s22 (model failure); FRX-ON-A s23 (budget limited); FRX-ON-A s27 (model failure); FRX-ON-A s28 (model failure); FRX-ON-I s20 (budget limited); FRX-ON-I s21 (success); FRX-ON-I s22 (success)\xspace}  
\newcommand{\VindRegNever}{2\xspace}  
\newcommand{\VindRegTextOnly}{5\xspace}  
\newcommand{\VindRegTextOnlyFail}{3\xspace}  
\newcommand{\VindRegTextOnlySucc}{2\xspace}  
\newcommand{\VindUncounted}{4\xspace}  
\newcommand{\WorldServices}{14\xspace}  

\newcommand{\envos}{EnvOS\xspace}

\newcommand{\code}[1]{\texttt{#1}}

\title{\textbf{EnvOS: Auditing Computer-Use Agents When the World Moves On}\\\large Phenomenon-Based Task Construction, Certification, and Evidence-Based Evaluation}
\author{%
\begin{tabular}[t]{c}Ankush Kadu\thanks{Equal contribution.}\\{\normalsize QpiAI, Bangalore, India}\\{\normalsize\texttt{ankush.k@qpiai.tech}}\end{tabular}\quad
\begin{tabular}[t]{c}Athira Gopal\footnotemark[1]\\{\normalsize QpiAI, Bangalore, India}\\{\normalsize\texttt{athira.g@qpiai.tech}}\end{tabular}\quad
\begin{tabular}[t]{c}Niharika Singh\footnotemark[1]\\{\normalsize QpiAI, Bangalore, India}\\{\normalsize\texttt{niharika.s@qpiai.tech}}\end{tabular}\\[0.8em]
\begin{tabular}[t]{c}Arpit Jain\\{\normalsize QpiAI, Bangalore, India}\\{\normalsize\texttt{arpit.j@qpiai.tech}}\end{tabular}\quad
\begin{tabular}[t]{c}Aswanth Krishnan\\{\normalsize QpiAI, Bangalore, India}\\{\normalsize\texttt{aswanth.k@qpiai.tech}}\end{tabular}}
\date{}

\begin{document}
\maketitle
{\renewcommand\thefootnote{}\footnotetext{Accepted at the NeurIPS 2026 Workshops on Agentic AI Benchmarks and Applications for Enterprise Tasks (AABA4ET) and on Continual Learning in the Era of Foundation Models and Embodied Agents (CL4FMAgents).}\footnotetext{Correspondence: ankush.k@qpiai.tech}}

\begin{abstract}
Computer-use agents built on large language models (LLMs) work in software that can change while they act: an appointment moves, a policy is revised, a record is updated. Testing agents under such changes is harder than it looks. When an agent fails, the cause may be the agent, but it may also be a change that did not fire as intended, an update the agent never received, or a grader that scored the episode wrongly. We introduce \textbf{EnvOS}, a framework for building and certifying environments that change mid-task. Before any agent runs, EnvOS checks that each change fires correctly, that the task remains solvable, and that the grader separates correct from incorrect behavior; after each run, separate code re-checks the verdict from archived evidence.

We apply it to \emph{stale information}, where a value the agent has already seen changes before the agent acts on it. In clinic booking, Claude Opus~5 succeeds on 0/16 tasks when nothing extra is shown, 2/16 when reminded to re-check the source, and 15/16 when the updated value is shown; the gain over the reminder holds after correcting for all tests ($p=0.00293$). In insurance-claim approval, Opus~5 re-checks on its own and succeeds on 10/10, whereas Claude Sonnet~5 rises from 1/10 to 9/10 when the current policy is shown again, a gain significant in its pre-registered test but not after correcting for all tests. Whether an agent caught a change thus varied with the model and the interface. We also stress-test the grading itself: on corrupted episodes designed blind from the rules, the EnvOS pipeline accepted none of 26 as clean, and on an externally generated set it caught 25 of 39, against 4 for simple log checks. Because every EnvOS environment is generated from a seed and graded by an audited automatic verifier, the same environments are natural candidates for training agents with reinforcement learning from verifiable rewards.
\end{abstract}

\section{Introduction}
\label{sec:intro}

Computer-use agents increasingly execute multi-step workflows over stateful applications, but evaluation commonly treats the application state as fixed once an episode begins. In deployed workflows, an appointment can move, a reimbursement policy can be revised, or a selected resource can become unavailable independently of the agent's actions. These changes are not merely visual perturbations: they modify the authoritative records against which later actions are judged. Existing work has expanded interactive evaluation to long-horizon tasks, temporal disruptions, and evolving environments \citep{yuan2026osworld,hui2026sttarena,froger2026gaia2,zou2026interruptbench}. A complementary methodological problem remains: how can such changes be instantiated as valid test cases and how can one establish that an observed failure belongs to the agent rather than to the benchmark machinery? The question is sharper when the same environments supply a training signal: reinforcement learning with verifiable rewards trains agents against programmatic checkers \citep{wang2026cuagym,wei2026opencomputer}, so an intervention that silently fails or a checker that accepts a corrupted episode yields a wrong reward, not only a wrong benchmark number \citep{zhu2025abc}.

The question involves two different sources of uncertainty. \emph{Intervention validity} concerns whether the environment underwent the intended state transition, at the intended point in the workflow, without unintended side effects or an impossible endpoint. \emph{Evaluation validity} concerns whether the policy received its assigned observations, whether attempted rollouts were classified under fixed rules, and whether the scorer's verdict is supported by archived evidence. A terminal pass/fail label alone cannot resolve these distinctions. For example, a booking agent may appear to use an outdated clinic location even when the update occurred before it saw the original value, or when the promised review information never reached the model. Either case invalidates the intended interpretation without requiring a failure of the agent's decision policy.

We present \emph{EnvOS}, a phenomenon-based framework linking task construction with executable certification and auditable outcome analysis (Figure~\ref{fig:pipeline}). A \emph{phenomenon} specifies an application-independent class of changes to authoritative state. An executable \emph{perturbation} implements one instance of that class, defining the state operator, triggering condition, observation channel, and single-execution guard. The resulting task, seeded world, perturbation, endpoint criteria, and verifier form a candidate evaluation artifact. EnvOS admits that artifact for model evaluation only after scoped tests of the state transition, observation consistency, feasibility, behavioral discrimination, and checker robustness. Prospective experiments additionally freeze their design and counting rules, isolate the policy from evaluator data, and reconstruct supported verdict terms through separate implementations.

\begin{figure}[t]
\vspace{-4pt}
\centering
\includegraphics[width=0.85\textwidth]{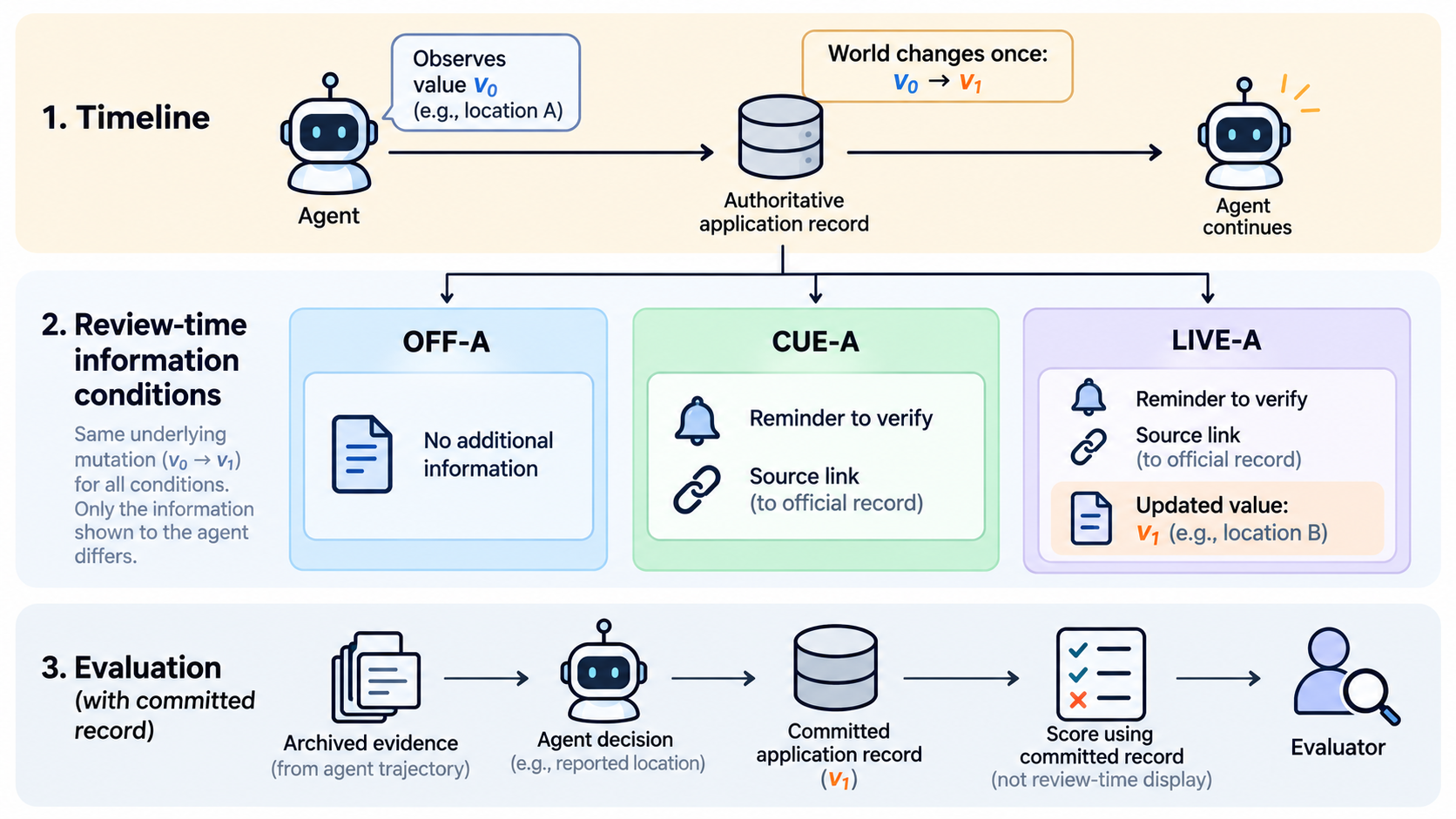}
\vspace{-6pt}
\caption{\textbf{Authoritative state and model observations during a stale-information intervention.} The application serves $v_0$ before a one-time exposure-conditioned mutation produces $v_1$. The review surface may disclose nothing, a link to the source, or the current value; these are observation conditions, not changes to the underlying mutation. Observation delivery is logged separately, and the evaluator scores committed application state rather than the disclosure review display.}
\label{fig:protocol}
\vspace{-6pt}
\end{figure}

The principal empirical analysis focuses on \emph{stale authoritative information}: the agent is first served a task-relevant value, the underlying application state changes on a later eligible request, and the agent subsequently makes a consequential decision. The timing is important: changing a value before it is served does not establish that the agent's earlier observation became stale. Figure~\ref{fig:protocol} separates the mutation of the authoritative record from the later interfaces through which its updated value may be disclosed. A recorded page serve, a delivered model observation, and the agent's comprehension are distinct events; the first two can be audited, whereas comprehension cannot be assumed from either.

For clarity, we retain the preregistered condition codes throughout. In the clinic experiment, OFF-A denotes no additional information at confirmation; CUE-A denotes a verification reminder and a link to the original record, with the updated location withheld; and LIVE-A denotes the same reminder and link with the current location shown. In the insurance experiment, OFF-A provides no additional policy information, whereas ON-A presents the current policy at approval. In both workflows, these conditions change the information presented to the agent, not the underlying state transition.

We investigate two applications of this evaluation design: clinic booking, where a session location changes, and insurance-claim approval, where a policy limit changes before the agent calculates an approved amount. In the principal paired comparisons the application-side change occurs in every condition; the manipulated variable is what appears at the review step. In clinic booking the agent receives either no additional review display, a review display pointing to the source page, or the current location itself. In insurance-claim approval the experiment compares review-time policy re-exposure with no additional disclosure. The original source remains accessible throughout. Consequently, the experiments estimate \emph{within-configuration review-information effects}, not the effect of making a static environment dynamic. Source revisits are reported as observed behavior, not automatically treated as a task instruction or a success criterion.

The evidence also exposes important differences across the tested configurations. In clinic booking, directly presenting the updated location improves Opus~5's success over providing a source link, including after paper-wide correction. In insurance approval, Opus~5 succeeds on the tested silent-change instances, while Sonnet~5 shows a block-local benefit from policy re-exposure that does not survive the retrospective paper-wide multiplicity adjustment. These results cannot be pooled into a controlled model ranking because task conditions and configurations differ. Separate archive challenges evaluate whether the same pipeline can detect deliberately corrupted records; their misses qualify, rather than disappear from, the reported evaluation claims.

\paragraph{Contributions.}
Our contributions are threefold. \textbf{(i) A phenomenon--perturbation formulation} that connects application-independent state changes to executable, exposure-conditioned task instances without conflating world state with model-visible notices. \textbf{(ii) A certification and evidence pipeline} encompassing intervention and feasibility checks, frozen execution rules, observation-delivery evidence, validity-first outcome scoring, independent reconstruction, and adversarial evaluator challenges. \textbf{(iii) An evidence-bounded empirical study} of two controlled stale-world workflows, together with evaluator challenge results that delineate what the outcome claims do and do not establish. Because instances are seeded and procedurally generated and outcomes are scored from committed application state, the certified environments and their verifiers are reusable as verifiable-reward environments for agentic RL, and the evaluator challenges show where such verifiers could mis-reward an episode. Supplementary phenomenon families are reported separately in Appendix~\ref{app:supplementary} and are not pooled with the principal experiments.

The remainder of the paper defines the framework and its intervention semantics, specifies the tasks and evaluation protocol, presents the prospective results and evaluator challenges, and discusses limitations. Implementation-level traces, generated evidence tables, and exploratory phenomenon-family analyses are retained in the appendices.

The upper row of Figure~\ref{fig:protocol} traces the authoritative state: a recorded $v_0$ serve precedes the one-time mutation to $v_1$, which determines the correct committed outcome. The lower row traces the information available to the policy, including the original delivered frame and the optional review disclosure. A review display does not cause or prove the state mutation, and a served page does not by itself establish that the model received or understood it. This distinction defines the evidence required for a legitimate stale-information episode.

\section{EnvOS Framework}
\label{sec:pipeline-overview}

\begin{figure}[t]
\centering
\includegraphics[width=0.96\textwidth]{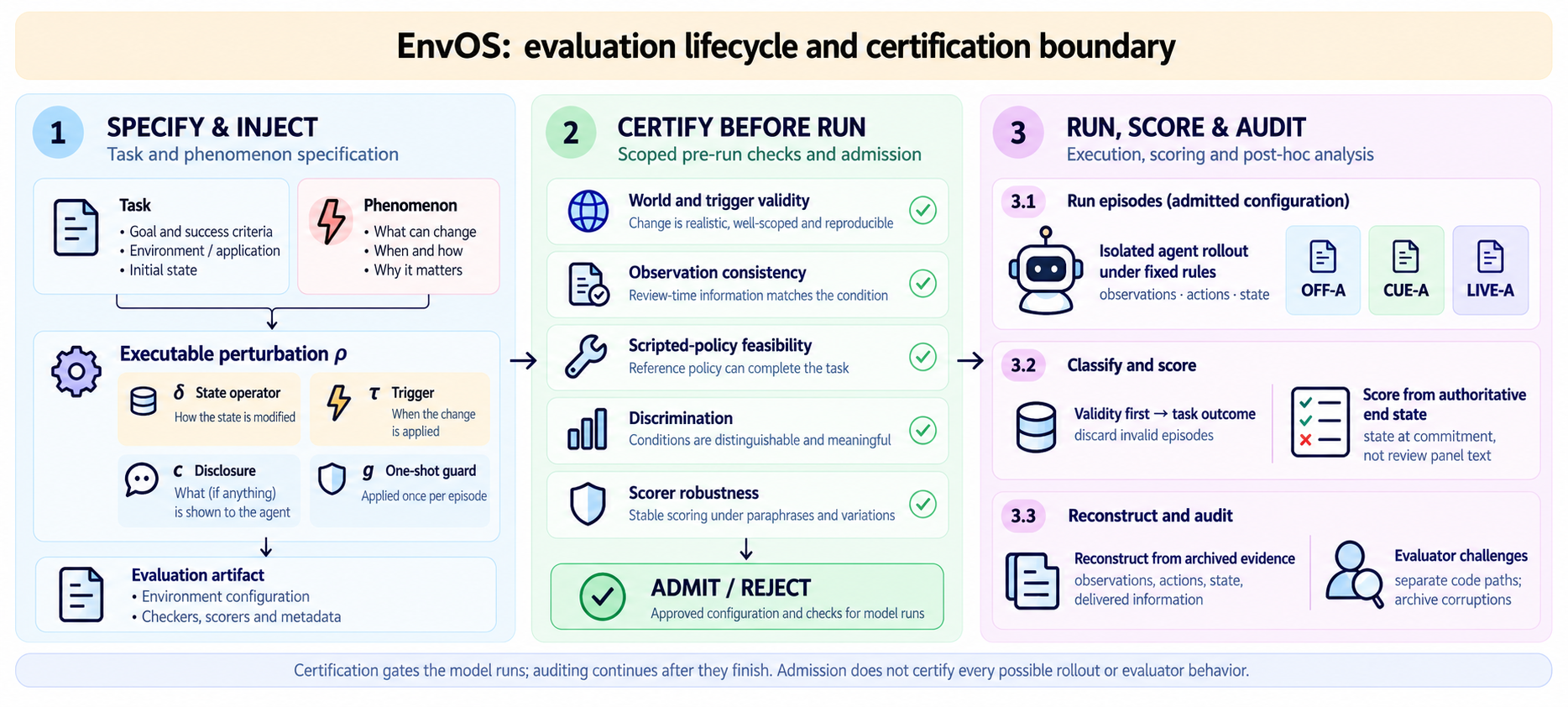}
\caption{\textbf{EnvOS evaluation lifecycle.} Task and phenomenon specifications define a candidate executable intervention. Pre-run checks assess the intervention, observation channel, feasibility, behavioral discrimination, and scorer robustness, determining whether the candidate is admitted to evaluation. Admitted instances are executed under fixed rules; their archived trajectories support validity classification, state-based task scoring, independently implemented reconstruction, and evaluator challenges. Admission applies to the specified checks and does not guarantee the validity of every subsequent episode.}
\label{fig:pipeline}
\end{figure}

EnvOS couples the construction of state-changing evaluation instances with executable assurance of their experimental validity. Its workflow has three stages (Figure~\ref{fig:pipeline}): constructing a candidate task and intervention, certifying the candidate against specified obligations, and executing admitted instances with independently auditable outcomes. The distinction is consequential. A functioning state mutation does not establish that a task remains solvable, that the assigned observations reach the agent, or that an endpoint checker correctly classifies the resulting behavior.

\subsection{Construction of Evaluation Instances}
A task specification determines the environment, initial state, user objective, and endpoint requirements. An associated phenomenon describes an application-independent class of environmental change relevant to that objective. EnvOS instantiates the phenomenon through an executable perturbation whose state operator, trigger, disclosure channel, and once-only guard are fixed with the task. The seeded world, perturbation, permitted observations, registered success invariants, and scorer jointly define a \emph{candidate evaluation artifact}. Author-specified task semantics are distinguished from automated instance construction, execution, and checking: the experiments do not claim unrestricted autonomous generation of new task objectives.

\subsection{Certification Before Model Execution}
Candidate instances undergo tests of five related properties. \emph{Intervention validity} checks the intended state transition, trigger boundary, and absence of unintended changes. \emph{Observation consistency} compares permitted disclosures with authoritative state and verifies the specified delivery path. \emph{Feasibility} requires a successful scripted route through the task. \emph{Behavioral discrimination} uses deliberately limited scripted policies to check that the constructed instance can expose the targeted deficiency. \emph{Scorer robustness} tests known corruptions, near-misses, and valid controls. Where applicable, scripted feasibility and discrimination checks are repeated on frozen generated instances rather than inferred from the template alone.

Each check returns \textsc{pass}, \textsc{violation}, or \textsc{monitor error}; an unassessable obligation does not count as a pass. Admission is conditional on the declared checks, not a proof that every future model trajectory will be correctly handled. The certification criteria and examples of their failure targets are specified in Section~\ref{sec:certification}.

\subsection{Controlled Execution and Adjudication}
Admitted configurations are evaluated with task instances, action budgets, validity criteria, outcome invariants, and analysis rules fixed before rollout. EnvOS archives application events, model-facing observations, agent actions, and committed state. A source page having been served is recorded separately from an observation actually delivered to the model; neither event establishes that the agent understood or used the information.

Run-level adjudication first determines whether the episode satisfies the registered validity conditions. Only valid episodes are assessed against applicable task invariants using the authoritative application state at the agent's consequential commit. This ordering prevents faulty interventions or missing model inputs from being counted as agent failures. Conversely, valid unsuccessful episodes, including those that exhaust the action budget, remain failures under the registered protocol.

\subsection{Evidence Reconstruction and Audit}
After the rollout, separate implementations reconstruct supported validity and endpoint terms from the preserved evidence. Unresolved terms remain unresolved when the archive lacks sufficient information. Sealed, rule-only, and externally constructed archive challenges then test whether the operational evaluator detects deliberately corrupted evidence and retains valid controls. Such challenges test evaluator reliability rather than agent capability; disagreements and false acceptances are reported alongside successful detections (Section~\ref{sec:audit}).

The division of evidence is summarized in Table~\ref{tab:lineage}. Pre-run certification asks whether a proposed test meets its declared contract; run-level verification asks whether one realized episode and its score are supported by the archive. These are complementary, not interchangeable, claims.

\begin{table}[t]
\centering\small
\caption{One EnvOS evaluation pipeline. The four assurance questions organize pre-run certification; the six audit stages track the evidence before, during, and after a particular model episode. Passing one part does not imply that another part was checked.}
\label{tab:lineage}
\begin{tabular}{@{}P{2.5cm}P{4.8cm}P{5.75cm}@{}}
\toprule
Assurance question & Candidate-instance checks & Run-level evidence and limits \\
\midrule
World validity & Authorized mutation, bounded scope, one-shot execution, anchored timing & State and access logs establish when the authoritative event actually fired \\
Observation consistency & Rendering comes from updated source state; available channels are consistent with intended condition & Served pages and delivered model frames are archived separately; neither establishes comprehension \\
Recovery validity & Reference completion route and declared deficient-policy discrimination, including frozen-instance feasibility & Actual model actions, rereads and commits are observed, not inferred from the scripted policy \\
Checker robustness & Scorer localization checks, negative mutations and positive controls & Validity-first inclusion, independently coded verdict reconstruction and external archive challenges test residual mistakes \\
\bottomrule
\end{tabular}
\end{table}

\subsection{Assurance Scope and Provenance}
\label{sec:lineage}
A versioned instance manifest connects the initial world, task, intervention, disclosure conditions, and evaluator. It preserves the distinction between obligations checked before rollout and evidence collected from a particular model episode. Operationally, it encompasses admission checks, policy isolation, frozen decision rules, validity-first counting, independent reconstruction, and evaluator challenges. The supplementary phenomenon corpus uses the same general assurance categories but is not assumed to share every implementation detail with the prospective experiments.

\section{Phenomena and Perturbations}
\label{sec:phen-object}

\subsection{Phenomenon Specification}
An environmental \emph{phenomenon} specifies a class of changes to authoritative application state, independent of any particular application or interface. A task defines the agent's intended outcome; the phenomenon defines an external event that may alter the state relevant to achieving that outcome. This separation prevents a model-visible notification from being treated as the underlying environmental event. In the two prospective workflows, the same broad phenomenon takes the form of a session-location update and a reimbursement-policy-limit update. Both affect application-owned records, rather than altering the user's request or inserting adversarial instructions.

A specification identifies the affected state variable, permitted transition, task relevance, admissible timing, and feasible post-change outcome. Its implementation is an \emph{executable perturbation}, which can differ across applications even when the phenomenon is conceptually shared. The prospective instances are generated within one synthetic workplace scaffold; they do not constitute evidence of portability to unrelated production applications.

\subsection{Executable Perturbation Model}
\label{sec:stale-specialization}
A perturbation is represented by
\begin{equation}
\rho=(\delta,\tau,c,g),
\end{equation}
where $\delta$ is an operator on authoritative state, $\tau$ identifies the triggering event, $c$ governs subsequent disclosure through permitted observation channels, and $g$ ensures that the intervention fires at most once. The state transition and its disclosure are distinct operations: the latter must reflect the former but does not constitute it. Researchers specify the task semantics, legally admissible changes, success criteria, and comparisons; EnvOS executes and checks the resulting configuration.

The focal \emph{stale-information} intervention requires a task-relevant value to have been served before a later, exposure-conditioned mutation supersedes it. The original serve arms the trigger; a subsequent eligible application request applies the update. Thus an update occurring before the old value was served would not establish the intended stale-world exposure. The harness additionally records the frame delivered to the model, because server-side service alone does not prove delivery. Figure~\ref{fig:protocol} illustrates this temporal distinction.

\begin{table}[t]
\centering\small
\caption{The phenomenon--perturbation mapping in the two prospective workflows. Entries specify the experimental roles, not the literal values of any generated instance.}
\label{tab:phenmap}
\begin{tabular}{@{}P{2.1cm}P{5.45cm}P{5.45cm}@{}}
\toprule
 & Clinic booking & Insurance claim approval \\
\midrule
Phenomenon & A booked session's previously served location is superseded & A policy's previously served reimbursement limit is superseded \\
Authoritative object & Target clinic session row & Target insurance policy row \\
Prior serve & Sessions page displays previous location & Policy page displays previous limit \\
Perturbation trigger & First eligible booking-wizard request after serve & First eligible form-entry request after serve \\
Mutation & Update the relevant stored location & Update the relevant stored limit \\
Disclosure & None, source link, or current location & None or re-exposed current policy record \\
Decision & Choose and submit compatible booking & Calculate and approve policy-constrained amount \\
Irrelevant control & Move an unrelated clinic session & Change an unrelated policy record \\
\bottomrule
\end{tabular}
\end{table}

\subsection{Intervention Admissibility}
The intervention contract requires four properties. First, the intended record must change at the specified request boundary, after the original-value serve and at most once. Second, the mutation must remain confined to the declared state scope; matched irrelevant-record controls test whether the machinery can run without changing the target answer. Third, observations must follow the assigned disclosure condition while the authoritative source remains accessible. Fourth, the post-change task must remain feasible, as witnessed by the registered reference policy on each admitted instance.

These requirements define a controlled state-changing task rather than an instruction to the agent to anticipate a hidden update. In both prospective workflows, the authoritative transition is held fixed across the principal paired conditions. Consequently, their comparisons identify differences associated with review-time information, not the causal effect of introducing the environmental change itself.

\section{Tasks and Evaluation Conditions}
\label{sec:tasks}\label{sec:applications}

\subsection{Clinic Booking}
The clinic task requires the agent to combine information from mail, calendar, and a booking portal to select and confirm a compatible appointment. The portal initially serves a session location; on a later eligible booking request, the authoritative session record is updated without an immediate announcement. The source record remains accessible. Success requires a correctly committed booking satisfying the applicable task invariants, rather than a particular source-revisiting action.

\subsection{Insurance Claim Approval}
The insurance task requires an approval amount consistent with the policy authoritative at the time of approval. Following the initial policy-page serve, an eligible request changes the stored reimbursement limit. The correct decision is
\begin{equation}
\mathrm{approved}=\min\bigl(\mathrm{claimed}-\mathrm{deductible},\ \mathrm{current\ limit}\bigr).
\end{equation}
The review-time policy record provides the current limit, not the computed answer. Some instances cross a calculation threshold, so a policy lookup alone is insufficient to guarantee correct approval. Unlike the clinic instructions, the claims task explicitly refers to the policy in force at approval; this difference is relevant when interpreting observed source revisits.

\subsection{Experimental Conditions}
The registered condition codes are defined in Table~\ref{tab:conditions}. The suffix \code{-A} indicates a task-relevant state change, \code{-I} an irrelevant-record change at the matched trigger, and \code{L0} a no-change control. In each principal paired contrast, the task-relevant record changes in both conditions; only the information provided before the final action differs. The authoritative source remains available throughout.

\begin{table}[h]
\centering\small
\caption{Experimental information conditions. In all principal comparisons, the application undergoes the same task-relevant state change and the original source remains accessible. Only the information supplied to the agent before its final decision differs.}
\label{tab:conditions}
\begin{tabular}{@{}p{3.15cm}p{1.9cm}p{6.95cm}@{}}
\toprule
Task & Code & Information available at the review step\\
\midrule
Clinic & OFF-A & No additional information at booking confirmation; the agent may revisit the original session record.\\
Clinic & CUE-A & A reminder identifies the session and provides a link for verification, without revealing the updated location.\\
Clinic & LIVE-A & The same reminder and link are accompanied by the updated session location.\\
Insurance & OFF-A & No additional policy information before approval; the original policy record remains accessible.\\
Insurance & ON-A & Current policy record is shown again, including the limit but not the approval amount.\\
Controls & -I, L0 & Irrelevant-record change at the trigger, or no change, respectively.\\
\bottomrule
\end{tabular}
\end{table}

In clinic booking, LIVE--CUE measures the effect of directly showing the current location beyond a verification reminder and source link. CUE--OFF measures the effect of supplying that reminder and link rather than no additional information. In insurance approval, ON--OFF compares re-exposure to the current policy record with no additional policy information. These are intervention-assignment effects under the action budget: they may jointly reflect salience, contradiction with an earlier value, and reduced verification effort.

\subsection{Endpoints and Trajectory Measures}
A task instance is a generated booking or claim scenario; an episode is one model rollout under one experimental condition. A \emph{commit} is the agent's consequential application action, distinct from EnvOS's earlier state mutation. Outcome measures are computed from the committed application state and registered invariants. Separate trajectory measures capture source revisits, link following, and policy re-queries. Such behaviors may help characterize the observed path but are not standalone success requirements. In particular, a correct outcome without revisiting the source is possible when the current value was supplied directly, and revisiting alone does not guarantee task success.

The two workflows probe distinct decision requirements. Clinic booking primarily tests selection under a superseded record value; insurance approval requires calculating a derived value from the updated policy. Their results are therefore analyzed by workflow and model configuration, not combined into a cross-model ranking.

\section{Intervention Timeline and Agent Rollout}
\label{sec:walkthrough}

In each stale-information episode, the application first serves the original task-relevant value and the harness records the observation delivered to the agent. The next eligible request, after that serve, triggers the authoritative mutation once. The assigned review-time information is then rendered from the updated state, and the agent continues under the fixed action budget. The same mutation occurs in all principal paired conditions, independently of whether the agent receives additional disclosure. Source revisiting remains possible in each condition.

At execution time the harness archives the state-change event, model-facing observations, browser actions, commit event, and final report. The outcome is not inferred from the displayed notice or from the model's verbal report: an episode first undergoes validity classification and, if counted, is scored from the committed record. Archived events subsequently support independent reconstruction and targeted corruption tests. These distinctions also apply to the insurance workflow, where the changed state enters an arithmetic endpoint rather than a location-selection endpoint.

\subsection{Episode-Level Evidence}
\label{sec:episode-evidence}
An episode contains separate evidence of the original page serve, delivered observation, state mutation, assigned disclosure, and final commit. This partition is essential because one event does not prove another: a server log does not establish model delivery, and model delivery does not establish comprehension or reliance. The evaluator tests the recorded sequence against the registered intervention and outcome rules. Illustrative request ordering and implementation-level traces are retained in Appendix~\ref{app:implementation-trace} rather than in the main methodological argument.

\subsection{Evaluation Harness}
The prospective experiments use a synthetic workplace testbed with \WorldServices{} containerized infrastructure services, including clinic and insurance applications and supporting mail, calendar, and database services. Only two workflows are prospectively evaluated. The policy receives screenshot and text observations through a controlled semantic-browser interface, acts through a fixed schema, and is invoked statelessly with a short memo and recent-action history. The clinic and claims tasks allow \ActionBudget{} and \ClmActionBudget{} actions, respectively. Matched conditions share instance data and instructions. This interface does not constitute unrestricted operating-system-level computer use; the relevant action and observation contracts are specified in Appendix~\ref{app:interface}.

\section{Certification and Evidence Auditing}
\label{sec:certification-overview}

EnvOS evaluates the test construction and the model episode separately. Pre-run certification checks whether the proposed intervention satisfies a declared contract, whereas post-run auditing determines whether the recorded trajectory supports its validity classification and outcome. Both are necessary: an admissible candidate may still produce an invalid rollout, and a seemingly correct final record cannot establish that the intended stale-world sequence occurred.

\subsection{Authoritative State and Observation Channels}
\label{sec:authority-channels}
The application database is the source of truth; a page or message shown to the policy is an observation derived from that state. For a stale-world episode, the original-value serve must precede the exposure-conditioned update, which must precede any review-time disclosure of the changed value. Serving and delivery are separate recorded events. The state operator changes the application-owned record; the disclosure channel changes only information presented to the agent. As a result, differences between the experimental conditions do not amount to different injected phenomena.

\begin{definition}[Phenomenon]
An application-independent specification of a class of authoritative-state changes and the conditions under which an instance of that class is admissible.
\end{definition}
\begin{definition}[Perturbation]\label{def:pert}
An environment-specific executable intervention $\rho=(\delta,\tau,c,g)$ comprising a state operator, triggering condition, observation channel, and once-only guard.
\end{definition}

\subsection{Perturbation Admissibility}
The validity requirements concern timing, state scope, observation delivery, and endpoint evaluation. Timing tests establish a prior serve and a subsequent single update at the declared request boundary. Scope tests rule out injector-induced changes to unrelated records. Observation checks establish whether the assigned information reached the policy through the admitted channel. Finally, outcome checks use the application state at commit rather than the disclosure text. An episode can satisfy some requirements and fail others; a terminal success flag therefore cannot certify the full intervention history.

\subsection{Exposure-Conditioned Interventions}
\label{sec:exposure}\label{sec:triggers}
An initial serve arms the trigger, and the next eligible agent request applies the mutation as an application-side transaction. The transaction is archived independently of the browser frame. Irrelevant-record controls exercise the same machinery while preserving the correct target decision. This construction establishes an auditable check--use interval without presuming that the agent noticed, retained, or understood the original value. The permitted source remains accessible after the update, but absence of a source revisit is not automatically an instruction violation.

For a valid episode, the scorer applies the five registered endpoint requirements where applicable: correct target state, a single commit, absence of collateral changes, required post-commit continuation, and consistency between the final report and committed state. Insurance approval does not require the continuation term. Process measures such as source revisits are recorded separately from these outcome invariants.

\subsection{Pre-run Certification}
\label{sec:certification}
Certification exercises the actual environment and candidate scorer before model evaluation. World and trigger tests check the event sequence, target scope, one-shot guard, and condition-equivalent neutral operations. Observation tests compare delivered content with the intended channel. Successful reference policies establish a feasible completion route; deliberately deficient policies test whether the instance can distinguish the specified behavior. These scripts are exercised on the frozen generated instances, since a valid template does not ensure that every generated booking or numerical claim remains discriminative.

Scorer tests include event-localization checks, deliberately corrupted records, near-misses, and valid positive controls. The relevant cases include an update firing too early or twice, missing promised observation delivery, a mismatched commit, and an incorrectly accepted or rejected archive. An unavailable check returns \textsc{monitor error} and blocks admission rather than being silently recorded as passing. These tests provide evidence for the specified fault classes; they are not a general guarantee of coverage against unknown defects.

\begin{table}[t]
\centering\small
\caption{Certification checks are admission evidence with specific failure targets. All checks are scoped to the tested scripted routes and generated instances.}
\label{tab:certificationdetails}
\begin{tabular}{@{}P{2.4cm}P{4.55cm}P{4.0cm}P{2.5cm}@{}}
\toprule
Check family & What is tested & Example invalid case & What passing does not prove\\
\midrule
World / trigger & old-value serve; anchored one-time update; condition-equivalent neutral operations & mutation fires before prior serve or fires twice & agent noticed the old value\\
Condition discrimination & reference and deficient scripted-policy trajectories and expected signatures & a value-only script passes when the value was hidden & arbitrary models will find the reference route\\
Instance feasibility & scripted completion and discrimination on each frozen generated instance & generated slot or decision is impossible & all future cases are feasible\\
Scoring instrumentation & event localization and registered target/commit invariants & scorer associates change with wrong request & every outcome field is reconstructable\\
Mutation / positive controls & deliberately corrupted and valid archives & damaged event enters clean; good archive is rejected & unknown corruption classes are covered\\
\bottomrule
\end{tabular}
\end{table}

A certificate applies to the candidate and exercised scripted routes. Actual model episodes require fresh delivery evidence, validity classification, and endpoint checks; neither model understanding nor the correctness of every possible trajectory follows from a successful pre-run test.

\subsection{Outcome Reconstruction}
\label{sec:reconstruction}
The archive separates authoritative state and commit records from delivered model observations and the action trajectory. One implementation reconstructs applicable endpoint invariants; a separate standard-library path re-derives validity classifications from raw events and frozen rules. Terms that cannot be supported by the archive are left unresolved. For the prospective clinic workflow, appointment-availability terms can be reconstructed from saved state, whereas calendar compatibility relies on specified generator and state-preservation conditions; the claims endpoint has broader independent reconstruction coverage (Table~\ref{tab:coverage}).

Here \emph{independent} refers to separate code paths developed by the same team, not an independent research group or independent source data. Agreement can reveal implementation disagreements but does not establish that their common archived evidence is free of errors. The evaluator challenges in Section~\ref{sec:audit} test this remaining exposure.

\subsection{Audit Coverage}
A six-stage operational audit records certification, policy isolation, frozen experimental rules, validity-first counting, independent re-derivation, and targeted challenges. Each stage addresses a distinct failure mode, as summarized below.

\begin{table}[t]
\centering\footnotesize
\caption{Audit stages, evidence and scope. Every successful check establishes a limited property rather than a general guarantee of benchmark correctness.}
\label{tab:auditchain}
\begin{tabular}{@{}P{1.6cm}P{4.35cm}P{4.15cm}P{3.0cm}@{}}
\toprule
Stage & Evidence & Fault exposed & Boundary \\
\midrule
Certify & scripted policies, trigger logs, per-instance feasibility and mutation controls & update fires early/twice; reference path impossible; scorer misses event & scripted paths only \\
Isolate & sealed OS user, allowed origins, canary access probes, observation-delivery records & policy leaks evaluator data or misses promised channel & unarchived provider-side returns remain \\
Freeze & preregistration, SHA-256 manifests, fixed templates & post-outcome changes to analysis or software & dated amendments are still possible and disclosed \\
Count & validity table, provider/harness log, reserves and stop rules & provider outage treated as model failure; invalid rerun cherry-picked & validity classification can itself err \\
Re-derive & separately implemented outcome scorer and standard-library validity checker & incorrect success invariant or eligibility verdict & same team; shared archive; unresolved terms \\
Challenge & corrupted archive copies, sealed expectations, baselines and external generator & detector false accepts and valid-control false rejects & construction access, disputes and coverage vary \\
\bottomrule
\end{tabular}
\end{table}

\subsection{Policy Isolation and Observation Delivery}
\label{sec:isolation}
During prospective execution the sealed model policy uses a separate operating-system account, restricted file permissions and specified browser origins. The harness records model-facing frames and permitted actions; canary and access probes test whether policy inputs remain separated from evaluator files and future-state evidence. These measures bound the policy's access under the declared harness configuration. They do not reconstruct all provider-side inputs, a limitation documented with the archived evidence inventory.

\section{Experimental Design}
\label{sec:design}

A reliable task score needs two decisions in order: first whether the episode belongs in the analysis, then whether the committed task outcome passes. The study also needs a comparison rule fixed before any model results are observed. EnvOS documents these decisions per model--workflow study block rather than searching across models for a single leaderboard score.

\subsection{Preregistration and Validity}
\label{sec:validity}\label{sec:freeze}

Before the first model rollout of a prospective block, the researchers freeze its task instances, conditions, contrasts, budgets, validity rules, reserves, stopping rules, statistical analysis, outcome invariants and interpretation templates. A release manifest hashes \ManifestPaths{} repository paths, relevant container images, generator, invoker and requested model identifier; the driver rejects a run if its live artifacts drift. Amendments and departures are dated and reported rather than silently incorporated (Appendices~\ref{app:rules} and~\ref{app:disclosures}).

Validity is counted before outcome. Provider rejection or failure to deliver the promised policy input can be excluded and replaced only under the registered reserve (\ReserveCell{} per cell; \ReserveBlock{} per block). A valid run that reaches its action budget counts as an unsuccessful run, not a replacement. A counted clean-control failure stops a block under the declared rule; that stopping decision alone does not prove a harness defect. The two workflows retain their distinct preregistered treatment of untriggered runs: claims counts them as failures under its table; clinic replaces them because the intended intervention did not occur. The claims blocks used \BlockClmBReplacementsSpent{} and \BlockClmBSReplacementsSpent{} replacements, respectively.

\subsection{Matched-Instance Comparisons}
\label{sec:stats}

Each generated instance is executed under every experimental condition for its block. A \emph{treatment-only success} means that instance succeeds in the reviewed-information condition but not the comparator; a \emph{control-only success} is the reverse; and a \emph{tie} means their success indicators match. The preregistered comparisons use an exact two-sided sign test on discordant pairs at $\alpha=\Alpha$, with marginal Newcombe 95\% intervals for the paired difference. Condition order is fixed and counterbalanced, not randomized.

The tests assume exchangeable directions of discordance under the null and no carry-over or period effects. Inference is conditional on the fixed instances and one sampled policy rollout per condition; reported intervals and $p$-values do not incorporate task-population variation or repeated-model-sampling variation. The total effects include any navigation effort saved by a review information at the fixed action budget. Within-order counts are descriptive, not evidence that period effects are absent. \ClmOrderClause{}

Prospective outcome interpretations are selected mechanically from templates written before the corresponding results, and void or incomplete blocks do not receive inferential readings (Appendix~\ref{app:readings}). Each block uses its own registered local multiplicity rule. \HolmMainSentence{} The retrospective twelve-test paper-wide Holm family is presented as sensitivity analysis; it is not substituted for the prospective claims registration (Appendix~\ref{app:multiplicity}).

\subsection{Inference and Assumptions}
Let $Y_i(c)\in\{0,1\}$ denote whether the agent completes generated instance $i$ under review condition $c$, with the same target-change phenomenon injected in every condition of the registered contrast. The observed paired difference for a treatment condition $c_1$ and reference condition $c_0$ is
\[
\widehat{\Delta}=\frac{1}{n}\sum_{i=1}^{n}\left(Y_i(c_1)-Y_i(c_0)\right)=\frac{N_{+}-N_{-}}{n},
\]
where $N_+$ counts treatment-only successes, $N_-$ counts reference-only successes, and the remaining $n-N_+-N_-$ are ties. The exact two-sided sign test is based only on discordant pairs. A positive result therefore supports a difference between review conditions \emph{for this fixed set of instances under the sampled rollouts and interface}. The estimate includes any saved navigation effort under the action limit, not merely a perceptual effect of seeing a value.

This notation also clarifies what is \emph{not} estimated. Because both $Y_i(c_1)$ and $Y_i(c_0)$ are observed after injection, the analysis does not contain a paired no-change/no additional information potential outcome for the same experiment. It cannot identify the average effect of the phenomenon occurring at all. Further, a no additional information failure does not demonstrate that the agent ignored a known update: by design the update was unannounced. Rechecking is a behavior visible in the trajectory, not a randomized intermediate variable, so the study cannot identify mediation through rechecking independently from other mechanisms such as contradiction, salience and reduced verification effort.

\section{Results}
\label{sec:experiments}

We report prospective matched-instance evaluations in clinic booking and insurance-claim approval. Both workflows use preregistered feasibility and validity criteria; task-relevant state changes are held fixed across the principal compared conditions. Table~\ref{tab:results} presents full-task outcomes and paired comparisons, and Figure~\ref{fig:results} displays instance-level results. Exploratory findings are reported separately in Appendix~\ref{app:corpus} and do not contribute to the confirmatory tests.

\begin{table}[t]
\centering
\caption{Prospective model results. Success requires the full evaluator endpoint; in clinic rows its
calendar-compatibility term rests on a stated design condition, not independent re-derivation. Paired counts:
treatment-only/control-only/ties; $p$: exact two-sided sign test (BC: pre-registered block-Holm value in
parentheses). `indep.\ endpoint': counted rollouts whose independent reconstruction (clinic: without calendar
compatibility) agrees with the evaluator. Blocks without results print their state; nets, intervals:
Table~\ref{tab:tests}. \ClmGCaption}
\label{tab:results}

\begingroup\setlength{\tabcolsep}{4pt}\small
\begin{tabular}{@{}lccclccc@{}}
\toprule
 & \multicolumn{3}{c}{arm successes} & paired & for/ag./ & local $p$ & indep. \\ \cmidrule(lr){2-4}
model (block) & OFF & ON/CUE & LIVE & contrast & tie & (block Holm) & endpoint \\
\midrule
\multicolumn{8}{@{}l}{\textit{insurance-claim approval}} \\
Opus~5 (CLM-B) & \BlockClmBClmOffAPassFrac & \BlockClmBClmOnAPassFrac & --- & ON$-$OFF & \BlockClmBOnOffRescue/\BlockClmBOnOffHarm/\BlockClmBOnOffTie & \BlockClmBOnOffP & \BlockClmBEndpointFull/\BlockClmBCounted \\
Sonnet~5 (CLM-BS) & \BlockClmBSClmOffAPassFrac & \BlockClmBSClmOnAPassFrac & --- & ON$-$OFF & \BlockClmBSOnOffRescue/\BlockClmBSOnOffHarm/\BlockClmBSOnOffTie & \BlockClmBSOnOffP & \BlockClmBSEndpointFull/\BlockClmBSCounted \\
Gemini (CLM-BG) & \ClmGOffAPassFrac & \ClmGOnAPassFrac & --- & ON$-$OFF & \ClmGRescue/\ClmGHarm/\ClmGTie & \ClmGP & \ClmGEndpointFull/\ClmGCounted \\
\midrule
\multicolumn{8}{@{}l}{\textit{clinic booking}} \\
Opus~5 (BC) & \BlockBCFRXOffAPassFrac & \BlockBCFRXCueAPassFrac & \BlockBCFRXLiveAPassFrac & LIVE$-$CUE & \BlockBCLiveCueRescue/\BlockBCLiveCueHarm/\BlockBCLiveCueTie & \BlockBCLiveCueP (\BlockBCLiveCueHolmLocal) & \BlockBCEndpointFull/\BlockBCCounted \\
 & & & & CUE$-$OFF & \BlockBCCueOffRescue/\BlockBCCueOffHarm/\BlockBCCueOffTie & \BlockBCCueOffP (\BlockBCCueOffHolmLocal) & \\
Sonnet~5 (BCS) & \multicolumn{7}{l}{\textit{\BlockBCSStateCell}} \\
\bottomrule
\end{tabular}
\endgroup

\end{table}

\begin{figure}[t]
\centering
\includegraphics[width=\textwidth]{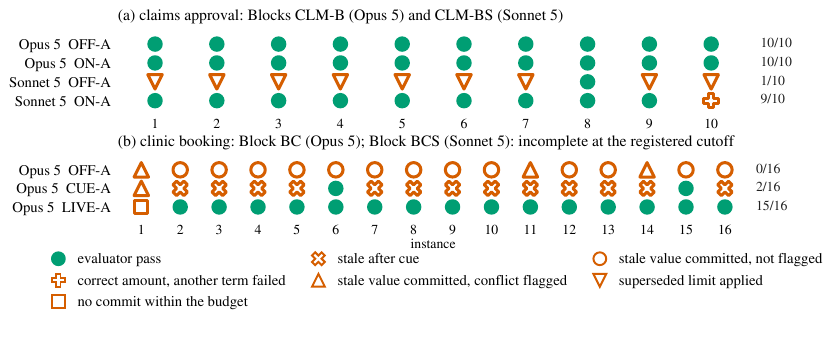}
\caption{\textbf{Matched experimental outcomes by instance.} Insurance-claim approval compares no additional policy information (OFF-A) against re-exposure to the current policy (ON-A); clinic booking compares no additional information (OFF-A), a verification reminder and source link (CUE-A), and direct disclosure of the updated location (LIVE-A). Each panel corresponds to a distinct registered model--workflow configuration; outcomes are not pooled across models.}
\label{fig:results}
\end{figure}

Per-instance outcomes in Figure~\ref{fig:results} preserve the pairing structure underlying the exact tests. The analysis treats each model--workflow configuration separately; discordant matched outcomes contribute to the sign test, whereas matching outcomes are ties. Table~\ref{tab:results} reports full-task success rather than successful information retrieval. The clinic comparison remains significant under the paper-wide adjustment; the Sonnet claims comparison is supported at its registered block-local level but not under the retrospective paper-wide family.

\subsection{Clinic Booking}
\label{sec:bc}

\paragraph{Design.}
The clinic block (BC) evaluates the contribution of disclosing the updated session location relative to providing a verification reminder and source link. It assigns the same \BlockBCInstances{} fresh instances to no additional information (OFF-A), a verification reminder with a source link (CUE-A), and direct display of the current location (LIVE-A) in a fixed counterbalanced order, with \BlockBCControls{} irrelevant-change controls (LIVE-I). The preregistered primary contrast is LIVE--CUE; the key secondary contrast CUE--OFF asks whether the verification reminder improves success over no additional information before confirmation. These are unconditional, block-local Holm-adjusted comparisons under the fixed \ActionBudget-action budget.

\paragraph{Results.}
\BlockBCReadingBody{} For transparency, the primary LIVE--CUE contrast has
\BlockBCLiveCueRescue{} treatment-only successes, \BlockBCLiveCueHarm{} control-only successes and
\BlockBCLiveCueTie{} ties (net \BlockBCLiveCueNetCI; exact $p=\BlockBCLiveCueP$;
block-local Holm $p=\BlockBCLiveCueHolmLocal$). The secondary CUE--OFF contrast has
\BlockBCCueOffRescue{} treatment-only successes, \BlockBCCueOffHarm{} control-only successes and
\BlockBCCueOffTie{} ties (net \BlockBCCueOffNetCI; block-local Holm $p=\BlockBCCueOffHolmLocal$).

\paragraph{Trajectory evidence.}
The source page was accessible under every condition. Only \BlockBCFRXCueALinkFollowed{} of
\BlockBCFRXCueAN{} verification-reminder trajectories followed the supplied link. This is a descriptive process
count, not a failure of an instruction to check or a separate definition of success. Independent
validity re-derivation agreed with the archive on \BlockBCVindAgree{} attempts; the scorer's
calendar-compatibility limitation is stated in the registered outcome above. The design's
pre-rollout enlargement and historical motivation are documented in Appendices~\ref{app:disclosures}
and~\ref{sec:historical}. The separately registered Sonnet clinic block (BCS) ended incomplete
at its cutoff and receives no inferential interpretation: \BlockBCSStateCell.

Directly providing the updated location may expose a mismatch with the agent's previously observed value and avoids requiring navigation back to the session record. A reminder with a source link leaves verification to the agent and can consume additional actions. The registered difference is the total effect of the interface choice at 48 actions. It is not an estimate of ``value understanding'' with navigation cost removed, and a failure to follow the cue link is not an independent failure under the task instruction. These mechanisms remain possible explanations rather than tested mediators.

\subsection{Insurance Claim Approval}
\label{sec:claims}

\paragraph{Design.}
The claims blocks evaluate approval accuracy when the current policy is re-exposed before the consequential decision. Each model is evaluated in its own block over the same \BlockClmBInstancePhrase{} with paired OFF-A and ON-A target-change conditions and additional ON-I and L0 controls. The blocks were registered before their model rollouts and follow a \BlockClmBOrderWord{} counterbalanced order. The interface presents the current policy record before approval, not the calculated approval amount; it combines a verification cue, value re-exposure, contradiction salience and saved lookup effort.

\paragraph{Claude Sonnet~5.}
\BlockClmBSReadingMainS{} \ClmBSPaperHolmR{} The registered local contrast supports higher full-task success with review-step re-exposure. Its retrospective twelve-test paper-wide Holm adjustment does not reject, so the two inference families must not be conflated. Sonnet applied the superseded limit in \BlockClmBSOffStale/\BlockClmBSClmOffAN{} episodes without additional policy information and requested the policy record after the change in \BlockClmBSOnReread/\BlockClmBSClmOnAN{} ON-A episodes. These are descriptive trajectory counts, not separate randomized tests of verification behavior. \ClmScorerScope{}

\paragraph{Claude Opus~5.}
\ClmOpusMain{} In these instances Opus already completed every claim without additional policy information correctly, so the registered rescue-opportunity gate was not met; that ceiling does not establish that the review display is generally ineffective. The full registered reading and process counts are in Appendix~\ref{app:claims}.

\paragraph{Gemini follow-up.}
\ClmGResultsR{} This is a separate preregistered model--invoker configuration on reused claims instances. A non-rejection does not establish absence of a benefit, and differences in provider, invocation and frame-delivery channel prevent isolation of vendor or model effects (Appendix~\ref{app:clmg}).

\section{Evaluator Audits}
\label{sec:caught}\label{sec:audit}

Agent outcomes are credible only if the evaluation can distinguish clean episodes from corrupted or unsupported records. An \emph{archive challenge} starts with a preserved episode and changes selected evidence under a known expected verdict. For example, altering the recorded timing so a purportedly mid-task mutation occurs before the old value was served should cause the validity layer to object; simply scoring the final booked row could miss the issue. An \emph{outcome scorer} reconstructs task invariants, \emph{validity reconstruction} re-applies eligibility rules, and \emph{evaluator replay} reproduces the archived evaluator's own checks. A simple raw-log checker serves as a limited comparator, not an independent oracle. The challenge is whether each layer recognizes the corruption within its stated scope.

\paragraph{Challenge design.}
The sealed, run-once challenge was built with repository access but was unseen by the Claude agents that built the pipeline. The rule-only challenge was specified from rule excerpts and the archive schema without implementation or outcome access, then instantiated by the authors and separately reviewed for fidelity. The external Gemini CLI challenge supplied rules, schema and archive copies to an agent that designed and generated its own cases. None was an independent-team audit. The challenges differ in their access, implementation, and the kinds of errors they can test; their original frozen expectations are kept even where disputed.

\ChallengeMainThree

\BlindMainThree

\ExtMainR

\paragraph{Challenge outcomes.}
A corruption caught and rejected is not the same as one receiving no verdict, and an expected verdict challenged by the authors is not automatically a proven evaluator defect. The external challenge's original labels include \ExtPipeFATotal{} of \ExtCorTotal{} designated corruptions counted clean and \ExtDisputes{} disputed expectations after scoring; its pipeline also rejected three of 16 controls, as detailed in Appendix~\ref{app:external}. We retain the original classification alongside the disputes rather than replacing the headline with a retrospectively restricted subset.

\paragraph{Raw-log baseline comparison.}
\MatchedMainFour

\begin{table}[tb]
\centering
\caption{\MatchedCaption}
\label{tab:matched}

\begingroup\footnotesize\setlength{\tabcolsep}{3pt}
\begin{tabular}{@{}P{3.9cm}P{1.1cm}P{2.0cm}P{6.3cm}@{}}
\toprule
checker & controls & corruptions (in scope) & would enter the results unflagged \\
\midrule
\multicolumn{4}{@{}l}{\emph{clinic: the sealed challenge's \MBClinicCtl{} controls and \MBClinicCorr{} corruptions, its recorded outputs re-tabulated}} \\
protocol, implemented reading & 10/10 & 15/18 (11/12) & notice delivery (\code{C01}); delivery of review-time information (\code{C02}); commit before change (\code{C08}) \\
protocol, registered reading & 10/10 & 16/18 (11/12) & commit before change (\code{C08}); truncated trajectory (\code{C15}) \\
simple baseline (\MBClinicBaseLines{} lines) & 9/10 & 11/18 (11/12) & 7, of which 6 outside its declared scope \\
\midrule
\multicolumn{4}{@{}l}{\emph{claims: \ValControls{} controls and \ValCorrupt{} corruptions we built from counted archives}} \\
operational pipeline & 10/10 & 16/19 (10/12) & duplicate commit in the access log only (\code{C09a}); truncated trajectory (\code{C15}); stale pages after the change (\code{C17p}) \\
simple baseline (\ValBaseLines{} lines) & 10/10 & 6/19 (6/12) & 13, of which 7 outside its declared scope \\
\bottomrule
\end{tabular}
\endgroup

\end{table}

Within the simple baseline's declared scope, the gap is smaller than an unrestricted total might suggest: on the clinic subset both approaches rejected 11 of 12 corruptions, while on claims the pipeline rejected 10 and the baseline 6 of 12 (Table~\ref{tab:matched}). The baseline and comparison cases were constructed post hoc by the authors with the pipeline in view, and the scores should not be interpreted as an unbiased detection-rate estimate. The external challenge is the more consequential test of independently constructed cases, albeit still produced by a model agent rather than an independent research team.

\subsection{Challenge Coverage and Failure Modes}
The archive challenges are better interpreted as \emph{tests of specific failure hypotheses} than as a single scalar score for evaluator quality. The first sealed challenge kept its builder's access and tested whether separate layers reproduced frozen expectations. A second challenge separated case specification from implementation and access to previous outcomes, but the cases were subsequently instantiated by the authors and therefore do not constitute end-to-end blind testing. The external challenge delegated both case design and generator construction to a different model agent, making it less entangled with the pipeline's design decisions but still not equivalent to validation by an independent research team.

Four outcomes must be separately reported. A \emph{detected corruption} is caught and appropriately barred from clean results; an \emph{undetected corruption} enters as clean; a \emph{no-verdict} case is not accepted clean but also does not produce the registered classification; and a \emph{control rejection} incorrectly challenges evidence designated valid. Disputed expected verdicts are additionally tracked without assuming either side of the dispute is right. The external challenge produced examples of all the consequential categories, including \ExtPipeFATotal{} of \ExtCorTotal{} designated corruptions accepted as clean, three of 16 controls rejected, and \ExtDisputes{} disputed expectations. Reporting only the caught cases would overstate reliability, while collapsing abstentions into successful detections would erase uncertainty.

The post-hoc simple-checker comparison asks a different question: which classes of defects are covered by the protocol's audit layers beyond a short baseline operating on the same raw evidence? Because the baseline and many mutations were constructed by the same authors after viewing the pipeline, the difference is diagnostic, not an unbiased population detection-rate estimate. The within-scope counts matter: for the narrow clinic subset, both pipeline and baseline catch 11 of 12; for claims, the pipeline catches 10 of 12 compared with six for the baseline. EnvOS's broader checking surface remains useful, but breadth does not imply that every individual check is more accurate.

\paragraph{Development defects.}
Certification, re-derivation and audits surfaced \NDefectsTotal{} defects, which now serve as regression cases. For \NDefectsModelAttributable{}, a previously scored archive already carried a model verdict under the defective code. Examples include timing, delivery and scoring mismatches that could otherwise become erroneous claims about agent behavior (Appendix~\ref{app:defects}). Every historical and prospective block is interpreted under the actual code that governed it; repairs do not retroactively make an earlier verdict more reliable. The remaining undetected corruptions, unevaluable archive terms and incomplete provider-side traces show why an evaluator should be regarded as tested within stated bounds, not certified correct in general.

\section{Related Work}
\label{sec:related}

\paragraph{Phenomenon-based evaluation.}
A phenomenon-based benchmark must distinguish the application state that changes from the model-visible information used to communicate that change. EnvOS encodes this distinction in an executable perturbation and makes certification a prerequisite to model measurement. Its most specific evaluated contribution is the combination of a prior-serve requirement, controlled review-step disclosure, and independent tests of timing, delivered observations and scoring. The goal is not to replace general-purpose benchmark generation or to claim that injections or evaluator tests themselves are unprecedented. Instead, the design is organized around which concrete evidence can support each conclusion about a sampled agent action. Supplementary phenomenon families illustrate other state-change classes under their own recorded execution and analysis conditions (Appendix~\ref{app:supplementary}).

\paragraph{Dynamic environments and adaptation.}
OSWorld~2.0 evaluates realistic computer workflows with mid-task information updates \citep{yuan2026osworld}. Gaia2 studies asynchronous changes and temporal constraints \citep{froger2026gaia2,are2025}; InterruptBench changes user goals during execution \citep{zou2026interruptbench}. StressWeb studies controlled interface and action-semantic perturbations \citep{bai2026stressweb}, while DynamicGUIBench and AgentHijack examine changes to dynamic interfaces, observation and control \citep{dynamicguibench2026,agenthijack2026}. STT-Arena is particularly close: it builds dynamic tool environments and validates injected conflicts, timing and plan disruption \citep{hui2026sttarena}. These efforts already establish controlled interventions and validation as important design ideas. EnvOS instead narrows its experimental target to an application value first served to a browser agent, then superseded, with matched review-step disclosure conditions and archive-level checks of that sequence. This is a benign time-of-check--time-of-use gap \citep{lilienthal2025toctou}, not a claim that the gap itself is newly identified.

\paragraph{Observation and stale-state reasoning.}
STALE studies recognition and downstream use of invalidated memories \citep{chao2026stale}; EnvProbe considers when an agent re-queries its environment \citep{song2026envprobe}; and work on agent--computer observation interfaces examines richer between-step observations \citep{li2026aoi}. EnvOS evaluates durable task actions under an unchanged instruction and contrasts no additional information, a link to accessible source information, and direct display of the current value. It does not claim that improved observation is new or that a record's prior serve proves comprehension.

\paragraph{Task generation and perturbation design.}
CUA-Gym co-generates task instructions, environment states and verifiable reward functions for computer-use training \citep{wang2026cuagym}. This is a substantive task-generation contribution; EnvOS does not replace it or claim that generation itself is novel. The two are complementary: CUA-Gym scales environments and rewards for training, whereas EnvOS certifies that a mid-episode change and its verifier behave as specified, a precondition for trusting either as a training signal. Our scientific unit is a \emph{mid-episode authoritative transition anchored to the agent's prior serve}, with separately controlled review-time disclosure and archived evidence supporting or withholding an outcome verdict. The present study uses authored phenomenon specifications and an implemented perturbation mechanism in a bounded synthetic testbed; it does not establish general autonomous discovery or synthesis of new change classes. Likewise, STT-Arena already injects dynamic spatio-temporal conflicts and validates key intervention properties. The distinction here is not ``first dynamic event'' but the specific exposure-and-delivery contract, matched disclosure contrasts and independent archive checks around the resulting full-task action.

\paragraph{Evaluation validity and verifier reliability.}
Multi-service environments such as TheAgentCompany and controlled task-generation systems motivate reproducible application workflows; generation alone is not the contribution here. OpenComputer investigates verifiable software worlds and state-based judging \citep{wei2026opencomputer}. ABC and log-based evaluation critiques document task, reward and attribution threats \citep{zhu2025abc,kirgis2026logs}; broader reviews question what an aggregate agent score measures \citep{notsingletarget2026,predictivevalidity2026}. EnvOS combines established techniques---state-based outcomes, preregistration, mutation testing, independent code paths and provenance hashes---around the specific timing, disclosure and scoring of an authoritative mid-task change. The comparison concerns focus and evidence coverage, not a claim that other benchmarks lack validation or that this feature combination is universally unprecedented (extended discussion: Appendix~\ref{app:related}).

\section{Limitations}
\label{sec:limitations}

The prospective evaluation covers two workflows in a synthetic workplace under a controlled semantic-browser interface, with fixed, counterbalanced instance orders and one sampled rollout per condition. These settings do not establish performance in unrestricted native computer use or naturally occurring change-time distributions. The principal contrasts manipulate review-time information while retaining the same authoritative mutation in both paired conditions; they do not estimate the causal effect of environmental change itself. Small paired samples and potential order effects further limit generalization, and non-rejection does not establish equivalence. We evaluate agents but do not train them; whether certified non-stationary environments improve RL-trained computer-use agents is untested.

Evidence coverage is also bounded. Served pages and delivered frames do not establish agent comprehension. In clinic booking, calendar compatibility is not independently reconstructed from the archived records, and certain provider-side inputs and historical endpoints are incomplete (Appendices~\ref{app:repro} and~\ref{app:claimsspec}). Independently implemented checkers were developed by the same team and rely on shared archived evidence. External archive challenges reveal residual false acceptance and control-rejection errors, with some expected verdicts disputed; the case sets are not population-level error-rate estimates. Supplementary phenomenon-family experiments use different settings and are kept separate from the prospective results (Appendix~\ref{app:supplementary}). \ClmGLimitR{} \OswNoGoR{}

\section{Conclusion}
\label{sec:conclusion}

EnvOS integrates phenomenon-based task construction, executable intervention certification, controlled agent evaluation, and evidence-based auditing into a single experimental workflow. By distinguishing authoritative state transitions from their observation channels and by evaluating episode validity before task success, the framework establishes a more explicit evidential basis for claims about agents operating in changing environments.

In the prospective stale-information study, the underlying mutation was fixed while review-time information varied. Direct presentation of the current clinic location improved booking success relative to a verification reminder and source link under the registered and paper-wide analyses. The insurance results show configuration-dependent behavior: Opus~5 succeeded without re-exposure on the tested instances, while Sonnet~5 benefited at its registered local level but not under the retrospective paper-wide adjustment. These contrasts concern review-time information, not a no-change baseline or general model ranking.

\ConclThree

Archive-based reconstruction and evaluator challenges reveal both useful verification coverage and remaining failure modes. The methodological contribution is therefore a pipeline in which the intervention, pre-run checks, model observations, committed outcome, and evaluator claims can be inspected separately, with residual uncertainty explicitly retained. A natural next step is to use certified EnvOS environments as non-stationary, verifiable-reward environments for reinforcement learning of computer-use agents, with certification and evaluator audits helping to guard the reward against hacking.

\label{endofbody}
\subsection*{Reproducibility statement}
Every quantitative statement in this paper is a macro emitted by a claim ledger (\code{claims.py}) that re-derives it from a hashed artifact and fails on
any missing or drifting input and on any macro used but not defined (Appendix~\ref{app:manifests}); in
Section~\ref{sec:experiments} and the conclusion, every sentence interpreting a prospective block is a template
selected by the ledger (Appendix~\ref{app:readings}), and the abstract and
Section~\ref{sec:intro} state the same conclusions in shortened prose with every number still emitted by the
ledger; \FreezeStatusText.
The figures are generated from the same artifacts. \ReproBCSentence{} It also contains \SuppSummary.
Evidence that cannot be reconstructed, such as slot occupancy at booking time in the historical blocks and the CLI's
unarchived defaults, is listed rather than inferred: Appendix~\ref{app:inventory} states the result of an offline run
of the supplement's reproduction script and what is permanently missing, and points to the supplement's complete
inventory, which ships as a generated provenance file.

\subsection*{Ethics statement}
The testbed is synthetic: every person, appointment, claim, message and record in the office-v1 world is
generated, and no human subjects, personal data or production systems are involved. The evaluated policies are
commercial Claude models accessed through the vendor's command-line interface (the Claude CLI) and, in Block
CLM-BG, Gemini 3.1 Pro Preview accessed through Google's REST API, under their providers' terms; the isolation described in \S\ref{sec:isolation} exists to protect the evaluation, not
to circumvent any provider control. The paper reports a non-replication of an earlier result of our own
(Block~A2); its account ships with the supplementary provenance file described in
Appendix~\ref{sec:nonreplication}, where the earlier evidence is retained rather than withdrawn.

\subsection*{AI use statement}
We disclose LLM involvement against the categories ICLR asks about. \textbf{Used:} method implementation (Claude
Code was used to author task modules, checkers and analysis code, all of which is covered by offline tests,
certification and the independent scorer); research methodology (GPT-6 was consulted on the experimental design,
the pre-registration addenda, the sizing of the blocks and the interpretation templates; every consult, answer and
disposition is archived); evaluation infrastructure (GPT-6, through the Codex CLI, built the second sealed archive
challenge of Section~\ref{sec:audit}, which no Claude agent read before its run; in a web chat, GPT-6 specified the
cases and expected verdicts of the rule-only case-design challenge of Section~\ref{sec:audit}, and a second chat
reviewed the fidelity of the authors' implementation of each case; Appendix~\ref{app:blind}; a Gemini CLI agent
(\texttt{\ExtConstructorModel}) designed the cases and expected verdicts of the externally built challenge and wrote the
generator that instantiated them, Appendix~\ref{app:external}); writing (GPT-6 critiqued
drafts of the abstract and of this manuscript; Claude Code drafted sections from the project's own documents).
\textbf{LLMs as the object of study:} the evaluated policies are LLM-based computer-use agents: Claude models and, in the claims block CLM-BG only,
Gemini 3.1 Pro Preview (\texttt{gemini-3.1-pro-preview}); no GPT-6 model result is reported (Appendix~\ref{app:disclosures}). \textbf{Not used:} no
measurement in this paper was produced by an LLM; every number is derived mechanically from archived artifacts by
the claim ledger, and no experimental outcome was generated, altered or selected by an LLM. The research question,
the decision of what to claim, and every acceptance of a result were the authors'. The authors take full
responsibility for all content, including any error introduced this way.


\appendix


The appendices put the final evidence first: the validity table and the rule reconciliation
(Appendices~\ref{app:validity}--\ref{app:rules}), the certificates, the scorer's coverage and the independent
re-derivations (Appendix~\ref{app:certificates}), then the prospective blocks---their records, the verbatim
interpretation templates, the paper-wide multiplicity family, the decisions of the final night and a diagnostic
invariant-completion fraction of the claims blocks (Appendices~\ref{app:bc}--\ref{app:osworld})---then specification and per-episode
records of the historical, hypothesis-generating blocks; the registration chronology with its errata follows them
(Appendix~\ref{app:precommit}). The frozen pre-registrations, the release manifests and the full claim ledger are
files, not typeset here (Appendices~\ref{app:precommit} and~\ref{app:manifests}).

\clearpage
\section*{Supplementary Evidence and Reproducibility}
The appendix is grouped below by what a reader might want to check. Nothing in the development-history group
carries a model-level claim: those sections record where the finalized protocol came from and what it replaced.

\smallskip
\noindent\begin{tabular}{@{}P{3.4cm}P{9.9cm}@{}}
\toprule
\textbf{To check} & \textbf{See} \\
\midrule
Experimental design & task text and generator (\S\ref{app:task}); the claims task, its instances and interface
census (\S\ref{app:claimsspec}); the cue-versus-value blocks (\S\ref{app:bc}); the claims blocks
(\S\ref{app:claims}); the non-Anthropic configuration follow-up (\S\ref{app:clmg}); the policy interface
compared with other harnesses (\S\ref{app:interface}); the protocol as a minimal abstraction
(\S\ref{app:abstraction}); observability, sensitivity and the within-episode timeline (\S\ref{app:design}) \\
\addlinespace
Reported results & the validity table (\S\ref{app:validity}); the paper-wide multiplicity family
(\S\ref{app:multiplicity}); the pre-scripted interpretation templates (\S\ref{app:readings}); independent-scorer coverage of every model episode
(\S\ref{app:coverage}); invariant completion and exposure attribution (\S\ref{app:osworld}) \\
\addlinespace
Evaluator reliability & certificates, mutation battery, independent scorer and leak audit
(\S\ref{app:certificates}); post-hoc checks against simple raw-log baselines (\S\ref{app:posthoc}); the
rule-only case-design challenge (\S\ref{app:blind}); the externally built challenge (\S\ref{app:external});
the full defect table (\S\ref{app:defects}) \\
\addlinespace
Reproduction & the policy interface and invoker specification (\S\ref{app:repro}); manifests and the claims ledger (\S\ref{app:manifests}); the video index (\S\ref{app:videos}) \\
\addlinespace
Development history & registered versus implemented rules (\S\ref{app:rules}); decisions disclosed late
(\S\ref{app:disclosures}); the historical predecessor blocks (\S\ref{sec:historical}); historical records and
registration provenance (\S\ref{app:percell}); the exploratory corpus (\S\ref{app:corpus}) \\
\bottomrule
\end{tabular}
\smallskip

Extended related work is in \S\ref{app:related}.

\section{Validity table}
\label{app:validity}
Section~2 of the pre-registration, generated from the frozen file. \ValidityNotCounted verdicts are not
counted and are replaced under the declared reserve; \ValidityCounted outcomes are counted. The historical blocks ran
success-first; only such validity failures were replaced there: \HistReplacedN{} historically (A2 \AtwoReplaced, BS
\BlockBSVoided; each independently re-derived as a harness error). Under the supersession family's pre-registration a
contrast whose control cell falls below \PreControlFloor is not headlined (moved from Section~\ref{sec:validity}).
\begin{center}
\small

\begingroup\setlength{\tabcolsep}{4pt}
\begin{tabular}{P{4.4cm}P{3.1cm}P{1.6cm}P{3.2cm}}
\toprule
event & verdict & counted & replaced \\
\midrule
provider/CLI rejects before any action & \texttt{infra\_\allowbreak{}rejected} & no & yes (fixed rule above) \\
action raised / app unreadable & \texttt{infra\_\allowbreak{}failure} & no & yes \\
scorer MONITOR\_ERROR & \texttt{verifier\_\allowbreak{}failure} & no & yes, after the checker is fixed and re-certified \\
arm served $\neq$ rung requested & \texttt{arm\_\allowbreak{}mismatch} & no & yes \\
injector fired outside its anchor & \texttt{invalid\_\allowbreak{}intervention} & no & yes \\
event log fired but the session row carries no $\Delta$S & \texttt{state\_\allowbreak{}delta\_\allowbreak{}lost} & no & yes \\
notice on screen but frame not delivered & \texttt{observation\_\allowbreak{}not\_\allowbreak{}delivered} & no & yes \\
agent reaches the 48-action budget without success & \texttt{budget\_\allowbreak{}limited} & yes (failure) & no \\
agent never opens an available notice (email arms only; n/a here) & \texttt{notice\_\allowbreak{}not\_\allowbreak{}opened} & yes & no \\
control fails for a behavioural reason & \texttt{model\_\allowbreak{}failure} & yes & no \\
success & \texttt{success} & yes & — \\
\bottomrule
\end{tabular}
\endgroup

\end{center}

\section{Registered versus implemented rules}
\label{app:rules}
This section distinguishes the registered rules from the rules implemented during execution, and records how
those differences affect the reported results.

Every validity or analysis rule, where the pre-registration registers it (line of the frozen file), where the
code implements it (file and line; the pre-image where the code later changed), the dated correction that
followed, and which archived values that correction re-derived. The five current model-block logs hold
\ModelAttempts attempts and \ModelCounted counted rollouts, each attempt matched one to one to a run directory
(\ModelRunDirs). Generated by \code{audit/rule\_reconciliation.py}, whose independent re-derivations of the
delivery checkpoints agree with the archived values on every row.
{
\begingroup\setlength{\tabcolsep}{2.5pt}\scriptsize
\begin{longtable}{P{1.65cm}P{1.15cm}P{2.05cm}P{2.75cm}P{2.4cm}P{2.6cm}}
\toprule
rule & blocks & registered & implemented & retrospective correction & archived values re-derived \\
\midrule
\endhead
Trigger anchor & A (superseded) & PRECOMMIT:19 & anchor 'second authenticated GET /\allowbreak{}' (code replaced; archived in superseded log rows) & ADD 3 (PRECOMMIT:203): exposure-conditioned anchor for new blocks; scorer after\_\allowbreak{}login fix (PRECOMMIT:213) & none persisted: archived C4 of superseded S-L2/\allowbreak{}S-C s0-s2 still reads False; 3 untriggered episodes stay uncounted; block superseded \\
Trigger anchor & A2, A2S & PRECOMMIT:206 & superseded\_\allowbreak{}session.py:442 & none & none \\
Trigger anchor & B, BS, BF & PRECOMMIT:274 & superseded\_\allowbreak{}session.py:449 & ADD 5 kept the late fire and redefined the reference policy (PRECOMMIT:280) -- before any rollout & none (prospective) \\
Validity before outcome (ladder order) & all & PRECOMMIT:41 & telemetry.py:32 (doc) vs telemetry.py:232 precedes telemetry.py:239 & ADD 11 discloses success-first order (PRECOMMIT:454); ADD 16 repeats it (PRECOMMIT:653) & no verdict re-derived; this script lists SUCCESS rows that validity-first would void \\
Silent-arm perception branch & B (row 1); BS, BF prospective & PRECOMMIT:42-54 (the \S{}2 table has no silent-arm row) & telemetry.py:253-254 & ADD 10 (PRECOMMIT:401) & changed\_\allowbreak{}FRX-OFF-A\_\allowbreak{}test\_\allowbreak{}id\_\allowbreak{}s20\_\allowbreak{}claude-opus-5: delivery\_\allowbreak{}ambiguous $\rightarrow$ model\_\allowbreak{}failure; pre-fix classifier would have voided 25 silent-arm model failures \\
Notice delivery (C3) & A2, A2S (announced arms) & PRECOMMIT:50 & preimage[2026-09-22T0004]/\allowbreak{}provenance.py:632 (pre-image; frame + ack on classified route) $\rightarrow$ provenance.py:663; verdict telemetry.py:266 & ADD 11 (ack on unclassified route, PRECOMMIT:438); ADD 16 (text or image, PRECOMMIT:619) & ADD 11: 5 A2 rows re-derived (changed\_\allowbreak{}S-L2\_\allowbreak{}s9 opus-5, changed\_\allowbreak{}S-L2\_\allowbreak{}s11 opus-5, cosmetic\_\allowbreak{}S-C\_\allowbreak{}s13 opus-5, cosmetic\_\allowbreak{}S-C\_\allowbreak{}s14 opus-5, cosmetic\_\allowbreak{}S-C\_\allowbreak{}s15 opus-5): C3 false$\rightarrow$true on 4, 4 signature changes; ADD 16: C3 false$\rightarrow$true on 2 rows (cosmetic\_\allowbreak{}S-C\_\allowbreak{}s12 opus-5, cosmetic\_\allowbreak{}S-C\_\allowbreak{}s15 opus-5); 0 verdict changes \\
Re-exposure delivery of review-time information & B, BS, BF (FRX-ON-*) & PRECOMMIT:305-308 & preimage[2026-09-22T1437]/\allowbreak{}superseded\_\allowbreak{}session.py:1934 (pre-image: screenshot required, checkpoint + signature only); never a verdict: telemetry.py:254 reads C3 only & ADD 16 redefines delivery as text or image after BS outcomes (PRECOMMIT:619); ADD 17 defect B discloses non-enforcement (PRECOMMIT:733) & ADD 16: review display checkpoint false$\rightarrow$true on 5 rows (changed\_\allowbreak{}FRX-ON-A\_\allowbreak{}s22 sonnet-5, changed\_\allowbreak{}FRX-ON-A\_\allowbreak{}s27 sonnet-5, changed\_\allowbreak{}FRX-ON-A\_\allowbreak{}s28 sonnet-5, cosmetic\_\allowbreak{}FRX-ON-I\_\allowbreak{}s21 sonnet-5, cosmetic\_\allowbreak{}FRX-ON-I\_\allowbreak{}s22 sonnet-5); signatures of BS ON-A s27, s28 STALE\_\allowbreak{}APPLIED\_\allowbreak{}UNFLAGGED $\rightarrow$ STALE\_\allowbreak{}AFTER\_\allowbreak{}REEXPOSURE; 0 verdict changes; ADD 17: none \\
Read-target (leak) rule & all (voids: A2 1, BS 3) & not in the \S{}2 table; harness rule (PRECOMMIT:508) & web\_\allowbreak{}rollout.py:910; override web\_\allowbreak{}rollout.py:1942 & ADD 12 (PRECOMMIT:495): errored reads and /\allowbreak{}dev/\allowbreak{}null no longer void & none: the 4 voided attempts stay uncounted (harness stopped them); cells re-run as replacements \\
Action budget & all & PRECOMMIT:51 & telemetry.py:218 & none & none \\
Provider limit & A (superseded), all later & PRECOMMIT:149 & telemetry.py:157 & none (ADD 2 predates every counted block) & none \\
Registered claim rule (all kept failures STALE\_\allowbreak{}APPLIED\_\allowbreak{}UNFLAGGED) & B, BS, BF, A2 & PRECOMMIT:89-90 & superseded\_\allowbreak{}session.py:1032; superseded\_\allowbreak{}session.py:2200 & ADD 17 defect A discloses the distractor-clinic flag (PRECOMMIT:705) & none: labels kept; \S{}6 rule NOT met for Block B (3 FLAGGED OFF-A rows) \\
Replacement reserve /\allowbreak{} stop rules & all & PRECOMMIT:35-39 & run\_\allowbreak{}block.py:97 & ADD 12 resumed BS after one repair cycle as registered & none \\
Primary contrast & A2 /\allowbreak{} B /\allowbreak{} BS; A2S, BF descriptive & PRECOMMIT:26; B PRECOMMIT:287; BS PRECOMMIT:349; A2S/\allowbreak{}BF PRECOMMIT:537 & stats.py:43 & ADD 13 withdrew A2S's test and downsized A2S/\allowbreak{}BF before their rollouts & none \\
\bottomrule
\end{longtable}
\endgroup
}

\paragraph{What the corrections changed.} ADDENDUM~10 re-derived one Block~B row; the classifier before it
would have voided \CorrTenCounterfactual model failures of the silent blocks, those of the re-exposure arm
included (every arm without a notice, the sense of ``silent-arm'' in the rule matrix above). ADDENDUM~11 re-derived \CorrElevenRows
Block~A2 rows and changed \CorrElevenSig signatures; ADDENDUM~16's delivery redefinition flipped
\CorrSixteenFlips checkpoints and changed \CorrSixteenSig signatures; neither changed a verdict. The
implemented classifier tested success before delivery, so a success was never checked for validity; that later
corrections left the verdicts unchanged shows what re-ordering changes on these archives, not that the success-first
procedure was valid. The
validity-first analysis covers \ValidityFirstOf success rows: the \VFirstCurrent counted successes of the five
current blocks and the \VFirstSuperseded counted successes of the superseded Block~A restart (ADDENDUM~3), which
\code{rule\_reconciliation.py} examines alongside them. Of these, \ValidityFirstVoid would have been voided under a
validity-first order with some archived version of the delivery checkpoint: \VFirstVoidCurrent current rows
(\VFirstVoidList) and \VFirstVoidSuperseded of the superseded block.

\paragraph{A manifest rebuilt mid-block.} A dated amendment can rebuild a block's manifest mid-block
(ADDENDA~\BSAmendAddenda: \BSPreAmendRows Block~BS attempts, \BSPreAmendCounted counted, ran under the manifest it
replaced), and a digest identifies an artifact without preserving it: that earlier manifest survives by digest
only, and Block~A2's task module is absent from the working tree (where it survives: Appendix~\ref{app:inventory}). Sonnet's review-information delivery rule: in each of its
\SonnetPanelTextFail stale FRX-ON-A failures the review information reached it as page text but it did not open the screenshot at
that step---undelivered under the pre-registration, a rule never enforced and redefined after these outcomes---while
in all \SonnetPanelShotSuccess rescues it opened the screenshot. The rule as registered voids those rows and the
successful FRX-ON-I controls (\VindRegDiffers rows under its literal reading, re-derived independently), beyond the
reserve; left unobserved, FRX-ON-A is \OrigBSFRXOnAFrac ($p=\OrigBSP$). Opus opened the screenshot at every review display
step, so its result is the same under either rule. The reported \BlockBSFRXOnAPassFrac, the narrower rule's and the
literal rule's figures (Table~\ref{tab:original-rules}) are different selected denominators; that they share a
paired $p$-value neither repairs the eligibility failure nor supplies the unobserved replacement outcomes.
\emph{Behaviour after assignment.} Opening the screenshot and reaching the review step are the agent's own behaviour,
not exogenous delivery events, so a rule that voids episodes on them can remove the very trajectories the review display is
meant to affect; the literal rule voids \VindRegNever episodes that never reached the review step, while an
action-cap termination elsewhere counts as a failure. Neither rule voids any Block~B row.

\paragraph{Amendments.} A design change is a dated addendum preceding the rollouts it governs; a defect found after
outcomes is a dated addendum that re-derives the archived values, keeping each old value beside the new (\PreAddenda
addenda to the supersession family's registration; chronology: Appendix~\ref{app:precommit}). Earlier blocks ran
under predecessor versions of the protocol; this appendix reconciles them (moved from Section~\ref{sec:validity}).

\paragraph{The frame-read rule as implemented (moved from Section~\ref{sec:isolation}).}
\label{app:readrule}
Frame delivery is admitted per channel: a model that reads frames through a file-read tool must read the frame it was
handed. As implemented, a successful read of a path the harness had not yet handed over voids the episode (before
ADDENDUM~12, an errored read too); re-reading an earlier handed frame is admitted, a reading chosen after seeing
archive data on which \VindPriorFrames counted Sonnet rows depend (Appendix~\ref{app:coverage}). For the other tool
calls a model emits (Sonnet~5: \NonReadCallsSonnet) the invoker archives only the name, so their containment rests on
the account's permissions; their availability was identical across paired arms (Section~\ref{sec:limitations}).

\paragraph{Common protocol requirements, workflow-specific rules and historical exceptions.}
Table~\ref{tab:protocol-rules} separates what every prospective block shares from what each workflow registers and
from how the historical blocks differed; each cell cites where the rule is registered or reconciled.
\begin{table}[h]
\centering
\caption{\textbf{The protocol as operated.} Common: every prospective block. Claims: CLM-B and CLM-BS
(PRECOMMIT\_CLAIMS). Clinic: BC and BCS (ADDENDA~18--19). Historical: Blocks A2, A2S, B, BS and BF (predecessor
versions; Appendix~\ref{sec:historical}).}
\label{tab:protocol-rules}
\begingroup\scriptsize\setlength{\tabcolsep}{3pt}
\begin{tabular}{@{}P{1.6cm}P{3.0cm}P{2.55cm}P{2.3cm}P{3.35cm}@{}}
\toprule
rule & common & claims & clinic & historical \\
\midrule
trigger & fires on the first eligible request after the old value was served (exposure-conditioned) & first form entry after the policy records page & first booking-wizard request after the sessions page & exposure-conditioned from Block~A2 on (ADDENDUM~3) \\
untriggered episode & registered per workflow & counted as a failure, reported separately & not counted; replaced & not counted; replaced \\
counting & validity first; only uncounted validity failures replaced, under a reserve (\ReserveCell{} per cell, \ReserveBlock{} per block) & the claims validity-first table; the generic classifier verdict is never read & the ADDENDUM~18 validity table & success first as implemented; the validity-first re-analysis is Table~\ref{tab:original-rules} \\
frame delivery & admitted per channel & file reads & file reads (an inline channel is defined for GPT-6 blocks) & read rule as implemented, one reading post hoc (above); ADDENDUM~5's review-screenshot rule not enforced \\
endpoint & invariants on world state and the agent's report & four (no step after the commit) & five & five, evaluator-scored; the nearest-slot term cannot be independently verified \\
independent audits & outcome scorer and standard-library validity re-derivation, claimed only once the artifact exists & both present & claimed when produced & archive re-analyses (Appendix~\ref{app:coverage}) \\
multiplicity & the registered local test & one contrast; the paper-wide family is a retrospective sensitivity analysis & block-local Holm over two contrasts, frozen before any outcome & retrospective slots in the paper-wide family \\
interpretation & sealed templates, selected mechanically (Appendix~\ref{app:readings}) & CLM templates & BC templates & frozen report rules; one post-outcome sentence (ADDENDUM~17) is a departure (Appendix~\ref{app:blockB}) \\
\bottomrule
\end{tabular}
\endgroup
\end{table}

\paragraph{Results under the original rules.} Table~\ref{tab:original-rules} recomputes every block under the
rules as first registered. A row an original rule voids is shown as unobserved; no replacement outcome is
imputed. Under ADDENDUM~5's screenshot rule, Block~BS would have had \BSInvalidUnderRule invalid attempts
against a block reserve of \BSReserveTotal, \BSReserveUsed of it already used, so at most \BSReserveLeft of
the new voids could have been replaced. Block~B is identical under every reading.
\begin{table}[h]
\centering
\caption{\textbf{Reported results and results under the original rules.} ``Validity first'' applies the
delivery rule before success, with the notice checkpoint as first archived, after ADDENDUM~11, and after
ADDENDUM~16. for/against/tied are paired discordances; unobserved rows are those the rule voids.}
\label{tab:original-rules}

\begingroup\setlength{\tabcolsep}{3pt}\scriptsize
\begin{tabular}{P{0.6cm}P{3.2cm}P{1.1cm}P{1.2cm}P{1.2cm}P{1.9cm}P{2.9cm}}
\toprule
block & analysis & control & treatment & ON-I / S-L0 & for/against/tied; $p$ & unobserved rows \\
\midrule
\textbf{B} & reported (implemented rules) & 0/10 & 10/10 & 3/3 & 10/0/0; 0.00195 & none \\
 & ADD.~5 screenshot rule as registered & 0/10 & 10/10 & 3/3 & 10/0/0; 0.00195 & none \\
 & ADD.~5, literal (also voids review information never rendered) & 0/10 & 10/10 & 3/3 & 10/0/0; 0.00195 & none \\
\midrule
\textbf{BS} & reported (implemented rules) & 0/10 & 6/10 & 2/3 & 6/0/4; 0.0312 & none \\
 & ADD.~5 screenshot rule as registered & 0/10 & 6/7 & 0/1 & 6/0/1; 0.0312 & FRX-ON-A s22, FRX-ON-A s27, FRX-ON-A s28, FRX-ON-I s21, FRX-ON-I s22 \\
 & ADD.~5, literal (also voids review information never rendered) & 0/10 & 6/6 & 0/0 & 6/0/0; 0.0312 & FRX-ON-A s22, FRX-ON-A s23, FRX-ON-A s27, FRX-ON-A s28, FRX-ON-I s20, FRX-ON-I s21, FRX-ON-I s22 \\
\midrule
\textbf{BF} & reported (implemented rules) & 0/4 & -- & 1/1 & -- & none \\
 & ADD.~5 screenshot rule as registered & 0/4 & -- & 1/1 & -- & none \\
 & ADD.~5, literal (also voids review information never rendered) & 0/4 & -- & 1/1 & -- & none \\
\midrule
\textbf{A2} & reported (implemented rules) & 8/8 & 7/8 & 2/2 & 1/0/7; 1 & none \\
 & validity first, C3 as first archived & 4/4 & 5/6 & 2/2 & 0/0/2; -- & S-C s12, S-C s13, S-C s14, S-C s15, S-L2 s9, S-L2 s11 \\
 & validity first, C3 after ADD.~11 & 6/6 & 7/8 & 2/2 & 1/0/5; 1 & S-C s12, S-C s15 \\
 & validity first, C3 after ADD.~16 & 8/8 & 7/8 & 2/2 & 1/0/7; 1 & none \\
\midrule
\textbf{A2S} & reported (implemented rules) & 2/4 & 4/4 & 1/1 & 0/2/2; 0.5 & none \\
 & validity first, C3 as first archived & 2/4 & 4/4 & 1/1 & 0/2/2; 0.5 & none \\
 & validity first, C3 after ADD.~11 & 2/4 & 4/4 & 1/1 & 0/2/2; 0.5 & none \\
 & validity first, C3 after ADD.~16 & 2/4 & 4/4 & 1/1 & 0/2/2; 0.5 & none \\
\bottomrule
\end{tabular}
\endgroup

\end{table}

\section{Certificates, mutation battery, independent scorer, leak audit}
\label{app:certificates}
\paragraph{Certificates.} Every certificate file of the \CertN conditions, with the blocks it governed (the
latest certificate of that level issued before the block's first rollout). Block~A2 ran under the original
S-* certificates, preserved in \code{P3v2/\_pre\_recert\_2026-09-22/}; they were re-issued under the module
edited for Block~B, and the re-issue governed Block~A2S. FRX-OFF-I was certified but no model block
ran it. Every check is tri-state; every certificate listed has zero failed checks and zero monitor errors;
each sha256 prefix is recorded in the addendum that admitted it, except the \CertGovUnrecorded re-issued
certificates, which no addendum records. Across the \CertGovN governing certificates the checks range
\CertGovChecksMin--\CertGovChecksMax (Block~A2's: \CertGovAtwoMin--\CertGovAtwoMax).
\begin{center}
\small

\begingroup\setlength{\tabcolsep}{3pt}\scriptsize
\begin{tabular}{lrrrllll}
\toprule
condition & checks & failed & mon.\ err. & issued (UTC) & sha256 & governed & sha in addendum \\
\midrule
S-L0 & 30 & 0 & 0 & 2026-09-18 04:30:10 & \texttt{e7f675db} & A2 & yes \\
S-L0 & 32 & 0 & 0 & 2026-09-22 10:01:50 & \texttt{e8894c3d} & A2S & no \\
S-C & 38 & 0 & 0 & 2026-09-18 04:46:55 & \texttt{5015a794} & A2 & yes \\
S-C & 40 & 0 & 0 & 2026-09-22 10:22:01 & \texttt{91f23bfe} & A2S & no \\
S-L2 & 48 & 0 & 0 & 2026-09-18 05:06:37 & \texttt{760de9ee} & A2 & yes \\
S-L2 & 50 & 0 & 0 & 2026-09-22 10:47:31 & \texttt{0dbc0b4e} & A2S & no \\
FRX-OFF-A & 40 & 0 & 0 & 2026-09-18 13:44:46 & \texttt{79e02eba} & B, BS, BF & yes \\
FRX-ON-A & 41 & 0 & 0 & 2026-09-18 14:05:27 & \texttt{2b643a35} & B, BS & yes \\
FRX-OFF-I & 37 & 0 & 0 & 2026-09-18 13:05:26 & \texttt{38be89f1} & none (certified, never run) & yes \\
FRX-ON-I & 37 & 0 & 0 & 2026-09-18 13:25:13 & \texttt{c304781a} & B, BS, BF & yes \\
\bottomrule
\end{tabular}
\endgroup

\end{center}
The \CertDistinctChecks distinct check names across the seven current certificate files:
{\small 
\begingroup\raggedright\sloppy
\texttt{aux\_\allowbreak{}rows\_\allowbreak{}restored\_\allowbreak{}on\_\allowbreak{}reset}, \texttt{boundary\_\allowbreak{}outcomes\_\allowbreak{}are\_\allowbreak{}the\_\allowbreak{}pre\_\allowbreak{}registered\_\allowbreak{}ones}, \texttt{clean\_\allowbreak{}control\_\allowbreak{}arms\_\allowbreak{}nothing}, \texttt{concurrent\_\allowbreak{}writer\_\allowbreak{}refused}, \texttt{dismissal\_\allowbreak{}reaches\_\allowbreak{}the\_\allowbreak{}access\_\allowbreak{}log\_\allowbreak{}ack\_\allowbreak{}field}, \texttt{early\_\allowbreak{}done\_\allowbreak{}does\_\allowbreak{}not\_\allowbreak{}suppress\_\allowbreak{}queued\_\allowbreak{}event}, \texttt{exposure\_\allowbreak{}conditioned\_\allowbreak{}arm\_\allowbreak{}waits\_\allowbreak{}for\_\allowbreak{}the\_\allowbreak{}exposure}, \texttt{filler\_\allowbreak{}facts\_\allowbreak{}identical\_\allowbreak{}across\_\allowbreak{}arms\_\allowbreak{}and\_\allowbreak{}clinic\_\allowbreak{}neutral}, \texttt{harness\_\allowbreak{}text\_\allowbreak{}passes\_\allowbreak{}the\_\allowbreak{}banned\_\allowbreak{}term\_\allowbreak{}scan}, \texttt{hidden\_\allowbreak{}state\_\allowbreak{}not\_\allowbreak{}policy\_\allowbreak{}reachable}, \texttt{instruction\_\allowbreak{}names\_\allowbreak{}no\_\allowbreak{}clinic}, \texttt{matrix\_\allowbreak{}collateral\_\allowbreak{}seed\_\allowbreak{}change}, \texttt{matrix\_\allowbreak{}covers\_\allowbreak{}at\_\allowbreak{}least\_\allowbreak{}three\_\allowbreak{}policies}, \texttt{matrix\_\allowbreak{}flagged\_\allowbreak{}stale}, \texttt{matrix\_\allowbreak{}late\_\allowbreak{}reread\_\allowbreak{}reference}, \texttt{matrix\_\allowbreak{}modal\_\allowbreak{}bypass}, \texttt{matrix\_\allowbreak{}null}, \texttt{matrix\_\allowbreak{}over\_\allowbreak{}apply}, \texttt{matrix\_\allowbreak{}path\_\allowbreak{}only\_\allowbreak{}reviser}, \texttt{matrix\_\allowbreak{}reference}, \texttt{matrix\_\allowbreak{}stale\_\allowbreak{}even\_\allowbreak{}when\_\allowbreak{}reexposed}, \texttt{matrix\_\allowbreak{}stale\_\allowbreak{}location}, \texttt{modal\_\allowbreak{}bypass\_\allowbreak{}fails\_\allowbreak{}to\_\allowbreak{}submit\_\allowbreak{}or\_\allowbreak{}classifies\_\allowbreak{}delivery\_\allowbreak{}ambiguous}, \texttt{near\_\allowbreak{}miss\_\allowbreak{}is\_\allowbreak{}not\_\allowbreak{}a\_\allowbreak{}boundary}, \texttt{no\_\allowbreak{}evaluator\_\allowbreak{}vocabulary\_\allowbreak{}on\_\allowbreak{}a\_\allowbreak{}reachable\_\allowbreak{}page}, \texttt{notice\_\allowbreak{}bytes\_\allowbreak{}derived\_\allowbreak{}from\_\allowbreak{}the\_\allowbreak{}mutation}, \texttt{notice\_\allowbreak{}renders\_\allowbreak{}until\_\allowbreak{}dismissed}, \texttt{one\_\allowbreak{}shot\_\allowbreak{}trigger}, \texttt{replay\_\allowbreak{}determinism\_\allowbreak{}reference}, \texttt{replay\_\allowbreak{}determinism\_\allowbreak{}stale\_\allowbreak{}location}, \texttt{reply\_\allowbreak{}format\_\allowbreak{}is\_\allowbreak{}the\_\allowbreak{}p1\_\allowbreak{}harnesss}, \texttt{report\_\allowbreak{}clinic\_\allowbreak{}term\_\allowbreak{}is\_\allowbreak{}scored\_\allowbreak{}by\_\allowbreak{}I5}, \texttt{report\_\allowbreak{}flag\_\allowbreak{}fixtures}, \texttt{rescue\_\allowbreak{}crossing}, \texttt{result\_\allowbreak{}line\_\allowbreak{}parser\_\allowbreak{}fixtures}, \texttt{service\_\allowbreak{}date\_\allowbreak{}identifies\_\allowbreak{}exactly\_\allowbreak{}one\_\allowbreak{}session}, \texttt{session\_\allowbreak{}page\_\allowbreak{}reader\_\allowbreak{}fixtures}, \texttt{session\_\allowbreak{}page\_\allowbreak{}shows\_\allowbreak{}the\_\allowbreak{}target\_\allowbreak{}row}, \texttt{signature\_\allowbreak{}collateral\_\allowbreak{}seed\_\allowbreak{}change}, \texttt{signature\_\allowbreak{}flagged\_\allowbreak{}stale}, \texttt{signature\_\allowbreak{}late\_\allowbreak{}reread\_\allowbreak{}reference}, \texttt{signature\_\allowbreak{}modal\_\allowbreak{}bypass}, \texttt{signature\_\allowbreak{}null}, \texttt{signature\_\allowbreak{}over\_\allowbreak{}apply}, \texttt{signature\_\allowbreak{}path\_\allowbreak{}only\_\allowbreak{}reviser}, \texttt{signature\_\allowbreak{}reference}, \texttt{signature\_\allowbreak{}stale\_\allowbreak{}even\_\allowbreak{}when\_\allowbreak{}reexposed}, \texttt{signature\_\allowbreak{}stale\_\allowbreak{}location}, \texttt{silent\_\allowbreak{}arm\_\allowbreak{}posts\_\allowbreak{}no\_\allowbreak{}notice}, \texttt{state\_\allowbreak{}before\_\allowbreak{}channel\_\allowbreak{}ordering}, \texttt{the\_\allowbreak{}claim\_\allowbreak{}is\_\allowbreak{}not\_\allowbreak{}a\_\allowbreak{}boundary}, \texttt{the\_\allowbreak{}firing\_\allowbreak{}request\_\allowbreak{}is\_\allowbreak{}located\_\allowbreak{}correctly}, \texttt{the\_\allowbreak{}policies\_\allowbreak{}that\_\allowbreak{}looked\_\allowbreak{}reached\_\allowbreak{}c3}, \texttt{the\_\allowbreak{}world\_\allowbreak{}fires\_\allowbreak{}the\_\allowbreak{}arm\_\allowbreak{}inside\_\allowbreak{}the\_\allowbreak{}agents\_\allowbreak{}own\_\allowbreak{}request}, \texttt{two\_\allowbreak{}sided\_\allowbreak{}over\_\allowbreak{}apply\_\allowbreak{}FRX-\allowbreak{}OFF-\allowbreak{}I\_\allowbreak{}vs\_\allowbreak{}FRX-\allowbreak{}OFF-\allowbreak{}A}, \texttt{two\_\allowbreak{}sided\_\allowbreak{}over\_\allowbreak{}apply\_\allowbreak{}FRX-\allowbreak{}ON-\allowbreak{}I\_\allowbreak{}vs\_\allowbreak{}FRX-\allowbreak{}ON-\allowbreak{}A}, \texttt{two\_\allowbreak{}sided\_\allowbreak{}over\_\allowbreak{}apply\_\allowbreak{}S-\allowbreak{}C\_\allowbreak{}vs\_\allowbreak{}S-\allowbreak{}L2}, \texttt{two\_\allowbreak{}sided\_\allowbreak{}reference\_\allowbreak{}FRX-\allowbreak{}OFF-\allowbreak{}I\_\allowbreak{}vs\_\allowbreak{}FRX-\allowbreak{}OFF-\allowbreak{}A}, \texttt{two\_\allowbreak{}sided\_\allowbreak{}reference\_\allowbreak{}FRX-\allowbreak{}ON-\allowbreak{}I\_\allowbreak{}vs\_\allowbreak{}FRX-\allowbreak{}ON-\allowbreak{}A}, \texttt{two\_\allowbreak{}sided\_\allowbreak{}reference\_\allowbreak{}S-\allowbreak{}C\_\allowbreak{}vs\_\allowbreak{}S-\allowbreak{}L2}, \texttt{two\_\allowbreak{}sided\_\allowbreak{}stale\_\allowbreak{}location\_\allowbreak{}FRX-\allowbreak{}OFF-\allowbreak{}I\_\allowbreak{}vs\_\allowbreak{}FRX-\allowbreak{}OFF-\allowbreak{}A}, \texttt{two\_\allowbreak{}sided\_\allowbreak{}stale\_\allowbreak{}location\_\allowbreak{}FRX-\allowbreak{}ON-\allowbreak{}I\_\allowbreak{}vs\_\allowbreak{}FRX-\allowbreak{}ON-\allowbreak{}A}, \texttt{two\_\allowbreak{}sided\_\allowbreak{}stale\_\allowbreak{}location\_\allowbreak{}S-\allowbreak{}C\_\allowbreak{}vs\_\allowbreak{}S-\allowbreak{}L2}, \texttt{v1\_\allowbreak{}was\_\allowbreak{}in\_\allowbreak{}a\_\allowbreak{}delivered\_\allowbreak{}observation}, \texttt{wizard\_\allowbreak{}lists\_\allowbreak{}both\_\allowbreak{}clinics}\par\endgroup
}

\paragraph{Feasibility.} On each block's own generated seeds, before any model ran: \FeasAtwoOk of \FeasAtwoN
scripted episodes as expected for Block~A2 and \FeasBOk of \FeasBN for Block~B. The \CertGovN certificates that
governed the historical model blocks carry \CertGovChecksMin--\CertGovChecksMax checks each, with zero failures.

\paragraph{Mutation battery.} \MutationsDev development mutations: \MutationsDevControls positive controls
the checker must accept and \MutationsDevCorruptions corruptions it must reject, all \MutationDevChecks
expectations met with zero monitor errors.

\paragraph{Held-out mutation challenge (sealed, not blind).} \MutationsHeldout held-out cases with frozen expectations, sealed by
sha256 (\MutationsHeldoutSha, re-verified before the run), were run once, on \HeldoutRunDate, after every model
block, against the checker as it then stood; the digest of every checker file is archived with the run. The
set was sealed against change but \emph{not} hidden: the held-out file had been read before the run, and
\code{audit/mutation\_battery.py}, last modified before the run, carries a case-specific adjudication that
labels one outcome below ``ADJUDICATED TEST DEFECT''. The result is therefore not blind evidence, and that
adjudication is \emph{not} used: the battery's own pass/fail is. \HeldoutMet of \HeldoutN exact frozen
expectations were met (Table~\ref{tab:heldout}). Of the \HeldoutCorrupt corruptions, \HeldoutCorruptInstantiated were
instantiated and \HeldoutCorruptRejected of those rejected; of the \HeldoutValid valid (invariant-preserving) cases,
\HeldoutValidInstantiated were instantiated and \HeldoutValidAccepted of those accepted, and
\HeldoutValidNotInstantiated could not be instantiated. The \HeldoutMissed misses, as the run's summary words them:
(1)~\code{heldout\_stale\_location\_via\_history\_actor}: ``the corruption (committed clinic rewritten to the
superseded one, plus a plausible `relocate' history row) WAS rejected: target\_state\_correct failed as
expected. final\_report\_correct ALSO failed, because the agent's report now disagrees with the rewritten record
--- correct checker behaviour that the frozen expectation omitted.'' It is counted as a miss against the
frozen expectation. (2)~\code{heldout\_notice\_file\_rewritten\_after\_commit}: ``could NOT be instantiated: the
mutation harness formats the orch payload with str.format and the JSON braces raise `ValueError: Invalid format
specifier' $\rightarrow$ MONITOR\_ERROR, which the tri-state protocol counts as not passing. A harness
(templating) defect, not a checker verdict. Not re-run: a repaired re-run would not be held-out evidence.''
This is one fixed challenge of \HeldoutN cases on one family; it does not establish general verifier
soundness. Of its two misses, one is additional correct checker behaviour and the other a mutation-harness failure,
and neither makes the run blind evidence. The development mutations are selected examples, and the \NDefectsTotal
defects we found (Appendix~\ref{app:defects}) give no denominator for defects missed on paths no scripted policy
takes; no blind estimate of detection or false-rejection \emph{rates} exists in this paper (the sealed challenge
below counts misses on constructed cases).
\begin{table}[h]
\centering
\caption{\textbf{The held-out mutation challenge, run once.} Expected: the invariants the frozen expectation
says must fail (I1 target state correct, I2 exactly one commit, I3 no collateral change, I4 workflow
continued from the commit, I5 final report correct); ``none'' marks a valid case the checker must accept.}
\label{tab:heldout}

\begingroup\setlength{\tabcolsep}{3pt}\scriptsize
\begin{tabular}{P{5.4cm}lP{2.2cm}P{2.2cm}l}
\toprule
held-out case & kind & expected to fail & observed to fail & met \\
\midrule
\texttt{valid\_\allowbreak{}reference\_\allowbreak{}passes} & noop & none & none & yes \\
\texttt{stale\_\allowbreak{}location\_\allowbreak{}via\_\allowbreak{}history\_\allowbreak{}actor} & sql & I1 & I1, I5 & \textbf{no} \\
\texttt{second\_\allowbreak{}active\_\allowbreak{}row\_\allowbreak{}same\_\allowbreak{}label} & sql & I2 & I2 & yes \\
\texttt{collateral\_\allowbreak{}slot\_\allowbreak{}shift\_\allowbreak{}on\_\allowbreak{}bystander} & sql & I3 & I3 & yes \\
\texttt{cancelled\_\allowbreak{}then\_\allowbreak{}reconfirmed\_\allowbreak{}row} & sql & I1, I2, I4, I5 & I1, I2, I4, I5 & yes \\
\texttt{report\_\allowbreak{}case\_\allowbreak{}and\_\allowbreak{}whitespace\_\allowbreak{}variant\_\allowbreak{}is\_\allowbreak{}still\_\allowbreak{}correct} & report & none & none & yes \\
\texttt{report\_\allowbreak{}names\_\allowbreak{}old\_\allowbreak{}clinic\_\allowbreak{}in\_\allowbreak{}note\_\allowbreak{}only} & report & none & none & yes \\
\texttt{report\_\allowbreak{}swapped\_\allowbreak{}fields} & report & I5 & I5 & yes \\
\texttt{session\_\allowbreak{}version\_\allowbreak{}bumped\_\allowbreak{}without\_\allowbreak{}move} & sql & none & none & yes \\
\texttt{notice\_\allowbreak{}file\_\allowbreak{}rewritten\_\allowbreak{}after\_\allowbreak{}commit} & orch & none & monitor error (not instantiated) & \textbf{no} \\
\bottomrule
\end{tabular}
\endgroup

\end{table}

\begin{table}[h]
\centering
\caption{\textbf{Auditing layers against the sealed archive challenge} (first and only run; expectations sealed by
digest before it). Counts are frozen expectations met; the per-case table is Table~\ref{tab:challenge}.}
\label{tab:layers}

\begingroup\scriptsize\setlength{\tabcolsep}{3pt}
\begin{tabular}{@{}P{2.5cm}P{1.05cm}P{4.1cm}P{5.55cm}@{}}
\toprule
layer & sealed challenge & what it establishes & known gap \\
\midrule
independent outcome scorer & \ChallengeScorerMet/\ChallengeN & every frozen expectation met on this suite & a term it cannot resolve is reported unresolved; clinic calendar compatibility is assumed; no historical success is verified in full \\
validity re-derivation (standard library) & \ChallengeVindMet/\ChallengeN & accepted \ChallengeVindValidMet/\ChallengeValid{} valid controls and rejected \ChallengeVindCorruptMet/\ChallengeCorrupt{} corruptions on this suite & blind spots: \ChallengeVindGaps \\
evaluator, archived verdict path & \ChallengeEvalMet/\ChallengeN & met \ChallengeEvalValidMet/\ChallengeValid{} valid controls but rejected \ChallengeEvalCorruptMet/\ChallengeCorrupt{} corruptions because archived validity/invariant claims were retained & re-derives the delivery checkpoints from raw logs but keeps the archived run-validity and invariant claims \\
\bottomrule
\end{tabular}
\endgroup

\end{table}

\paragraph{Second, sealed archive challenge (unseen by the pipeline's Claude authors).} \ChallengeClause{}
\ChallengeLedgerClause{} \ChallengeAppendixText
\begin{table}[h]
\centering
\caption{\textbf{The sealed archive challenge, first and only run, per tool and case class.} ``met'': every field
the frozen expectation assesses for that tool was met; ``no'' names the fields that were not (for the validity
re-derivation, \emph{validity} is its verdict under the rules as implemented and \emph{validity (registered)} under
the rules as registered). The outcome scorer assesses I1--I5 only; the validity re-derivation assesses validity only;
the evaluator's invariants are its archived claims. From \code{CHALLENGE\_RESULT.json}, pinned by sha256 in the ledger.}
\label{tab:challenge}

\begingroup\setlength{\tabcolsep}{3pt}\scriptsize
\begin{tabular}{P{5.2cm}rP{2.3cm}P{2.3cm}P{2.3cm}}
\toprule
case class & cases & outcome scorer & validity re-derivation & evaluator (replay) \\
\midrule
valid controls & 10 & 10 of 10 & 10 of 10 & 10 of 10 \\
\texttt{C01} success-before-validity: notice & 1 & met & \textbf{no}: validity & \textbf{no}: validity \\
\texttt{C02} missing review-time information delivery & 1 & met & \textbf{no}: validity & \textbf{no}: validity \\
\texttt{C03} change before old value served & 1 & met & met & \textbf{no}: validity \\
\texttt{C04} change fired twice & 1 & met & met & \textbf{no}: validity \\
\texttt{C05} change in nonagent request & 1 & met & met & \textbf{no}: validity \\
\texttt{C06} notice on silent level & 1 & met & met & \textbf{no}: validity \\
\texttt{C07} review information shown on OFF level & 1 & met & met & \textbf{no}: validity \\
\texttt{C08} commit before change & 1 & met & \textbf{no}: validity, validity (registered) & \textbf{no}: validity \\
\texttt{C09} duplicate commit & 1 & met & met & \textbf{no}: I2 \\
\texttt{C10} collateral row change & 1 & met & met & \textbf{no}: I3 \\
\texttt{C11} report disagrees with committed record & 1 & met & met & \textbf{no}: I5 \\
\texttt{C12} stale clinic with forged history & 1 & met & met & \textbf{no}: I1 \\
\texttt{C13} incomplete slot evidence & 1 & met & met & \textbf{no}: I1 \\
\texttt{C14} forged complete slot evidence & 1 & met & met & \textbf{no}: I1 \\
\texttt{C15} truncated trajectory & 1 & met & \textbf{no}: validity (registered) & \textbf{no}: validity, I1, I2, I4, I5 \\
\texttt{C16} event/access firing request disagreement & 1 & met & met & \textbf{no}: validity \\
\texttt{C17} event/page state delta disagreement & 1 & met & met & \textbf{no}: validity, I1 \\
\texttt{C18} success-before-validity: exposure & 1 & met & met & \textbf{no}: validity \\
\midrule
all cases & 28 & 28 met & 24 met & 10 met \\
\bottomrule
\end{tabular}
\endgroup

\end{table}

\paragraph{Independent scorer.} \ScorerRunsAudited certification episodes re-scored from archived
state on \ScorerAuditDate by an implementation with \ScorerEvaluatorImports imports from the evaluator
(counted on the released scorer source, digest \ScorerSourceSha; the audit record names the scorer by
path and carries no digest of it): \ScorerContradictions contradictions, \ScorerBehaviour behaviour
notes, and \ScorerUnresolved unresolved terms on \ScorerRunsWithUnresolved episodes of two deficient
scripted policies, \code{null} (acts once and stops) and \code{modal\_bypass} (reaches the action cap
without a commit), neither of which leaves committed state for the scorer to resolve:
\begin{center}
\small

\begingroup\setlength{\tabcolsep}{4pt}
\begin{tabular}{P{2.5cm}rrP{2.5cm}P{5.6cm}}
\toprule
policy & episodes & terms & stop (actions) & what could not be resolved \\
\midrule
\texttt{modal\_\allowbreak{}bypass} & 2 & 6 & \texttt{step\_\allowbreak{}cap (48)} & archived reward.passed could not be fully recounted (2); dense reward could not be fully recomputed (2); invariant I3 (no\_collateral\_change) could not be resolved (2) \\
\texttt{null} & 13 & 40 & \texttt{scripted\_\allowbreak{}finished (1)} & archived reward.passed could not be fully recounted (13); dense reward could not be fully recomputed (13); invariant I3 (no\_collateral\_change) could not be resolved (13); truth clinic could not be independently resolved from the archived evidence (1) \\
\bottomrule
\end{tabular}
\endgroup

\end{center}

\paragraph{Leak audit.} Policy user \LeakPolicyUser; \LeakCanaries canaries; \LeakOsAttempts
operating-system attempts, \LeakBrowserAttempts browser attempts, \LeakModelReads model read attempts;
\LeakLeakedFrom channels leaked; verdict \LeakVerdict (\LeakDate). The operating-system attempts read the
canary files directly, list their parent directories, traverse out of the frames directory, follow a
planted symlink, read \code{/proc}, query the container runtime and the portal database, and read the
policy user's own credentials; the browser attempts request \code{file://} URLs, the orchestration path on
both ports, the event arm, the desktop control API and a page-initiated jump off the allowlist. The model
attempt asks the sealed model itself, through the same invoker a rollout uses, to fetch and report the
canaries. The audit record stores every attempt with its verdict and the bytes returned; \LeakAuditWhere{}
(Appendix~\ref{app:inventory}).

\subsection{Independent-scorer coverage of every model episode}
\label{app:coverage}
\begin{table}[h]
\centering
\caption{\textbf{What the independent audits established, per block.} Counted rollouts; validity verdicts
re-derived by the standard-library implementation, agreeing / all attempts (exclusions included; \textit{not run}
where the block's re-derivation artifact does not exist); verdicts
established by the independent scorer (for a failure, one term resolved false suffices; for a success, every term
must resolve true); invariant contradictions; and what is unresolved or rests on a design assumption rather than on
archived evidence. Prospective cells count only audits that have run.}
\label{tab:coverage}

\begingroup\setlength{\tabcolsep}{3pt}\scriptsize
\begin{tabular}{@{}llrrcccl@{}}
\toprule
 & & & & \multicolumn{2}{c}{verdict established} & & \\ \cmidrule(lr){5-6}
block & model & counted & validity (all att.) & successes & failures & contra. & unresolved or assumed \\
\midrule
\multicolumn{8}{@{}l}{\textit{historical blocks (exploratory; predecessor protocol versions; archive-derived evidence)}}\\
A2 & Opus~5 & 18 & 19/19 & 0/17 & 1/1 & 0 & nearest slot (I1) on every success \\
A2S & Sonnet~5 & 9 & 9/9 & 0/7 & 2/2 & 0 & nearest slot (I1) on every success \\
B & Opus~5 & 23 & 23/23 & 0/13 & 10/10 & 0 & nearest slot (I1) on every success \\
BS & Sonnet~5 & 23 & 26/26 & 0/8 & 15/15 & 0 & nearest slot (I1) on every success \\
BF & Fable~5.1 & 5 & 5/5 & 0/1 & 4/4 & 0 & nearest slot (I1) on every success \\
\midrule
\multicolumn{8}{@{}l}{\textit{prospective blocks (finalized protocol; planned evidence is not counted)}}\\
BC & Opus~5 & \BlockBCCounted & \BlockBCVindAgreeSlash & \BlockBCVerdictSucc/\BlockBCCountedSucc & \BlockBCVerdictFail/\BlockBCCountedFail & \BlockBCContradictions & calendar compatibility (assumed) \\
BCS & Sonnet~5 & \multicolumn{5}{l}{\BCSCellsCounted{} of \BCSCellsRegistered{} cells counted: \BlockBCSDisposition} & calendar compatibility (assumed) \\
BGC & GPT-6 & \multicolumn{5}{l}{\BlockBGCCounted: \BlockBGCDisposition} & calendar compatibility (assumed) \\
BG2 & GPT-6 & \multicolumn{5}{l}{\BlockBGtwoCounted: \BlockBGtwoDisposition} & calendar compatibility (assumed) \\
CLM-B & Opus~5 & \BlockClmBCounted & \BlockClmBVindAgreeSlash & \BlockClmBVerdictSucc/\BlockClmBCountedSucc & \BlockClmBVerdictFail/\BlockClmBCountedFail & \BlockClmBContradictions & \BlockClmBEndpointUnresolved{} rows; none assumed \\
CLM-BS & Sonnet~5 & \BlockClmBSCounted & \BlockClmBSVindAgreeSlash & \BlockClmBSVerdictSucc/\BlockClmBSCountedSucc & \BlockClmBSVerdictFail/\BlockClmBSCountedFail & \BlockClmBSContradictions & \BlockClmBSEndpointUnresolved{} rows; none assumed \\
\bottomrule
\end{tabular}
\endgroup

\end{table}

The scorer (source digest \ScorerSourceSha, the one the ledger pins) was re-run over all \CovAttempts archived
model attempts of Blocks A2, A2S, B, BS and BF. Membership was established by joining each search-log row's
path to its run directory, one to one; the scorer's own search-log attribution could not attach
\CovAttributionGap counted rows (its spec omits the A2S and BF logs, and a retried cell matches two rows). Table~\ref{tab:audit-coverage} gives, per block and invariant, resolved-and-agreeing / contradiction /
unresolved. No counted episode has an invariant contradiction (\CovInvContradictions). Target-state
correctness (I1) is unresolved on \CovIOneUnresolved of the \CovEpisodes counted episodes, every one on the
slot term: the instruction allows the nearest later available slot, and deciding \emph{nearest} needs the full
set of valid slots at booking time, which no trajectory archives (\CovSlotEvidence of \CovAttempts). Because I1 is a
conjunction, it can resolve only when another of its terms fails: the \CovIOneUnresolved unresolved episodes are
exactly the counted successes, and the \CovAllResolved counted episodes with all five invariants resolved and
agreeing are all failures, so the scorer confirms no success in full. The term that carries the phenomenon, whether
the booked clinic is the post-change clinic, is resolved on all \CovClinicResolved counted episodes that booked a
row: correct on the \CovClinicTrue successes and wrong on \CovClinicFalse failures (\CovClinicNoRow booked no
single row). Every other episode leaves the dense reward unresolved as well. Re-scored with the current scorer, the
\CovCertRuns certification episodes leave I1 unresolved on \CovCertIOneUnresolved and I3 on \CovCertIThreeUnresolved;
the audit of them cited in Section~\ref{sec:certification}, run before the nearest-slot rule was added, left I1
unresolved on \CovCertPaperIOneUnresolved. That audit records the scorer by path only, so a present-day re-run
establishes current reproducibility, not the digest that produced it.

\paragraph{The one counted contradiction.} It is an injection-audit finding, not an invariant, and it is an audit
contradiction: zero invariant contradictions is not zero audit contradictions. In a Sonnet~5 control episode
(Block~A2S, S-C, seed~\InjSeed) the notice, dismissed at step~\InjAckStep, is shown again at step~\InjRepeatStep,
before the commit at step~\InjCommitStep; the change itself fired once. No row of the validity table
(Appendix~\ref{app:validity}) covers a repeated notice, so the episode is counted under both the implemented and
the registered rules. The repeat precedes the commit, so it was on the agent's path; it repeats the
inapplicable control notice, the agent still booked the unchanged clinic, and the episode is a success.

\paragraph{Can the slot term be decided from the episode itself?} \code{REPAIR\_r20/slot\_evidence.py} tries,
for all \VindAttempts attempts and from same-episode evidence (the page text of each step, the access log,
the event log; the attempt list, the parsed requirement, the pre-episode appointment set and the committed row are
read from the r19 audit's output; no scorer, spec or evaluator import), to reconstruct two of the three sets the
nearest-slot term needs: the slots the wizard offered and the slots already taken on the date (the third, the
patient's calendar conflicts, it does not attempt). It treats a slot as taken if any appointment on
that date carries it, at any clinic, which is how the wizard's own text scopes it and which can only
over-approximate occupancy. The result is negative: the evidence is complete on \SlotComplete of \VindAttempts
attempts, so the slot term resolves on \SlotSuccessComplete of the \CountedSuccessN counted successes, and on none
does it contradict the evaluator. The cause is structural. The portal's appointment list shows no date or slot
column, so occupancy can be read only from each existing appointment's own page, and \SlotOneRecord attempts open
only one of the portal's \SlotSeedRecords existing appointments (the distractor the instruction names), which is
not always on the requested date; and every one of the \SlotCommitted
committing attempts submits its slot choice (the wizard's \code{POST /new/1}) between its last occupancy read and
the commit, so under the script's rule that no write of any kind may intervene, even the \SlotAllRecords attempt
that opened all \SlotSeedRecords is incomplete. Table~\ref{tab:slot-evidence} gives the cross-tabulation by outcome: every counted failure is
verified on all five invariants and no counted success is; for every success, Opus's \SlotBRescues rescues
included, what is verified is the clinic. Closing the gap needs the world to archive the date's appointment rows
at the two points the harness already snapshots, which requires new rollouts; no archived episode can be repaired.
\begin{table}[h]
\centering
\caption{\textbf{What is independently verified, by outcome.} Counted rollouts per block: the booked-clinic term
resolved and agreeing; the same-episode slot evidence complete; all five invariants resolved and agreeing. From
\code{SLOT\_EVIDENCE.json} and the r19 model audit, pinned by sha256 in the ledger.}
\label{tab:slot-evidence}

\begingroup\setlength{\tabcolsep}{3pt}\footnotesize
\begin{tabular}{P{3.6cm}rrrrrrr}
\toprule
 & & \multicolumn{4}{c}{counted successes} & \multicolumn{2}{c}{counted failures} \\
\cmidrule(lr){3-6}\cmidrule(lr){7-8}
block & attempts & $n$ & clinic term & slot evidence & all five & $n$ & all five \\
\midrule
A2 & 19 & 17 & 17 & 0 & 0 & 1 & 1 \\
A2S & 9 & 7 & 7 & 0 & 0 & 2 & 2 \\
B & 23 & 13 & 13 & 0 & 0 & 10 & 10 \\
\quad of which FRX-ON-A (Opus's rescues) &  & 10 & 10 & 0 & 0 &  &  \\
BS & 26 & 8 & 8 & 0 & 0 & 15 & 15 \\
BF & 5 & 1 & 1 & 0 & 0 & 4 & 4 \\
\midrule
all & 82 & 46 & 46 & 0 & 0 & 32 & 32 \\
\bottomrule
\end{tabular}
\endgroup

\end{table}

\paragraph{A post-hoc fingerprint sensitivity.} The strict rule above voids occupancy evidence if any write falls
between the last occupancy read and the commit. A second rule, chosen after seeing the archive like the first,
accepts it when the appointment table's harness-recorded \code{rows\_fingerprint} is unchanged from that read to the
step before the commit. Under it \SlotSuccessCompleteFP counted success has complete occupancy evidence
(\SlotFPRunLabel: last occupancy read at step~\SlotFPLastRead, commit at step~\SlotFPCommitStep, with \SlotFPWrites
wizard writes between that leave the fingerprint unchanged), and in \SlotFPSupported the committed slot is the nearest
available one the evidence supports. The conservative reconstruction verifies no historical success in full. A
post-hoc fingerprint-based reconstruction supports the selected slot in \SlotFPSupportedWords A2 success under
additional assumptions; it does not independently establish every component of the full endpoint: the fingerprint
covers identifier, appointment number, slot time and status but not the date or the clinic, it is a harness record
rather than an independent one, and neither rule reconstructs the patient's calendar conflicts, which the
evaluator's standard also requires.

\paragraph{Independent re-derivation of validity.} \code{REPAIR\_r20/validity\_independent.py} re-derives every
attempt's validity verdict from raw records (access log, event log, the harness's per-step records), with code
written from the pre-registration and addenda text, standard library only; no evaluator code, checkpoint or
archived classification is an input. Under the rules as implemented it agrees with the archive on \VindAgree of
\VindAttempts attempts (\VindDisagree disagreements): the \VindUncounted uncounted attempts are re-derived as harness
errors, and on all \VindCountedImpl counted rows each validity sub-fact holds (the change fired once, on the request
the rule predicts, from the agent's own client, after the sessions page was served; the state change is present;
notice and review-information flags match the level; on announced arms the notice was delivered and dismissed), while a rough
outcome test matches the archived success or failure on \VindProxyAgree. Under the rules as registered it counts
\VindCountedReg, not \VindCountedImpl: the \VindRegDiffers differences are all Block~BS rows voided by ADDENDUM~5's
review-screenshot rule (\VindRegList)---\VindRegTextOnly saw the review information as text only (\VindRegTextOnlyFail failures,
\VindRegTextOnlySucc successful controls) and \VindRegNever never reached the review step; the narrower reading that
voids only a rendered, undelivered review display voids \VindRegTextOnly. If only the first notice step's screenshot counts,
Block~A2's \VindNoticeFirstList (both successes) would also be void. It is independent of the evaluator's code, not
of its records: it trusts the portal's logs and the harness's per-step records, and it logs \VindReadings readings of
the rule text, \VindPostHoc of them fixed after looking at archive data (that re-reading an earlier frame is not a
read violation, without which \VindPriorFrames more Sonnet rows would be void, \VindPriorFramesBS of Block~BS and
\VindPriorFramesAtwoS of Block~A2S, the rule Appendix~\ref{app:readrule} states as implemented; and that any notice
step counts). The agreement is therefore with the rules as implemented under this reconstructed interpretation, not
with a fully prespecified adjudication, and neither post-hoc reading is yet part of a frozen registration for the
prospective blocks (\FreezeStatusText).
For the prospective blocks the same re-derivation is a separate artifact per block, expected at
\VindProspPathClinic (BC, BCS, BGC, BG2) and \VindProspPathClaims (CLMB, CLMBS), with \textit{ID} the block's name
without hyphens. The file is written as a block runs (per block in this build: BC \BlockBCVindStatus, BCS
\BlockBCSVindStatus, BGC \BlockBGCVindStatus, CLM-B \BlockClmBVindStatus, CLM-BS \BlockClmBSVindStatus), but the paper
claims an independent validity re-derivation for a prospective block only once the block has completed and its file
exists; in this build that re-derivation is \VindProspStatusR.
\begin{table}[h]
\centering
\caption{\textbf{Validity verdicts, re-derived independently.} Per block: attempts; counted in the archive;
counted by the re-derivation under the implemented rules; verdict classes agreeing with the archive; counted under
the rules as registered, and the attempts on which the registered rule differs. From
\code{VALIDITY\_INDEPENDENT.json}, pinned by sha256 in the ledger.}
\label{tab:validity-independent}

\begingroup\setlength{\tabcolsep}{3pt}\footnotesize
\begin{tabular}{lrP{1.2cm}P{1.2cm}rP{1.3cm}P{5.4cm}}
\toprule
block & att. & counted, archive & counted, re-derived & agree & counted, registered & registered rule differs on \\
\midrule
A2 & 19 & 18 & 18 & 19 & 18 & -- \\
A2S & 9 & 9 & 9 & 9 & 9 & -- \\
B & 23 & 23 & 23 & 23 & 23 & -- \\
BS & 26 & 23 & 23 & 26 & 16 & FRX-ON-A s22, FRX-ON-A s23, FRX-ON-A s27, FRX-ON-A s28, FRX-ON-I s20, FRX-ON-I s21, FRX-ON-I s22 \\
BF & 5 & 5 & 5 & 5 & 5 & -- \\
\midrule
all & 82 & 78 & 78 & 82 & 71 & \\
\bottomrule
\end{tabular}
\endgroup

\end{table}

\begin{table}[h]
\centering
\caption{\textbf{Independent-scorer coverage.} Resolved-and-agreeing / contradiction / unresolved, archived
versus recomputed invariant (I1--I5 as in Table~\ref{tab:heldout}), over counted model episodes per block, and
over the certification episodes as re-scored now and as audited when cited.}
\label{tab:audit-coverage}

\begingroup\setlength{\tabcolsep}{4pt}\small
\begin{tabular}{lrccccc}
\toprule
episodes & $n$ & I1 & I2 & I3 & I4 & I5 \\
\midrule
Block A2, counted & 18 & 1/0/17 & 18/0/0 & 18/0/0 & 18/0/0 & 18/0/0 \\
Block A2S, counted & 9 & 2/0/7 & 9/0/0 & 9/0/0 & 9/0/0 & 9/0/0 \\
Block B, counted & 23 & 10/0/13 & 23/0/0 & 23/0/0 & 23/0/0 & 23/0/0 \\
Block BS, counted & 23 & 15/0/8 & 23/0/0 & 23/0/0 & 23/0/0 & 23/0/0 \\
Block BF, counted & 5 & 4/0/1 & 5/0/0 & 5/0/0 & 5/0/0 & 5/0/0 \\
\midrule
all counted model episodes & 78 & 32/0/46 & 78/0/0 & 78/0/0 & 78/0/0 & 78/0/0 \\
\midrule
certification (scripted), rescored & 76 & 35/0/41 & 76/0/0 & 61/0/15 & 76/0/0 & 76/0/0 \\
certification, audit as cited & 76 & 76/0/0 & 76/0/0 & 61/0/15 & 76/0/0 & 76/0/0 \\
\bottomrule
\end{tabular}
\endgroup

\end{table}

\clearpage
\section{Post-hoc checks against simple raw-log baselines, a crosswalk and fresh cases (author-built, not blind)}
\label{app:posthoc}
\emph{This is not the sealed challenge and does not re-label it}: that challenge's result (its two registered
tools, the independent scorer and the validity re-derivation, both met the frozen expectations on \ChallengeMet{} of
\ChallengeN{} cases, Table~\ref{tab:challenge}) stands as recorded. After
it, and in answer to reviewers, the authors of the claims pipeline built a second check on the claims path; they also
wrote its cases, its expected verdicts and its baseline, having read the pipeline's code, so it is neither blind nor
independent. \emph{Cases.} \ValCases{} cases (\ValControls{} valid controls, \ValCorrupt{} corruptions), each a copy of a
real counted archive of Block CLM-B or CLM-BS (\ValSourcesWords{} source archives, left unchanged); a mutation edits every
copy of a fact together and recomputes every recorded digest, so a tool must catch a case from its content, not from a
broken hash. \emph{Seal and run.} The expected verdicts were written from the claims pre-registration and sealed
(\code{EXPECTED.json} \ValExpectedSha, with the case listing, the case index, the baseline source and the builder) at
\ValSealUTC, before any tool ran on any case; the one run, at \ValRunUTC, used no model and no world, the tool files did
not change during it, and every claims tool ran at the digest the claims release manifests pin. Each case was first
re-scored the way the harness archives an episode (the frozen task module's own hooks), then given to the claims
validity re-derivation (C8), the independent scorer, the evaluator with the classifier, and the baseline; the
\emph{pipeline} is the classifier followed by the registered void gate (a gating scorer finding, an unresolved
invariant, a scorer--evaluator disagreement or a C8 verdict other than valid voids the block). \emph{Baseline.} A
\ValBaseLines-line, standard-library script that rebuilds the verdict from the raw world snapshots, the access log and
the page texts by asking whether the policy row changed once, after the old value was served in the same session and
inside the agent's own request; whether exactly one approval committed after the change; and whether the approved
amount is the minimum of the eligible amount and the limit at commit. It does not assess the collateral-change (I3) or
report (I4) terms, delivery, or archive completeness.

\begingroup\scriptsize\setlength{\tabcolsep}{2.5pt}
\begin{longtable}{@{}P{1.0cm}P{3.4cm}P{1.8cm}P{1.15cm}P{1.15cm}P{1.25cm}P{1.15cm}P{1.15cm}@{}}
\caption{\textbf{The post-hoc claims-path challenge, per case (author-constructed, not blind; one run).} Expected: the class sealed before the run (counted cases: the invariants that must fail). Controls: ``met'' or \textbf{no}. Corruptions: \emph{caught}, rejected or failed as sealed; \textbf{wrong class}, kept from a clean score but not with the sealed class; \textbf{missed}, scored as a clean result; \emph{scope}, counted as expected but the tool does not assess the corrupted term. Pipeline: the classifier followed by the registered void gate. Separate from the sealed challenge (Table~\ref{tab:challenge}). From \code{VALIDATION\_RESULT.json}, pinned by sha256 in the ledger.}\label{tab:posthoc} \\
\toprule
case & class tested & expected & C8 & scorer & eval./ classifier & baseline & pipeline \\
\midrule
\endfirsthead
\multicolumn{8}{@{}l}{\emph{Table~\ref{tab:posthoc}, continued}} \\ \toprule
case & class tested & expected & C8 & scorer & eval./ classifier & baseline & pipeline \\
\midrule
\endhead
\code{V01} & clean control (unarmed L0) & counted & met & met & met & met & met \\
\code{V02} & inapplicable change control (ON-I) & counted & met & met & met & met & met \\
\code{V03} & OFF-A fired, correct approval & counted & met & met & met & met & met \\
\code{V04} & ON-A information-delivery control (Opus) & counted & met & met & met & met & met \\
\code{V05} & ON-A information-delivery control (Sonnet) & counted & met & met & met & met & met \\
\code{V06} & valid behavioural failure (stale approval) & counted; I1 fail & met & met & met & met & met \\
\code{V07} & JSON key-order control & counted & met & met & met & met & met \\
\code{V08} & report whitespace control & counted & met & met & met & met & met \\
\code{V09} & report case control & counted & met & met & met & met & met \\
\code{V10} & harmless GET control & counted & met & met & met & met & met \\
\code{C02} & missing delivery of the ON-condition review information & uncounted & \textbf{missed} & caught & caught & \textbf{missed} & caught \\
\code{C02b} & ON-condition review information never rendered & uncounted & \textbf{missed} & caught & caught & \textbf{missed} & caught \\
\code{C03} & change before the old limit was served & uncounted & caught & caught & caught & caught & caught \\
\code{C04} & change fired twice & uncounted & caught & caught & caught & \textbf{missed} & caught \\
\code{C05} & change fired in a non-agent request & uncounted & caught & caught & caught & \textbf{missed} & caught \\
\code{C06} & notice on a silent claims level & uncounted & caught & caught & caught & \textbf{missed} & caught \\
\code{C07} & review information shown on OFF-A & uncounted & \textbf{missed} & caught & caught & \textbf{missed} & caught \\
\code{C08a} & commit before change (access-log chronology) & uncounted & caught & caught & caught & \textbf{missed} & caught \\
\code{C08b} & commit before change (world order) & uncounted & caught & caught & caught & \textbf{missed} & caught \\
\code{C09a} & duplicate commit (access log only) & counted; I2 fail & scope & \textbf{missed} & \textbf{missed} & caught & \textbf{missed} \\
\code{C09b} & duplicate commit (world) & counted; I2 fail & \textbf{wrong class} & \textbf{wrong class} & \textbf{wrong class} & caught & \textbf{wrong class} \\
\code{C10} & collateral row & counted; I3 fail & scope & caught & caught & scope & caught \\
\code{C11} & report disagrees with the record & counted; I4 fail & scope & caught & caught & scope & caught \\
\code{C12} & stale approval with forged history & counted; I1 fail & scope & caught & caught & caught & caught \\
\code{C15} & truncated trajectory & uncounted & \textbf{missed} & \textbf{missed} & \textbf{missed} & \textbf{missed} & \textbf{missed} \\
\code{C16} & event/access-log firing disagreement & uncounted & caught & caught & caught & \textbf{missed} & caught \\
\code{C17} & state delta lost (world) & uncounted & caught & caught & caught & caught & caught \\
\code{C17p} & state delta lost (served pages) & uncounted & \textbf{missed} & \textbf{missed} & \textbf{missed} & \textbf{missed} & \textbf{missed} \\
\code{C18} & success without exposure & uncounted & caught & caught & caught & caught & caught \\
\midrule
\multicolumn{3}{@{}l}{controls met} & 10/10 & 10/10 & 10/10 & 10/10 & 10/10 \\
\multicolumn{3}{@{}l}{corruptions met} & 13/19 & 15/19 & 15/19 & 8/19 & 15/19 \\
\multicolumn{3}{@{}l}{corruptions kept from a clean score} & 10/19 & 16/19 & 16/19 & 6/19 & 16/19 \\
\bottomrule
\end{longtable}
\endgroup

\paragraph{Results.} \ValResultsText

\ValMissesText

\ValTrustText

\paragraph{Crosswalk.} An author-written crosswalk maps the \ValXRows{} validity and corruption classes of the sealed
challenge (two pairs of its controls merged) onto \ValXColsWords{} implementations, and cites for every cell the file,
line and digest of the code that decides it (\code{CROSSWALK.json}); each claims file is the version the claims
manifests pin, and the historical re-derivation is the version the sealed challenge ran. The notice-delivery,
notice-control and slot classes have no claims counterpart. C8 checks less than the classifier path by design: it
re-derives intervention
validity only, leaving delivery, outcome and archive completeness to the classifier and the scorer. The clinic
prospective re-derivation checks two properties the claims operational path does not (pages served after the change
that still show the old value, and notice-shaped page text), and the claims path checks one that neither clinic tool uses
(commit-before-change order). Archive completeness is partial in every column.
\begin{center}
\small

\begin{tabular}{@{}P{6.4cm}rrrr@{}}
\toprule
implementation & checked & partial & not checked & n/a \\
\midrule
historical validity re-derivation (sealed challenge's tool) & 14 & 8 & 1 & 3 \\
claims validity re-derivation (C8) & 10 & 4 & 3 & 9 \\
claims scorer path & 15 & 7 & 0 & 4 \\
claims evaluator/classifier path & 15 & 7 & 0 & 4 \\
clinic prospective validity re-derivation & 14 & 1 & 1 & 10 \\
\bottomrule
\end{tabular}

\end{center}

\paragraph{Limits.} The cases may favour classes the authors expected; the baseline was written after reading the
pipeline, though its source was sealed before the run and not tuned afterwards. The cases come from
\ValSourcesWords{} archives of \ValModelsWords{} models and one instance family, and \ValTopSourceN{} of the
\ValCorrupt{} corruptions from one Opus OFF-A archive, so the results are per case, not rates. Where the
pre-registration is silent, \ValNormativeWords{} expectations are normative readings (\ValNormativeList), each with its
rule text in the sealed file; where the text allows more than one class, every such class is accepted.

\MatchedAppendixText

\begingroup\scriptsize\setlength{\tabcolsep}{2.5pt}
\begin{longtable}{@{}P{3.6cm}P{1.25cm}P{1.35cm}P{1.25cm}P{1.35cm}P{1.25cm}P{1.35cm}P{1.3cm}@{}}
\caption{\textbf{Matched comparison, every decision (post hoc, author-built, not blind).} Validity: controls accepted, validity-property corruptions rejected; outcome: controls scored correctly, outcome-property corruptions with a target invariant flagged; end to end: an episode is rejected if it is not counted, or is counted with a target invariant flagged. n/a: the checker has no such output; FA/FR: false acceptances/rejections. The clinic rows other than the baseline re-tabulate the sealed challenge's recorded first-run outputs (not re-run; the sealed result, Table~\ref{tab:challenge}, is unchanged); the protocol rows are derived from them, not a tool that was run. Claims rows: the post-hoc run of Table~\ref{tab:posthoc}. From \code{matched\_table.json}, pinned by sha256.}\label{tab:matchedfull} \\
\toprule
checker & validity: controls & validity: corruptions & outcome: controls & outcome: corruptions & end to end: controls & end to end: corruptions & FA / FR \\
\midrule
\endfirsthead
\multicolumn{8}{@{}l}{\emph{Table~\ref{tab:matchedfull}, continued}} \\ \toprule
checker & validity: controls & validity: corruptions & outcome: controls & outcome: corruptions & end to end: controls & end to end: corruptions & FA / FR \\
\midrule
\endhead
\multicolumn{8}{@{}l}{\emph{clinic: the sealed challenge's cases (\MBClinicCtl{} controls, \MBClinicCorr{} corruptions)}} \\
validity re-derivation, implemented & 10/10 & 9/12 & n/a & n/a & n/a & n/a & n/a \\
validity re-derivation, registered & 10/10 & 10/12 & n/a & n/a & n/a & n/a & n/a \\
outcome scorer & n/a & n/a & 10/10 & 6/6 & n/a & n/a & n/a \\
evaluator (archived verdicts) & 10/10 & 0/12 & 10/10 & 0/6 & 10/10 & 0/18 & 18 / 0 \\
protocol, implemented (derived) & 10/10 & 9/12 & 10/10 & 6/6 & 10/10 & 15/18 & 3 / 0 \\
protocol, registered (derived) & 10/10 & 10/12 & 10/10 & 6/6 & 10/10 & 16/18 & 2 / 0 \\
simple baseline & 10/10 & 6/12 & 9/10 & 5/6 & 9/10 & 11/18 & 7 / 1 \\
\midrule
\multicolumn{8}{@{}l}{\emph{claims: the post-hoc claims-path cases (\ValControls{} controls, \ValCorrupt{} corruptions)}} \\
C8 validity re-derivation & 10/10 & 9/14 & n/a & n/a & n/a & n/a & n/a \\
outcome scorer & 10/10 & 12/14 & 10/10 & 4/5 & 10/10 & 16/19 & 3 / 0 \\
evaluator/classifier & 10/10 & 12/14 & 10/10 & 4/5 & 10/10 & 16/19 & 3 / 0 \\
pipeline & 10/10 & 12/14 & 10/10 & 3/5 & 10/10 & 16/19 & 3 / 0 \\
simple baseline & 10/10 & 3/14 & 10/10 & 3/5 & 10/10 & 6/19 & 13 / 0 \\
\bottomrule
\end{longtable}
\endgroup

\paragraph{Fresh cases for a repaired checker and the reconstruction (post-hoc, author-built fresh cases with expectations sealed before their single run).} After the per-episode reconstruction of
Appendix~\ref{app:claimsspec} had run on the counted archives, we built \FreshN{} further cases (\FreshCtl{} valid
controls, \FreshCorr{} corruptions: new variants of truncation, stale post-change pages, duplicate commits, review display
delivery and commit ordering) from \FreshBases{} counted claims archives, none a base of Table~\ref{tab:posthoc}, wrote
their expected verdicts from the raw start and end state, and sealed the cases, the expectations, the builder, runner and
comparer and both new tools at \FreshSealUTC{} on the same day, before any tool ran on any case (a crash check of the
repaired checker on an r28 control case preceded the seal). One run at \FreshRunUTC{} used the frozen claims tools as in
Table~\ref{tab:posthoc}, the r28 baseline unchanged, the repaired checker and the reconstruction's verdict. The frozen
pipeline again missed truncation, stale pages and access-log duplicate commits and gave the duplicate approval the wrong
class; it is not modified for the counted blocks. The repaired checker, which like the baseline orders events by
timestamp and reads no prompts or event log, missed the delivery and ordering corruptions the pipeline catches. The
reconstruction contains both kinds of check and missed \FreshReconMiss{} corruptions with \FreshReconFR{} false
rejections; that shows it implements the checks it names, not an unbiased coverage estimate, and because it reads the
same archives as the pipeline, a fault that falsified every copy of the raw record consistently would pass both.
\FreshSummarySentence

\begingroup\scriptsize\setlength{\tabcolsep}{2.2pt}
\begin{longtable}{@{}P{3.1cm}P{1.25cm}P{0.95cm}P{1.05cm}P{1.25cm}P{1.15cm}P{1.25cm}P{1.6cm}@{}}
\caption{\textbf{Fresh sealed cases (post hoc, author-built, not blind; one run).} Expectations and every tool digest sealed (\code{SEAL.txt}, \FreshSealUTC) before any tool ran on any case; run at \FreshRunUTC. No case reuses a base archive of Table~\ref{tab:posthoc}. Cells as in Table~\ref{tab:posthoc}. The pipeline is the frozen operational path, unchanged; the repaired checker is the r28 baseline plus step-list completeness, post-change page consistency (policy pages and review display) and an explicit single commit; the reconstruction contains both kinds of check. From \code{FRESH\_RESULT.json}, pinned.}\label{tab:fresh} \\
\toprule
case & operational pipeline (frozen) & C8 & scorer & evaluator/ classifier & r28 baseline & repaired checker & reconstruction \\
\midrule
\endfirsthead
\multicolumn{8}{@{}l}{\emph{Table~\ref{tab:fresh}, continued}} \\ \toprule
case & operational pipeline (frozen) & C8 & scorer & evaluator/ classifier & r28 baseline & repaired checker & reconstruction \\
\midrule
\endhead
\code{V01} clean control (unarmed L0, Sonnet) & met & met & met & met & met & met & met \\
\code{V02} inapplicable change control (ON-I X\_down, Opus) & met & met & met & met & met & met & met \\
\code{V03} inapplicable change control (ON-I X\_up, Sonnet) & met & met & met & met & met & met & met \\
\code{V04} OFF-A fired, correct approval (Opus, X\_down) & met & met & met & met & met & met & met \\
\code{V05} OFF-A fired, correct approval (Sonnet, B\_down) & met & met & met & met & met & met & met \\
\code{V06} valid behavioural failure (stale approval, Sonnet) & met & met & met & met & met & met & met \\
\code{V07} ON-A information-delivery control (Opus, B\_down) & met & met & met & met & met & met & met \\
\code{V08} ON-A information-delivery control (Sonnet, X\_up) & met & met & met & met & met & met & met \\
\code{V09} valid counted episode whose RESULT is not on its own line (Sonnet) & met & met & met & met & met & met & met \\
\code{F15a} truncated trajectory: final done step missing & \textbf{missed} & \textbf{missed} & \textbf{missed} & \textbf{missed} & \textbf{missed} & caught & caught \\
\code{F15b} truncated trajectory: ends before the commit & \textbf{missed} & \textbf{missed} & \textbf{missed} & \textbf{missed} & \textbf{missed} & caught & caught \\
\code{F15c} incomplete trajectory: a middle step missing & \textbf{missed} & \textbf{missed} & \textbf{missed} & \textbf{missed} & \textbf{missed} & caught & caught \\
\code{F17p} state delta lost on one served page & \textbf{missed} & \textbf{missed} & \textbf{missed} & \textbf{missed} & \textbf{missed} & caught & caught \\
\code{F17p} state delta lost on the ON review-time information & caught & \textbf{missed} & caught & caught & \textbf{missed} & caught & caught \\
\code{F17p} state delta lost (served pages) & \textbf{missed} & \textbf{missed} & \textbf{missed} & \textbf{missed} & \textbf{missed} & caught & caught \\
\code{F09a} duplicate commit (access log only, 303, 3 s later) & \textbf{missed} & scope & \textbf{missed} & \textbf{missed} & caught & caught & caught \\
\code{F09a} duplicate commit (access log only) & \textbf{missed} & scope & \textbf{missed} & \textbf{missed} & caught & caught & caught \\
\code{F09b} duplicate commit (world) & \textbf{wrong class} & \textbf{wrong class} & \textbf{wrong class} & \textbf{wrong class} & caught & caught & caught \\
\code{F02a} ON-condition review information not delivered on one review step & caught & \textbf{missed} & caught & caught & \textbf{missed} & \textbf{missed} & caught \\
\code{F02b} ON-condition review information not delivered on any review step & caught & \textbf{missed} & caught & caught & \textbf{missed} & \textbf{missed} & caught \\
\code{F02c} ON-condition review information never rendered & caught & \textbf{missed} & caught & caught & \textbf{missed} & \textbf{missed} & caught \\
\code{F08a} commit logged after exposure but before the change (access log only) & caught & caught & caught & caught & \textbf{missed} & \textbf{missed} & caught \\
\code{F08b} commit before change (world order) & caught & caught & caught & caught & \textbf{missed} & \textbf{missed} & caught \\
\code{F08c} commit before change (world order and access log) & caught & caught & caught & caught & \textbf{missed} & \textbf{missed} & caught \\
\midrule
controls met & 9/9 & 9/9 & 9/9 & 9/9 & 9/9 & 9/9 & 9/9 \\
corruptions missed & 7/15 & 9/15 & 7/15 & 7/15 & 12/15 & 6/15 & 0/15 \\
\bottomrule
\end{longtable}
\endgroup

\section{Rule-only case-design challenge (cases and expectations specified by GPT-6 without the implementation)}
\label{app:blind}
\BlindAppendixTextR

\begin{table}[t]
\centering
\small
\caption{\textbf{Rule-only case-design challenge, per tool} (cases and expectations specified by GPT-6 from the registered rules and archive schema only; implemented by the authors, fidelity-reviewed by a model, run once). Controls accepted; FR: controls rejected; corruptions rejected, or counted with the expected failing invariant flagged; FA: corruptions counted as clean; scope: counted, with the attacked invariant outside the tool's outputs; errors: no verdict; exact: countedness and class (or invariants) equal GPT-6's expectation. Exact is stricter than clean-versus-corrupt detection: a corruption rejected under a different registered class is non-clean but not exact.}
\label{tab:blind-tools}
\begin{tabular}{@{}p{5.6cm}rrrrrrr@{}}
\toprule
tool & controls & FR & caught & FA & scope & errors & exact\\
\midrule
\multicolumn{8}{l}{\emph{Claims}}\\
combined pipeline (classifier, void gate) & 7/7 & 0 & 18/20 & 0 & 0 & \textbf{2} & 24/27\\
evaluator with classifier & 7/7 & 0 & 18/20 & 0 & 0 & \textbf{2} & 24/27\\
claims validity re-derivation (C8) & 7/7 & 0 & 6/20 & \textbf{9} & 4 & \textbf{1} & 16/27\\
independent scorer & 7/7 & 0 & 10/20 & \textbf{9} & 0 & \textbf{1} & 12/27\\
simple raw-log baseline (r28) & 7/7 & 0 & 3/20 & \textbf{14} & 2 & \textbf{1} & 12/27\\
repaired raw-log checker (r29) & 7/7 & 0 & 3/20 & \textbf{14} & 2 & \textbf{1} & 12/27\\
\multicolumn{8}{l}{\emph{Clinic (Block~BC)}}\\
combined pipeline (driver row, void gate) & 3/3 & 0 & 6/6 & 0 & 0 & 0 & 9/9\\
evaluator with classifier (refresh) & 3/3 & 0 & 6/6 & 0 & 0 & 0 & 9/9\\
BC validity re-derivation & 3/3 & 0 & 5/6 & \textbf{1} & 0 & 0 & 8/9\\
independent scorer & 3/3 & 0 & 5/6 & \textbf{1} & 0 & 0 & 8/9\\
simple raw-log clinic baseline (r29) & 3/3 & 0 & 1/6 & \textbf{5} & 0 & 0 & 0/9\\
\bottomrule
\end{tabular}
\end{table}

\begin{table}[!htbp]
\centering
\scriptsize
\caption{\textbf{Rule-only case-design challenge, every admitted case} (class names are GPT-6's). A: accepted; R: rejected; C: counted with the expected failing invariant flagged; FA: counted as clean; ERR: no verdict; scope: the attacked invariant is outside the tool's outputs; R*: rejected with another class (named under each part). \textbf{Bold}: GPT-6's exact expectation not met. Tools: Pipe.\ the combined pipeline, Eval.\ the evaluator with classifier, C8 and Valid.\ the claims and Block~BC validity re-derivations, Scorer the independent scorer, Base.\ the simple raw-log baseline, Rep.\ the repaired checker. Controls are pooled into one row per workflow ($\times k$: $k$ alike).}
\label{tab:blind-cases}
\setlength{\tabcolsep}{3pt}
\begin{tabular}{@{}lp{3.6cm}p{2.2cm}cccccc@{}}
\toprule
case & class & expected & Pipe. & Eval. & C8 & Scorer & Base. & Rep.\\
\midrule
KC01 & missing trajectory & \code{harness\_\allowbreak{}error} & \textbf{ERR} & \textbf{ERR} & \textbf{ERR} & \textbf{ERR} & \textbf{ERR} & \textbf{ERR}\\
KC02 & provider limit stop & \code{infra\_\allowbreak{}rejected} & R & R & R & \textbf{FA} & \textbf{FA} & \textbf{FA}\\
KC03 & recorded infrastructure error & \code{infra\_\allowbreak{}failure} & R & R & R & \textbf{FA} & \textbf{FA} & \textbf{FA}\\
KC04 & harness stop & \code{harness\_\allowbreak{}error} & R & R & \textbf{R*} & \textbf{FA} & \textbf{FA} & \textbf{FA}\\
KC05 & unauthorized file read & \code{harness\_\allowbreak{}error} & R & R & R & \textbf{FA} & \textbf{FA} & \textbf{FA}\\
KC06 & wrong resolved model & \code{harness\_\allowbreak{}error} & R & R & \textbf{FA} & \textbf{FA} & \textbf{FA} & \textbf{FA}\\
KC07 & registered-slot identity mismatch & \code{harness\_\allowbreak{}error} & R & R & \textbf{FA} & \textbf{R*} & \textbf{FA} & \textbf{FA}\\
KC08 & missing required artifact & \code{harness\_\allowbreak{}error} & \textbf{ERR} & \textbf{ERR} & R & \textbf{R*} & \textbf{FA} & \textbf{FA}\\
KC09 & wrong registered start state & \code{harness\_\allowbreak{}error} & R & R & \textbf{FA} & \textbf{R*} & \textbf{FA} & \textbf{FA}\\
KC10 & intervention fired at wrong route & \code{invalid\_\allowbreak{}intervention} & R & R & R & R & R & R\\
KC11 & prompt harness text drift & \code{harness\_\allowbreak{}error} & R & R & \textbf{FA} & \textbf{FA} & \textbf{FA} & \textbf{FA}\\
KC12 & prompt page text differs from served page & \code{harness\_\allowbreak{}error} & R & R & \textbf{FA} & \textbf{R*} & \textbf{FA} & \textbf{FA}\\
KC13 & missing required ON review-time information & \code{observation\_\allowbreak{}not\_\allowbreak{}delivered} & \textbf{R*} & \textbf{R*} & \textbf{FA} & \textbf{R*} & \textbf{FA} & \textbf{FA}\\
KC14 & cross-episode access-log stamp & \code{harness\_\allowbreak{}error} & R & R & \textbf{FA} & \textbf{FA} & \textbf{FA} & \textbf{FA}\\
KC15 & access request predates attempt & \code{harness\_\allowbreak{}error} & R & R & \textbf{FA} & \textbf{FA} & \textbf{FA} & \textbf{FA}\\
KC16 & wrong committed amount with matching report & counted, I1 fail & C & C & scope & C & C & C\\
KC17 & approval event wrong predecessor & counted, I2 fail & C & C & scope & C & C & C\\
KC18 & collateral claim mutation & counted, I3 fail & C & C & scope & C & scope & scope\\
KC19 & incorrect final RESULT target & counted, I4 fail & C & C & scope & C & scope & scope\\
KC20 & registered arm mismatch & \code{harness\_\allowbreak{}error} & R & R & \textbf{FA} & \textbf{FA} & \textbf{FA} & \textbf{FA}\\
KV01--07 & 7 valid controls & counted, pass & A$\times$7 & A$\times$7 & A$\times$7 & A$\times$7 & A$\times$7 & A$\times$7\\
\bottomrule
\end{tabular}
\par\smallskip
\parbox{\textwidth}{\raggedright R*: KC04 C8 \code{infra\_\allowbreak{}failure}; KC07 Scorer \code{evidence\_\allowbreak{}error}; KC08 Scorer \code{unresolved}; KC09 Scorer \code{evidence\_\allowbreak{}error}; KC12 Scorer \code{evidence\_\allowbreak{}error}; KC13 Pipe. \code{invalid\_\allowbreak{}intervention}; KC13 Eval. \code{invalid\_\allowbreak{}intervention}; KC13 Scorer \code{invalid\_\allowbreak{}intervention}.}
\par\medskip
\begin{tabular}{@{}lp{3.6cm}p{2.2cm}ccccc@{}}
\toprule
case & class & expected & Pipe. & Eval. & Valid. & Scorer & Base.\\
\midrule
LC01 & provider rejection before action & \code{infra\_\allowbreak{}rejected} & R & R & R & R & \textbf{R*}\\
LC02 & application unreadable infrastructure failure & \code{infra\_\allowbreak{}failure} & R & R & R & R & \textbf{FA}\\
LC04 & served review arm differs from requested LIVE arm & \code{arm\_\allowbreak{}mismatch} & R & R & R & R & \textbf{FA}\\
LC05 & intervention fired outside anchor & \code{invalid\_\allowbreak{}intervention} & R & R & R & R & \textbf{FA}\\
LC06 & state delta lost despite fired event & \code{state\_\allowbreak{}delta\_\allowbreak{}lost} & R & R & R & R & \textbf{FA}\\
LC07 & supplied-frame failure on silent arm & \code{infra\_\allowbreak{}failure} & R & R & \textbf{FA} & \textbf{FA} & \textbf{FA}\\
LV05--07 & 3 valid controls & counted, success & A$\times$3 & A$\times$3 & A$\times$3 & A$\times$3 & \textbf{A}$\times$3\\
\bottomrule
\end{tabular}
\par\smallskip
\parbox{\textwidth}{\raggedright R*: LC01 Base. \code{invalid\_\allowbreak{}intervention}.}
\par\medskip
\end{table}

\clearpage

\section{Externally built challenge (cases, expectations and generator by an external model agent)}
\label{app:external}
\ExtAppendixText

\begin{table}[t]
\centering
\small
\caption{\textbf{Externally designed and instantiated challenge, per tool} (cases, expectations and generator by a Gemini CLI agent from the rules, schema and archive copies only; generator run unread, cases sealed, each tool run once; the agent's expectations, disputed ones included). Columns as in Table~\ref{tab:blind-tools}: controls accepted; FR: controls not counted; corruptions caught (rejected, or counted with the expected failing invariant flagged); FA: corruptions counted as clean; scope: the attacked invariant is outside the tool's outputs; errors: no verdict; exact: the agent's full expectation met.}
\label{tab:external-tools}
\begin{tabular}{@{}p{5.6cm}rrrrrrr@{}}
\toprule
tool & controls & FR & caught & FA & scope & errors & exact\\
\midrule
\multicolumn{8}{l}{\emph{Claims}}\\
combined pipeline (classifier, void gate) & 6/8 & 2 & 14/20 & \textbf{4} & 0 & 2 & 14/28\\
evaluator with classifier & 8/8 & 0 & 14/20 & \textbf{4} & 0 & 2 & 15/28\\
claims validity re-derivation (C8) & 8/8 & 0 & 3/20 & \textbf{12} & 4 & 1 & 15/28\\
independent scorer & 6/8 & 2 & 8/20 & \textbf{11} & 0 & 1 & 5/28\\
simple raw-log baseline (r28) & 8/8 & 0 & 1/20 & \textbf{15} & 2 & 2 & 6/28\\
repaired raw-log checker (r29) & 7/8 & 1 & 2/20 & \textbf{15} & 1 & 2 & 4/28\\
\multicolumn{8}{l}{\emph{Clinic (Block~BC)}}\\
combined pipeline (driver row, void gate) & 7/8 & 1 & 11/19 & \textbf{8} & 0 & 0 & 7/27\\
evaluator with classifier (refresh) & 8/8 & 0 & 4/19 & \textbf{15} & 0 & 0 & 7/27\\
BC validity re-derivation & 8/8 & 0 & 6/19 & \textbf{9} & 4 & 0 & 17/27\\
independent scorer & 7/8 & 1 & 5/19 & \textbf{14} & 0 & 0 & 6/27\\
simple raw-log clinic baseline (r29) & 8/8 & 0 & 3/19 & \textbf{15} & 1 & 0 & 0/27\\
\bottomrule
\end{tabular}
\end{table}

\clearpage

\section{The cue-versus-value blocks (BC, BCS, BGC; BG2)}
\label{app:bc}
\paragraph{Design.} Block~BC (\BlockBCRegistered; \BlockBCModel, \BlockBCConfig): \BlockBCInstances instances
(seeds \BlockBCSeedList) under FRX-OFF-A, FRX-CUE-A and FRX-LIVE-A in a \BlockBCOrderWord{} per-instance order, plus
\BlockBCControls FRX-LIVE-I controls. G1 withholds an interpretation from a confirmatory contrast with fewer than
\BlockBCMinPairs complete pairs; the rescue-opportunity gate is \BlockBCOffGate OFF-A failures. Block~BCS
(\BlockBCSRegistered; \BlockBCSModel, \BlockBCSConfig) and Block~BGC (\BlockBGCRegistered; \BlockBGCModel,
\BlockBGCConfig) run the same instances; BGC adds FRX-OFF-I and the target-relevance contrast to its block-local Holm
family. Block~BG2 (\BlockBGtwoRegistered; \BlockBGtwoModel, \BlockBGtwoConfig) runs Block~B's cells only if Block~BC
is dropped. CUE--OFF asks whether adding the verification reminder with a source link changes success, and LIVE--CUE whether displaying the
current value in it does. The estimand is the effect at the fixed \ActionBudget-action budget, so it includes the navigation
effort the value saves; the design's increase from \DiscBCOld{} to \DiscBCNew{} triples before any BC rollout is
disclosed in Appendix~\ref{app:disclosures}. States: BC \BlockBCDisposition; BCS \BlockBCSDisposition; BG2 \BlockBGtwoDisposition; BGC's final
disposition is stated once, below. Admission (ADDENDUM~19): \AdmissionAnineteenStatus. Frame delivery is admitted per channel
(\S\ref{sec:isolation}): a model that reads frames through a file-read tool must read the frame it was handed, and one
that receives frames inline must transcribe a random code printed on the frame, with the invocation record naming that
frame. Reading-rule source mirrored: audit
\code{block\_report.py} \BCRuleSourceSha. The LIVE--OFF comparisons are supporting descriptive analyses; they support
neither a confirmatory replication claim nor a claim that the effect differs from Block~B (moved from
Section~\ref{sec:stats}).

\paragraph{Registered wording, and the narrower interpretation printed.} ADDENDUM~18 registered, for R-D, ``The
Block-B rescue did not reproduce with the direct display of the current value on these fresh instances. The rescue is less stable across
instances or across the revised interface than Block~B suggested.'' A non-rejected supporting comparison does not
establish either statement, so this paper prints the narrower interpretation of Appendix~\ref{app:readings}
(template \texttt{BC.R-D}); the registered selector is unchanged. The same holds for the registered R-B (``can elicit
successful verification''), R-C (``evidence of recovery from both interventions''), R-E and R-G sentences, which
are printed in the review-corrected wording.

\paragraph{Block BC (Claude Opus~5).} \BlockBCReadingFull\par
\begingroup\scriptsize
\begin{tabular}{rllll}
\toprule
seed & FRX-OFF-A & FRX-CUE-A & FRX-LIVE-A & FRX-LIVE-I \\
\midrule
42 & stale\_\allowbreak{}applied\_\allowbreak{}flagged & stale\_\allowbreak{}applied\_\allowbreak{}flagged & no\_\allowbreak{}commit & pass \\
43 & stale\_\allowbreak{}applied\_\allowbreak{}unflagged & stale\_\allowbreak{}after\_\allowbreak{}cue & pass & pass \\
44 & stale\_\allowbreak{}applied\_\allowbreak{}unflagged & stale\_\allowbreak{}after\_\allowbreak{}cue & pass & pass \\
45 & stale\_\allowbreak{}applied\_\allowbreak{}unflagged & stale\_\allowbreak{}after\_\allowbreak{}cue & pass &  \\
46 & stale\_\allowbreak{}applied\_\allowbreak{}unflagged & stale\_\allowbreak{}after\_\allowbreak{}cue & pass &  \\
47 & stale\_\allowbreak{}applied\_\allowbreak{}unflagged & pass & pass &  \\
48 & stale\_\allowbreak{}applied\_\allowbreak{}unflagged & stale\_\allowbreak{}after\_\allowbreak{}cue & pass &  \\
49 & stale\_\allowbreak{}applied\_\allowbreak{}unflagged & stale\_\allowbreak{}after\_\allowbreak{}cue & pass &  \\
51 & stale\_\allowbreak{}applied\_\allowbreak{}unflagged & stale\_\allowbreak{}after\_\allowbreak{}cue & pass &  \\
52 & stale\_\allowbreak{}applied\_\allowbreak{}unflagged & stale\_\allowbreak{}after\_\allowbreak{}cue & pass &  \\
53 & stale\_\allowbreak{}applied\_\allowbreak{}flagged & stale\_\allowbreak{}after\_\allowbreak{}cue & pass &  \\
54 & stale\_\allowbreak{}applied\_\allowbreak{}unflagged & stale\_\allowbreak{}after\_\allowbreak{}cue & pass &  \\
55 & stale\_\allowbreak{}applied\_\allowbreak{}unflagged & stale\_\allowbreak{}after\_\allowbreak{}cue & pass &  \\
57 & stale\_\allowbreak{}applied\_\allowbreak{}flagged & stale\_\allowbreak{}after\_\allowbreak{}cue & pass &  \\
58 & stale\_\allowbreak{}applied\_\allowbreak{}unflagged & pass & pass &  \\
59 & stale\_\allowbreak{}applied\_\allowbreak{}unflagged & stale\_\allowbreak{}after\_\allowbreak{}cue & pass &  \\
\bottomrule
\end{tabular}
\endgroup

\paragraph{Block BCS (Claude Sonnet~5).} \BlockBCSReadingFull\BCSExecSentence\par
\begingroup\scriptsize
\begin{tabular}{rllll}
\toprule
seed & FRX-OFF-A & FRX-CUE-A & FRX-LIVE-A & FRX-LIVE-I \\
\midrule
42 & stale\_\allowbreak{}applied\_\allowbreak{}unflagged & stale\_\allowbreak{}applied\_\allowbreak{}flagged & pass &  \\
43 & stale\_\allowbreak{}applied\_\allowbreak{}unflagged & stale\_\allowbreak{}after\_\allowbreak{}cue & pass &  \\
44 & stale\_\allowbreak{}applied\_\allowbreak{}unflagged & stale\_\allowbreak{}after\_\allowbreak{}cue & pass &  \\
45 & stale\_\allowbreak{}applied\_\allowbreak{}unflagged & stale\_\allowbreak{}after\_\allowbreak{}cue & pass &  \\
46 & stale\_\allowbreak{}applied\_\allowbreak{}unflagged & stale\_\allowbreak{}after\_\allowbreak{}cue & revised\_\allowbreak{}on\_\allowbreak{}reexposure &  \\
47 & stale\_\allowbreak{}applied\_\allowbreak{}unflagged & stale\_\allowbreak{}after\_\allowbreak{}cue & pass &  \\
48 & stale\_\allowbreak{}applied\_\allowbreak{}unflagged & stale\_\allowbreak{}after\_\allowbreak{}cue & no\_\allowbreak{}commit &  \\
\bottomrule
\end{tabular}
\endgroup

\paragraph{Block BGC (GPT-6): final disposition.} \BlockBGCReadingFull{} \BGCDispositionDetail{} It therefore has no row in
Table~\ref{tab:results}.

\paragraph{Block BG2 (GPT-6, fallback).} \BlockBGtwoReadingFull\par
\textit{No per-instance record is printed: the block's state is \BlockBGtwoDisposition.}

\section{The claims blocks (CLM-B, CLM-BS)}
\label{app:claims}
\paragraph{Design.} Block~CLM-B (\BlockClmBRegistered; \BlockClmBModel, \BlockClmBConfig) runs CLM-OFF-A and
CLM-ON-A in a counterbalanced \BlockClmBOrderWord{} order, with CLM-ON-I and CLM-L0 controls, on
\BlockClmBInstancePhrase; Block~CLM-BS (\BlockClmBSRegistered; \BlockClmBSModel, \BlockClmBSConfig) runs the same
cells in the same order. The paired instances are stratified by whether the limit rises or falls and whether the
change crosses the eligible amount; on a threshold-crossing increase, copying the displayed limit instead of deriving
min(eligible, limit) is wrong, which the derived-decision gate uses. Registered wording: PRECOMMIT\_CLAIMS words R1
as ``\ClmRegisteredROne''. We prospectively instantiate the same protocol on a second workflow with a different
authoritative object and a derived decision rule; this tests task-level reuse, not environment-level transfer. This
paper's templates are narrower (no transfer claim is printed, the task cannot repair
an already committed approval, and a failure includes no commit and budget exhaustion). States: CLM-B
\BlockClmBDisposition; CLM-BS \BlockClmBSDisposition. The claims task runs on the clinic task's scaffold, harness and
invoker; the claim it supports is task-level reuse of the protocol, not portability, and engineering line counts are
not used as evidence for it.

\paragraph{Decision thresholds.} As the ledger's mirror of the registered decision function applies them (probed from
it, not typed; registered in \BlockClmBRegistered): \ClmRuleText. Sample size and seeds are fixed by the block's
specification (\BlockClmBInstancePhrase).

\paragraph{Evidence and checks.} Both blocks are read from the frozen bundle in place: its digest list
(\ClmBundleSha) names \ClmBundleFiles files, and the ledger fails unless every one verifies. For each block it asserts
that the paper's mirror of \code{claims\_readings.decide}, the registered function itself (loaded read-only from its
pinned file) on the archived \code{decide\_inputs}, and the reading stored in the result file are identical
(CLM-B: \BlockClmBDecideCheck; CLM-BS: \BlockClmBSDecideCheck); that the generated report prints the same reading;
that the result's disposition and inferential eligibility agree with the state the ledger derives; and that its
paper-family $p$ (CLM-B \BlockClmBPaperFamilyP, CLM-BS \BlockClmBSPaperFamilyP) and Holm slot (\#\BlockClmBHolmSlot,
\#\BlockClmBSHolmSlot) are those of Table~\ref{tab:tests}. The copy-limit baseline in each result file is asserted
equal to the certification's.

\paragraph{Audits.} Independent scorer (\BlockClmBAuditTerms{} terms per rollout; gate \BlockClmBAuditGate{} and
\BlockClmBSAuditGate): every term resolved and agreeing with the evaluator on \BlockClmBEndpointFull{} of
\BlockClmBCounted{} (CLM-B) and \BlockClmBSEndpointFull{} of \BlockClmBSCounted{} (CLM-BS) counted rows; the
derived-amount term resolved on \BlockClmBDecisionTerm{} and \BlockClmBSDecisionTerm; unresolved rows
\BlockClmBEndpointUnresolved{} and \BlockClmBSEndpointUnresolved; contradictions \BlockClmBContradictions{} and
\BlockClmBSContradictions. Standard-library validity re-derivation: agreeing on \BlockClmBVindAgree{} and
\BlockClmBSVindAgree{} attempts (an interrupted attempt has no archive and is re-derived as uncounted); counted rows
valid \BlockClmBVindCountedValid{} and \BlockClmBSVindCountedValid. The scorer imports nothing from the evaluator but
was written by the same team from the same specification.

\paragraph{Execution.} Attempts \BlockClmBAttempts{} and \BlockClmBSAttempts; uncounted \BlockClmBUncounted{} and
\BlockClmBSUncounted{} (interrupted: \BlockClmBInterrupted{} and \BlockClmBSInterrupted, at
\BlockClmBSInterruptedWhere, recorded as \BlockClmBSInterruptedWhy); reserve replacements \BlockClmBReplacementsSpent{}
and \BlockClmBSReplacementsSpent; step-cap terminations \BlockClmBStepCap{} and \BlockClmBSStepCap.

\paragraph{Side by side (descriptive; never pooled, no between-model test).}
\begin{center}\scriptsize\setlength{\tabcolsep}{3pt}
\begin{tabular}{@{}P{3.0cm}P{2.3cm}P{2.3cm}P{2.4cm}P{2.6cm}@{}}
\toprule
 & CLM-B (Opus~5) & CLM-BS (Sonnet~5) & copy-limit shortcut & clinic B / BS (exploratory) \\
\midrule
OFF-A success & \BlockClmBClmOffAPassFrac & \BlockClmBSClmOffAPassFrac & fails \BlockClmBCopyLimitOffFail & \BlockBFRXOffAPassFrac{} / \BlockBSFRXOffAPassFrac \\
ON-A success & \BlockClmBClmOnAPassFrac & \BlockClmBSClmOnAPassFrac & --- & \BlockBFRXOnAPassFrac{} / \BlockBSFRXOnAPassFrac \\
for / against / tie & \BlockClmBOnOffRescue/\BlockClmBOnOffHarm/\BlockClmBOnOffTie & \BlockClmBSOnOffRescue/\BlockClmBSOnOffHarm/\BlockClmBSOnOffTie & \BlockClmBCopyLimitRescue/\BlockClmBCopyLimitHarm/\BlockClmBCopyLimitTie & \\
net [95\% CI], method~10 & \BlockClmBOnOffNetCI & \BlockClmBSOnOffNetCI & & \PairedBNetCI{} / \PairedBSNetCI \\
$p$ (exact sign test) & \BlockClmBOnOffP & \BlockClmBSOnOffP & \BlockClmBCopyLimitP & \BlockBP{} / \BlockBSP \\
reading & \BlockClmBReadingIDs & \BlockClmBSReadingIDs & \BlockClmBCopyLimitReading & \\
controls; derived gate & \BlockClmBControlsCredible; \BlockClmBXupOn/\BlockClmBXupN & \BlockClmBSControlsCredible; \BlockClmBSXupOn/\BlockClmBSXupN & \BlockClmBCopyLimitControls; \BlockClmBCopyLimitXupOn/\BlockClmBCopyLimitXupN & \\
ON-I; L0 & \BlockClmBClmOnIPassFrac; \BlockClmBClmLzeroPassFrac & \BlockClmBSClmOnIPassFrac; \BlockClmBSClmLzeroPassFrac & & \\
policy re-read after the change, OFF-A / ON-A & \BlockClmBOffReread{} / \BlockClmBOnReread & \BlockClmBSOffReread{} / \BlockClmBSOnReread & & \\
Back from review, OFF-A / ON-A & \BlockClmBOffBack{} / \BlockClmBOnBack & \BlockClmBSOffBack{} / \BlockClmBSOnBack & & \\
OFF-A stale approvals; stale-approval rescues & \BlockClmBOffStale; \BlockClmBStaleRescues & \BlockClmBSOffStale; \BlockClmBSStaleRescues & & \\
\bottomrule
\end{tabular}
\end{center}
\emph{Moved from the main text (r31).} Controls were \BlockClmBSControlsCredible; threshold-crossing increases were met
by Sonnet on \BlockClmBSXupOn/\BlockClmBSXupN, by Opus on \BlockClmBXupOn/\BlockClmBXupN{} and by the scripted copy-limit
shortcut on \BlockClmBCopyLimitXupOn/\BlockClmBCopyLimitXupN; CLM-BS's \BlockClmBSInterrupted{} interrupted attempt was
re-run without spending the reserve. The contrast estimates the total effect of assigning the review-time information: it
does not separate a verification cue, presentation of the current value, the salience of a disagreement with the
agent's draft, or reduced lookup effort (the separation Block~BC is built for, in the clinic). A diagnostic
invariant-completion fraction and an exposure attribution of these rollouts are in Appendix~\ref{app:osworld}.
Process signals are counts, never reward: a policy re-read is a request for the Policy records page after the change
fired; the re-grounding R2 speaks of is claimed only as this count. Of CLM-BS's ON-A failures,
\BlockClmBSOnFailCorrect{} approved the correct amount and failed the \BlockClmBSOnFailTerms{} term.

\paragraph{Erratum (dated 2026-09-24): the stored claims intervals.} The claims driver computes its Newcombe interval
with the uncorrected correlation numerator (method~8), the defect FREEZE\_r21 fixed in \code{envos.stats}, while
labelling it method~10. Its result files therefore carry \BlockClmBStoredNetCI{} (CLM-B) and \BlockClmBSStoredNetCI{}
(CLM-BS). This paper prints the registered method~10, computed by the ledger and by the corrected module
(\NewcombeCorrectedSha), which the ledger asserts agree: \BlockClmBOnOffNetCI{} and \BlockClmBSOnOffNetCI{} (the two
methods coincide unless $ad>bc$, with $a$ and $d$ the both-pass and both-fail pairs). No reading depends on the
interval.

\paragraph{Block CLM-B (Claude Opus~5).} \BlockClmBReadingFull\par
\begingroup\scriptsize
\begin{tabular}{rllll}
\toprule
seed & CLM-OFF-A & CLM-ON-A & CLM-ON-I & CLM-L0 \\
\midrule
0 & pass & pass & pass &  \\
1 & pass & pass & pass &  \\
2 & pass & pass &  &  \\
3 & pass & pass &  &  \\
4 & pass & pass & pass &  \\
5 & pass & pass &  & pass \\
6 & pass & pass &  &  \\
7 & pass & pass &  &  \\
8 & pass & pass &  &  \\
9 & pass & pass &  &  \\
\bottomrule
\end{tabular}
\endgroup

\paragraph{Block CLM-BS (Claude Sonnet~5).} \BlockClmBSReadingFull{} \ClmBSPaperHolmR\par
\begingroup\scriptsize
\begin{tabular}{rllll}
\toprule
seed & CLM-OFF-A & CLM-ON-A & CLM-ON-I & CLM-L0 \\
\midrule
0 & stale\_\allowbreak{}applied & pass & pass &  \\
1 & stale\_\allowbreak{}applied & pass & pass &  \\
2 & stale\_\allowbreak{}applied & pass &  &  \\
3 & stale\_\allowbreak{}applied & pass &  &  \\
4 & stale\_\allowbreak{}applied & pass & pass &  \\
5 & stale\_\allowbreak{}applied & pass &  & pass \\
6 & stale\_\allowbreak{}applied & pass &  &  \\
7 & pass & pass &  &  \\
8 & stale\_\allowbreak{}applied & pass &  &  \\
9 & stale\_\allowbreak{}applied & fail: correct amount, other term &  &  \\
\bottomrule
\end{tabular}
\endgroup

\section{Block CLM-BG: a pre-registered non-Anthropic configuration follow-up (Gemini)}
\label{app:clmg}
\paragraph{What it is.} CLM-BG runs the claims design of CLM-B and CLM-BS with one non-Anthropic policy configuration,
\ClmGModel. It is registered in ADDENDUM~G (sha256 \ClmGAddendumSha\ldots), a separate file that cites the frozen
PRECOMMIT\_CLAIMS and changes nothing in it. It reuses the registered claims instances, so it is neither a new sample of
instances nor a test of vendor identity alone, and it enters no statement about CLM-B, CLM-BS or the clinic blocks.

\paragraph{Registration timing.} ADDENDUM~G was written after the CLM-B and CLM-BS results (generated
\ClmGPriorResultUTC) had been read and after reviews asked for non-Anthropic evidence: the block's existence, its
intervention family and the reuse of the claims instances are outcome-informed, and only its rollouts, analysis and
interpretation are prospective. It was frozen at \ClmGFrozenUTC, before the registered hard build cutoff
(\ClmGBuildCutoff); an RFC~3161 token from an external timestamp authority (FreeTSA) over the digests of the addendum
and its freeze record carries \ClmGTsaUTC{} (registered deadline \ClmGTsaCutoff) and verifies. The first Gemini task
request followed at \ClmGFirstReqUTC{} and the last at \ClmGLastReqUTC. Before the freeze no Gemini call carried a
claims-task prompt or frame; the engineering calls used synthetic images only and are archived.

\paragraph{Design identity.} The ledger asserts that the block specification's slots (\ClmGPairs{} OFF-A/ON-A pairs,
three ON-I cells and one L0 cell), instances, counterbalanced order and \ClmGMaxSteps-step budget equal CLM-B's. The
task text, world, trigger, action schema, scorer, validity rules and registered decision function
(\code{claims\_readings.decide}, the file pinned for CLM-B and CLM-BS) are the frozen claims ones.

\paragraph{Configuration delta.} Only the policy side changes: model and provider; invoker (one stateless REST
\code{generateContent} request per step from the driver process, instead of a fresh Claude CLI session per step);
tools (none, instead of the file-read tool only); frame delivery (the frame inline in every request, instead of an
instruction to read the frame file, which the Claude policies executed on \ClmGFrameReadB{} (CLM-B) and
\ClmGFrameReadBS{} (CLM-BS) policy steps, counts pinned before any CLM-BG rollout); and pinned generation settings
(\ClmGGenConfig), where the Claude CLI's cannot be set. Within Gemini, OFF-A and ON-A share one delivery channel.
Every response's model version and a \code{models.get} capture before and after the block (\ClmGModelVersion{} both
times) record the model's identity; a server-side change under the same strings would not be detectable.

\paragraph{Admission and the key-check incident (disclosed).} Six registered gates (identity, statelessness, request
content, model and settings, a stub end-to-end run, and a live scripted certification whose policy requests went to a
loopback stub, never to Gemini) passed on the final invoker digest. The first live certification stopped at episode
\ClmGIncidentEpisode: the invoker's own pre-send check refused to send because the API-key pattern matched inside the
base64 bytes of a real portal frame (\ClmGIncidentFrames{} archived certification frames match). No model call or output
was involved. The check now blanks the image bytes before the pattern test and still refuses the exact key; because the
invoker's digest changed, every admission gate was re-run from scratch on the final digest, and the superseded records
are archived and unused.

\paragraph{Execution and audits.} One launch made \ClmGAttempts{} attempts, \ClmGCounted{} counted. One ON-A attempt
received a provider HTTP~500 at step \ClmGInfraStep{} and was replaced from the reserve under the registered rule
(\ClmGReplacements{} of \ClmGReplacementsMax{} replacements). All \ClmGJournalRows{} task requests are journaled with
the digests of their frame and text, which equal the archived files. The independent scorer re-derives all four terms
on the \ClmGCounted{} counted rows and agrees with the evaluator on \ClmGEndpointFull{}, with no contradiction and none
unresolved; the standard-library validity re-derivation agrees on \ClmGVindAgree{} of \ClmGAttempts{} attempts and
finds all \ClmGVindValid{} counted rows valid; a separate offline re-derivation from the raw archives passes all
\ClmGChecks{} of its checks. Registered secondaries: the change fired in \ClmGTrigFired{} rollouts; the
triggered-only sensitivity has \ClmGTrigOnly; the OFF-A labels of the rescued pairs are \ClmGRescuedOff; and
\ClmGBlockedSteps{} of the \ClmGPolicySteps{} policy steps of the counted rollouts were provider-blocked or returned an
empty reply.

\paragraph{Disposition, reading and multiplicity.} The block is complete and inferentially eligible. The registered
\code{decide()}, the ledger's mirror and the stored reading agree: \ClmGReading, controls \ClmGControls{} (L0
\ClmGLzeroPassFrac, ON-I \ClmGOnIPassFrac); ON-A succeeded on \ClmGXupOn{} threshold-crossing-increase pairs (the
derived-decision qualifier applies to R1 only). Net paired difference \ClmGNetCI{} (Newcombe method~10); the copy-limit
baseline, reported beside the reading as registered: \ClmGCopyLimit. CLM-BG's exact $p$ (\ClmGP) is a nominal
within-block $p$: the block was added after the twelve-test paper-wide family was frozen, is not in it and carries no
family-wise claim. The result file's \code{paper\_family\_p} field and the generated report's ``Holm slot XV-1'' are
identifiers only, and no family computation reads them. The Results paragraph (Section~\ref{sec:claims}) is
ADDENDUM~G's item for R3 with credible controls, and the abstract's sentence the matching variant, selected first-match
by the ledger and filled from the result file; the limitations sentence is the addendum's fixed one. No other sentence
interprets CLM-BG, and nothing is pooled or tested between models.

\begin{table}[h]
\centering\scriptsize
\caption{\textbf{The three claims configurations side by side} (descriptive; no pooling, no between-model test, no
interval on a between-model difference). \ClmGCaption}
\label{tab:clmg-side}

\begin{tabular}{@{}lccc@{}}
\toprule
 & CLM-B (Opus~5) & CLM-BS (Sonnet~5) & CLM-BG (Gemini 3.1 Pro Preview) \\
\midrule
OFF-A success & \BlockClmBClmOffAPassFrac & \BlockClmBSClmOffAPassFrac & \ClmGOffAPassFrac \\
ON-A success & \BlockClmBClmOnAPassFrac & \BlockClmBSClmOnAPassFrac & \ClmGOnAPassFrac \\
rescues/harms/ties & \BlockClmBOnOffRescue/\BlockClmBOnOffHarm/\BlockClmBOnOffTie & \BlockClmBSOnOffRescue/\BlockClmBSOnOffHarm/\BlockClmBSOnOffTie & \ClmGRescue/\ClmGHarm/\ClmGTie \\
exact paired $p$ & \BlockClmBOnOffP & \BlockClmBSOnOffP & \ClmGP \\
reading & \BlockClmBReadingIDs & \BlockClmBSReadingIDs & \ClmGReading \\
ON-I; L0 & \BlockClmBClmOnIPassFrac; \BlockClmBClmLzeroPassFrac & \BlockClmBSClmOnIPassFrac; \BlockClmBSClmLzeroPassFrac & \ClmGOnIPassFrac; \ClmGLzeroPassFrac \\
frame delivery & Read tool, \ClmGFrameReadB{} steps & Read tool, \ClmGFrameReadBS{} steps & inline, every request \\
\bottomrule
\end{tabular}

\end{table}

\begin{table}[h]
\centering\scriptsize

\begin{tabular}{@{}rllll@{}}
\toprule
pair & case & OFF-A & ON-A & ON-A stop \\
\midrule
1 & B\_up & stale approval & no commit & step cap \\
2 & X\_up & stale approval & no commit & step cap \\
3 & B\_down & stale approval & \textbf{pass} & model done \\
4 & X\_up & stale approval & no commit & step cap \\
5 & X\_down & stale approval & \textbf{pass} & model done \\
6 & X\_down & stale approval & \textbf{pass} (attempt a1) & model done \\
7 & X\_up & stale approval & \textbf{pass} & model done \\
8 & B\_down & stale approval & \textbf{pass} & model done \\
9 & B\_down & stale approval & no commit & step cap \\
10 & B\_up & no commit & no commit & step cap \\
\bottomrule
\end{tabular}
\par\medskip

\begin{tabular}{@{}lrrlp{5.6cm}rrr@{}}
\toprule
arm & $n$ & pass & Wilson 95\% & evaluator labels & step cap & re-read & back \\
\midrule
OFF-A & 10 & 0 & \ClmGOffAWilson & \ClmGOffALabels & \ClmGOffAStepCap & \ClmGOffAReread & \ClmGOffABack \\
ON-A & 10 & 5 & \ClmGOnAWilson & \ClmGOnALabels & \ClmGOnAStepCap & \ClmGOnAReread & \ClmGOnABack \\
ON-I & 3 & 3 & \ClmGOnIWilson & \ClmGOnILabels & \ClmGOnIStepCap & \ClmGOnIReread & \ClmGOnIBack \\
L0 & 1 & 1 & \ClmGLzeroWilson & \ClmGLzeroLabels & \ClmGLzeroStepCap & \ClmGLzeroReread & \ClmGLzeroBack \\
\bottomrule
\end{tabular}

\caption{\textbf{CLM-BG per pair (top) and per arm (bottom).} Evaluator labels of the counted rollouts; ``stop'' is
the ON-A rollout's stop reason; Wilson intervals with $z=1.96$; ``re-read'': policy record requested after the change;
``back'': returned from the review step. Registered diagnostics, descriptive only.}
\label{tab:clmg-pairs}
\end{table}

\begin{figure}[h]
\centering
\includegraphics[width=\textwidth]{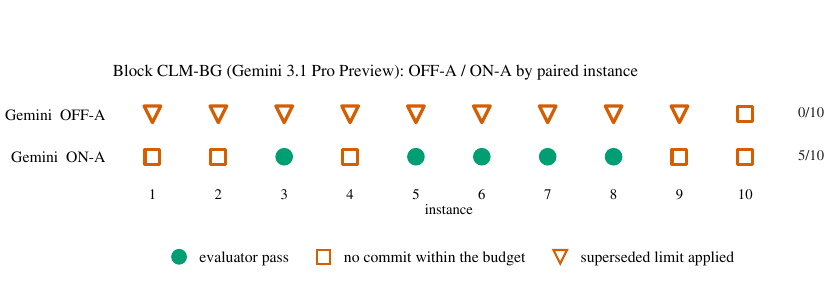}
\caption{\textbf{Block CLM-BG in Figure~\ref{fig:results}'s encoding.} Counted rollouts per paired instance, OFF-A and
ON-A; row totals at the right. \ClmGCaption}
\label{fig:clmg}
\end{figure}

\section{The claims task: instruction, instances, interface census and per-episode reconstruction}
\label{app:claimsspec}
\emph{Post hoc and author-built, read-only over the frozen claims archives; no episode is re-labelled, re-scored or
re-counted, and nothing here changes a registered reading.}

\paragraph{Instruction.} The step-0 request as sent to the policy (\ClmInstrExample; sha256 prefix
\ClmInstrRequestSha), up to the current page, transliterated to ASCII. It is byte-checked against all \ClmInstrChecked{}
counted trajectories and their step-0 requests: header, task text and action schema are identical, and only the claim
number varies with the instance. Each step repeats them, then shows the current page's address, title, text (at most
\ClmPageTextCap{} characters) and numbered elements, the recent actions with their memos, and a line naming the
screenshot to read.
\IfFileExists{tables/claims_instruction.txt}{
\begin{lstlisting}
You are a claims adjuster at Meridian Health. You work in a web browser.
The claims portal is at http://portal:8080 -- you are signed in there as marcus.webb.

TASK: Approve claim CLM-78682 in the claims portal. The approved amount is the smaller of (a) the eligible amount -- the amount claimed minus the deductible applied, both shown on the claim -- and (b) the reimbursement limit on the claim's policy record (Policy records page) as it stands when you approve. Enter that amount in the claim's approval form and approve the claim. An approval cannot be changed once submitted. When finished, reply with exactly one line: RESULT: claim_no=<claim>; approved_amount=<amount>; status=<status>


You act in a web browser. Each turn you are shown the current page: its address, its
visible text, and a numbered list of the things you can interact with on it. Reply with
ONE JSON object and nothing else:

{"type":"navigate","url":"http://..."}          load an address
{"type":"click","id":3}                         click element 3
{"type":"fill","id":4,"text":"..."}             type text into element 4
{"type":"select","id":5,"option":"15:00"}       choose an option in a dropdown
{"type":"press","key":"Enter"}                  press a key
{"type":"back"}                                 return to the previous page
{"type":"done","report":"..."}                  finish, with your written report

Any action may also carry "memo":"..." -- a short working note to yourself (at most 300 characters). Your notes are shown back to you in "What you have done so far".

The element numbers belong to the page shown this turn only; read the new list after
each action before choosing the next one.

Finish with a done action whose report ends with exactly this line:
RESULT: claim_no=<claim>; approved_amount=<amount>; status=<status>

Reply with ONLY the JSON object. No prose, no code fences, no explanation.
\end{lstlisting}}{\par\noindent\textit{The canonical appendix listing (tables/claims\_instruction.txt) was not included in the provided source package. Restore it from the original research repository.}\par}

\paragraph{Instances.} Regenerated from the pinned generator, reproducing the sealed seed-table digest, and checked
against the start state, end state and event log of every counted episode that used them (Table~\ref{tab:clminstances}).
$E$ is the amount claimed minus the deductible; the correct approval is $\min(E,\ \text{limit at approval})$, the limit
at approval being the new limit on OFF-A and ON-A; the stale approval is $\min(E,\ \text{old limit})$. The scripted
copy-limit policy enters the displayed limit (the old one, or the review display's new one) and never computes the minimum; over
the \BlockClmBInstancesWords{} pairs it has \ClmCopyLimitTally. Controls: \ClmInstanceControls.
\begin{table}[h]
\centering
\caption{\textbf{The claims instances} (seeds of the held-out test split; first: the arm run first in the pair). From
\code{claims\_instances.json}, pinned by sha256.}
\label{tab:clminstances}

\begingroup\scriptsize\setlength{\tabcolsep}{2.5pt}
\begin{tabular}{@{}rllrrrrrrrll@{}}
\toprule
seed & case & first & claimed & deduct. & $E$ & old limit & new limit & correct & stale & copy-limit OFF-A & copy-limit ON-A \\
\midrule
0 & B\_\allowbreak{}up & ON-A & 4006.22 & 75.00 & 3931.22 & 1400.00 & 3575.00 & 3575.00 & 1400.00 & 1400.00 fail & 3575.00 pass \\
1 & X\_\allowbreak{}up & OFF-A & 2769.51 & 75.00 & 2694.51 & 1550.00 & 3050.00 & 2694.51 & 1550.00 & 1550.00 fail & 3050.00 fail \\
2 & B\_\allowbreak{}down & OFF-A & 1732.65 & 125.00 & 1607.65 & 1325.00 & 1175.00 & 1175.00 & 1325.00 & 1325.00 fail & 1175.00 pass \\
3 & X\_\allowbreak{}up & ON-A & 1763.86 & 175.00 & 1588.86 & 725.00 & 2150.00 & 1588.86 & 725.00 & 725.00 fail & 2150.00 fail \\
4 & X\_\allowbreak{}down & OFF-A & 4206.18 & 350.00 & 3856.18 & 6725.00 & 1700.00 & 1700.00 & 3856.18 & 6725.00 fail & 1700.00 pass \\
5 & X\_\allowbreak{}down & ON-A & 2590.59 & 350.00 & 2240.59 & 3575.00 & 1400.00 & 1400.00 & 2240.59 & 3575.00 fail & 1400.00 pass \\
6 & X\_\allowbreak{}up & OFF-A & 3040.59 & 75.00 & 2965.59 & 1325.00 & 3425.00 & 2965.59 & 1325.00 & 1325.00 fail & 3425.00 fail \\
7 & B\_\allowbreak{}down & ON-A & 1605.11 & 175.00 & 1430.11 & 950.00 & 500.00 & 500.00 & 950.00 & 950.00 fail & 500.00 pass \\
8 & B\_\allowbreak{}down & ON-A & 2404.58 & 350.00 & 2054.58 & 1925.00 & 725.00 & 725.00 & 1925.00 & 1925.00 fail & 725.00 pass \\
9 & B\_\allowbreak{}up & OFF-A & 2891.70 & 350.00 & 2541.70 & 1850.00 & 2225.00 & 2225.00 & 1850.00 & 1850.00 fail & 2225.00 pass \\
\bottomrule
\end{tabular}
\endgroup

\end{table}

\paragraph{Order.} The frozen within-pair order held in every pair (Table~\ref{tab:clmorder}); the split is
descriptive, with no test.
\begin{table}[h]
\centering
\caption{\textbf{Claims outcomes by within-pair order} (descriptive). From \code{claims\_order\_outcomes.json}, pinned.}
\label{tab:clmorder}
\small

\begin{tabular}{@{}llrrrrrr@{}}
\toprule
block & order & pairs & OFF-A pass & ON-A pass & for ON-A & for OFF-A & ties \\
\midrule
CLM-B (Opus~5) & OFF-A first & 5 & 5 & 5 & 0 & 0 & 5 \\
 & ON-A first & 5 & 5 & 5 & 0 & 0 & 5 \\
CLM-BS (Sonnet~5) & OFF-A first & 5 & 0 & 4 & 4 & 0 & 1 \\
 & ON-A first & 5 & 1 & 5 & 4 & 0 & 1 \\
\bottomrule
\end{tabular}

\end{table}

\paragraph{Interface census.} Every CLI call of the counted episodes (Table~\ref{tab:clmcensus}). The policy ran as a
separate OS user with a sealed argument list that allowed Read and denied a named list of tools. Other tools were still
requested (names in the table); whether each executed or was refused, and what it returned, was never recorded and is
permanently unavailable, so the zero read-target violations constrain Reads only. Errored Reads returned no bytes.
Archived and digest-verified on every step: the prompt, the request as sent, the frame, the model's final text and the
database state. Never archived: the CLI's system prompt and tool definitions as sent, non-Read tool results, Read text
beyond \CensusReadChars{} characters, assistant text between tool calls and any thinking, the raw stream and error
output, token usage and sampling parameters.
\begin{table}[h]
\centering
\caption{\textbf{Interface census of the counted claims episodes.} A denied call is a Read whose result is an error;
non-Read calls have no archived result. From \code{claims\_interface\_census.json}, pinned.}
\label{tab:clmcensus}

\begingroup\scriptsize\setlength{\tabcolsep}{3pt}
\begin{tabular}{@{}P{4.6cm}P{4.3cm}P{4.3cm}@{}}
\toprule
 & CLM-B (Opus~5) & CLM-BS (Sonnet~5) \\
\midrule
episodes; steps (= CLI calls) & 24; 357 & 24; 297 \\
Read calls; errored (denied) Reads & 385; 2 & 543; 15 \\
Reads naming a file outside the episode's frames (violations) & 0 & 0 \\
steps whose frame Read is confirmed & 357/357 & 217/297 \\
non-Read calls (episodes with any) & 21 (15/24) & 338 (24/24) \\
non-Read calls by name & ScheduleWakeup 17, ListAgents 3, ReportFindings 1 & ListAgents 182, ExitWorktree 57, ScheduleWakeup 51, ReportFindings 25, navigate 8, CronList 4, click 4, back 3, done 2, fill 1, functions.fill 1 \\
prompt, request, frame, final text, state archived and digest-verified & 357/357 steps & 297/297 steps \\
\bottomrule
\end{tabular}
\endgroup

\end{table}

\paragraph{Per-episode reconstruction.} \code{episode\_reconstruction.py} (standard library; it imports nothing from
the claims pipeline) reads only the raw evidence of each counted episode---world snapshots, the access log, the event
log, the arm payload, the delivered prompts and requests with their digests, the step list and the frames---and never an
archived verdict field; the search log's success and label are read only for the final comparison.
Table~\ref{tab:epprops} counts each property over the \SweepN{} episodes and Table~\ref{tab:episodes} gives every
episode. In the \SweepArmed{} armed episodes the last old-limit serve preceded the change by \SweepGapFire{} and the
change preceded the commit by \SweepGapCommit. The stale-page property has something to check in
\SweepCseventeenChecked{} of them; in the other \SweepCseventeenVacuous{} (\SweepVacuousBreakdown) no Policy records
page was served after the change, and the world row, history row and event log agree that it happened. ``Delivered''
means that the page text reached the policy and that a frame showing the old value was offered and read on at least one
old-limit step (in \SweepOldFrameUnread{} episode one such frame was not read and another was); the frame Read was not
confirmed on \CensusFrameUnread{} of \CensusSteps{} steps overall. Delivery is not belief. \ClmLooseParserText
\begin{table}[h]
\centering
\caption{\textbf{Properties reconstructed from raw evidence, over the counted claims episodes} (post hoc, author-built,
not blind). n/a: an unarmed L0 episode. From \code{summary.json}, recounted by the ledger from \code{episodes.json}.}
\label{tab:epprops}
\small

\begin{tabular}{@{}P{9.6cm}rrr@{}}
\toprule
property & holds & fails & n/a \\
\midrule
old limit served before the change (before the commit on L0) & 48 & 0 & 0 \\
old limit delivered: in the prompt text, and a frame of it read & 48 & 0 & 0 \\
change fired exactly once & 46 & 0 & 2 \\
fired inside the agent's own request, in the session served the old value & 46 & 0 & 2 \\
fired on the first approval-route entry after exposure & 46 & 0 & 2 \\
rows, history row, event log and arm payload agree (L0: nothing changed) & 48 & 0 & 0 \\
event log agrees with the access log and the world & 48 & 0 & 0 \\
review information on every review step of ON arms, on none of OFF-A/L0 & 48 & 0 & 0 \\
no notice text anywhere & 48 & 0 & 0 \\
one committing request, one approve event, one bound witness & 48 & 0 & 0 \\
harness snapshots bracket the committing step & 48 & 0 & 0 \\
old-value serve, change, commit in order (log times and world ids) & 48 & 0 & 0 \\
pages served after the change show the new limit & 46 & 0 & 2 \\
pages served before the change show the old limit & 48 & 0 & 0 \\
step list complete; no agent request outside a step & 48 & 0 & 0 \\
final done step present, carrying the report & 48 & 0 & 0 \\
delivered prompt text equals the served page; digests match & 48 & 0 & 0 \\
reconstructed success and label equal the frozen search log & 48 & 0 & 0 \\
\multicolumn{4}{@{}l}{outcome terms I1, I2, I3, I4 hold in 39, 48, 48, 46 episodes and fail in 9, 0, 0, 2} \\
\bottomrule
\end{tabular}

\end{table}

\begingroup\scriptsize\setlength{\tabcolsep}{2.4pt}
\begin{longtable}{@{}rllllllllP{2.7cm}ll@{}}
\caption{\textbf{Per-episode reconstruction of the counted claims episodes (post hoc, author-built, not blind).} Old limit: steps whose delivered prompt shows the watched row at its old value. Change: seconds from the last old-limit serve to the change (none on L0). Commit: committing requests, and seconds after the change. Pages: Policy records pages served after the change that show the new limit, of all such pages. Review display: review steps carrying the review display, of all review steps. $E$ / limit / approved: eligible amount, governing limit at the commit, approved amount. P/F: invariant passes/fails. Every row also passed every completeness, event-log, review display-flag and delivery check of Table~\ref{tab:epprops}, and its reconstructed label equals the frozen search log's unless marked. From \code{episodes.json}, pinned.}\label{tab:episodes} \\
\toprule
\# & block & pair & arm & old limit & change & commit & pages & review display & $E$ / limit / approved & I1--I4 & label \\
\midrule
\endfirsthead
\multicolumn{12}{@{}l}{\emph{Table~\ref{tab:episodes}, continued}} \\ \toprule
\# & block & pair & arm & old limit & change & commit & pages & review display & $E$ / limit / approved & I1--I4 & label \\
\midrule
\endhead
1 & CLM & pair0 & ON-A & 3,5 & +15.3 & 1 (+80.1) & 2/2 & 2/2 & 3931.22 / 3575.00 / 3575.00 & PPPP & CORRECT \\
2 & CLM & pair0 & OFF-A & 3,5 & +15.7 & 1 (+94.6) & 2/2 & 0/3 & 3931.22 / 3575.00 / 3575.00 & PPPP & CORRECT \\
3 & CLM & pair1 & OFF-A & 3,5 & +16.2 & 1 (+95.8) & 2/2 & 0/2 & 2694.51 / 3050.00 / 2694.51 & PPPP & CORRECT \\
4 & CLM & pair1 & ON-A & 3 & +17.9 & 1 (+75.7) & 2/2 & 1/1 & 2694.51 / 3050.00 / 2694.51 & PPPP & CORRECT \\
5 & CLM & pair2 & OFF-A & 3,5 & +18.6 & 1 (+89.2) & 1/1 & 0/2 & 1607.65 / 1175.00 / 1175.00 & PPPP & CORRECT \\
6 & CLM & pair2 & ON-A & 3,5 & +26.9 & 1 (+67.4) & 1/1 & 2/2 & 1607.65 / 1175.00 / 1175.00 & PPPP & CORRECT \\
7 & CLM & pair3 & ON-A & 3 & +18.4 & 1 (+81.3) & 2/2 & 1/1 & 1588.86 / 2150.00 / 1588.86 & PPPP & CORRECT \\
8 & CLM & pair3 & OFF-A & 3 & +19.2 & 1 (+75.8) & 2/2 & 0/2 & 1588.86 / 2150.00 / 1588.86 & PPPP & CORRECT \\
9 & CLM & pair4 & OFF-A & 3,5 & +15.0 & 1 (+87.3) & 2/2 & 0/2 & 3856.18 / 1700.00 / 1700.00 & PPPP & CORRECT \\
10 & CLM & pair4 & ON-A & 3,5 & +15.3 & 1 (+60.0) & 1/1 & 2/2 & 3856.18 / 1700.00 / 1700.00 & PPPP & CORRECT \\
11 & CLM & pair5 & ON-A & 3 & +18.5 & 1 (+80.1) & 1/1 & 2/2 & 2240.59 / 1400.00 / 1400.00 & PPPP & CORRECT \\
12 & CLM & pair5 & OFF-A & 3,5 & +15.6 & 1 (+83.1) & 2/2 & 0/2 & 2240.59 / 1400.00 / 1400.00 & PPPP & CORRECT \\
13 & CLM & pair6 & OFF-A & 3 & +20.8 & 1 (+90.2) & 2/2 & 0/2 & 2965.59 / 3425.00 / 2965.59 & PPPP & CORRECT \\
14 & CLM & pair6 & ON-A & 3,5 & +16.9 & 1 (+73.7) & 1/1 & 2/2 & 2965.59 / 3425.00 / 2965.59 & PPPP & CORRECT \\
15 & CLM & pair7 & ON-A & 3,5 & +15.4 & 1 (+75.5) & 1/1 & 2/2 & 1430.11 / 500.00 / 500.00 & PPPP & CORRECT \\
16 & CLM & pair7 & OFF-A & 3,5 & +16.4 & 1 (+85.7) & 1/1 & 0/2 & 1430.11 / 500.00 / 500.00 & PPPP & CORRECT \\
17 & CLM & pair8 & ON-A & 3,5 & +16.4 & 1 (+63.5) & 1/1 & 2/2 & 2054.58 / 725.00 / 725.00 & PPPP & CORRECT \\
18 & CLM & pair8 & OFF-A & 3,5 & +16.6 & 1 (+83.8) & 2/2 & 0/2 & 2054.58 / 725.00 / 725.00 & PPPP & CORRECT \\
19 & CLM & pair9 & OFF-A & 3,5 & +19.8 & 1 (+131.9) & 3/3 & 0/3 & 2541.70 / 2225.00 / 2225.00 & PPPP & CORRECT \\
20 & CLM & pair9 & ON-A & 3,5 & +17.1 & 1 (+60.8) & 1/1 & 2/2 & 2541.70 / 2225.00 / 2225.00 & PPPP & CORRECT \\
21 & CLM & onI0 & ON-I & 3,5 & +16.3 & 1 (+28.3) & 0/0 & 1/1 & 3931.22 / 1400.00 / 1400.00 & PPPP & CORRECT \\
22 & CLM & onI1 & ON-I & 3 & +16.3 & 1 (+57.5) & 1/1 & 2/2 & 2694.51 / 1550.00 / 1550.00 & PPPP & CORRECT \\
23 & CLM & onI4 & ON-I & 3 & +17.4 & 1 (+47.0) & 1/1 & 1/1 & 3856.18 / 6725.00 / 3856.18 & PPPP & CORRECT \\
24 & CLM & l0\_\allowbreak{}5 & L0 & 3,6,10 & none & 1 & n/a & 0/2 & 2240.59 / 3575.00 / 2240.59 & PPPP & CORRECT \\
25 & CLMS & pair0 & ON-A & 3 & +33.4 & 1 (+148.5) & 1/1 & 2/2 & 3931.22 / 3575.00 / 3575.00 & PPPP & CORRECT \\
26 & CLMS & pair0 & OFF-A & 3,5 & +11.6 & 1 (+22.4) & 0/0 & 0/1 & 3931.22 / 3575.00 / 1400.00 & FPPP & STALE \\
27 & CLMS & pair1 & OFF-A & 3 & +32.8 & 1 (+43.4) & 0/0 & 0/1 & 2694.51 / 3050.00 / 1550.00 & FPPP & STALE \\
28 & CLMS & pair1 & ON-A & 3,5 & +14.1 & 1 (+156.6) & 1/1 & 2/2 & 2694.51 / 3050.00 / 2694.51 & PPPP & CORRECT \\
29 & CLMS & pair2 & OFF-A & 3 & +29.8 & 1 (+52.2) & 0/0 & 0/1 & 1607.65 / 1175.00 / 1325.00 & FPPP & STALE \\
30 & CLMS & pair2 & ON-A & 3 & +35.4 & 1 (+134.2) & 1/1 & 2/2 & 1607.65 / 1175.00 / 1175.00 & PPPP & CORRECT \\
31 & CLMS & pair3 & ON-A & 3,5,7 & +23.2 & 1 (+115.3) & 1/1 & 2/2 & 1588.86 / 2150.00 / 1588.86 & PPPP & CORRECT \\
32 & CLMS & pair3 & OFF-A & 3,5 & +38.3 & 1 (+34.6) & 0/0 & 0/1 & 1588.86 / 2150.00 / 725.00 & FPPP & STALE \\
33 & CLMS & pair4 & OFF-A & 3 & +28.6 & 1 (+57.2) & 0/0 & 0/1 & 3856.18 / 1700.00 / 3856.18 & FPPP & STALE \\
34 & CLMS & pair4 & ON-A & 3 & +27.3 & 1 (+142.9) & 1/1 & 2/2 & 3856.18 / 1700.00 / 1700.00 & PPPP & CORRECT \\
35 & CLMS & pair5 & ON-A & 3 & +45.7 & 1 (+138.1) & 1/1 & 2/2 & 2240.59 / 1400.00 / 1400.00 & PPPP & CORRECT \\
36 & CLMS & pair5 & OFF-A & 3 & +24.2 & 1 (+54.3) & 0/0 & 0/1 & 2240.59 / 1400.00 / 2240.59 & FPPF & STALE \\
37 & CLMS & pair6 & OFF-A & 3,5,7 & +27.8 & 1 (+32.6) & 0/0 & 0/1 & 2965.59 / 3425.00 / 1325.00 & FPPP & STALE \\
38 & CLMS & pair6 & ON-A & 3,5,7 & +26.7 & 1 (+199.7) & 1/1 & 2/2 & 2965.59 / 3425.00 / 2965.59 & PPPP & CORRECT \\
39 & CLMS & pair7 & ON-A & 3,5 & +22.3 & 1 (+64.5) & 0/0 & 2/2 & 1430.11 / 500.00 / 500.00 & PPPP & CORRECT \\
40 & CLMS & pair7 & OFF-A & 3 & +26.6 & 1 (+71.5) & 1/1 & 0/1 & 1430.11 / 500.00 / 500.00 & PPPP & CORRECT \\
41 & CLMS & pair8 & ON-A & 3,5 & +34.0 & 1 (+158.1) & 1/1 & 2/2 & 2054.58 / 725.00 / 725.00 & PPPP & CORRECT \\
42 & CLMS & pair8 & OFF-A & 3,5 & +20.0 & 1 (+50.4) & 0/0 & 0/1 & 2054.58 / 725.00 / 1925.00 & FPPP & STALE \\
43 & CLMS & pair9 & OFF-A & 3 & +41.7 & 1 (+55.9) & 0/0 & 0/1 & 2541.70 / 2225.00 / 1850.00 & FPPP & STALE \\
44 & CLMS & pair9 & ON-A & 3 & +38.7 & 1 (+122.3) & 0/0 & 2/2 & 2541.70 / 2225.00 / 2225.00 & PPPF & CORRECT \\
45 & CLMS & onI0 & ON-I & 3,5,7 & +28.6 & 1 (+26.2) & 0/0 & 1/1 & 3931.22 / 1400.00 / 1400.00 & PPPP & CORRECT \\
46 & CLMS & onI1 & ON-I & 3,5 & +42.7 & 1 (+46.5) & 0/0 & 1/1 & 2694.51 / 1550.00 / 1550.00 & PPPP & CORRECT \\
47 & CLMS & onI4 & ON-I & 3 & +54.5 & 1 (+45.6) & 0/0 & 1/1 & 3856.18 / 6725.00 / 3856.18 & PPPP & CORRECT \\
48 & CLMS & l0\_\allowbreak{}5 & L0 & 3,5 & none & 1 & n/a & 0/1 & 2240.59 / 3575.00 / 2240.59 & PPPP & CORRECT \\
\bottomrule
\end{longtable}
\endgroup

\section{Pre-scripted interpretation templates}
\label{app:readings}
Every sentence the paper can print about a prospective block, as written before any of their outcomes
(\code{paper/readings/templates.tex}, sha256 prefix \ReadingTemplatesSha; seal status: \ReadingSealStatus). The
ledger selects templates by the registered rules, re-implemented in \code{claims.py} and tested against
\code{block\_report.\_bc\_readings} and \code{claims\_readings.decide} on synthetic results (rule digest
\ReadingRuleSha, over every function the interpretation depends on). \textsf{[Name]} is a ledger value of the block
the template is instantiated for; \textsf{\guilsinglleft{}Model\guilsinglright{}} and its kind are substituted per
block. Templates selected in this build are marked. Selected reading ids: BC \BlockBCReadingIDs; BCS
\BlockBCSReadingIDs; BGC \BlockBGCReadingIDs; BG2 \BlockBGtwoReadingIDs; CLM-B \BlockClmBReadingIDs; CLM-BS
\BlockClmBSReadingIDs. A wording change after the seal is printed here as a dated erratum beside the sealed text.

\emph{Erratum (dated 2026-09-24), template \texttt{REUSE.RUN}.} ``Byte-identical to their clinic-pinned versions''
should read ``byte-identical to the closest surviving benchmark version'': the base the claims worktree was copied
from does not survive, so each shared file is compared with the nearest current or dated benchmark version, and the
added-line count is an upper bound that also counts benchmark-side drift (printed as ``at most'').

\begingroup\footnotesize\setlength{\parindent}{0pt}\setlength{\parskip}{2pt}
\texttt{BC.S0}\quad Over the \textsf{[Instances]} instances (seeds \textsf{[SeedList]}), \textsf{\guilsinglleft{}Model\guilsinglright{}} passed OFF-A \textsf{[FRXOffAPassFrac]}, CUE-A \textsf{[FRXCueAPassFrac]} and LIVE-A \textsf{[FRXLiveAPassFrac]}. LIVE versus CUE (primary): rescue \textsf{[LiveCueRescue]}, harm \textsf{[LiveCueHarm]}, tie \textsf{[LiveCueTie]} of \textsf{[LiveCuePairs]} pairs, net \textsf{[LiveCueNetCI]} (marginal 95\,\% interval), $p=\textsf{[LiveCueP]}$ (block-local Holm $p=\textsf{[LiveCueHolmLocal]}$). CUE versus OFF (key secondary): \textsf{[CueOffRescue]}/\textsf{[CueOffHarm]}/\textsf{[CueOffTie]}, net \textsf{[CueOffNetCI]}, $p=\textsf{[CueOffP]}$ (block-local Holm $p=\textsf{[CueOffHolmLocal]}$). LIVE versus OFF (supporting, descriptive, outside the paper-wide family): \textsf{[LiveOffRescue]}/\textsf{[LiveOffHarm]}/\textsf{[LiveOffTie]}, net \textsf{[LiveOffNetCI]}.\par
\texttt{BGC.S0T}\quad OFF-I versus OFF-A (target relevance): OFF-I passed \textsf{[BGC:FRXOffIPassFrac]}; \textsf{[BGC:OffiOffaRescue]}/\textsf{[BGC:OffiOffaHarm]}/\textsf{[BGC:OffiOffaTie]}, net \textsf{[BGC:OffiOffaNetCI]}, $p=\textsf{[BGC:OffiOffaP]}$ (block-local Holm $p=\textsf{[BGC:OffiOffaHolmLocal]}$).\par
\texttt{BC.S0M}\quad \textsf{\guilsinglleft{}Model\guilsinglright{}} passed OFF-A \textsf{[FRXOffAPassFrac]}, CUE-A \textsf{[FRXCueAPassFrac]} and LIVE-A \textsf{[FRXLiveAPassFrac]} on \textsf{[Instances]} instances (Table~\ref{tab:tests} gives every contrast).\par
\texttt{BC.R-A} \textbf{[selected: BC]}\quad At the registered block-local level, displaying the current location increased success relative to the verification reminder and source link. The CUE--OFF contrast did not reject; this does not establish absence of a cue effect. Contradiction, salience, and reduced verification effort remain possible explanations.\par
\texttt{BC.R-B}\quad At the registered block-local level, the verification reminder with a source link increased success relative to no additional information. The LIVE--CUE contrast did not reject; this establishes neither equivalence nor absence of an additional value-display effect.\par
\texttt{BC.R-B.E}\quad CUE versus OFF rejected in CUE-A's direction at the registered block-local level (\textsf{[CueOffRescue]}/\textsf{[CueOffHarm]}/\textsf{[CueOffTie]}, net \textsf{[CueOffNetCI]}, block-local Holm $p=\textsf{[CueOffHolmLocal]}$). The registered rescue-opportunity threshold was not met. The LIVE--CUE contrast did not reject; this establishes neither equivalence nor absence of an additional value-display effect.\par
\texttt{BC.R-C}\quad Both adjacent contrasts rejected in the treatment direction at the registered block-local level; their paired differences and intervals quantify the two increments.\par
\texttt{BC.R-C.E}\quad Both adjacent contrasts rejected in the treatment direction at the registered block-local level (LIVE versus CUE net \textsf{[LiveCueNetCI]}, block-local Holm $p=\textsf{[LiveCueHolmLocal]}$; CUE versus OFF net \textsf{[CueOffNetCI]}, block-local Holm $p=\textsf{[CueOffHolmLocal]}$). The registered rescue-opportunity threshold was not met.\par
\texttt{BC.R-D}\quad The supporting LIVE--OFF analysis did not establish a positive effect of review-time information on these instances. This does not establish absence of an effect or a difference from Block~\textsf{\guilsinglleft{}RefBlock\guilsinglright{}}.\par
\texttt{BC.R-E}\quad OFF-A failed on \textsf{[OffFail]} of \textsf{[FRXOffAN]} counted rows; \textsf{[FRXOffAN]} of \textsf{[OffPlanned]} planned rows were observed. Fewer than \textsf{[OffGate]} counted failures were available, so the registered rescue-opportunity threshold was not met. At least one counted OFF-A row succeeded, so the universal OFF-A failure observed in Block~\textsf{\guilsinglleft{}RefBlock\guilsinglright{}} did not recur. Post-change source re-reading occurred in \textsf{[OffReread]} of \textsf{[FRXOffAN]} OFF-A episodes.\par
\texttt{BC.R-E.0}\quad OFF-A failed on \textsf{[OffFail]} of \textsf{[FRXOffAN]} counted rows; \textsf{[FRXOffAN]} of \textsf{[OffPlanned]} planned rows were observed. Fewer than \textsf{[OffGate]} counted failures were available, so the registered rescue-opportunity threshold was not met. This threshold statement alone does not establish that universal failure failed to recur.\par
\texttt{BC.R-F.cue}\quad At this budget the verification reminder with a source link alone outperformed the value display. No mechanism is inferred.\par
\texttt{BC.R-F.off}\quad The verification reminder with a source link reduced success relative to no additional information. No mechanism is inferred.\par
\texttt{BC.R-G}\quad Unresolved at the fixed budget. R-G describes the combined cue/value interpretation only and does not negate an independently eligible contrast rejection; each eligible contrast, its direction, interval, and multiplicity status is reported separately. An absent or incomplete contrast is unknown, not non-rejected, and a contrast without discordant pairs has raw $p=1$: insufficient evidence of a difference, not equivalence.\par
\texttt{BC.ELIG}\quad The \textsf{\guilsinglleft{}Contrast\guilsinglright{}} comparison, which is eligible, rejected in the \textsf{\guilsinglleft{}Dir\guilsinglright{}} direction at the registered block-local level (net \textsf{[\textsf{\guilsinglleft{}C\guilsinglright{}}NetCI]}, $p=\textsf{[\textsf{\guilsinglleft{}C\guilsinglright{}}P]}$).\par
\texttt{BC.G1}\quad The \textsf{\guilsinglleft{}Contrast\guilsinglright{}} comparison's complete pairs (\textsf{[\textsf{\guilsinglleft{}C\guilsinglright{}}Pairs]}) fall below its registered minimum (\textsf{[\textsf{\guilsinglleft{}C\guilsinglright{}}MinPairs]}), so it is unknown and receives no confirmatory interpretation. Other eligible contrasts retain their registered analyses.\par
\texttt{BC.A19} \textbf{[selected: BC]}\quad The LIVE--OFF comparison is a supporting descriptive analysis; it supports neither a confirmatory replication claim nor a claim that the effect differs from Block~\textsf{\guilsinglleft{}RefBlock\guilsinglright{}}.\par
\texttt{BC.GZP}\quad Block~\textsf{\guilsinglleft{}Block\guilsinglright{}} has run, but its validity evidence is incomplete (\textsf{[GzeroWhy]}); under its registered validity gate no result is interpreted until that evidence exists.\par
\texttt{BC.VOID}\quad Block~\textsf{\guilsinglleft{}Block\guilsinglright{}} is void under its registered validity gate (\textsf{[GzeroWhy]}). It contributes no inferential result; its rows are listed in Appendix~\ref{app:bc} as a record only.\par
\texttt{BC.PEND}\quad Block~\textsf{\guilsinglleft{}Block\guilsinglright{}} (\textsf{\guilsinglleft{}Model\guilsinglright{}}) had not reported when this version was built. Its arms, instances, contrasts and the interpretations its result will select are fixed in Appendix~\ref{app:readings}; every number it contributes prints as pending until its result file exists.\par
\texttt{BC.NOTRUN} \textbf{[selected: BGC]}\quad Block~\textsf{\guilsinglleft{}Block\guilsinglright{}} was not run (\textsf{[Disposition]}). It contributes no inferential result; its paper-wide slots are non-rejection placeholders.\par
\texttt{BC.INCOMPLETE} \textbf{[selected: BCS]}\quad Block~\textsf{\guilsinglleft{}Block\guilsinglright{}} started but did not complete (\textsf{[Disposition]}). It contributes no inferential result; its rows are listed in Appendix~\ref{app:bc} as a record only.\par
\texttt{BC.QI}\quad The LIVE-I controls passed \textsf{[FRXLiveIPassFrac]}; nothing is attributed to the review display from unmatched controls.\par
\texttt{BC.QE}\quad The independent scorer reconstructs appointment availability from archived rows; calendar compatibility is not re-derived but rests on stated design conditions (the generator and state preservation), whose checks are reported separately. Under that condition, every outcome term is resolved and agreeing with the evaluator on \textsf{[EndpointFull]} of \textsf{[Counted]} counted rollouts (Table~\ref{tab:coverage}).\par
\texttt{BC.QE0}\quad The independent scorer had not been run on this block's counted rollouts when this version was built, so no outcome term is independently re-derived; calendar compatibility rests on stated design conditions in any case.\par
\texttt{BC.QV}\quad Independent code re-derived this block's validity verdicts (\textsf{[VindPath]}), agreeing with the archive on \textsf{[VindAgree]} attempts.\par
\texttt{BC.QV0}\quad No independent re-derivation of this block's validity verdicts exists (it is expected at \textsf{[VindPath]}); its validity verdicts are the evaluator's alone.\par
\texttt{BC.QP}\quad Strict recovery occurred in \textsf{[FRXCueARecovered]} of \textsf{[FRXCueAN]} CUE-A and \textsf{[FRXLiveARecovered]} of \textsf{[FRXLiveAN]} LIVE-A runs, and \textsf{[FRXCueALinkFollowed]} of \textsf{[FRXCueAN]} CUE-A runs followed the reminder's source link (descriptive; never a reading condition).\par
\texttt{BGC.R-T+}\quad Under this trigger and interface, GPT-6 succeeded more often when an irrelevant session moved than when the target moved. This supports a target-relevance effect on success under the tested trigger and interface; it is not an estimate of a general effect of dynamic environments.\par
\texttt{BGC.R-T-}\quad GPT-6 succeeded more often under the target change than under the irrelevant one; no mechanism is inferred.\par
\texttt{BGC.R-TB}\quad GPT-6 failed often even when the target did not move (FRX-OFF-I failed on \textsf{[BGC:OffIFail]} of \textsf{[BGC:OffIN]} counted rows; \textsf{[BGC:OffiOffaPairs]} matched pairs), so its OFF-A failures are not shown to be specific to supersession; its cue and value contrasts are read with that caveat.\par
\texttt{BGC.R-TC}\quad The target-relevance contrast is unresolved: insufficient evidence of a difference, not equivalence (FRX-OFF-I failed on \textsf{[BGC:OffIFail]} of \textsf{[BGC:OffIN]} counted rows; \textsf{[BGC:OffiOffaPairs]} matched pairs).\par
\texttt{BGC.T-X}\quad No reading was registered for the target-relevance contrast before the block ran; it is reported as numbers only.\par
\texttt{QH.survive} \textbf{[selected: BC]}\quad The \textsf{\guilsinglleft{}Contrast\guilsinglright{}} rejection survives the paper-wide Holm analysis over \textsf{[HolmPaperM]} claim-bearing tests (adjusted $p=\textsf{[\textsf{\guilsinglleft{}C\guilsinglright{}}HolmPaper]}$).\par
\texttt{QH.local} \textbf{[selected: CLM-BS]}\quad The \textsf{\guilsinglleft{}Contrast\guilsinglright{}} contrast rejects under its registered block-local analysis but not under the paper-wide Holm analysis (adjusted $p=\textsf{[\textsf{\guilsinglleft{}C\guilsinglright{}}HolmPaper]}$ over \textsf{[HolmPaperM]} tests).\par
\texttt{QH.pending}\quad The paper-wide Holm-adjusted $p$ of the \textsf{\guilsinglleft{}Contrast\guilsinglright{}} contrast awaits the rest of the family (it is at most \textsf{[\textsf{\guilsinglleft{}C\guilsinglright{}}BonfPaper]}).\par
\texttt{BG2.S0}\quad GPT-6 on Block~B's instances: OFF-A \textsf{[BGtwo:FRXOffAPassFrac]}, ON-A \textsf{[BGtwo:FRXOnAPassFrac]} (\textsf{[BGtwo:OnOffRescue]}/\textsf{[BGtwo:OnOffHarm]}/\textsf{[BGtwo:OnOffTie]}, net \textsf{[BGtwo:OnOffNetCI]}, $p=\textsf{[BGtwo:OnOffP]}$).\par
\texttt{BG2.OPP-NOT}\quad The registered rescue-opportunity gate was not met (OFF-A failed on \textsf{[BGtwo:OffFail]} of \textsf{[BGtwo:FRXOffAN]}).\par
\texttt{BG2.OPP-MET}\quad OFF-A failed on \textsf{[BGtwo:OffFail]} of \textsf{[BGtwo:FRXOffAN]}, which meets the registered rescue-opportunity gate.\par
\texttt{BG2.EFF-ON}\quad At the registered local level, the review-time information increased GPT-6's full-task success on Block~B's instances.\par
\texttt{BG2.EFF-OFF}\quad At the registered local level, the review-time information reduced GPT-6's full-task success; no mechanism is inferred.\par
\texttt{BG2.EFF-NONE}\quad The paired contrast did not reject: insufficient evidence of a difference, not zero.\par
\texttt{BG2.PEND}\quad Block~BG2, the predeclared fallback that runs only if Block~BC is dropped, had not reported when this version was built.\par
\texttt{BG2.G1}\quad The paired contrast has \textsf{[BGtwo:OnOffPairs]} complete pairs, fewer than its registered minimum of eight, so it is unknown and receives no confirmatory interpretation; its paper-wide slot is a non-rejection placeholder.\par
\texttt{BG2.GZP}\quad Block~BG2 has run, but its validity or result-agreement evidence is incomplete; under its registered admissibility rule no result is interpreted until that evidence exists.\par
\texttt{BG2.VOID}\quad Block~BG2 is void under its registered validity gate. It contributes no inferential result.\par
\texttt{BG2.NOTRUN} \textbf{[selected: BG2]}\quad Block~BG2 was not run (\textsf{[BGtwo:Disposition]}).\par
\texttt{BG2.INCOMPLETE}\quad Block~BG2 started but did not complete (\textsf{[BGtwo:Disposition]}); it contributes no inferential result.\par
\texttt{CLM.S0} \textbf{[selected: CLM-B, CLM-BS]}\quad Over \textsf{[OnOffPairs]} sealed instance pairs (seeds \textsf{[SeedList]}), \textsf{\guilsinglleft{}Model\guilsinglright{}} met the full approval endpoint on \textsf{[ClmOffAPassFrac]} without additional information at review and \textsf{[ClmOnAPassFrac]} with it; of the pairs, \textsf{[OnOffRescue]} favoured the review display, \textsf{[OnOffHarm]} the arm without it, and \textsf{[OnOffTie]} neither; net \textsf{[OnOffNetCI]} (marginal 95\,\% interval), $p=\textsf{[OnOffP]}$. Controls: ON-I \textsf{[ClmOnIPassFrac]}, L0 \textsf{[ClmLzeroPassFrac]}.\par
\texttt{CLM.R1} \textbf{[selected: CLM-BS]}\quad The review-time information increased full-task success at the registered local level. The OFF-A failure categories and observed pre-commit revisions are reported separately (Appendix~\ref{app:claims}).\par
\texttt{CLM.R1-NC}\quad A local paired success difference was observed, but the registered control-credibility criterion was not met; no transfer claim is made.\par
\texttt{CLM.R1-DG}\quad The derived-decision gate was not met (\textsf{[XupOn]}/\textsf{[XupN]} threshold-crossing increases succeeded). This result does not establish performance beyond the scalar-copy shortcut.\par
\texttt{CLM.R2} \textbf{[selected: CLM-B]}\quad Without the review display, the agent succeeded on \textsf{[OffPass]}/\textsf{[ClmOffAN]} instances under silent supersession; the registered rescue-opportunity threshold was not met.\par
\texttt{CLM.R2-H}\quad The rescue-opportunity gate was not met. Nevertheless, ON-A reduced full-task success relative to OFF-A (\textsf{[OnOffRescue]}/\textsf{[OnOffHarm]}/\textsf{[OnOffTie]}, net \textsf{[OnOffNetCI]}, local $p=\textsf{[OnOffP]}$).\par
\texttt{CLM.R3}\quad Without the review display, the agent failed on \textsf{[OffFail]}/\textsf{[ClmOffAN]} instances under silent supersession; a positive effect of review-time information was not established at the registered local level.\par
\texttt{CLM.R3-H}\quad At the registered local level, ON-A reduced full-task success relative to OFF-A (\textsf{[OnOffRescue]}/\textsf{[OnOffHarm]}/\textsf{[OnOffTie]}, net \textsf{[OnOffNetCI]}, local $p=\textsf{[OnOffP]}$).\par
\texttt{CLM.R3-NC}\quad The registered control-credibility criterion was not met (L0 \textsf{[ClmLzeroPassFrac]}, ON-I \textsf{[ClmOnIPassFrac]}), so no transfer claim is made either way.\par
\texttt{CLM.R4}\quad The block failed its registered readiness or execution requirements (\textsf{[RfourWhy]}). It provides no admissible model-effect result; all attempts and exclusions remain in the engineering record.\par
\texttt{CLM.QCL} \textbf{[selected: CLM-B, CLM-BS]}\quad The scripted copy-limit shortcut, run on the same cells before any model, is assigned \textsf{[CopyLimitReading]} under the same rules.\par
\texttt{CLM.QE} \textbf{[selected: CLM-B, CLM-BS]}\quad Every outcome term the independent scorer re-derives is resolved and agreeing with the evaluator on \textsf{[EndpointFull]} of \textsf{[Counted]} counted rollouts.\par
\texttt{CLM.QE0}\quad The independent scorer had not been run on this block's counted rollouts when this version was built.\par
\texttt{CLM.QV} \textbf{[selected: CLM-B, CLM-BS]}\quad The standard-library validity re-derivation agrees with the archive on \textsf{[VindAgree]} attempts.\par
\texttt{CLM.QV0}\quad No independent validity re-derivation of this block exists (Appendix~\ref{app:coverage}); its validity verdicts are the evaluator's alone.\par
\texttt{CLM.QL}\quad The claims task shares the scaffold, harness, invoker and vendor with the clinic task: it is a second decision rule for the same intervention, not an independent environment, and no cue or announcement was varied in it.\par
\texttt{CLM.PEND}\quad Block~\textsf{\guilsinglleft{}Block\guilsinglright{}} (\textsf{\guilsinglleft{}Model\guilsinglright{}}) had not reported when this version was built; its interpretations and their qualifiers are fixed in Appendix~\ref{app:readings}.\par
\texttt{CLM.NOTRUN}\quad Block~\textsf{\guilsinglleft{}Block\guilsinglright{}} was not run (\textsf{[Disposition]}); it contributes no inferential result, and its paper-wide slot is a non-rejection placeholder.\par
\texttt{CLM.INCOMPLETE}\quad Block~\textsf{\guilsinglleft{}Block\guilsinglright{}} started but did not meet its registered completeness requirements (\textsf{[Disposition]}); it contributes no inferential result.\par
\texttt{SHORT.BC}\quad \textsf{\guilsinglleft{}Model\guilsinglright{}} on the same instances (Block~\textsf{\guilsinglleft{}Block\guilsinglright{}}): OFF-A \textsf{[FRXOffAPassFrac]}, CUE-A \textsf{[FRXCueAPassFrac]}, LIVE-A \textsf{[FRXLiveAPassFrac]}; block-local Holm $p=\textsf{[LiveCueHolmLocal]}$ (LIVE--CUE) and $p=\textsf{[CueOffHolmLocal]}$ (CUE--OFF); \textsf{[ReadingShort]}.\par
\texttt{SHORT.CLM}\quad \textsf{\guilsinglleft{}Model\guilsinglright{}} on CLM-B's frozen slots (Block~\textsf{\guilsinglleft{}Block\guilsinglright{}}): OFF-A \textsf{[ClmOffAPassFrac]}, ON-A \textsf{[ClmOnAPassFrac]}, $p=\textsf{[OnOffP]}$; \textsf{[ReadingShort]}. The two models share instances, so their difference is descriptive.\par
\texttt{SHORT.STATE}\quad \textsf{\guilsinglleft{}Model\guilsinglright{}} on the same instances (Block~\textsf{\guilsinglleft{}Block\guilsinglright{}}): \textsf{[ReadingShort]}.\par
\texttt{FRAG.R-A}\quad at the block-local level the value display increased success relative to the verification reminder with a source link\par
\texttt{FRAG.R-B}\quad at the block-local level the verification reminder with a source link increased success relative to no additional information\par
\texttt{FRAG.R-B.E}\quad CUE versus OFF rejected in CUE-A's direction at the block-local level; the rescue-opportunity threshold was not met\par
\texttt{FRAG.R-C}\quad both adjacent contrasts rejected in the treatment direction at the block-local level\par
\texttt{FRAG.R-C.E}\quad both adjacent contrasts rejected in the treatment direction at the block-local level; the rescue-opportunity threshold was not met\par
\texttt{FRAG.R-D}\quad the supporting LIVE--OFF analysis did not establish a positive effect of review-time information\par
\texttt{FRAG.R-E}\quad OFF-A failed on only \textsf{[OffFail]} of \textsf{[FRXOffAN]} counted rows, below the rescue-opportunity threshold\par
\texttt{FRAG.R-E.0}\quad OFF-A failed on \textsf{[OffFail]} of \textsf{[FRXOffAN]} counted rows (of \textsf{[OffPlanned]} planned), below the rescue-opportunity threshold\par
\texttt{FRAG.R-F.cue}\quad the verification reminder with a source link outperformed the value display at the block-local level\par
\texttt{FRAG.R-F.off}\quad the verification reminder with a source link reduced success relative to no additional information at the block-local level\par
\texttt{FRAG.R-G}\quad the registered rules leave the combined interpretation unresolved\par
\texttt{FRAG.ELIG}\quad the eligible \textsf{\guilsinglleft{}Contrast\guilsinglright{}} comparison rejected in the \textsf{\guilsinglleft{}Dir\guilsinglright{}} direction at the block-local level\par
\texttt{FRAG.G1}\quad the \textsf{\guilsinglleft{}Contrast\guilsinglright{}} comparison is incomplete and receives no confirmatory interpretation\par
\texttt{FRAG.R-T+}\quad success was higher when an irrelevant session moved than when the target moved (block-local)\par
\texttt{FRAG.R-T-}\quad success was higher when the target moved than when an irrelevant session moved (block-local)\par
\texttt{FRAG.R-TB}\quad the target-relevance contrast did not reject, and GPT-6 also failed often when the target did not move\par
\texttt{FRAG.R-TC}\quad the target-relevance contrast is unresolved (not rejected)\par
\texttt{FRAG.T-X}\quad no reading was registered for the target-relevance contrast\par
\texttt{FRAG.PEND}\quad pending when this version was built\par
\texttt{FRAG.NOTRUN}\quad not run (\textsf{[Disposition]})\par
\texttt{FRAG.INCOMPLETE}\quad started but not completed (\textsf{[Disposition]}); no inference\par
\texttt{FRAG.GZP}\quad validity evidence incomplete; no result interpreted\par
\texttt{FRAG.VOID}\quad void under its validity gate; no conclusion\par
\texttt{FRAG.CLM.R1}\quad the review display increased full-task success at the registered local level\par
\texttt{FRAG.CLM.R1-NC}\quad a local paired difference was observed but the control-credibility criterion was not met; no transfer claim\par
\texttt{FRAG.CLM.R1-DG}\quad the derived-decision gate was not met\par
\texttt{FRAG.CLM.R2}\quad OFF-A succeeded often, below the rescue-opportunity threshold\par
\texttt{FRAG.CLM.R3}\quad a positive effect of review-time information was not established\par
\texttt{FRAG.CLM.HARM}\quad the review display reduced full-task success at the registered local level\par
\texttt{FRAG.CLM.R4}\quad readiness or execution requirements not met; no admissible result\par
\texttt{FRAG.PW.survive}\quad surviving the paper-wide Holm analysis\par
\texttt{FRAG.PW.local}\quad not significant under the paper-wide Holm analysis\par
\texttt{FRAG.PW.pending}\quad paper-wide Holm pending\par
\texttt{APP.pending}\quad Its finalized implementation is frozen before each of two pre-registered prospective applications.\par
\texttt{APP.pending.draft}\quad Its finalized implementation is to be frozen before the first rollout of each of two prospective applications, whose registrations are not yet frozen.\par
\texttt{APP.two}\quad Its finalized implementation is evaluated prospectively in two workflows, clinic booking and insurance-claim approval.\par
\texttt{APP.bc}\quad Its finalized implementation is evaluated prospectively in clinic booking, with a second pre-registered application in insurance-claim approval.\par
\texttt{APP.clm}\quad Its finalized implementation is evaluated prospectively in insurance-claim approval, with a pre-registered application in clinic booking.\par
\texttt{APP.none}\quad Neither of its two pre-registered prospective applications yielded an admissible result.\par
\texttt{ABS.BC.lead}\quad In clinic booking, counterbalanced no additional information, cue-and-link and current-value conditions on \textsf{[BC:InstancesWords]} fresh instances (Claude Opus~5)\par
\texttt{ABS.BC.PEND}\quad had not reported when this version was built.\par
\texttt{ABS.BC.NOTRUN}\quad were not run (\textsf{[BC:Disposition]}).\par
\texttt{ABS.BC.INCOMPLETE}\quad started but did not complete (\textsf{[BC:Disposition]}) and support no inference.\par
\texttt{ABS.BC.VOID}\quad are void under their registered validity gate and support no model conclusion.\par
\texttt{ABS.BC.GZP}\quad have run, but their validity evidence is incomplete, so no result is interpreted.\par
\texttt{ABS.BC.R-C}\quad rejected both adjacent contrasts in the treatment direction at the registered block-local level\par
\texttt{ABS.BC.R-C.E}\quad rejected both adjacent contrasts in the treatment direction at the registered block-local level\par
\texttt{ABS.BC.R-A}\quad showed at the registered block-local level that displaying the current location increased success relative to the verification reminder with a source link, with no rejection of cue versus no additional information\par
\texttt{ABS.BC.R-B}\quad showed at the registered block-local level that the verification reminder with a source link increased success relative to no additional information, with no rejection of value versus cue\par
\texttt{ABS.BC.R-B.E}\quad rejected cue versus no additional information in the cue's direction at the registered block-local level\par
\texttt{ABS.BC.R-F.cue}\quad showed at the registered block-local level that the verification reminder with a source link outperformed the value display\par
\texttt{ABS.BC.R-F.off}\quad showed at the registered block-local level that the verification reminder with a source link reduced success relative to no additional information\par
\texttt{ABS.BC.R-D}\quad did not establish a positive effect of the direct display of the current value over no additional information in a supporting descriptive analysis\par
\texttt{ABS.BC.R-E}\quad found fewer counted no additional information failures than the registered rescue-opportunity threshold requires (OFF-A failed on \textsf{[BC:OffFail]} of \textsf{[BC:FRXOffAN]} counted rows), and at least one counted no additional information row succeeded\par
\texttt{ABS.BC.R-E.0}\quad found fewer counted no additional information failures than the registered rescue-opportunity threshold requires (OFF-A failed on \textsf{[BC:OffFail]} of \textsf{[BC:FRXOffAN]} counted rows, of \textsf{[BC:OffPlanned]} planned)\par
\texttt{ABS.BC.R-G}\quad left the combined interpretation unresolved under the registered rules\par
\texttt{ABS.BC.ELIG}\quad although the eligible \textsf{\guilsinglleft{}Contrast\guilsinglright{}} comparison rejected in the \textsf{\guilsinglleft{}Dir\guilsinglright{}} direction at the block-local level\par
\texttt{ABS.BC.G1}\quad (the \textsf{\guilsinglleft{}Contrast\guilsinglright{}} comparison was incomplete and receives no confirmatory interpretation)\par
\texttt{ABS.PW.all}\quad and every block-local rejection survives the paper-wide Holm analysis\par
\texttt{ABS.PW.none}\quad but no block-local rejection survives the paper-wide Holm analysis\par
\texttt{ABS.PW.some}\quad and only the \textsf{\guilsinglleft{}Contrast\guilsinglright{}} rejection survives the paper-wide Holm analysis\par
\texttt{ABS.PW.pending}\quad pending the paper-wide Holm analysis\par
\texttt{ABS.CLM.lead}\quad In insurance-claim approval, where the approved amount is derived,\par
\texttt{ABS.CLM.PEND}\quad the registered comparison with and without the review-time information (Claude Opus~5) had not reported when this version was built.\par
\texttt{ABS.CLM.NOTRUN}\quad the registered comparison with and without the review-time information was not run (\textsf{[ClmB:Disposition]}).\par
\texttt{ABS.CLM.INCOMPLETE}\quad the registered comparison started but did not meet its completeness requirements and supports no inference.\par
\texttt{ABS.CLM.R4}\quad the registered block failed its readiness or execution requirements and provides no admissible model-effect result.\par
\texttt{ABS.CLM.R1}\quad the review-time information increased full-task success for Claude Opus~5 at the registered local level ($p=\textsf{[ClmB:OnOffP]}$)\par
\texttt{ABS.CLM.R1-NC}\quad a local paired success difference was observed, but the registered control-credibility criterion was not met; no transfer claim is made\par
\texttt{ABS.CLM.R1-DG}\quad ; the derived-decision gate was not met, so performance beyond the scalar-copy shortcut is not established\par
\texttt{ABS.CLM.R2}\quad without additional information at review Claude Opus~5 succeeded on \textsf{[ClmB:OffPass]} of \textsf{[ClmB:ClmOffAN]} instances under silent supersession; the registered rescue-opportunity threshold was not met\par
\texttt{ABS.CLM.R3}\quad without additional information at review Claude Opus~5 failed on \textsf{[ClmB:OffFail]} of \textsf{[ClmB:ClmOffAN]} instances under silent supersession; a positive effect of review-time information was not established at the registered local level\par
\texttt{ABS.CLM.HARM}\quad ; nevertheless the review display reduced full-task success at the registered local level\par
\texttt{ABS.AUDIT.hist}\quad Independent audits re-derive every validity verdict of the historical blocks (\textsf{[VindAgree]} of \textsf{[VindAttempts]} agree under the rules as implemented) and the booked clinic of every historical booking, but verify no historical success in full\par
\texttt{ABS.AUDIT.pending}\quad ; prospective coverage is reported per block, and planned evidence collection is not counted as verification.\par
\texttt{ABS.AUDIT.some}\quad ; on the prospective blocks, with clinic calendar compatibility resting on stated design conditions, every outcome term is resolved and agreeing with the evaluator on \textsf{[ProspEndpointFull]} of \textsf{[ProspCounted]} counted rollouts.\par
\texttt{CONCL.fixed}\quad These contrasts estimate within-configuration effects; unrejected comparisons do not establish absence or equivalence.\par
\texttt{CHAL.PEND}\quad A second archive challenge, built by a different model and sealed before any Claude agent read it (seal record \textsf{[ChallengeSealSha]}), had not been run when this version was built.\par
\texttt{CHAL.RUN}\quad A second, sealed archive challenge built by a different model met \textsf{[ChallengeMet]} of \textsf{[ChallengeN]} frozen expectations on its first and only run (\textsf{[ChallengeCorruptRejected]} of \textsf{[ChallengeCorruptInstantiated]} instantiated corruptions rejected; \textsf{[ChallengeValidAccepted]} of \textsf{[ChallengeValidInstantiated]} valid cases accepted); its \textsf{[ChallengeMissed]} misses are listed in Appendix~\ref{app:certificates} as its comparison output words them.\par
\texttt{QCONFIG}\quad GPT-6 received frames inline; Claude received frames through file reads. Each configuration passed its specified delivery and isolation checks. Within-configuration paired contrasts under a counterbalanced arm order estimate intervention effects. Cross-configuration differences include model, invoker, delivery-channel, and inference-setting differences and do not isolate vendor or model effects.\par
\texttt{REUSE.PEND}\quad The count of reused and newly implemented components (a deterministic comparison with the clinic-pinned versions) had not been produced when this version was built.\par
\texttt{REUSE.RUN}\quad \textsf{[ClmReuseUnchanged]} of \textsf{[ClmReuseShared]} shared components are byte-identical to their clinic-pinned versions, the others gained \textsf{[ClmReuseAddedLines]} lines, and \textsf{[ClmNewFiles]} new files (\textsf{[ClmNewLines]} lines) implement the task, its checks and its audits.\par
\endgroup

\section{The paper-wide multiplicity family}
\label{app:multiplicity}
A block's execution disposition and inferential eligibility govern every manuscript output: failed-admission, void
and R4 blocks contribute no inferential result, their reserved multiplicity slots taking $p=1$ only as non-rejection
placeholders displayed as `not admissible'; started-but-incomplete blocks are distinguished from unrun ones; and only
contrasts meeting their registered completeness and validity requirements are interpreted (moved from
Section~\ref{sec:stats}). Every claim-bearing test enters one paper-wide Holm family of \HolmPaperMWords tests. The
family is \HolmFixedClause{} and never shrinks; \HolmPaperStatus. Its historical slots are
retrospective, and the family does not make the exploratory search that preceded those blocks prospective. Readings are
selected by the local registered analyses; the paper-wide adjusted $p$ is a qualifier. \HolmClaimsRetroSentence{}
\HolmBCProspectiveSentence The Newcombe intervals are
marginal 95\,\% intervals, not simultaneous family-adjusted intervals, and the paper-wide adjustment does not repair
Block~BS's eligibility problems (delivery-confounded, Appendix~\ref{app:rules}).
\begin{center}

\begingroup\setlength{\tabcolsep}{2.6pt}\scriptsize
\begin{tabular}{@{}rP{1.8cm}P{1.6cm}P{1.3cm}P{1.8cm}P{1.4cm}lll@{}}
\toprule
\# & block (model) & contrast & role & registered in & status & $p$ & $\min(1, mp)$ & Holm (paper) \\
\midrule
1 & A2 (Opus~5) & S-C vs S-L2 & hist. primary & PRECOMMIT S0--S6, ADD.~3 & observed (retrospective) & 1.00 & \BonfPaperAtwo & \HolmPaperAtwo \\
2 & B (Opus~5) & OFF-A vs ON-A & hist. primary & ADD.~5 & observed (retrospective) & 0.00195 & \BonfPaperB & \HolmPaperB \\
3 & BS (Sonnet~5) & OFF-A vs ON-A & hist. primary & ADD.~6 & observed (retrospective) & 0.0312 & \BonfPaperBS & \HolmPaperBS \\
4 & BC (Opus~5) & CUE vs LIVE & primary & ADD.~18 & observed & 0.000244 & \BonfPaperBCLiveCue & \HolmPaperBCLiveCue \\
5 & BC (Opus~5) & OFF vs CUE & key sec. & ADD.~18 & observed & 0.5 & \BonfPaperBCCueOff & \HolmPaperBCCueOff \\
6 & CLM-B (Opus~5) & OFF-A vs ON-A & primary & PRECOMMIT\_\allowbreak{}CLAIMS & observed & 1.00 & \BonfPaperClmB & \HolmPaperClmB \\
7 & CLM-BS (Sonnet~5) & OFF-A vs ON-A & repl. of 6 & ADD.~19; PRECOMMIT\_\allowbreak{}CLAIMS ADD.~1 & observed & 0.00781 & \BonfPaperClmBS & \HolmPaperClmBS \\
8 & BCS (Sonnet~5) & CUE vs LIVE & repl. primary & ADD.~19 & not admissible & \textit{not admissible} & \BonfPaperBCSLiveCue & \HolmPaperBCSLiveCue \\
9 & BCS (Sonnet~5) & OFF vs CUE & repl. key sec. & ADD.~19 & not admissible & \textit{not admissible} & \BonfPaperBCSCueOff & \HolmPaperBCSCueOff \\
10 & BGC or BG2 [BGC] (GPT-6) & CUE vs LIVE, or OFF-A vs ON-A & primary & ADD.~19 & not admissible & \textit{not admissible} & \BonfPaperSlotTen & \HolmPaperSlotTen \\
11 & BGC (GPT-6) & OFF vs CUE & key sec. & ADD.~19 & not admissible & \textit{not admissible} & \BonfPaperBGCCueOff & \HolmPaperBGCCueOff \\
12 & BGC (GPT-6) & OFF-A vs OFF-I & target rel. & ADD.~19 & not admissible & \textit{not admissible} & \BonfPaperBGCOffiOffa & \HolmPaperBGCOffiOffa \\
\bottomrule
\end{tabular}
\endgroup

\end{center}
\begin{table}[h]
\centering
\caption{\textbf{Every claim-bearing test, with its local and paper-wide standing.} for/ag./tie: discordant pairs
favouring the treatment (for Block~A2, the hypothesised failure), favouring the control, and tied; net: paired
difference with a Newcombe marginal 95\% interval, not family-adjusted; $p$: exact two-sided sign test. Holm
(block): within the block's registered family. Holm (paper): over all \HolmPaperMWords tests (\HolmPaperStatus;
$^\ast$until every slot resolves: the range the adjusted $p$ can take whatever the unresolved slots hold); $^\dagger$historical: hypothesis-generating
(predecessor protocol) and retrospective in the family. A slot whose block is not run or not admissible is a
non-rejection placeholder, shown as such. Not in the family: the LIVE--OFF
comparisons (descriptive), every control cell, the A2S and BF screens, the sensitivity filters and every
cross-model comparison. The for/ag./tie counts are evaluator verdicts; of rows 1--3's \TestsHistSucc paired-arm successes the independent scorer
confirms the clinic on \TestsHistClinic and all five invariants on \TestsHistFull. Row~2's rescue reading departs
from the registered claim rule (Appendix~\ref{app:blockB}). \QconfigNote}
\label{tab:tests}

\begingroup\setlength{\tabcolsep}{1.8pt}\scriptsize
\begin{tabular}{@{}r@{\hspace{3pt}}lllcclll@{}}
\toprule
\# & block, model & contrast & role & for/ag./tie & net [95\% CI] & $p$ & Holm (block) & Holm (paper) \\
\midrule
1$^\dagger$ & A2, Opus~5 & S-C vs S-L2 & hist. primary & 1/0/7 & \PairedAtwoNetCI & \PairedAtwoP & --- & \HolmPaperAtwo \\
2$^\dagger$ & B, Opus~5 & OFF-A vs ON-A & hist. primary & 10/0/0 & \PairedBNetCI & \PairedBP & --- & \HolmPaperB \\
3$^\dagger$ & BS, Sonnet~5 & OFF-A vs ON-A & hist. primary & 6/0/4 & \PairedBSNetCI & \PairedBSP & --- & \HolmPaperBS \\
4 & BC, Opus~5 & LIVE--CUE & primary & \BlockBCLiveCueRescue/\BlockBCLiveCueHarm/\BlockBCLiveCueTie & \BlockBCLiveCueNetCI & \BlockBCLiveCueP & \BlockBCLiveCueHolmLocal & \HolmPaperBCLiveCue \\
5 & BC, Opus~5 & CUE--OFF & key sec. & \BlockBCCueOffRescue/\BlockBCCueOffHarm/\BlockBCCueOffTie & \BlockBCCueOffNetCI & \BlockBCCueOffP & \BlockBCCueOffHolmLocal & \HolmPaperBCCueOff \\
6 & CLM-B, Opus~5 & OFF-A--ON-A & primary & \BlockClmBOnOffRescue/\BlockClmBOnOffHarm/\BlockClmBOnOffTie & \BlockClmBOnOffNetCI & \BlockClmBOnOffP & \BlockClmBOnOffHolmLocal & \HolmPaperClmB \\
7 & CLM-BS, Sonnet~5 & OFF-A--ON-A & repl. of 6 & \BlockClmBSOnOffRescue/\BlockClmBSOnOffHarm/\BlockClmBSOnOffTie & \BlockClmBSOnOffNetCI & \BlockClmBSOnOffP & \BlockClmBSOnOffHolmLocal & \HolmPaperClmBS \\
8 & BCS, Sonnet~5 & LIVE--CUE & repl. primary & \multicolumn{4}{l}{\textit{not admissible} (incomplete at the registered cutoff)} & \HolmPaperBCSLiveCue \\
9 & BCS, Sonnet~5 & CUE--OFF & repl. key sec. & \multicolumn{4}{l}{\textit{not admissible} (incomplete at the registered cutoff)} & \HolmPaperBCSCueOff \\
10 & BGC\,/\,BG2$\to$BGC, GPT-6 & LIVE--CUE & primary & \multicolumn{4}{l}{\textit{not admissible} (not run; Appendix~\ref{app:bc})} & \HolmPaperSlotTen \\
11 & BGC, GPT-6 & CUE--OFF & key sec. & \multicolumn{4}{l}{\textit{not admissible} (not run; Appendix~\ref{app:bc})} & \HolmPaperBGCCueOff \\
12 & BGC, GPT-6 & OFF-I--OFF-A & target rel. & \multicolumn{4}{l}{\textit{not admissible} (not run; Appendix~\ref{app:bc})} & \HolmPaperBGCOffiOffa \\
\bottomrule
\end{tabular}
\endgroup

\end{table}
\paragraph{Facts that follow from the family before any new result.} (1) Block~B's paper-wide Holm-adjusted $p$ is
at most $m\,p = \HolmPaperBBound$, below $\alpha$ whatever the other slots hold. (2) Block BS cannot reject under this
fixed paper-wide Holm family because A2's $p=1$ occupies a later rank: its adjusted $p$ is at least
\HolmPaperBSMin. (3) Block~A2 never rejects. (4) The first Holm step needs $p \le \HolmPaperFirst$; with at most
\HolmPaperBCPairs pairs (Block~BC's size) the favourable splits that pass it are \HolmPaperFirstSplits. (5)
\HolmPaperRCText{} \BCRCReachText{} (6) A contrast that rejects under its registered block-local analysis but
not under the paper-wide Holm analysis is printed with exactly that qualification. \HolmPaperSurvivalFacts

\paragraph{Erratum (dated 2026-09-24): Newcombe intervals of the historical blocks.} The historical paired intervals
were labelled Newcombe method~10 but computed with the uncorrected correlation numerator, $ad-bc$ (method~8;
FREEZE\_r21 round~3). All \NewcombeErratumNWords of them (Table~\ref{tab:paired}, the no-action-cap row included) are
recomputed with the corrected module (\code{envos/stats.py}, \NewcombeCorrectedSha, imported read-only) and with the
ledger's own method-10 implementation, which agree. The correction changes the correlation only when $ad>bc$, which
holds for \NewcombeErratumApplies of them, so \NewcombeErratumChanged of the printed intervals change (largest change
of any bound \NewcombeErratumMaxDiff); the label ``method~10'' now describes the computation. The claims driver's
stored intervals have the same defect (Appendix~\ref{app:claims}).

\section{Decisions taken in the final night, disclosed}
\label{app:disclosures}
Six procedural decisions were taken during the night before the experiment freeze. Each preceded the relevant freeze
and every outcome it could affect; each record is pinned by the ledger (\code{FREEZE\_r21/CODE\_FINAL.md}, the
round-3 pre-images of the spec and the templates, the superseded seal record, the claims block log).
\begin{center}
\begingroup\scriptsize\setlength{\tabcolsep}{3pt}
\begin{tabular}{@{}P{1.9cm}P{3.2cm}P{3.9cm}P{4.2cm}@{}}
\toprule
decision & when, relative to outcomes & what changed & inferential consequence \\
\midrule
transport timeout & before any BC-family model rollout; certification rerun from scratch & only the Docker transport timeout, \DiscTimeoutOld{}~s $\to$ \DiscTimeoutNew{}~s, after a certification episode exceeded it under host overload; code refrozen & none: no task rule, threshold, action budget or model setting changed; the claims blocks ran from their own pinned code \\
branch deadline & before any BC outcome and before the branch rule was frozen & the fallback deadline, \DiscDeadlineOld{} $\to$ \DiscDeadlineNew{} UTC, because certification and feasibility were projected to finish minutes before it & none: the branch criterion was unchanged \\
template wording & before any outcome and before the experimental freeze & \DiscTemplatesWord{} sealed templates said ``randomized'' for the fixed seeded order; the word became ``counterbalanced'', the original seal retained and the templates resealed & none: no order assignment changed \\
an interrupted attempt & during CLM-BS, before that attempt had a valid archive or an outcome & \DiscInterruptedWord{} process terminated by the operator session; under the frozen crash-recovery rule the world baseline was verified, the attempt logged as interrupted and the cell rerun & none: no outcome existed to discard, and the reserve was not spent \\
BC design size & before BC's freeze and any BC rollout & \DiscBCOld{} $\to$ \DiscBCNew{} triples, after exact-test arithmetic showed one registered two-contrast interpretation unreachable at \DiscBCOld; the original design is retained in the record & a registered interpretation becomes reachable; no BC outcome informed the change, and no extension is permitted after rollout begins \\
second-vendor admission & after the original GPT-6 admission failed on a criterion specific to the Claude file-read channel; no GPT-6 task block ran before quota exhaustion & admission amended to define an inline-frame channel separately & no GPT-6 model result is reported, and neither admission attempt is used as model evidence (configurations and the per-channel rule: Appendix~\ref{app:bc}; status: \AdmissionAnineteenStatus) \\
\bottomrule
\end{tabular}
\endgroup
\end{center}
\BCSHaltPointer

\section{Diagnostic invariant-completion fraction and exposure attribution of the claims blocks}
\label{app:osworld}
Beside binary success we describe the claims blocks by two diagnostics built from existing evidence only---the
archived independent-scorer records and validity terms, read through the ledger; no rollout was run for this table,
and nothing in it is a test or a reading. \emph{Diagnostic invariant-completion fraction}: of the independent
scorer's \BlockClmBAuditTerms{} outcome terms per counted rollout, the fraction met, pooled over an arm's rollouts.
This diagnostic fraction is specific to EnvOS's endpoint decomposition and is not comparable to partial scores
reported by other computer-use benchmarks. \emph{Exposure attribution} (labels after OSWorld~2.0
\citep{yuan2026osworld}): \emph{exposure} means the archived validity record shows the target limit changed after the
superseded value had been delivered to the policy and before the commit. \emph{Handled}: exposed, and the approved
amount at commit is the one derived from the limit then in force. \emph{Failed after exposure}: exposed, amount wrong.
\emph{Untested}: the arm moves no target record (ON-I moves an irrelevant one, L0 nothing), or exposure is not
evidenced. A handled rollout can still miss the full endpoint on another term, which the fraction shows. For Sonnet~5
the fraction was \OswClmBSOffAPartial{} without additional information at review and \OswClmBSOnAPartial{} with it, and its OFF-A rollouts
were \OswClmBSOffAFailed{} failed-after-exposure and \OswClmBSOffAHandled{} handled.
\begin{table}[h]
\centering
\caption{\textbf{Claims blocks: diagnostic invariant-completion fraction and exposure attribution} (counted rollouts;
ledger-derived from the archived audit records; descriptive only, not comparable to other benchmarks' partial
scores).}
\label{tab:osworld}

\begingroup\setlength{\tabcolsep}{4pt}\scriptsize
\begin{tabular}{@{}llrrrrrr@{}}
\toprule
model & arm & $n$ & full endpoint & invariant-completion fraction & Handled & Failed after exposure & Untested \\
\midrule
Opus~5 & OFF-A & \OswClmBOffAN & \OswClmBOffASucc & \OswClmBOffAPartial & \OswClmBOffAHandled & \OswClmBOffAFailed & \OswClmBOffAUntested \\
Opus~5 & ON-A & \OswClmBOnAN & \OswClmBOnASucc & \OswClmBOnAPartial & \OswClmBOnAHandled & \OswClmBOnAFailed & \OswClmBOnAUntested \\
Opus~5 & ON-I & \OswClmBOnIN & \OswClmBOnISucc & \OswClmBOnIPartial & \OswClmBOnIHandled & \OswClmBOnIFailed & \OswClmBOnIUntested \\
Opus~5 & L0 & \OswClmBLzeroN & \OswClmBLzeroSucc & \OswClmBLzeroPartial & \OswClmBLzeroHandled & \OswClmBLzeroFailed & \OswClmBLzeroUntested \\
\midrule
Sonnet~5 & OFF-A & \OswClmBSOffAN & \OswClmBSOffASucc & \OswClmBSOffAPartial & \OswClmBSOffAHandled & \OswClmBSOffAFailed & \OswClmBSOffAUntested \\
Sonnet~5 & ON-A & \OswClmBSOnAN & \OswClmBSOnASucc & \OswClmBSOnAPartial & \OswClmBSOnAHandled & \OswClmBSOnAFailed & \OswClmBSOnAUntested \\
Sonnet~5 & ON-I & \OswClmBSOnIN & \OswClmBSOnISucc & \OswClmBSOnIPartial & \OswClmBSOnIHandled & \OswClmBSOnIFailed & \OswClmBSOnIUntested \\
Sonnet~5 & L0 & \OswClmBSLzeroN & \OswClmBSLzeroSucc & \OswClmBSLzeroPartial & \OswClmBSLzeroHandled & \OswClmBSLzeroFailed & \OswClmBSLzeroUntested \\
\bottomrule
\end{tabular}
\endgroup

\end{table}

\section{The policy interface compared with other harnesses}
\label{app:interface}
Our policy interface is not intended to reproduce OSWorld-style native computer
use. It is a controlled semantic browser harness: each step exposes the current screenshot together with clipped
visible page text and an indexed element list, accepts a fixed browser-action schema, and provides only self-authored
working memory through a bounded memo/history channel. This reduces low-level perception and pointer-control variance
so the experiment can isolate response to state changes. It is more instrumented than a pure pixel benchmark but also
more constrained: a fixed action schema, no shell, a navigation allowlist, an isolated user, bounded memo/history and
an action cap (Table~\ref{tab:interface}). The structured text and element list may raise absolute task success
relative to pixel-only computer use, while the stateless per-step process may lower it relative to an agent with
persistent conversational memory; we therefore avoid ranking the harness globally as easier or harder. Our results
are fixed-instance intervention contrasts within one interface under the stated assumptions (\S\ref{sec:stats}), not
directly comparable leaderboard scores: because \envos estimates matched intervention contrasts under one fixed
interface rather than absolute computer-use capability, cross-benchmark score comparability is not required for them;
interface transfer remains an external-validity question.

\begin{table}[h]
\centering
\caption{\textbf{Interfaces compared.} \envos beside OSWorld~2.0 \citep{yuan2026osworld} and WebArena-style
browser benchmarks \citep{zhou2024webarena}. Representative interfaces used in the cited evaluations; this table is
not an exhaustive description of every agent configuration those benchmarks support. The paired arms of an \envos
block share every row of this table, so the interface cancels within a contrast; it does not cancel across
benchmarks.}
\label{tab:interface}
\begingroup\scriptsize\setlength{\tabcolsep}{3pt}
\begin{tabular}{@{}P{1.55cm}P{2.55cm}P{2.3cm}P{1.95cm}P{1.75cm}P{2.75cm}@{}}
\toprule
 & observation & action & memory & budget & runs, scoring, purpose \\
\midrule
\envos & screenshot + clipped page text + indexed element list & fixed browser-action schema; no shell; allowlist; sealed, isolated user & self-authored memo ($\le$\MemoMaxChars{} chars), last \HistoryWindow{} actions & \ActionBudget{} / \ClmActionBudget{} actions & paired arms per instance, one run per cell; binary endpoint re-derived independently; matched intervention effects \\
OSWorld~2.0 & screenshot only & Claude: native computer-use tool; others: pyautogui code & agent-dependent & 500 steps, 3\,s pause per action & one run per model; binary and checkpoint partial score; exposure attribution; end-to-end capability \\
WebArena-style & structured browser state (accessibility tree) & browser actions & agent-dependent & per benchmark & task success; web-task capability \\
\bottomrule
\end{tabular}
\endgroup
\end{table}

\section{The protocol as a minimal abstraction, and continuous change}
\label{app:abstraction}
\paragraph{Transition rule.} Given action history $H_t$, a perturbation applies the transition
$(\delta\circ T)(s_t,a_t)$ when $\tau(H_t)\land\neg g_t$, and otherwise applies $T(s_t,a_t)$, with
$\delta$, $\tau$ and $g$ as in Definition~\ref{def:pert}.

\paragraph{The abstraction.} Stripped of the testbed, an \envos intervention is a tuple $(v, E, \tau, c, V, I, R)$:
an authoritative variable $v$ on which the correct action depends; an exposure predicate $E$ over the archived
request and frame record (the old value of $v$ was served in a frame delivered to the policy); a trigger $\tau$ that
applies the transition of $v$ inside the first eligible request after $E$ holds, in one transaction with its event
record; a matched inapplicable control $c$ that fires the same trigger on a variable irrelevant to the correct action;
a validity table $V$ mapping each attempt's archived evidence to counted or not counted before its outcome is scored;
invariants $I$ evaluated on authoritative state at commit time and on the agent's report; and an interpretation map
$R$ from registered test outcomes to pre-written sentences, frozen with $V$ and $I$ before rollout. The checks of
Section~\ref{sec:certification} test these components: certification tests $\tau$ and $c$ with scripted policies; mutation
batteries and the archive challenges test $V$ and $I$; the independent scorer and validity re-derivation re-implement
$I$ and $V$; and the claim ledger executes $R$. Nothing in the tuple names the semantic-browser interface. What depends
on it is how $E$ is evidenced (here, request logs and delivered text frames; in a screenshot-only harness, the frame
that rendered the value), and portability to another harness is untested (Section~\ref{sec:limitations}).

\paragraph{Asynchronous and continuous change.} The tuple assumes a discrete transition. For state that changes
asynchronously or continuously (a price feed, a queue position), the same structure applies when the environment
versions $v$: $E$ becomes ``version $k$ was served'', $\tau$ publishes version $k{+}1$ on the first eligible request
after that, and $I$ scores the commit against the version current at commit time, so that a commit acting on version
$k$ is distinguishable from one acting on a later version. A continuously drifting quantity additionally needs a
decision-relevant threshold crossing (as the claims task's threshold-crossing instances use) or a stated tolerance, so
that the correct action changes discretely. Without versioned authoritative state neither the exposure certificate nor
the commit-time scorer can be built, and the protocol does not apply. We have not implemented this extension.

\section{Task text and generator}
\label{app:task}
The instruction of instance seed~0 (split \code{test\_id}) as frozen in \S4 of the pre-registration
(sha256 prefix \PrecommitTaskTextSha of the frozen block; transliterated to ASCII below). Every arm of one
instance runs the same bytes: on the archived trajectories of Block~A2, \AtwoInstructionIdenticalInstances
of \AtwoInstructionInstances instances carry one instruction sha256 across all their rollouts
(\AtwoInstructionDistinct distinct texts over the block; two seeds may draw the same patient, label and
date). The release manifest pins, separately, the task module's default instruction text
(\InstructionSha), not a per-instance text.
\IfFileExists{tables/task_text.txt}{
\begin{lstlisting}
Two things. First, open Marie Lefevre's appointment APT-20448 in the booking portal and note its status. Then book Grace Adeyemi into the influenza vaccination session on Thursday 20 August 2026 at 15:00, using appointment number APT-20451. Her contact email is in her intake message in the clinic mailbox (Mailpit); her calendar for that day is in Nextcloud -- look at it for a clash before you choose the time. The portal's Clinic sessions page lists where each session runs. If 15:00 is unavailable, book the nearest later available slot. Then confirm the appointment in the portal.
Finish your reply with exactly one line in this form:
RESULT: appointment_no=<...>; clinic=<clinic>; slot=<HH:MM>; status=<status shown in the portal>; note=<APT-20448 status>
\end{lstlisting}}{\par\noindent\textit{The canonical appendix listing (tables/task\_text.txt) was not included in the provided source package. Restore it from the original research repository.}\par}
\paragraph{Generator.} One salted digest of (seed, split) draws every element of an instance: the patient,
the distractor appointment, the session date and the control session two days later, the appointment
number, the intake mail and calendar entry, and the ordered (old, new) clinic pair from a pool of four
names. The pair is assigned by a Latin-square rule, so each of the twelve ordered pairs occurs once in
twelve consecutive seeds and the dropdown position of the old and the new clinic is counterbalanced across
the block. The neutral work --- a distractor record to read, a mailbox lookup and a calendar check --- is
drawn from (seed, split) alone, so it is identical for every level of an instance by construction. The
certificate check \code{filler\_facts\_identical\_across\_arms\_and\_clinic\_neutral} scans the intake message
and the calendar entry for any clinic name and the frozen cue tokens, on the module's own instance
(\code{S-train-0}, in all \NeutralCertFiles certificate files); regenerating the \NeutralInstances evaluated
instances, \NeutralMailCalendarHits name a clinic. The check does not cover the distractor appointment: that
record carries a clinic from the same pool, on \DistractorIsNewClinic of the \NeutralInstances instances the
new one (\DistractorNewRows counted rows), which is how the report-flag rule came to read the distractor's
location as a flag (ADDENDUM~17; Table~\ref{tab:defect-checks}). Five salted splits (train, development, held-out test, compositional, sealed) partition the seed
space. Three uses of a test instance are distinct here. \emph{Generation}: every instance is generated from a
held-out test seed. \emph{First-model evaluation}: a family's first model runs instances no model had run
(Block~A2: seeds \AtwoSeedList; Block~B: seeds \BSeedList; Block~BC: seeds \BlockBCSeedList). \emph{Later-model replication}: Blocks
A2S, BS and BF re-use those instances (A2S a subset of A2's, BS all of B's, BF \BlockBFFRXOffAN of B's), so their
instances are known to the experimenter, not fresh. A seed whose treatment outcome has become known is burnt for
any new first-model block.

\section{Policy interface and invoker specification}
\label{app:repro}
{\sloppy
This appendix states the harness-authored part of the agent--environment interface of the P3v2 experiments as
the archived rollouts were run. It is not the whole interface: what was not archived (the CLI's built-in system
prompt and tool definitions, its decoding defaults, per-step element lists and page titles, the results of
non-\code{Read} tool calls) is marked \emph{unavailable} where it arises. It is derived from the source files whose sha256 the release manifest of each block
pins, not from the current tree. Table~\ref{tab:repro-versions} lists the pinned digests. Of the
pinned files, 19 of 20 (block, file)
pairs are available byte-for-byte. Each of the 82 rollouts in the five search logs records the digests of the code it ran. All match the block manifest except as follows: 4 rollouts of Block~BS (1 of them counted) ran harness \code{2439d3a38953}, the version before that block's manifest. They ran under the block's first
manifest (\BSPreAmendManifest), which the dated amendment applying ADDENDA~\BSAmendAddenda rebuilt as
\BSManifestNow; each row records the digests it ran, but no file with the first manifest's digest survives. That version is also available and is listed below. The harness versions used (\texttt{2439d3a38953}, \texttt{b9f4fdb5b8e4}, \texttt{fe4b9bb76d40}) differ only in the read-target check and in delivery
bookkeeping. Their prompt constants and the functions \code{build\_prompt}, \code{\_history\_line},
\code{with\_agent\_notes} and \code{\_trim\_note} are AST-identical. The two browser versions differ only
in the port configuration: observation, clipping and rendering are identical. The two task-module
versions that are available differ only in the delivery-certification function (Addendum~16). The
instruction, generator, step budget and history window are identical. $^\dagger$The task-module version
used by Block~A2 (\code{17f6c7cd72f2}) was not in the working tree when this specification was derived from the pinned sources (\AtwoModuleWhere).
For A2 the instruction text is taken from the archive instead (item~1). This specification was derived before that
module was recovered, so for A2 it describes the module's checkpoint and delivery-certification logic as it then
stood only through the archived checkpoints, and does not restate its declared history window (item~4). Repository paths are written
relative to \code{/envos}.

\begin{table}[h]\centering\scriptsize
\caption{Pinned code identity per block (sha256, first 12 hex). ``CLI'' is the policy CLI binary.}
\label{tab:repro-versions}
\resizebox{\linewidth}{!}{\begin{tabular}{llrlllll}\toprule
Block & requested model & step cap & harness & browser & invoker & task module & CLI \\\midrule
A2 & \texttt{claude-opus-5} & 48 & \texttt{2439d3a38953} & \texttt{657b2381dfe2} & \texttt{99c919f1fca4} & \texttt{17f6c7cd72f2}$^\dagger$ & \texttt{15e2d05148f8} \\
A2S & \texttt{claude-sonnet-5} & 48 & \texttt{fe4b9bb76d40} & \texttt{1c1010d38219} & \texttt{095a8a4081a8} & \texttt{5c38ce9c2153} & \texttt{15e2d05148f8} \\
B & \texttt{claude-opus-5} & 48 & \texttt{2439d3a38953} & \texttt{657b2381dfe2} & \texttt{99c919f1fca4} & \texttt{5c38ce9c2153} & \texttt{15e2d05148f8} \\
BS & \texttt{claude-sonnet-5} & 48 & \texttt{b9f4fdb5b8e4} & \texttt{657b2381dfe2} & \texttt{99c919f1fca4} & \texttt{5c38ce9c2153} & \texttt{15e2d05148f8} \\
BF & \texttt{claude-fable-5-1} & 48 & \texttt{fe4b9bb76d40} & \texttt{657b2381dfe2} & \texttt{095a8a4081a8} & \texttt{5c38ce9c2153} & \texttt{15e2d05148f8} \\
\bottomrule\end{tabular}}
\end{table}

\paragraph{(1) Policy prompt, verbatim.} Each policy step is \emph{one fresh, stateless} CLI process that
receives the text below as its single user prompt. There is no \code{--continue} or \code{--resume}, so no
conversation is carried from one step to the next. Placeholders are in angle brackets. The
\code{RESULT} line of the action specification is the instance form built by
\code{result\_form\_for\_task}, and the \code{memo} paragraph is inserted by \code{with\_agent\_notes}
because the task module declares \code{AGENT\_NOTES = True}. The last sentence is appended
by the invoker, and \code{<FRAME PATH>} is the step's screenshot inside the per-episode observation
directory (\code{/srv/envos-obs/<16 hex>/step\_NN\_before.png}). Non-ASCII characters are
transliterated: the em dash becomes \code{--}.
\begin{lstlisting}
You are an office assistant for a clinic. You work in a web browser.
The booking portal is at http://portal-health:8080 -- sign in there as scheduler with the password sched2026.

TASK: <INSTRUCTION: see below>


You act in a web browser. Each turn you are shown the current page: its address, its
visible text, and a numbered list of the things you can interact with on it. Reply with
ONE JSON object and nothing else:

{"type":"navigate","url":"http://..."}          load an address
{"type":"click","id":3}                         click element 3
{"type":"fill","id":4,"text":"..."}             type text into element 4
{"type":"select","id":5,"option":"15:00"}       choose an option in a dropdown
{"type":"press","key":"Enter"}                  press a key
{"type":"back"}                                 return to the previous page
{"type":"done","report":"..."}                  finish, with your written report

Any action may also carry "memo":"..." -- a short working note to yourself (at most 300 characters). Your notes are shown back to you in "What you have done so far".

The element numbers belong to the page shown this turn only; read the new list after
each action before choosing the next one.

Finish with a done action whose report ends with exactly this line:
RESULT: appointment_no=<number>; clinic=<clinic>; slot=<HH:MM>; status=<status>; note=<status of <distractor label>>

Reply with ONLY the JSON object. No prose, no code fences, no explanation.

CURRENT PAGE
address: <page URL>
title: <page title>
--- page text ---
<page text>
--- end page text ---
--- things you can interact with ---
<ELEMENT LIST>
--- end ---

What you have done so far:
<history line for step 0>
<history line for step 1>
...

Your next action:

The current screen is the image at <FRAME PATH>. Read it first.
\end{lstlisting}

The \code{<INSTRUCTION>} is the instance's instruction (Appendix~\ref{app:task}). It is archived verbatim in
every trajectory. For 82 of the 82 archived model rollouts, including all of Block~A2, it
equals, byte for byte, the text that the generator \code{generate\_variant\_params(seed, split)} produces
today. The CLI also prepends its own built-in system prompt and tool definitions. That text is not part
of the harness and was not archived, so it is \emph{unavailable}. The CLI version that supplied it is
pinned (item~5).

\paragraph{(2) Action schema.} This is the JSON object listed in the prompt above. The schema allows 7
action types, with an optional \code{memo} on any action. A reply is parsed by
removing code fences and then taking the \emph{first} brace-delimited substring (non-greedy
\code{\textbackslash{}\{.*?\textbackslash{}\}}). That substring must be a JSON object with a
\code{type} key. A reply that does not parse ends the episode (\code{unparsed\_reply}), and so does a
\code{done} action. Element ids index the list shown on that same step, and the browser resolves them
positionally. Across the 2548 archived model steps, the actions carried these keys outside the
schema: none. The harness
ignores extra keys, but they are echoed back in the history. Each action has a
15\,s timeout. The episode is capped at
48 policy steps in every block.

\paragraph{(3) Observation per step.} Each step delivers two channels, both captured \emph{before} the action:
\begin{itemize}[leftmargin=*,itemsep=1pt,topsep=2pt]
\item \textbf{Text}, inside the prompt. It carries the page URL and title, and the page's
\code{document.body.innerText} clipped to 3000 characters (the overflow becomes
\code{... [N more chars]}). Table cells are tab-separated. The text was clipped on
0 of 2548 archived steps. The prompt then lists the interactive
elements, at most 400 across frames and shadow roots, one per line as
\code{[id] role "label"}. A line may add \code{options=[...]} (at most 40), \code{value="..."} (at most
200 characters) or \code{href=...}. Labels are clipped to 200 characters, and a table row's own text is
appended to its controls' labels (at most 120 characters). The archived pages had a median of
8 elements and a maximum of 172. The
per-step element lists and page titles were not archived. Only their count is, so a rendered element
block cannot be reproduced from the archive.
\item \textbf{Image}. This is a PNG screenshot of the whole virtual desktop, taken by the desktop's control
server, at 1920x1080~px (82 runs). It reaches the model only if the
model calls the CLI's \code{Read} tool on \code{<FRAME PATH>}. The archive records this per step. The
frame was read on 2305 of 2548 model steps (per model in Table~\ref{tab:repro-models}).
\end{itemize}
The page settles adaptively before each observation. The page signature is polled every
0.1\,s until 5 consecutive polls are unchanged. The wait is
never shorter than 1.0\,s and never longer than 5.0\,s.
Navigation is restricted to the three task origins (\texttt{http://portal-health:8080}, \texttt{http://mailpit:8025}, \texttt{http://nextcloud}).

\paragraph{(4) Memo and history semantics.} The memo is part of the action. After parsing, it is truncated
to its first 300 characters. The full raw reply is archived untruncated. Nothing is scored
from a memo. The history block lists the last 48 executed actions, one line each:
\code{step i: <json.dumps(action)> -> ok=<True|False|None> @ <last 46 characters of the URL after the
action>}. The history window (48) is not smaller than the step cap (48), so
\emph{every} earlier action and \emph{every} earlier memo is echoed back at every step. The agent's memory is
therefore the full list of its own memos and actions, not only the latest memo. It carries no earlier page
text or image. \code{json.dumps} escapes non-ASCII characters in the echo (e.g.\ \code{\textbackslash{}u2014}). The done step is not echoed, because the episode ends there. Example: the first history line of
the Opus FRX-ON-A instance-20 episode (285 characters):
\begin{lstlisting}
step 0: {"type": "navigate", "url": "http://portal-health:8080", "memo": "Priya email: priya.sharma@example.org (from Mailpit intake). Next: login portal, check APT-20448 status, sessions page, then Nextcloud calendar for Thu 20 Aug 2026."} -> ok=True @ http://portal-health:8080/login
\end{lstlisting}

In the archive, 2476 of 2548 model actions carried a memo. 126 were cut at
300 characters, and 0 archived memos differ from the
truncated raw reply. The archived Block~A2 episodes had at most 39
steps, so the echo there was complete under any window of at least 38
lines. The harness default is 40 and the later task module declares 48, but A2's own
module value is \emph{unavailable}.

\begin{table}[h]\centering\scriptsize
\caption{Per requested model, over all 82 archived model rollouts: policy steps; steps whose frame
the model Read; actions with a memo; memo length median/max (characters); memos truncated at
300; wall time per policy call median/max (s).}
\label{tab:repro-models}
\begin{tabular}{lrrrrrr}\toprule
model & steps & frame read & memo & memo len & truncated & call s \\\midrule
\texttt{claude-fable-5-1} & 133 & 133 & 128 & 271 / 300 & 20 & 8.4 / 17.3 \\
\texttt{claude-opus-5} & 1148 & 1145 & 1110 & 224 / 300 & 88 & 6.3 / 24.9 \\
\texttt{claude-sonnet-5} & 1267 & 1027 & 1238 & 128 / 300 & 18 & 14.1 / 226.1 \\
\bottomrule\end{tabular}
\end{table}

\paragraph{(5) Invoker settings and model identity.} All 2548 archived
model steps record the sealed invoker (2548 steps). Each
step runs the pinned CLI binary under
\code{sudo -n -u <policy-user> env -i HOME=<policy-home> CLAUDE\_CONFIG\_DIR=<policy-home>/.claude
PATH=/usr/local/bin:/usr/bin:/bin TERM=dumb LANG=C.UTF-8}. The policy user is unprivileged and can read
only the observation directory. The arguments are:
\begin{lstlisting}
<cli> -p <prompt> --model <model-id>
  --allowedTools Read
  --disallowedTools Bash BashOutput KillShell Write Edit MultiEdit NotebookEdit Glob Grep WebFetch WebSearch Task Agent TodoWrite ToolSearch Skill Workflow SlashCommand Monitor SendMessage Artifact EnterWorktree CronCreate RemoteTrigger DesignSync
  --permission-mode default
  --setting-sources ''  --strict-mcp-config
  --output-format stream-json --verbose
\end{lstlisting}

Each call has a 300\,s timeout. A failed call or a provider error is retried on the
same prompt and frame after 20\,s and again after
60\,s. A provider usage limit is never retried: it ends the episode as
infrastructure. \textbf{Decoding, reasoning and output-length settings} are not passed: the argument
list sets no temperature, effort, thinking budget, maximum tokens, maximum turns or system-prompt flag, and
no environment variable is set beyond those listed. They are therefore the CLI's and the provider's
defaults for the model id. Those default \emph{values} were not recorded and are \emph{unavailable}.
Token usage was not recorded for these calls (0 steps with usage).
Within one call the model may call tools several times: \texttt{claude-fable-5-1} made more than one call on 1 step and none on 0; \texttt{claude-opus-5} made more than one call on 17 steps and none on 3; \texttt{claude-sonnet-5} made more than one call on 912 steps and none on 237. \textbf{Tool calls.} The argument
list auto-approves only \code{Read} and denies 25 tools by name. The model also emitted calls to
other tool names: \texttt{claude-fable-5-1} 1 call on 1 step (\texttt{ReportFindings}~1); \texttt{claude-opus-5} 6 calls on 6 steps (\texttt{ScheduleWakeup}~5, \texttt{ReportFindings}~1); \texttt{claude-sonnet-5} 1591 calls on 801 steps (\texttt{ListAgents}~812, \texttt{ScheduleWakeup}~315, \texttt{ReportFindings}~184, \texttt{ExitWorktree}~168, \texttt{CronList}~52, \texttt{navigate}~38, \texttt{select}~6, \texttt{done}~4, \texttt{fill}~4, \texttt{click}~3, \texttt{Bash}~2, \texttt{Navigate}~1, \texttt{PushNotification}~1, \texttt{back}~1). The
invoker archives the \emph{name} of every tool call, but a \emph{result} (executed or refused) only for
\code{Read} calls (\texttt{claude-fable-5-1}: 133 ok; \texttt{claude-opus-5}: 1 error, 1155 ok; \texttt{claude-sonnet-5}: 81 error, 3602 ok). Whether each of the non-\code{Read} calls was refused or executed is
therefore \emph{unavailable} from the archive. The isolation that bounds them is the operating-system
account: it can read only the observation directory, and it has no repository, run-directory or docker
access.
\textbf{CLI version.} Every block's manifest pins the binary's sha256 \code{15e2d05148f8}. The file on disk today has the same digest. Its name is
\code{claude-2.1.274}, and run offline it reports \code{2.1.274 (Claude Code)}.
\textbf{Model identity.} Requested ids: \texttt{claude-fable-5-1}, \texttt{claude-opus-5}, \texttt{claude-sonnet-5}. The evidence of the
served model is the \code{model} field of the CLI's \code{system/init} stream event, recorded on
every step. It equals the requested id on every step of every archived
rollout:
\texttt{claude-fable-5-1}: 133 of 133 steps; \texttt{claude-opus-5}: 1148 of 1148 steps; \texttt{claude-sonnet-5}: 1267 of 1267 steps.
The per-episode label \code{policy\_resolved\_model\_id}, which is derived from those init
events, agrees on all 82 rollouts. The init field is the CLI's own report
of the model it was configured to call. It is not a server attestation. Server-side identity evidence (the \code{model} field of each API response) was not parsed
or archived, so it is \emph{unavailable}.

\paragraph{(6) The FRX-ON-A review display, as rendered.} On FRX-ON-A (and FRX-ON-I), the portal's review-and-submit
step appends one table below the draft, rendered from the \emph{live} \code{clinic\_session} row for
the selected service and date. The table has a title, a header row and one data row. It carries no
sentence: no statement that anything changed and no explicit verification instruction. It does
place the current location beside a draft naming the old one, and that visible disagreement may itself cue a
re-check (Appendix~\ref{sec:blockB}). Its template is
\code{templates/wizard.html}, pinned in the manifests (\code{bb26ee02bca1}), with the title
taken from the deployed spec. Every page that carried the review information in the 25
archived ON rollouts (FRX-ON-A: 19, FRX-ON-I: 6) matched
1 template once the session id, date and location were abstracted. A
scan for 16 cue words (e.g.\ \emph{moved, changed, updated, please, check, verify}) found
0 hits in the review information text. The review display's text channel, with cells
tab-separated as in the prompt:
\begin{lstlisting}
Session details for the selected date
Session    Service    Date    Location
<session id>    Influenza    <session date>    <location>
\end{lstlisting}

Below is the full page text of the review step, as the policy received it on step 20 of the Opus
FRX-ON-A instance-20 episode (names are the generator's fictitious ones). The review-time information is the last table
before the buttons, and it shows the new location while the draft above still names the old one:
\begin{lstlisting}
Northgate Health -- Vaccination Scheduling
Signed in as Clinic Scheduler
All appointments New submission Clinic sessions Switch user
1. Patient details
2. Clinic and time
3. Review and submit
Review and submit
Date    2026-08-20
Appointment number    APT-20455
Clinic    Northgate Main
Contact email    priya.sharma@example.org
Notes    None
Patient full name    Priya Sharma
Time slot    15:00
Vaccine    Influenza
Documents    0 attached
Session details for the selected date
Session    Service    Date    Location
SES-0820-FLU    Influenza    2026-08-20    Riverside Annexe

Submit Back
\end{lstlisting}

\par}

\section{Observability, sensitivity and the within-episode timeline}
\label{app:design}

\begin{table}[h]
\centering
\caption{\textbf{What the agent could know, and when.} Per condition: whether the authoritative row changes,
what announces it, where the current value is visible after the change, and the scripted evidence that the
condition is solvable and discriminating --- a legal policy that scores 1.0 and a declared deficient policy
that does not. The trigger column is the certified anchor: the change fires on the agent's own request, after
it has been served the sessions page. Derived from the certificates, the audit spec and the blocks'
feasibility artifacts; no cell is typed.}
\label{tab:observability}

\begingroup\setlength{\tabcolsep}{2.5pt}\scriptsize
\begin{tabular}{P{1.45cm}P{1.25cm}P{1.05cm}P{2.75cm}P{2.55cm}P{2.05cm}}
\toprule
level & change & announced & current value visible at (trigger) & legal policy (1.0) & deficient policy \\
\midrule
\code{S-L0} & none & -- & sessions page (unchanged) (--) & \code{reference} & \code{--} \\
\code{S-C} & control & modal notice & sessions page; notice (the agent's first \path{/} after exposure) & \code{reference} & \code{over\_apply (0.8)} \\
\code{S-L2} & applicable & modal notice & sessions page; notice (the agent's first \path{/} after exposure) & \code{reference} & \code{stale\_location (0.8)} \\
\code{FRX-OFF-A} & applicable & none & sessions page (re-read) (the agent's first \path{/new/} after exposure) & \code{late\_reread\_ref} & \code{reference (0.8)} \\
\code{FRX-ON-A} & applicable & none & sessions page (re-read); review step (the agent's first \path{/new/} after exposure) & \code{late\_reread\_ref} & \code{reference (0.8)} \\
\code{FRX-OFF-I} & control & none & sessions page (re-read) (the agent's first \path{/new/} after exposure) & \code{--} & \code{--} \\
\code{FRX-ON-I} & control & none & sessions page (re-read); review step (the agent's first \path{/new/} after exposure) & \code{late\_reread\_ref} & \code{--} \\
\bottomrule
\end{tabular}
\endgroup

\end{table}

\begin{table}[h]
\centering
\caption{\textbf{Sensitivity of the paired results to analysis choices} (historical blocks; exploratory, predecessor protocol). Each block is shown as reported
under the rules actually implemented (which counted Sonnet's \SonnetPanelTextFail FRX-ON-A rows whose review display
reached it as text only; ADDENDUM~5 would have voided and replaced them), then with rows whose intervention
never reached a delivered observation in text or image removed (the ADDENDUM~16 definition), then with
action-cap rows removed, then both; a filtered row takes its pair partner with it. For Block~BS a last row
applies ADDENDUM~5's frame rule as registered, voided rows left unobserved (Appendix~\ref{app:rules}).
Deletion is not the registered replacement procedure, whose outcomes are unobserved. Cells count each arm over
the rows an analysis keeps; the exact test uses the complete pairs only, so in the ADD.~5 row the control cell
also holds the rows whose partner that rule voids (``unpaired''). Computed by
\code{audit/sensitivity\_analysis.py}, whose reported column is checked against each block's result file
(Block~B \SensBMatchesResultFile, Block~BS \SensBSMatchesResultFile, Block~A2 \SensAtwoMatchesResultFile).}
\label{tab:sensitivity}

\begingroup\setlength{\tabcolsep}{3.5pt}\small
\begin{tabular}{P{1.15cm}P{2.45cm}P{1.35cm}P{2.6cm}P{2.6cm}P{1.5cm}}
\toprule
block & analysis & complete pairs & control cell & treatment cell & exact $p$ \\
\midrule
\textbf{B} & reported (implemented rules) & 10 & \code{FRX-OFF-A} 0/10 & \code{FRX-ON-A} 10/10 & 0.00195 \\
 & delivered, text or image (ADD.~16) & 10 & \code{FRX-OFF-A} 0/10 & \code{FRX-ON-A} 10/10 & 0.00195 \\
 & no action-cap rows & 10 & \code{FRX-OFF-A} 0/10 & \code{FRX-ON-A} 10/10 & 0.00195 \\
 & both filters & 10 & \code{FRX-OFF-A} 0/10 & \code{FRX-ON-A} 10/10 & 0.00195 \\
\midrule
\textbf{BS} & reported (implemented rules) & 10 & \code{FRX-OFF-A} 0/10 & \code{FRX-ON-A} 6/10 & 0.0312 \\
 & delivered, text or image (ADD.~16) & 9 & \code{FRX-OFF-A} 0/9 & \code{FRX-ON-A} 6/9 & 0.0312 \\
 & no action-cap rows & 8 & \code{FRX-OFF-A} 0/8 & \code{FRX-ON-A} 5/8 & 0.0625 \\
 & both filters & 8 & \code{FRX-OFF-A} 0/8 & \code{FRX-ON-A} 5/8 & 0.0625 \\
 & ADD.~5 frame rule, voided rows unobserved (retro.) & 7 & \code{FRX-OFF-A} 0/10 (3 unpaired) & \code{FRX-ON-A} 6/7 & 0.0312 \\
\midrule
\textbf{A2} & reported (implemented rules) & 8 & \code{S-C} 8/8 & \code{S-L2} 7/8 & 1 \\
 & delivered, text or image (ADD.~16) & 8 & \code{S-C} 8/8 & \code{S-L2} 7/8 & 1 \\
 & no action-cap rows & 8 & \code{S-C} 8/8 & \code{S-L2} 7/8 & 1 \\
 & both filters & 8 & \code{S-C} 8/8 & \code{S-L2} 7/8 & 1 \\
\bottomrule
\end{tabular}
\endgroup

\end{table}

\begin{table}[h]
\centering
\caption{\textbf{Paired statistics.} T / C: pairs in which only the treatment / only the control arm
succeeded. Net difference P(treatment)$-$P(control) over matched instances, ties in the denominator, with a Newcombe (1998, method~10) hybrid-score marginal 95\% interval (not exact, not family-adjusted); exact two-sided sign
test on the discordant pairs. The envos \code{effect} field of the result files is the discordant-pair
\emph{proportion}, which is not a net effect (it reads \AtwoEffect for Block~A2, whose net is \PairedAtwoNet). Holm is a
\emph{retrospective} correction over the \HolmFamily pre-declared primary tests, which were registered at
$\alpha=\Alpha$ each; it is superseded by the paper-wide family of Table~\ref{tab:tests} and kept for comparison. The last row drops the pairs in which either arm hit the action cap (retrospective).
Block~A2S's test was withdrawn before any rollout (ADDENDUM~13); its row is descriptive. All rows are historical and
exploratory (predecessor protocol); the intervals are recomputed under the dated erratum of
Appendix~\ref{app:multiplicity}.}
\label{tab:paired}

\begingroup\setlength{\tabcolsep}{3pt}\footnotesize
\begin{tabular}{P{1.9cm}P{2.2cm}rcccc}
\toprule
block & treatment vs control & pairs & T / C / tied & net [95\% CI] & exact $p$ & Holm \\
\midrule
A2 & S-L2 vs S-C & 8 & 0 / 1 / 7 & $-0.13$ [$-0.47$, $0.22$] & 1 & 1 \\
B & FRX-ON-A vs FRX-OFF-A & 10 & 10 / 0 / 0 & $+1.00$ [$0.61$, $1.00$] & 0.00195 & 0.00586 \\
BS & FRX-ON-A vs FRX-OFF-A & 10 & 6 / 0 / 4 & $+0.60$ [$0.20$, $0.83$] & 0.0312 & 0.0625 \\
A2S & S-L2 vs S-C & 4 & 2 / 0 / 2 & $+0.50$ [$-0.10$, $0.85$] & 0.5 (descr.) & -- \\
BS, no action-cap pairs & FRX-ON-A vs FRX-OFF-A & 8 & 5 / 0 / 3 & $+0.63$ [$0.17$, $0.86$] & 0.0625 (retro.) & -- \\
\bottomrule
\end{tabular}
\endgroup

\end{table}

\begin{figure}[h]
\centering
\includegraphics[width=\textwidth]{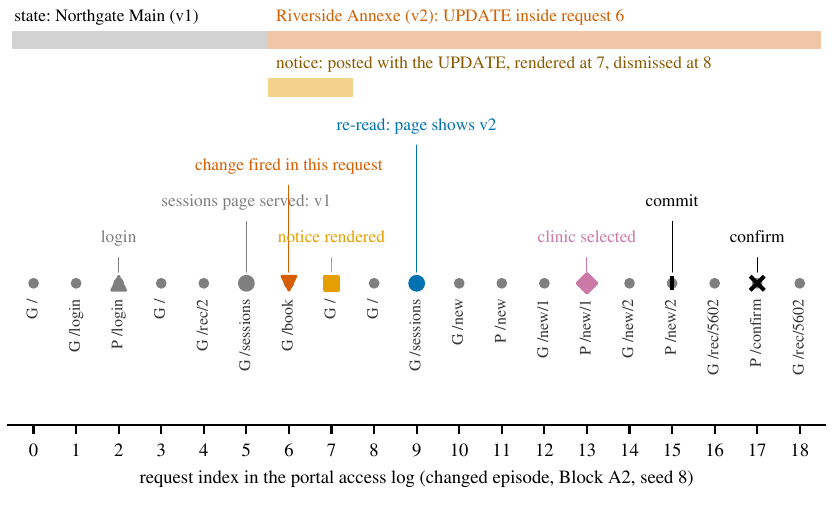}
\caption{\textbf{State precedes channel.} One changed episode (Block A2, seed 8) on the axis of the
agent's own requests to the portal. The authoritative row changes inside the request the exposure
condition selects, the notice is posted with the same mutation and rendered on the next request, and the
agent's re-read of the sessions page shows the new location before the clinic is selected and committed.
Every marker is read from the episode's access log, event log and checkpoints.}
\label{fig:timeline}
\end{figure}

\begin{figure}[h]
\centering
\includegraphics[width=\textwidth]{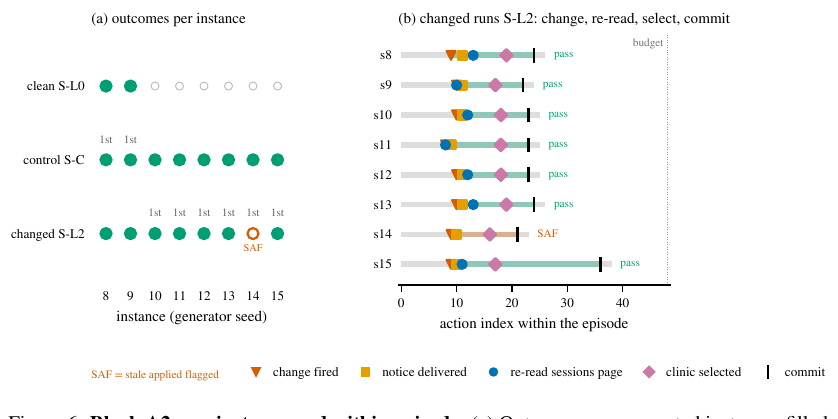}
\caption{\textbf{Block A2, per instance and within episode.} (a) Outcomes per generated instance:
filled, pass; open, fail (the failure's signature code is written under it and spelled out in the key);
``1st'' marks the arm the seeded order ran first. (b) For each changed run, the action index at which the
change fired, the notice frame was delivered, the agent re-read the sessions page, selected the clinic in
the wizard, and committed; the shaded segment spans change to commit. \AtwoSuccessReread of \AtwoSuccessN
successful runs re-read the authoritative page at or after the request that fired the change and before
selecting (the page served by the firing request already carries the new location, so the re-read and the
change can coincide); the one failure did not. Dotted line: the \ActionBudget-action budget.}
\label{fig:blockA2}
\end{figure}

\begin{figure}[h]
\centering
\includegraphics[width=\textwidth]{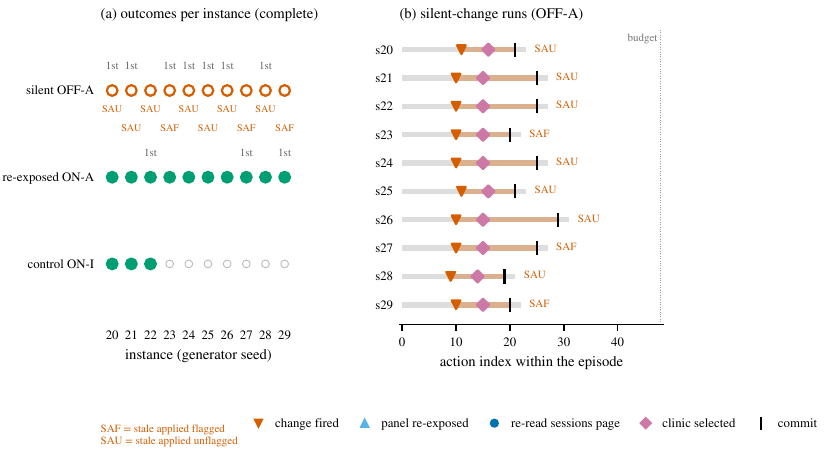}
\caption{\textbf{Block B: silent supersession, with and without re-exposure} (Claude Opus~5,
\BlockBPairs paired instances). Same encoding as Figure~\ref{fig:blockA2}; the review-information marker on FRX-ON
arms is the review step re-rendering the live row. FRX-OFF-A \BlockBFRXOffAPassFrac, FRX-ON-A
\BlockBFRXOnAPassFrac, $p=\BlockBP$.}
\label{fig:blockB}
\end{figure}

\clearpage
\section{Historical, hypothesis-generating evidence (predecessor protocol)}
\label{sec:historical}

Every block in this section ran under a predecessor version of the protocol (success-first counting, an unenforced
review-information delivery rule; Appendix~\ref{app:rules}); none is pooled with the prospective blocks, and the paper's causal
conclusions do not rest on them.

\paragraph{Silent change and the direct display of the current value (Block~B).}
\label{sec:blockB}
Historical Block~B motivated the prospective experiments but is not treated as confirmatory evidence under the
finalized protocol. Its evaluator recorded a \PairedBTreatOnly/\BPairs{} paired change in booked-clinic correctness
between no additional information and value-review-time information conditions, and archive re-scoring independently verifies the booked-clinic term on
all \BPairsWords{} pairs; however, the block did not satisfy one preserved pre-registered claim condition, so we
report the contrast as hypothesis-generating historical evidence rather than as a confirmatory rescue claim. Its
successes are evaluator passes whose booked clinic is independently verified; none is verified on the full endpoint,
because the historical archives do not hold the slot occupancy the nearest-slot term needs (\IndepFullBFRXOnA{} of
\BlockBFRXOnAPass; Appendix~\ref{app:coverage}). The claim condition, the post-outcome sentence that read the contrast
from the primary test alone, and the alternative explanations the block cannot separate---among them cueing a
re-check versus delivering the value, the question Block~BC was built for---are in the supplementary provenance file
(Appendix~\ref{app:blockB}).
Descriptively, no OFF-A run of the silent blocks requested the sessions page after the change (\RequeryOffAFrac),
although a legal scripted policy that does so solves \FeasBOffAReferenceSolved{} of \FeasBOffAReferenceN{} OFF-A
instances.

\paragraph{Other predecessor blocks.} A Sonnet~5 block on Block~B's instances, a Fable~5.1 screen, an announced-change
block whose result did not reproduce an earlier failure of ours (Block~A2), the exploratory corpus behind the
re-verification hypothesis and their sensitivity analyses are reported, with their limitations, in
the supplementary provenance file (Appendix~\ref{app:percell}) and Appendices~\ref{app:design}
and~\ref{app:corpus}.

\section{Historical records and registration provenance (supplementary files)}
\label{app:percell}\label{app:episodes}\label{sec:blockA2}\label{sec:nonreplication}\label{app:blockB}%
\label{app:alternatives}\label{app:precommit}\label{tab:main}\label{tab:chronology}
The historical blocks are motivation and provenance only (Appendix~\ref{sec:historical}); none carries a claim of this
paper. Their per-cell rates, the per-episode tables of the announced (A2, A2S) and silent (B, BS, BF) predecessor
blocks with the detailed accounts moved there from earlier versions of this paper (Block~A2's non-replication, Block~B's
report labels and registered claim rule, and the alternative explanations of its effect of review-time information), and the full registration
chronology with its errata are therefore no longer typeset. They ship as a generated supplementary file
(\code{supp\_provenance/EnvOS\_historical\_provenance.tex} and its PDF, listed with their digests in
\code{supp\_provenance/MANIFEST.json}), which the claim ledger writes from the same pinned artifacts, macros and generated
tables as this paper, so every number in it is a ledger entry. For orientation: the supersession family's
pre-registration (\code{PRECOMMIT\_P3v2.md}) is frozen by sha256 (\PrecommitSha) with \PreAddenda dated addenda below the
freeze marker, and the claims pre-registration is a file of the claims blocks' frozen evidence bundle (digest list
\ClmBundleSha); both ship as files, byte for byte. Registered-versus-implemented rules are reconciled in
Appendix~\ref{app:rules}, and the final prospective specifications are in Appendices~\ref{app:bc}--\ref{app:claims}.

\section{Exploratory corpus (Stack B)}
\label{app:corpus}
This section documents the earlier exploratory corpus. Its records explain how the protocol developed; they do
not support quantitative model-effect claims in this paper.

The corpus was produced on a separately built stack with its own harness, and is reported in
Appendix~\ref{sec:historical} as the source of the organising hypothesis, never as a model-level claim. Its
structure, re-derived by the claim ledger from the corpus's per-arm and per-rollout files
(\code{audit/failure\_mining/STACKB/}, where the corpus's own ledger script agrees on every count): \StackBArms arms
across \StackBEnvsWords environments (a clinic, a hotel and a travel desk), of which \StackBPhenArms are hand-written
phenomena --- among them a selected resource taken between selection and commit, a write silently altered after
the commit, a fact revised with no notification, an entity merged under a new name, a policy reissued off the
agent's path and an acknowledgement lost --- and \StackBControlArms are controls (clean baselines and cosmetic
changes). Claude Opus~5 ran all \StackBArms arms, \StackBOpusRollouts rollouts in all, from \StackBKMin to
\StackBKMax per arm (\StackBKThreeArms arms at three); Claude Sonnet~5 ran \StackBSonnetArms arms,
\StackBSharedArms of them also run on Opus~5, so the two models together cover \StackBUnionArms arms.

\paragraph{Validity ledger.} Four properties keep it exploratory, and each is a reason no rate from it is
reported. (i) No arm carries a certificate in the sense of Section~\ref{sec:certification}: the stack's own test
suites do not establish that the change fired once, inside the agent's own request, after the fact it supersedes
was served. (ii) Its triggers are hand-coded: heuristics on the agent's own actions (a page load followed by a
delay, a view or reservation of the target, a submit) or, in the travel desk's cold-start arm, the episode's start, never certified
against the value the agent was served,
so an episode in which the agent never saw the superseded value is indistinguishable, from the outcome alone,
from one in which it did --- the ambiguity the exposure-conditioned trigger of \S\ref{sec:triggers} removes. (iii) Several
cells hit a turn cap, and capped episodes were handled inconsistently (counted as failures in some cells, voided
and re-run in another), so a denominator does not uniformly separate ``acted on a stale belief'' from ``ran out of
turns''. (iv) The corpus's verifier was corrected during its own life, and its versions are not recorded:
every archived Opus~5 run record carries a re-score field (\StackBRescoreField of \StackBOpusRollouts), of which
\StackBRescoreFlips record a stored failure re-scored as a pass, and a later fix to the hotel task's verifier turned
some Sonnet~5 failures on that task into passes and is reported by the corpus's own document, not re-derived
here, to leave that task's archived Opus~5 rollouts unchanged. Per-arm pass counts are therefore left to the corpus's own document; the
ledger re-derives only the structure above, from files whose digests it pins.

\section{Full defect table}
\label{app:defects}
The defects include a restarted block in which \SupersededBlockNotTriggered{} of \SupersededBlockEpisodes{} episodes
never triggered the change, a provider session-limit reply scored as a model decision, and a classifier that would have
voided \CorrTenCounterfactual{} model failures of the historical silent blocks; some showed only in model evidence, which
no scripted policy produces.

\begin{table}[h]
\centering
\caption{\textbf{Certification challenge.} For each of the \NDefectsTotal defects: the inference it permitted,
the executable check that rejects it now, and the corrected conclusion; ``rows (model)'' counts archived
search-log rows scored under the defect and how many already carried a model verdict.}
\label{tab:defect-checks}

\begingroup\setlength{\tabcolsep}{3pt}\scriptsize
\begin{tabular}{P{0.3cm}P{3.3cm}P{4.2cm}P{3.7cm}P{1.0cm}}
\toprule
\# & inference the defect permitted & executable check that rejects it now & corrected conclusion & rows (model) \\
\midrule
1 & the clean control fails, so the family is hard & clean-control stop rule (\code{run\_block}: a counted clean failure halts the block) + the memo channel declared by the task & harness lacked working memory; attempt voided; Block~A2 clean \AtwoCleanPass/\AtwoCleanN & 1 (1) \\
2 & a session-limit reply is a model failure & provider-limit signal in the sealed invoker \(\to\) \code{infra\_rejected}, block pauses, reserve untouched & not counted; no failure rate moved & 1 (1) \\
3 & a changed-arm success with no change is evidence of recovery & exposure-conditioned trigger (\code{requires\_seen}) + validity ladder branch \code{not\_triggered} + pause rule & \SupersededBlockNotTriggered of \SupersededBlockEpisodes episodes were not instances; re-run as Block~A2 & 11 (8) \\
4 & correct runs read as unexposed & scorer places the change on the firing request; C4 recomputed; independent scorer cross-checks & exposure certified in \AtwoVOneExposed/\AtwoChangedN changed runs & 3 (3) \\
5 & the family's reference policy solves the silent arm, so its failures are model deficits & per-level certification of the expected matrix before any model: the ordinary \code{reference} must score 0.8 and the \code{late\_reread\_reference} 1.0 & trigger timing kept (first wizard request after exposure); reference redefined to re-read the source inside the wizard (\FeasBOffAReferenceSolved/\FeasBOffAReferenceN at 1.0), ordinary reference declared deficient; certificate \CertFRXOffAChecks/\CertFRXOffAChecks & -- \\
6 & a 6-pair block is adequate for the rescue test & pre-registered power arithmetic (\code{envos.stats}) checked at freeze & 6 unanimous pairs do reject ($p=\PSignSixZero$), but one tie or one pair against forfeits rejection; enlarged to \BPairs pairs before any Block~B rollout, which tolerates one pair against ($p=\PSignBOneAgainst$) & -- \\
7 & a silent-arm stale booking is not a counted failure & ladder branch 5 skips \code{silent\_arm} checkpoints; re-derived from archive & Block~B row 1 re-derived from \code{delivery\_ambiguous} to \code{model\_failure}; OFF-A \BlockBFRXOffAPassFrac & -- \\
8 & the agent never dismissed the notice & independent scorer: raw access log vs C3, both directions & labels of \AtwoRederivedSignatures rows corrected; verdicts unchanged; dismissal \AtwoChangedDismissedLog/\AtwoChangedN & 5 (5) \\
9 & an errored read voids the episode & read-target rule voids only a successful read of a path the harness did not hand over & \DefectIRows Sonnet attempts stayed void, cells replaced under the fixed rule; Block~BS completed & 3 (0) \\
10 & a silent-arm run labelled flagged told the user about the change & archive audit re-reads every report and quotes the sentence behind each flag (\code{FLAG\_AUDIT}) & labels kept; no silent-arm report states the move; \S6 not met as registered for Block~B & 3 (3) \\
11 & Sonnet's rescue-arm failures ignored a delivered review display & delivery standard applied to the re-exposure checkpoint; channel recorded per step & review information reached Sonnet as text only in all \DefectKRows; confounded with delivery; Opus unaffected & 3 (3) \\
\bottomrule
\end{tabular}
\endgroup

\end{table}

The defects divide by what caught them: design-time gates, validity rules on a block's first rows, and archived
model evidence alone, because no scripted policy produces them (an unauthenticated first request, a dismissal from
the sessions page, a report naming the distractor's clinic, and a review information read as text alone). No fix was validated by
re-running a model: classification, the checkpoints and the signature are pure functions of the archive, so every
corrected row was re-derived from it with the previous value kept beside the new.

\NDefectsTotal defects with archived evidence. For \NDefectsModelAttributable of them an archived
search-log row scored under the defect already carried a model verdict (\code{success} or
\code{model\_failure}): they would have been reported as model outcomes. The other \NDefectsBlockB were
caught before any model-scored row existed under them (at Block~B's design or certification time, or on a
not-counted verdict after its first rollout). The last column counts the archived rows scored under each
defect and, in brackets, those with a model verdict; the count \NDefectsModelAttributable is derived from
that column, not from the order of this table.
\begingroup\small
\let\tabular\longtable\let\endtabular\endlongtable

\begingroup\setlength{\tabcolsep}{4pt}
\begin{tabular}{P{0.3cm}P{6.3cm}P{4.3cm}P{1.9cm}}
\toprule
\# & defect & archived evidence & rows (model) \\
\midrule
1 & no working memory across stateless per-step calls; the clean control failed and the stop rule halted the block & \path{P3v2/_void_blockA_attempt1_2026-09-17/}; \path{P3v2/_void_blockA_attempt1_2026-09-17/search_log_blockA.json}; \path{PRECOMMIT_P3v2.md} & 1 (1) \\
2 & a provider session-limit reply was scored as a model failure & \path{P3v2/_void_blockA_attempt1_2026-09-17/}; \path{P3v2/_void_blockA_attempt1_2026-09-17/search_log_blockA.json}; \path{PRECOMMIT_P3v2.md} & 1 (1) \\
3 & the trigger depended on the agent revisiting a page, so episodes were not instances of the phenomenon & \path{P3v2/_superseded_blockA_2026-09-17/}; \path{P3v2/_superseded_blockA_2026-09-17/search_log_blockA.json}; \path{PRECOMMIT_P3v2.md} & 11 (8) \\
4 & the scorer located the change one request early, so the exposure check read false on correct episodes & \path{P3v2/_superseded_blockA_2026-09-17/}; \path{P3v2/_superseded_blockA_2026-09-17/search_log_blockA.json}; \path{PRECOMMIT_P3v2.md} & 3 (3) \\
5 & the silent arm as first frozen defeated every policy that re-reads before the wizard, including the certified reference & \path{PRECOMMIT_P3v2.md} & -- \\
6 & six pairs could reject only if every pair crossed one way & \path{PRECOMMIT_P3v2.md} & -- \\
7 & the classifier read a silent arm's absent perception checkpoint as 'not certified' and would have discarded every silent-arm failure as invalid & \path{PRECOMMIT_P3v2.md} & -- \\
8 & the perception checkpoint dropped a dismissal made from a page its route table did not name, so announced-arm failures dismissing there would have been discarded as invalid (found by the independent scorer on the model rows) & \path{P3v2/FINAL_REPORT_blockA2_ERRATUM.md}; \path{P3v2/reward_audit_runs/REWARD_AUDIT_defect8_detected_2026-09-21.json}; \path{P3v2/search_log_blockA2.json}; \path{PRECOMMIT_P3v2.md} & 5 (5) \\
9 & the read-target validity rule voided episodes on Reads that obtained nothing (errored reads, /dev/null), so a model with that habit could not complete a block & \path{P3v2/search_log_blockBS.json}; \path{audit/staged/repair12_receipt.json}; \path{PRECOMMIT_P3v2.md} & 3 (0) \\
10 & the pre-registered report-flag rule counts the new clinic's name wherever a report mentions it, so silent-arm reports naming the distractor appointment's clinic were labelled as flagging the change (found by an archive audit of the model reports) & \path{P3v2/FLAG_AUDIT.json}; \path{P3v2/search_log_blockB.json}; \path{PRECOMMIT_P3v2.md} & 3 (3) \\
11 & the pre-registered rule voiding a re-exposure row whose review observation frame was not delivered was never enforced, and delivery was redefined as text or image after the outcomes were known & \path{P3v2/search_log_blockBS.json}; \path{PRECOMMIT_P3v2.md} & 3 (3) \\
\bottomrule
\end{tabular}
\endgroup

\endgroup

\section{Extended related work}
\label{app:related}
Section~\ref{sec:related} positions this work against four lines; this appendix says what each shares
with \envos and where it stops.

\paragraph{Benchmarks that make the environment move.} OSWorld~2.0 \citep{yuan2026osworld} is the
closest in ambition: 31 services, long-horizon workflows, mid-task messages and a simulated user, and it
reports the symptom this paper studies. Gaia2 \citep{froger2026gaia2}, built in the ARE platform
\citep{are2025}, lets asynchronous events change state while time flows, and pairs each scenario with a
write-action verifier --- the closest existing work to our invariant scoring. Real-Time Reasoning Gym \citep{wen2026realtime} keeps the environment evolving while the agent is still
reasoning and trades reasoning depth against latency; our change is one discrete intervention rather than
continuous evolution of the environment. In $\tau^2$-bench \citep{barres2025tau2bench} a simulated
user changes shared state through its own tools as part of a cooperative task, and InterruptBench
\citep{zou2026interruptbench} injects announced revisions of the user's goal into long-horizon WebArena-Lite
tasks; our changes alter world state rather than the goal, and include silent arms with no message at all.
DynamicGUIBench \citep{dynamicguibench2026} changes interfaces between actions and answers with a
video-input agent; AgentHijack \citep{agenthijack2026} corrupts perception and control. StressWeb
\citep{bai2026stressweb} compares perturbed with clean runs of the same tasks and finds agents that
confidently report success on tasks they failed --- the signature our silent arms also show --- but its
perturbations act on layout, interaction semantics and execution, not on authoritative backend state.
STT-Arena \citep{hui2026sttarena} is the closest design: a trigger fires when a condition on state and
spatio-temporal context holds, applies an effect that can invalidate the agent's plan, and, for solvable tasks, is scored by
rule-based checks on the final state; the benchmark is compared with non-dynamic environments in aggregate,
and it names stale-state execution as a dominant failure mode. Its environments are Python tool APIs rather
than a browser, its trigger conditions are defined on state and context rather than on what the agent has
observed, and we found no per-instance paired test in it. Our design combines a change fired only after the
agent has been served the value it supersedes, with that exposure certified from archives; inapplicable controls
sharing the changed arm's trigger (and, for an announced change, its notice); and a matched re-exposure test.
Mitigation has also been studied: \citet{li2026aoi}, discussed below, test observation-interface changes at
benchmark scale. These benchmarks answer ``do agents fail when the world moves''; the design here asks, under one particular change fired after one particular observation,
whether one interface element---the review-time information---removes the failure. The change itself is present in
both arms, so its own causal effect is not estimated (in the historical blocks FRX-OFF-I was certified but never
run on a model; Block~BGC adds it). The two are complementary: breadth of setting against control of one mechanism.

\paragraph{Stale belief as an object of study.} Reading a value and committing after it has been
superseded is the classic time-of-check-to-time-of-use race. \citet{lilienthal2025toctou} study it in LLM
agents as a security vulnerability (TOCTOU-Bench, 66 tasks) and propose defences such as state-integrity
monitoring; we measure it as a behavioural failure under a benign change fired inside the agent's own
check-use window. Knowledge-conflict studies \citep{xie2024knowledgeconflict} show that LLMs can adopt
coherent contradicting evidence yet show confirmation bias toward what they already believe, in static
question answering. STALE
\citep{chao2026stale} asks whether a model notices that a stored memory has been invalidated, in question
answering, without actions or a world to act on. Our silent arms are the acting counterpart of both: the belief is formed by reading a page, the
invalidation is a real row changing, and the cost is a durable write. EnvProbe \citep{song2026envprobe}
approaches the same problem from the policy side, deciding when an agent should re-query a
belief; our re-exposure arm is the environment-side counterpart. In the historical, hypothesis-generating clinic
Block~B (predecessor protocol; Appendix~\ref{sec:blockB}), putting the current value on Opus~5's path immediately
before the commit coincided with a \PairedBTreatOnly/\BPairs{} paired change in evaluator-recorded booked-clinic
correctness, consistent with the interface carrying a burden the policy was not carrying; those successes are not
verified on the full endpoint and are not confirmatory. \RelatedClaimsAnchor{} \citet{li2026aoi} reach a
compatible conclusion at benchmark scale: richer between-step observation (keyframes, audio transcription,
narration) gains 17 to 48 points on 100 dynamic browser tasks, with a 50-task static control. Our
re-exposure arm is narrower and pre-registered: one element, the current value on the agent's path
immediately before the commit, tested on matched instances with an exact paired test.

\paragraph{Verifier validity and evaluation auditing.} TheAgentCompany \citep{xu2025theagentcompany}
self-hosts a reproducible workplace of services with simulated colleagues and checkpoint-based partial
credit; our testbed is a workplace of self-hosted services in the same style. OpenComputer \citep{wei2026opencomputer} builds
verifiable software worlds with state verifiers and shows hard-coded verifiers agree with human
adjudication better than a model judge, an empirical validation of outcome verifiers on static tasks. ABC \citep{zhu2025abc}
turns task-setup and reward-design flaws found in existing agentic benchmarks into a checklist and shows
such flaws can shift measured performance by up to 100\% relative; our defect taxonomy is evidence of the
same kind for one dynamic protocol, with protocol-specific checks for trigger timing relative to exposure
and delivery of the evidence.
\citet{kirgis2026logs} argue that outcome-only agent evaluation is not credible and give a taxonomy of
threats documented through execution-log analysis --- the argument this paper partially executes, for the
particular case of a dynamic intervention, where the verdict is not the only thing needing certification:
the intervention, its timing relative to the agent's own observations, and the delivery of the evidence
all do. Related critiques \citep{notsingletarget2026,predictivevalidity2026} question what a single
leaderboard number means at all. Our defect taxonomy (\S\ref{sec:caught}) is the empirical form of that
argument: \NDefectsModelAttributable of \NDefectsTotal defects had already produced rows carrying model
verdicts.

\paragraph{Conventions we inherit.} The testbed follows the task and scoring conventions of the
computer-use, web-agent and tool-use benchmarks \citep{xie2024osworld,zhou2024webarena,trivedi2024appworld,
yao2024taubench}; the exact paired test is McNemar's \citep{mcnemar1947} and the intervals are Wilson's
\citep{wilson1927}. Neither statistical choice is novel here; what is pre-registered is which single
contrast they are applied to.

\section{Manifests and the claims ledger}
\label{app:manifests}
\paragraph{Release manifests.} Block~A2: \ManifestPaths pinned repository paths, \ManifestImages
container image ids, $\lambda=\LambdaShape$, reward version \RewardVersion, protocol version
\ProtocolVersion, manifest sha256 \AtwoManifestSha. Block~B: \ManifestBPaths pinned paths. A manifest pins the sha256
of every path it names, as hashed at freeze time; the rollout driver recomputes every one before each episode and
refuses to start on any difference, which is what ``the implementation did not change during the block'' means here.
The manifests are distributed as files; their path-by-path digest tables are not typeset.

\paragraph{Disposition of every registered block.} Every block the pre-registrations name and what became
of it (table below): completed (A2, B, BS; A2S and BF as the screens ADDENDUM~13 downsized
them to), void or superseded (Block~A attempt~1; the Block~A restart), never started (D; P1MS), cancelled before
any rollout (BR, BRS, by the consult decision ADDENDUM~9 deferred to; no addendum recorded it, as ADDENDUM~9
required), failed admission (BG, whose leak audit did not pass; P2S/P2SS, whose re-certification was rejected
with \PtwoRecertFailed failed checks), and started but unevaluable (P1M). No outcome of any block that ran is
omitted from this paper. The clean-control stop rule makes a stop a ``harness finding'' by rule, not by diagnosis:
a valid clean task can fail by chance and would stop a block the same way. For Block~A attempt~1, the one block it
stopped, the attribution to the harness rests on ADDENDUM~2's separate diagnosis (the policy had no working
memory), not on the stop. Counts are read from the search logs; the category is the authors' reading of the
cited artifact.
\begin{center}

\begingroup\setlength{\tabcolsep}{3pt}\scriptsize
\begin{tabular}{P{1.4cm}P{1.7cm}rrP{4.6cm}P{3.0cm}}
\toprule
block & disposition & att. & cnt. & reason & artifact \\
\midrule
A (attempt 1) & void & 1 & 1 & clean-control failure stopped the block (ADD.~2); harness lacked working memory & \path{P3v2/_void_blockA_attempt1_2026-09-17/search_log_blockA.json} \\
A (restart) & superseded & 12 & 8 & paused on the infrastructure rule; trigger replaced (ADD.~3); never pooled & \path{P3v2/_superseded_blockA_2026-09-17/search_log_blockA.json} \\
A2 & completed & 19 & 18 & pre-registered primary test (Section~\ref{sec:blockA2}) & \path{P3v2/BLOCK_A2_RESULT.json} \\
A2S & completed (screen) & 9 & 9 & downsized, test withdrawn before any rollout (ADD.~13) & \path{P3v2/BLOCK_A2S_RESULT.json} \\
B & completed & 23 & 23 & pre-registered primary test (Appendix~\ref{sec:blockB}) & \path{P3v2/BLOCK_B_RESULT.json} \\
BS & completed & 26 & 23 & one repair cycle (ADD.~12) & \path{P3v2/BLOCK_BS_RESULT.json} \\
BF & completed (screen) & 5 & 5 & downsized before any rollout (ADD.~13); silent arm and one control only & \path{P3v2/BLOCK_BF_RESULT.json} \\
D & never started & 0 & 0 & demoted to development, its rollouts released to Block~B (ADD.~4) & \path{jobs/continuation.json} \\
BR & cancelled & 0 & 0 & consult round 13 skipped it before any rollout; ADD.~9 required an addendum saying so, none was written & \path{jobs/continuation.json} \\
BRS & cancelled & 0 & 0 & as BR & \path{jobs/continuation.json} \\
BG & failed admission & 0 & 0 & leak audit did not pass (the model's frame read failed); provider quota then exhausted & \path{jobs2/world2.json} \\
P1M & unevaluable & 4 & 1 & both intervention rollouts exhausted the budget before the committing request; paused & \path{P1/search_log_blockP1M.json} \\
P1MS & never started & 0 & 0 & blocked behind P1M & \path{jobs2/world2.json} \\
P2S & failed admission & 0 & 0 & re-certification rejected; admission gate never passed & \path{jobs2/world2.json} \\
P2SS & failed admission & 0 & 0 & as P2S & \path{jobs2/world2.json} \\
\bottomrule
\end{tabular}
\endgroup

\end{center}

\paragraph{Claims ledger.} \emph{What it checks.} Every macro this paper uses is an entry of the ledger
(\code{claims.json}, written by \code{claims.py}) that records its value, the artifacts it is derived from with their
sha256, the statistical unit and the aggregation; \code{python claims.py --verify} re-derives every entry, fails on
any drift of a pinned artifact and on any macro used but not defined, and checks the abstract's generated clauses, the
main text's guarded wording and the interpretation templates' seal. \emph{What is pinned.} Frozen artifacts are read
in place against their digests (the claims blocks' evidence bundle against its digest list, \ClmBundleSha); a live
artifact (a search log its block was still writing; the scorer source) is read through a dated snapshot under
\code{paper/snapshots/}, taken by the build, whose digest is the one pinned; an entry read with no snapshot has its
drift reported, not failed, and the verification fails if the manuscript uses it, directly or inside another macro
(in this build, \LiveUsedText). \emph{How to reproduce.} \code{./build.sh} snapshots the live artifacts, regenerates
the ledger, every table and every figure, compiles the paper and runs the verification; \code{./build.sh --stress}
builds a temporary copy with synthetic worst-case prospective results and checks its page limit. \emph{What is
missing.} Evidence that was never archived, or was overwritten, cannot be re-derived by any build; the next paragraph
lists it. The per-macro ledger table (value, source artifact, digest prefix) is distributed with
\code{claims.json} rather than typeset.

\paragraph{Supplement inventory and evidence gaps.}\label{app:inventory} \emph{Records:} \SuppRecordsShip. The packaged
supplement holds \SuppFiles files (\SuppZipMB\,MB zipped). \SuppReproResult{} Its complete inventory---what is packaged,
what exists but is not packaged and why, and what was not yet produced---is a generated supplementary file
(\code{supp\_provenance/}, Appendix~\ref{app:percell}).\SuppClaimsAfter{} \emph{Permanently missing} (never archived, or overwritten; no package can supply them, and only new rollouts can
replace what they would have shown): Block~BS's first manifest (\BSPreAmendManifest), known by digest only; the
\SuppIrrecoverableText; the first outputs of Block~A2's mutation battery and leak audit,
overwritten later that day; the first transfer run's output; the scorer digest of the certification audit; the
CLI's built-in system prompt, tool definitions, decoding and reasoning defaults and token usage; the results of
non-\code{Read} tool calls, including whether Sonnet's \NonReadCallsSonnet other calls executed; per-step page titles and
element lists, so the observation a policy received cannot be reconstructed exactly; server-side model identity; and
slot occupancy at booking time and calendar compatibility in the historical blocks, so no historical success can be
verified in full.

\section{Video index}
\label{app:videos}
Videos are rendered from the archived frames of real episodes --- no episode is re-run to make one ---
and each manifest records the frame digests and the source trajectory. The pairs that show the paper's
result directly are \code{blockB\_s20} and \code{blockB\_s21}: the same instance under FRX-OFF-A and
FRX-ON-A, where Claude Opus~5 commits the superseded clinic in the first and, in the second, goes back
to change its selection after the review step re-renders the live row. \code{blockBS\_s22} is the
counterpart in which Sonnet~5 is \emph{not} rescued. \code{blockA2\_s8} and \code{blockA2\_s14} show the
announced arm, the second being the block's single failure.
\begin{center}
\small

\begingroup\setlength{\tabcolsep}{4pt}
\begin{tabular}{P{5.4cm}rP{2.0cm}P{3.8cm}}
\toprule
file & duration (s) & sha256 & episodes \\
\midrule
\texttt{ambiguous\_\allowbreak{}commit\_\allowbreak{}failure\_\allowbreak{}demo.mp4} & 59.0 & \texttt{c3814b95} (trajectory) & claude-fable-5-1 AC-L4 changed None \\
\texttt{blockA2\_\allowbreak{}s14\_\allowbreak{}control\_\allowbreak{}vs\_\allowbreak{}changed.mp4} & 198.04 & \texttt{0bdacf63} (video) & S-C S-test\_\allowbreak{}id-14 APPLIED\_\allowbreak{}CORRECTLY; S-L2 S-test\_\allowbreak{}id-14 STALE\_\allowbreak{}APPLIED\_\allowbreak{}FLAGGED \\
\texttt{blockA2\_\allowbreak{}s8\_\allowbreak{}control\_\allowbreak{}vs\_\allowbreak{}changed.mp4} & 208.04 & \texttt{182d1726} (video) & S-C S-test\_\allowbreak{}id-8 APPLIED\_\allowbreak{}CORRECTLY; S-L2 S-test\_\allowbreak{}id-8 APPLIED\_\allowbreak{}CORRECTLY \\
\texttt{blockBS\_\allowbreak{}s20\_\allowbreak{}OFF-\allowbreak{}A\_\allowbreak{}vs\_\allowbreak{}ON-\allowbreak{}A\_\allowbreak{}sonnet5.mp4} & 290.04 & \texttt{573303be} (video) & FRX-OFF-A S-test\_\allowbreak{}id-20 STALE\_\allowbreak{}APPLIED\_\allowbreak{}UNFLAGGED; FRX-ON-A S-test\_\allowbreak{}id-20 REVISED\_\allowbreak{}ON\_\allowbreak{}REEXPOSURE \\
\texttt{blockBS\_\allowbreak{}s22\_\allowbreak{}OFF-\allowbreak{}A\_\allowbreak{}vs\_\allowbreak{}ON-\allowbreak{}A\_\allowbreak{}sonnet5\_\allowbreak{}not\_\allowbreak{}rescued.mp4} & 271.04 & \texttt{ebd34ddc} (video) & FRX-OFF-A S-test\_\allowbreak{}id-22 STALE\_\allowbreak{}APPLIED\_\allowbreak{}UNFLAGGED; FRX-ON-A S-test\_\allowbreak{}id-22 STALE\_\allowbreak{}APPLIED\_\allowbreak{}FLAGGED \\
\texttt{blockB\_\allowbreak{}s20\_\allowbreak{}OFF-\allowbreak{}A\_\allowbreak{}vs\_\allowbreak{}ON-\allowbreak{}A.mp4} & 203.04 & \texttt{12f8f746} (video) & FRX-OFF-A S-test\_\allowbreak{}id-20 STALE\_\allowbreak{}APPLIED\_\allowbreak{}UNFLAGGED; FRX-ON-A S-test\_\allowbreak{}id-20 REVISED\_\allowbreak{}ON\_\allowbreak{}REEXPOSURE \\
\texttt{blockB\_\allowbreak{}s21\_\allowbreak{}OFF-\allowbreak{}A\_\allowbreak{}vs\_\allowbreak{}ON-\allowbreak{}A.mp4} & 239.04 & \texttt{9c2fa9a3} (video) & FRX-OFF-A S-test\_\allowbreak{}id-21 STALE\_\allowbreak{}APPLIED\_\allowbreak{}UNFLAGGED; FRX-ON-A S-test\_\allowbreak{}id-21 REVISED\_\allowbreak{}ON\_\allowbreak{}REEXPOSURE \\
\texttt{p1\_\allowbreak{}ambiguous\_\allowbreak{}commit\_\allowbreak{}AC-\allowbreak{}L3\_\allowbreak{}fable\_\allowbreak{}rlstyle.mp4} & 110.04 & \texttt{86ca27ea} (video) & AC-L3 ambiguous\_\allowbreak{}commit\_\allowbreak{}booking ? \\
\texttt{p3\_\allowbreak{}named\_\allowbreak{}target\_\allowbreak{}s0\_\allowbreak{}opus5\_\allowbreak{}rlstyle.mp4} & 118.04 & \texttt{fe613332} (video) & P3-L2 superseded\_\allowbreak{}clinic\_\allowbreak{}after\_\allowbreak{}ack STALE\_\allowbreak{}APPLIED\_\allowbreak{}UNFLAGGED \\
\texttt{p3v2\_\allowbreak{}blockA2\_\allowbreak{}opus5\_\allowbreak{}s14\_\allowbreak{}failure.mp4} & 151.04 & \texttt{81823a6e} (video) & S-C S-test\_\allowbreak{}id-14 DELIVERY\_\allowbreak{}AMBIGUOUS; S-L2 S-test\_\allowbreak{}id-14 STALE\_\allowbreak{}APPLIED\_\allowbreak{}FLAGGED \\
\texttt{p3v2\_\allowbreak{}blockA2\_\allowbreak{}opus5\_\allowbreak{}s8.mp4} & 159.04 & \texttt{9254f31b} (video) & S-C S-test\_\allowbreak{}id-8 APPLIED\_\allowbreak{}CORRECTLY; S-L2 S-test\_\allowbreak{}id-8 APPLIED\_\allowbreak{}CORRECTLY \\
\texttt{p3v2\_\allowbreak{}superseded\_\allowbreak{}session\_\allowbreak{}scripted\_\allowbreak{}demo.mp4} & 157.04 & \texttt{7c4424ba} (video) & S-L2 S-test\_\allowbreak{}id-0 APPLIED\_\allowbreak{}CORRECTLY; S-L2 S-test\_\allowbreak{}id-0 STALE\_\allowbreak{}APPLIED\_\allowbreak{}UNFLAGGED \\
\texttt{superseded\_\allowbreak{}clinic\_\allowbreak{}failure\_\allowbreak{}demo.mp4} & 75.0 & \texttt{e61d438f} (trajectory) & claude-fable-5-1 P3-L2 changed STALE\_\allowbreak{}APPLIED\_\allowbreak{}UNFLAGGED \\
\texttt{superseded\_\allowbreak{}clinic\_\allowbreak{}failure\_\allowbreak{}demo\_\allowbreak{}opus5.mp4} & 75.0 & \texttt{b6766eb2} (trajectory) & claude-opus-5 P3-L2 changed STALE\_\allowbreak{}APPLIED\_\allowbreak{}UNFLAGGED \\
\bottomrule
\end{tabular}
\endgroup

\end{center}

\clearpage
\setcounter{section}{0}
\renewcommand{\thesection}{S\arabic{section}}
\section{Supplementary phenomenon families, certification checks and exploratory evaluations}
\label{app:supplementary}

\paragraph{Scope and provenance of the supplementary corpus.}
EnvOS also has exploratory evaluations covering additional authoritative-state phenomena in clinic, hotel and travel environments. They use a Claude Opus~5 configuration with a separate shell/GUI execution interface and a variant-level rather than a matched-instance statistical design. Their role in this manuscript is to document the reach of the phenomenon--perturbation abstraction, the bounded four-layer certification suite, and trajectory-level diagnostic examples. They are \emph{not} extra cells in the prospective clinic and insurance review-disclosure experiments: no count from this appendix enters Table~\ref{tab:results}, the registered paired tests, the twelve-test sensitivity family or the main evaluator-challenge totals. Some of these exploratory results were originally reported in \citep{gopal2026envos}; they are retained here with their separate provenance, statistical limitations and observed defects rather than silently reinterpreted as prospective replications.

\subsection{General phenomenon model and executable perturbations}
The framework defines a \emph{phenomenon} as an application-independent class of real change to authoritative world state for reasons other than the agent's intended action. It calls an environment-specific implementation a \emph{perturbation} $\rho=(\delta,\tau,c,g)$: a state operator $\delta:\mathcal S\to\mathcal S$, a trigger predicate $\tau$ over action history, a channel $c$ through which the changed information may be observed, and a one-shot guard $g$. The agent acts in a partially observed environment $E=(\mathcal S,\mathcal A,\mathcal O,T,\Omega,s_0)$ and sees rendered observations $\Omega(s)$, not the complete authoritative state. When the trigger fires, the perturbation modifies that state; the observation-channel output is generated only after the update is recorded. This enforces the common principle \emph{state precedes channel}.

For the supplementary environments, task construction uses a manifest binding the base task, application state, phenomenon, perturbation, observation channel, checker and reward. The base-task generator and discriminator, both implemented through Claude Code, iteratively construct a task setup, completion state and verifier, adapted from the CUA-Gym dual-sandbox procedure. A specified phenomenon is then introduced and evaluated with a reference policy $\pi^\star$ and a deficient policy $\pi^-$. The former demonstrates a valid path, while the latter deliberately omits one recovery discipline. Human authors specify scenarios, invariants and policies; the generation procedure does not autonomously invent unrestricted new evaluation scenarios.

\begin{table}[ht]
\centering\small
\caption{Phenomenon families and application-level perturbation examples in the exploratory corpus. These examples are not additional prospective review-disclosure conditions.}
\label{tab:prior-phenomena}
\begin{tabular}{@{}P{3.0cm}P{5.05cm}P{4.9cm}@{}}
\toprule
Phenomenon & Causal situation & Example realization in an application \\
\midrule
Contested Resource & A previously selected resource is taken before commitment & A chosen clinic slot is booked by someone else; a selected hotel room becomes unavailable \\
Ghost Commit & An accepted write stores a different value & The requested appointment time or hotel room differs from the committed record \\
Mandate Drift & A task-relevant authoritative fact changes in place & A request's vaccine changes; a hotel booking policy is reissued \\
Cold-Start Inconsistency & Related application records disagree before the first action & A travel environment starts with inconsistent cross-application state \\
\bottomrule
\end{tabular}
\end{table}

\subsection{Four-layer certification in the supplementary environments}
For these environments, certification is one blocking admission step before model evaluation, implemented as 389 executable assertions across 19 suites in its three environments. These tests constitute a \emph{bounded certifier}, not a proof that an admitted evaluation is globally sound.

\paragraph{Layer 1: world validity.} The intended application state must change, and unlisted entities must remain unchanged. Checks compare serialized state outside the declared entity set, test the one-shot guard, verify the state-before-channel ordering, ensure monitoring probes do not trigger the change, and detect a contaminated starting world.

\paragraph{Layer 2: observation equivalence.} When multiple channels present the changed record, the certifier parses their emitted text into canonical facts and checks that the channels are consistent about the state. This prevents an alleged channel comparison from accidentally comparing different underlying facts. It is not the same as the present study's proof that an individual model received a particular served frame.

\paragraph{Layer 3: recovery and discriminative validity.} A reference policy must solve both clean and changed tasks; a deficient policy must solve the clean task but fail the changed one. Writing $R_d$ for the normalized invariant score, the admission requirement is
\[
R_d(\pi^\star\mid E)=1,\quad R_d(\pi^\star\mid E_\rho)=1,\quad
R_d(\pi^-\mid E)=1,\quad R_d(\pi^-\mid E_\rho)<1.
\]
This gives a constructive witness of solvability and a change-sensitive failure for that \emph{specified pair of policies}. It does not prove that an arbitrary agent notices the change or that the instance measures only one latent capability.

\paragraph{Layer 4: checker robustness.} Each named failure signature is exercised on a positive example and a paired near-miss that must not receive the same signature. Certification can consequently discover both incorrect rejection of valid behavior and overly permissive labels. A test suite that passes is still limited to those supplied positive and negative cases.

\subsection{Trajectory diagnosis and reward in the exploratory corpus}
Each supplementary environment has five task invariants $I_j$, scored from terminal authoritative state and provenance. It reports both strict success and a diagnostic dense score,
\[
\mathrm{Success}=\prod_{j=1}^{5}I_j,\qquad R_d=\frac15\sum_{j=1}^{5}I_j.
\]
Checkpoints identify broad trajectory stages (acquisition, change, detection, attempted repair, terminal state), while failure signatures label recurring mechanisms. These signals distinguish, for example, an agent that never revisited a silently modified record from one that recognized a discrepancy but applied a forbidden repair. Dense reward is diagnostic, not evidence that partial completion is fully correct. The primary stale-world experiments retain state-based success invariants and descriptive trajectory signatures, but do not inherit the supplementary corpus's dense scores or inferential sampling scheme.

\subsection{Exploratory environments, variants and certified outcomes}
The exploratory corpus evaluates a Claude Opus~5 configuration driven through Claude Code with shell and GUI interaction tools, in three multi-application environments. It reports 32 variants and 93 rollouts overall. The variant, not each rollout, is its primary statistical unit; repeated rollouts of one variant share the same initial state, perturbation and harness. The corpus includes both synthetic mocks and one independent third-party mail service. The certification process identified five nondiscriminative instances, six checker false-negatives and contamination of early clinic episodes by an uncleared Mailpit store.

\begin{table}[ht]
\centering\small
\caption{Descriptive performance in four supplementary phenomenon families. The leak-affected variant was excluded from these family counts. These rates are \emph{not} the main paper's prospective stale-world results.}
\label{tab:prior-results}
\begin{tabular}{@{}P{4.3cm}P{5.3cm}cc@{}}
\toprule
Phenomenon family & Recovery discipline & Environments & Successful rollouts \\
\midrule
Contested Resource & Recheck resource availability before committing & 2 & 1/5 \\
Ghost Commit & Verify committed value and repair non-destructively & 2 & 0/6 \\
Mandate Drift & Revisit a previously acquired authoritative fact & 2 & 1/9 \\
Cold-Start Inconsistency & Check inherited cross-application state & 1 & 1/3 \\
\midrule
Pooled four-family subset & --- & 3 & 3/23 \\
\bottomrule
\end{tabular}
\end{table}

The exploratory corpus reports 26 changed variants after excluding the leak-affected variant. Seventeen variants were handled successfully in 53 of 55 rollouts; the remaining nine variants form the four families in Table~\ref{tab:prior-results}, succeeding in three of 23 rollouts. Together these account for 78 changed rollouts. The aggregate unperturbed controls achieved 11/12 successes; cosmetic controls achieved 5/6. The four difficult families were selected from observed failures, so their comparison is descriptive rather than a preregistered confirmatory contrast. The exploratory corpus's unselected variant-level test gives $p=0.137$, with no comparison rejecting at $\alpha=0.05$. These values must not be added to the prospective stale-world instance counts or interpreted as replicated stale-world effect of review-time informations.

\paragraph{Certification findings are not detection-rate estimates.}
Five nondiscriminative instances had allowed the deficient policy to succeed even after the change. Six checker false-negatives rejected correct behavior because literal string matching stood in for the intended semantic condition; the checkers were revised and archived rollouts re-scored. The mail service carried messages across 11 early clinic rollouts, including a particularly affected variant excluded from primary analysis. A clean-world assertion covered 82 subsequent rollouts. Sensitivity analyses for early contamination are reported; they do not establish that such contamination was harmless merely from unchanged aggregate directions. These observed failures illustrate what the certification layers caught during construction; the proportions do not measure sensitivity to unknown future defects.

\subsection{Ghost Commit: a detection--recovery distinction}
The exploratory Ghost Commit case changes when an accepted write is altered. With the same underlying state operator and report channel, the alteration occurs either while a selection is still revisable or at submission when the interface no longer permits amendment. In the clinic realization, the agent re-reads and detects the altered value in all compared rollouts. It recovers successfully in 7/7 pre-commitment rollouts and 0/3 post-commitment rollouts; the cosmetic control succeeds in 3/3. In the post-commitment cases the agent attempts a prohibited delete-and-recreate repair while an allowed reporting/escalation route remains available. The difference localizes failure \emph{after detection}, not at initial observation.

The exploratory analysis also reports a two-sided Fisher exact $p=0.008$ on these rollout counts but marks it \emph{descriptive only}: each timing condition is a single frozen variant, so the repeated runs cannot act as independent variant-level experimental units. The comparison changes which repair operations are available and includes a process-rule invariant; it is not a matched multi-variant confirmatory estimate. The hotel Ghost Commit realization reproduces the failure pattern on a different application but has no corresponding pre-commitment timing arm. The individual trajectory and screenshot evidence belongs to that exploratory dataset, not to the prospective disclosure sample.

\subsection{Why perturbation certification alone does not establish prior-value exposure}
A valid perturbation and a sound reference/deficient-policy contrast do \emph{not} certify that an agent was previously served the value being superseded. The supplementary rates concern episodes where an injected change fired, not necessarily episodes in which a model had demonstrably acquired an old belief. The primary stale-world experiments add the prior-serve anchor and separately archived delivered observations to target that narrower phenomenon. The main experiments additionally study the information supplied at review, whereas the supplementary failure-family analyses address broader recovery disciplines. The use of distinct model interfaces, tasks and statistical units prevents pooling the two result sets or treating the supplementary corpus as an independent validation of the current scorer. This is the evidence boundary between descriptive supplementary evaluations and the prospective disclosure results.

\section{Illustrative event trace and request ordering}
\label{app:implementation-trace}
This explanatory material was moved from the main text to keep the research narrative centered on the phenomenon, certification gate, experimental controls, and verified endpoints. It is a schematic description, not evidence that the archive contains every displayed field in every historical run.

\subsection{An event-level trace and its evidence sources}
The following schematic makes the ordering constraints concrete without relying on an individual generated episode's literal field values. Let $v_0$ be the location shown in the original sessions record and $v_1$ the updated location. Application state is $S_t$, observations delivered to the policy are $O_t$, and the eventual booking action is $a_{\mathrm{commit}}$.

\begin{table}[t]
\centering\small
\caption{Illustrative timeline for one valid injected clinic episode. The table distinguishes application evidence from policy delivery and final task scoring.}
\label{tab:trace}
\begin{tabular}{@{}P{1.1cm}P{5.7cm}P{6.2cm}@{}}
\toprule
Event & Executed operation & Evidence and interpretation \\
\midrule
$e_0$ & Initialize synthetic task instance with session location $v_0$. & Pre-state snapshot identifies target and irrelevant records. \\
$e_1$ & Agent request serves the sessions page containing $v_0$. & Application access log proves the old-value serve. \\
$e_2$ & Harness delivers screenshot/text showing $v_0$ to the policy. & Frame and channel logs establish delivery, not comprehension. \\
$e_3$ & First eligible booking-wizard request follows. & Anchored request ID and once-only guard determine trigger. \\
$e_4$ & Perturbation atomically changes the stored location to $v_1$. & Transaction/event log and authoritative row snapshot confirm mutation. \\
$e_5$ & Agent reaches review: OFF, CUE or LIVE information is rendered. & Delivery of review-time information is evaluated independently from the mutation. \\
$e_6$ & Agent may revisit source, continue and commit a booking. & Browser trajectory, committed row and final report are archived. \\
$e_7$ & Validity classifier and outcome scorer run, followed by independent reconstruction. & A clean endpoint needs applicable invariants and corroborating evidence. \\
\bottomrule
\end{tabular}
\end{table}

The serve and the delivered observation are intentionally different checks. An access log can show that a server returned $v_0$ without proving that the agent was shown that response; delivery records can show a frame reached the policy without proving that the agent attended to it. Moreover, a later screenshot can display $v_1$ while the database transaction never occurred if the review display logic is defective; the authoritative state transition must have its own record. These distinctions explain why EnvOS cannot be reduced to a screenshot-based scorer or a benchmark message injected into the conversation.

\subsection{The execution loop}
Each policy invocation receives the declared task, a current browser observation and a bounded history of previous actions and self-authored memos; it returns one action. The harness executes that action, captures the next observation and records application events. The perturbation is integrated into the application request path rather than appended as a post-hoc model instruction. The following pseudocode is explanatory; the released source and archived event schema in the appendices define the implemented details.

\begin{lstlisting}[basicstyle=\ttfamily\footnotesize,breaklines=true,frame=single]
# Application request handler (event ordering is server-side)
def handle_request(request):
    if old_value_was_served and eligible(request) and not fired:
        atomic_mutation_of_authoritative_record()
        record_mutation(request.id)
        fired = True
    response = application_serve(request)
    if response_contains_original_value(response) and not fired:
        record_old_value_serve(request.id)
        old_value_was_served = True
    return response

# Controlled browser and policy loop
for step in range(action_budget):
    observation = capture_and_record_delivered_frame()
    action = invoke_stateless_policy(task, observation, history)
    result = browser_execute(action)  # uses handle_request
    archive(action, result, state_checkpoints)
    if task_committed(result):
        break

validity = classify_validity_before_outcome(archive)
if validity.is_counted:
    outcome = score_application_state_at_commit(archive)
independent_checkers.reconstruct(archive, validity, outcome)
\end{lstlisting}
The mutation is represented inside the application-side request handler, before the handler serves the triggered response; the client loop does not inject a changed value into a model prompt. This pseudocode describes event order and evidence responsibilities, not a verbatim implementation or a full specification of all eligible routes.

\end{document}